\documentclass[double]{article}
\usepackage{times}

\usepackage{amsmath}
\usepackage{amsfonts}
\usepackage{amssymb}
\usepackage{amsthm}

\usepackage{graphicx}
\usepackage{subcaption}
\usepackage{float}
\usepackage{adjustbox}

\usepackage{authblk}
\usepackage{hyperref}

\usepackage{array}
\newcolumntype{L}[1]{>{\raggedright\arraybackslash}p{#1}}

\begin{document}

\title{\Large Profiling Privacy Preservation Against Gradient Inversion Attacks in Tabular Federated Learning}

\author{Ivo {\"O}sterberg Nilsson$^{1,x}$, Maximilian Birr Engvall$^{1,x}$, Viktor Valadi$^{2}$, Teddy Lazebnik$^{1,3,*}$\\
\(^1\) Department of Computing, J{\"o}nk{\"o}ping University, J{\"o}nk{\"o}ping, Sweden\\
\(^2\) Scaleout Systems, Uppsala, Sweden\\ 
\(^3\) Department of Information Systems, University of Haifa, Haifa, Israel\\
\(^x\) These authors contributed equally.\\
\(^*\) Corresponding author: teddy.lazebnik@ju.se 
}

\date{ }

\maketitle 

\begin{abstract}
\noindent
Federated learning (FL) enables multiple data holders to train machine learning models collaboratively without centralizing raw data, making it useful in privacy sensitive domains such as healthcare and institutional data sharing. FL keeps data local to clients while communicating only model updates, such as gradients or model deltas. Nevertheless, these updates can expose private client data through gradient inversion attacks (GIAs). We study this risk for tabular FL under an honest-but-curious server threat model across FL protocols, client batch sizes, training stages, attacker assumptions, model architectures, and binary classification, multiclass classification, and regression tasks. We use MIMIC-IV and complementary benchmark datasets. Our evaluation distinguishes numerical and categorical recovery, baseline recoverability, feature level recovery, and exact match rate (EMR). We evaluate FedSGD gradients and FedAvg model deltas with an exposure aligned protocol, comparing attacked models after matched client data exposure rather than matched communication rounds. We compare multilayer perceptron (MLP), ResNet, and FT-Transformer models, and isolate architecture effects through an MLP grid over width, depth, activation, normalization, and dropout. The results show that small client batches and updates representing few distinct records are most vulnerable. Larger local batches and stronger aggregation reduce reconstruction but do not eliminate leakage. FT-Transformer is consistently harder to invert than one-hot baselines, while reconstructability also varies substantially within the MLP family. These findings identify architecture as a practical privacy variable in tabular FL. We also show that aggregate reconstruction accuracy can overstate complete record recovery in sparse data, making EMR and baseline comparisons essential.
\end{abstract}

\thispagestyle{empty}
% Begin using page numbers and a header
\pagestyle{plain}
\setcounter{page}{1}% reset page number to 1

\section{Introduction}
Federated learning (FL) enables multiple data holders to train machine learning models collaboratively without centralizing raw data \cite{mcmahan2017communication,kairouz2021advances,yang2019federated}. By keeping data local and exchanging gradients or model updates, FL supports learning in settings where centralized data sharing is undesirable or infeasible because of privacy, governance, or institutional constraints. As a result, FL has become especially attractive in domains such as healthcare, in which data are highly sensitive and often distributed across organizations \cite{antunes2022federated,xu2021federated,vaid2021federated,pan2024adaptive}.

Nevertheless, FL should not be conflated with complete privacy. Even when raw data never leaves the client, the gradients and model updates communicated during training may themselves reveal sensitive information \cite{zhu2019deep,zhao2020idlg,geiping2020inverting,huang2021evaluating}. Gradient inversion attacks (GIAs) are a prominent attack family in this space, in which an adversary, or an honest-but-curious server, attempts to reconstruct private training examples from shared gradients or related updates \cite{zhu2019deep,geiping2020inverting,huang2021evaluating,vero2023tableak}. Early work showed that gradients can leak both inputs and labels \cite{zhu2019deep,zhao2020idlg}, and later work showed that stronger optimization can recover substantially more detailed inputs from gradients \cite{geiping2020inverting}.

Despite this risk, the empirical understanding of GIAs is still dominated by image domains \cite{zhu2019deep,geiping2020inverting,yin2021gradinversion,hatamizadeh2022gradvit}. For tabular data, the problem is different. Image reconstructions can often be inspected visually, which makes attack success intuitive to assess. Tabular reconstructions instead require evaluation that distinguishes numerical tolerances from categorical exact matches, respects column semantics and constraints, and separates partial feature recovery from exact row recovery. These challenges make both optimization and evaluation harder, and they limit how directly conclusions from image benchmarks transfer to structured data regimes that are common in real world FL deployments. Vero et al.\ highlighted this gap by formulating a reconstruction attack for tabular FL and showing that both FedSGD and FedAvg can leak sensitive tabular records under realistic settings \cite{vero2023tableak}.

At the same time, tabular neural networks (NNs) have become increasingly important in mainstream machine learning \cite{gorishniy2021revisiting,borisov2022deep,shmuel2025comprehensive}. Their architectures and input representations can change the constraints that gradients impose on the original features. Models that consume direct one-hot encodings expose categorical variables at the input layer, whereas transformer style tabular models typically map categorical variables through learned embeddings before applying attention and feedforward transformations. This changes the differentiable relationship between the observed FL training update and the input features, and may therefore change how easily an attacker can reconstruct tabular records from gradients or model updates.

This architectural question is timely because recent tabular neural architectures increasingly replace direct feature processing with learned feature interactions. TabNet uses sequential attention to select features, TabTransformer applies self attention to categorical embeddings, and FT-Transformer tokenizes both numerical and categorical features before processing them with transformer blocks \cite{arik2021tabnet,huang2020tabtransformer,gorishniy2021revisiting}. These designs were developed primarily to improve predictive modeling, but they also change how gradients relate to individual numerical and categorical features. Their privacy consequences therefore have yet to be investigated.

This issue is becoming practically relevant because neural tabular models are increasingly studied in federated settings, including structured clinical prediction tasks \cite{lindskog2023federated,lindskog2022federated,vaid2021federated,pan2024adaptive}. Much of this work evaluates predictive performance, communication efficiency, or generalization across sites. Far less is known about how the chosen tabular architecture changes the reconstructability of client data from the communicated FL update.

These trends motivate our study of when neural tabular models are vulnerable to gradient inversion and how architecture affects that vulnerability. We investigate GIAs against tabular FL models under two common update protocols, FedSGD and FedAvg. We evaluate three complementary model families: a one-hot multilayer perceptron (MLP), a one-hot ResNet, and FT-Transformer. This allows us to compare models that expose categorical variables directly as one-hot inputs with a transformer based model that first maps categorical variables through learned embeddings.

This study makes five empirical and methodological contributions: (i) we introduce an exposure aligned experimental protocol for FL training; (ii) we provide a broad empirical evaluation of tabular gradient inversion across task types, update protocols, and model families; (iii) we evaluate tabular gradient inversion in settings that move closer to realistic deployment conditions, including a clinical MIMIC-IV in-hospital mortality prediction task; (iv) we show that architecture and input representation are practical privacy variables in tabular FL; and (v) we identify architecture dependent leakage patterns within the MLP family. These results show that architecture affects reconstructability not only across model families, but also within ordinary tabular NN design choices. To the best of our knowledge, this is the first gradient inversion study of tabular FL with neural models on MIMIC-IV and the first to study gradient inversion against a transformer based tabular architecture, FT-Transformer.

Together, these experiments show that tabular FL leakage is concentrated in weakly aggregated settings, especially small client batches, direct one-hot input representations, and strong label knowledge assumptions. More privacy aware design choices reduce reconstruction quality but do not provide formal privacy guarantees. The resulting analysis provides an empirical view of the privacy and utility tradeoff and empirical guidance for evaluating gradient inversion risk in NN based tabular FL deployments.

The remainder of the study is organized as follows. Section~\ref{sec:background} provides a computational background on FL, privacy leakage, and tabular neural learning. Section~\ref{sec:methodology} formalizes the setting and notation, presents the threat model, datasets, preprocessing steps, models, attack pipeline, and evaluation metrics. Section~\ref{sec:exp} describes the experimental protocol. Section~\ref{sec:results} reports the empirical findings, and Section~\ref{sec:discussion} discusses their implications, limitations, and recommendations for privacy aware tabular FL design. Additional robustness, sensitivity, and implementation validation results are reported in the appendix. 

\section{Background}
\label{sec:background}
In standard FL, clients keep raw data local while communicating model updates to a coordinating server \cite{mcmahan2017communication,kairouz2021advances}. The privacy relevance of these updates depends on the FL protocol and on what the server can observe \cite{kairouz2021advances,bonawitz2017practical}. In FedSGD, the server observes a gradient tied closely to one local mini batch. In FedAvg, the server observes a model update that may combine several local steps and epochs \cite{mcmahan2017communication}. Prior work on FedAvg leakage shows that local training complicates inversion because the attacker must reconstruct through several unobserved local optimizer steps rather than match a single batch gradient directly \cite{dimitrov2022data}.

FL does not by itself provide a formal privacy guarantee. Secure aggregation can prevent the server from seeing individual client updates by revealing only an aggregate over many clients \cite{bonawitz2017practical}, and differential privacy can add formal protection by clipping and perturbing gradients or model updates \cite{abadi2016deep,mcmahan2018learning}. These defenses change the information available to the attacker. In contrast, our threat model studies the standard noncryptographic setting in which an honest-but-curious server observes individual client gradients or model deltas before aggregation. This setting isolates the privacy risk of ordinary FL communication rather than systems protected by secure aggregation or differential privacy.

GIAs challenge the assumption that gradients are safe to share. Deep Leakage from Gradients reconstructed private training examples by optimizing dummy inputs and labels so that their gradients match a shared gradient \cite{zhu2019deep}. iDLG showed that, for differentiable models trained with cross-entropy over one-hot labels, labels can often be inferred directly from gradients \cite{zhao2020idlg}. Inverting Gradients strengthened this line of work through a magnitude invariant objective and improved optimization, recovering high fidelity images even from trained networks and identifying strong input recovery relationships for fully connected layers \cite{geiping2020inverting}. R-GAP added a recursive analytical perspective and connected gradient leakage to architectural rank properties \cite{zhu2021rgap}.

Later work showed that gradient inversion success is not a fixed property of FL, but depends on the training configuration and the information available to the attacker. Evaluations of GIAs and defenses have shown that success depends strongly on label knowledge, batch size, model state, and the observed client update \cite{huang2021evaluating}. Other work has shown that leakage estimates can change substantially in more realistic FL settings, especially when local training dynamics, private batch normalization statistics, dropout behavior, architecture, input resolution, or post processing defenses are considered \cite{huang2021evaluating,scheliga2023dropout,hatamizadeh2023unsafe,valadi2026practical}. These findings motivate experiments that vary attacker label knowledge, attack budget, client batch size, training stage, client heterogeneity, and FedAvg local computation in order to distinguish stable leakage patterns from artifacts of a single favorable attack setting.

Beyond the original convolutional image settings, later work has examined both transformer architectures and other modalities. GradViT demonstrated that vision transformers can be vulnerable to gradient inversion and argued that attention based architectures introduce their own leakage behavior \cite{hatamizadeh2022gradvit}. In language, attacks such as LAMP reconstruct text from gradients by combining gradient matching with language model priors and mixed continuous and discrete optimization \cite{balunovic2022lamp}. Together, these studies show that gradient leakage is architecture and representation dependent. However, their settings differ from tabular FL. Images and text have strong spatial or linguistic priors, while tabular records contain heterogeneous numerical and categorical columns whose correctness cannot be judged by visual or semantic plausibility alone.

Tabular data introduces reconstruction and evaluation challenges that differ from standard image benchmarks. Categorical variables create a mixed discrete and continuous optimization problem, while numerical and categorical attributes require different evaluation rules. Feature level metrics are computed after aligning reconstructed and attacked rows, since gradient matching does not identify row order within the reconstructed batch. TabLeak addressed the reconstruction side of this problem through softmax relaxation, pooled ensembling, and entropy based uncertainty estimation, and introduced tabular reconstruction metrics for mixed feature types \cite{vero2023tableak}.

The choice of model architecture is especially important for tabular learning. Revisiting Deep Learning Models for Tabular Data showed that simple MLP style models remain strong baselines, that a ResNet architecture is a competitive tabular baseline, and that FT-Transformer is a strong transformer adaptation for tabular data, while also emphasizing that no single method universally dominates gradient boosted decision trees across datasets \cite{gorishniy2021revisiting}.

Neural tabular models are also increasingly studied in federated settings \cite{lindskog2023federated,lindskog2022federated}, including structured clinical prediction tasks \cite{vaid2021federated,pan2024adaptive}. These studies mainly evaluate predictive utility, communication behavior, or performance under client and institutional heterogeneity. They do not directly evaluate whether the chosen tabular architecture changes how easily client records can be reconstructed from the communicated FL update. This leaves open the privacy question studied here, namely whether embedded tabular representations reduce gradient inversion risk or mainly change the form of the attack.

Architecture also matters for privacy because gradients constrain the input through the model representation. One-hot MLP and ResNet models expose categorical features directly in the input representation, whereas FT-Transformer maps ordinal categorical inputs through learned embeddings before attention and feedforward layers. This changes the differentiable relationship available to an attacker and requires a relaxation that is aware of the model representation. Existing transformer leakage work shows that attention based models can leak in vision and language settings \cite{hatamizadeh2022gradvit,balunovic2022lamp}. In tabular FL, the corresponding question is how embedded categorical representations change inversion relative to direct one-hot inputs. This gap motivates a direct comparison between one-hot and embedded tabular architectures, together with tests of FT-Transformer categorical attack parameterizations and analyses of whether dropout accounts for the observed architecture level leakage differences.

\section{Methodology}
\label{sec:methodology}
The experimental methodology begins with the honest-but-curious server threat model and then defines the FL setting, training protocols, datasets and preprocessing pipeline, model families, attack setup, and evaluation metrics.

\subsection{Threat model}
\label{sec:threat-model}
We study privacy leakage under a centralized FL threat model with an \textit{honest-but-curious} server in a standard noncryptographic FL setting. The server coordinates training, distributes the global model, receives individual client updates, and applies the prescribed aggregation rule. It follows the protocol but uses the received updates to infer information about the private data that produced them. This setting isolates the privacy risk of standard FL training, where raw records remain local but individual client updates are still observed by the coordinating server. Below, we outline the elements of the GIA under this setting.

\subsubsection{Adversary position and observability}
The training occurs over discrete communication rounds \(t \in [1, T]\) where \(T < \infty\). At communication round \(t\), the server holds the current global model parameters \(w_t\) and receives a client update from each participating client. We denote the observed update from client \(c\) by \(u_c^{(t)}\). The exact form of \(u_c^{(t)}\) depends on the specific FL protocol.

In FedSGD, the observed update is the gradient computed on the local client batch \(B_c^{(t)}\): \(u_c^{(t)} = g_c^{(t)} = \nabla_w \mathcal{L}(w_t; B_c^{(t)})\). Thus, each attacked FedSGD gradient corresponds directly to a single client mini batch in the default setting. In contrast, under FedAvg, the observed update is the model delta after local client training: \(u_c^{(t)} = \Delta_c^{(t)} = w_{t,c}^{\mathrm{local}} - w_t\). In this case, the model delta may summarize several local optimization steps or local epochs before communication. The attacker therefore observes a post-training model delta rather than a raw one-step gradient. This difference is important because FedAvg can either reduce leakage by mixing information across more local examples or preserve leakage when the same small local data are reused for several local steps.

\subsubsection{Adversary knowledge}
The server is assumed to know the training pipeline that it coordinates, which includes the model architecture, current model parameters, loss function, optimizer, learning rates, local batch size, number of local steps or local epochs, aggregation rule, and the public input schema used to construct the model. In particular, the server is assumed to know the feature order, feature types, numerical and categorical positions, categorical cardinalities, and model input representation.

Conversely, the attacker does not know the private client records and does not receive direct access to the underlying raw client dataset. Reconstruction is performed only from the observed client update and the public training specification described above. Client local normalization statistics, client marginals, and evaluation baselines are not available to the attacker. They are retained only by the evaluator for decoding reconstructions, computing metrics, and interpreting leakage.

The reconstruction attack uses the known model representation. For one-hot MLP and ResNet models, the reconstructed variables are optimized directly in the model input space. For FT-Transformer, categorical features are optimized in an expanded continuous space where each categorical feature is represented by one block over its possible categories. Each block is converted to probabilities either by simplex normalization or by a softmax over logits, and the resulting probabilities are multiplied by the learned embedding matrix. This allows gradients to pass through the categorical embedding interface during reconstruction.

\subsubsection{Label knowledge}
The main experiments use a strong label known attacker. Under this setting, the attacker is given the true labels or target values associated with the attacked batch and optimizes only the input features. This assumption provides an upper bound estimate of feature leakage and is consistent with prior gradient inversion evaluations in which label information may be inferred, leaked through gradients, or otherwise available to the attacker.

Due to the fact that label knowledge may not always be available in practice, we also evaluate label unknown attacks as a sensitivity analysis. In the label unknown setting, the attacker does not receive the true labels or target values. The attack uses dummy labels and optimizes only the input features. This distinction allows us to separate leakage caused by the feature gradients themselves from leakage that depends on a stronger label aware adversary.

\subsubsection{Adversary objective}
The primary adversary objective is record reconstruction. Given an observed client update \(u_c^{(t)}\), the attacker attempts to recover the tabular feature matrix \(X_c^{(t)}\) used by client \(c\) at round \(t\). The attack produces a reconstructed batch \(\hat{X}_c^{(t)}\) by solving an optimization problem that makes the gradient or model delta induced by the reconstructed batch match the observed client update.

The secondary objective is distributional inference. Even when exact record reconstruction is incomplete, the reconstructed batch may still reveal client level feature structure, such as categorical marginals, sparse history indicators, or numerical ranges. We therefore distinguish record level recovery from feature level and distributional leakage throughout the evaluation.

Because the observed update is invariant to the order of rows within the attacked batch, reconstruction is evaluated after aligning reconstructed rows to true rows with Hungarian matching. The matching procedure and reconstruction metrics are described in the evaluation protocol.

The concrete optimization procedure used to produce \(\hat{X}_c^{(t)}\) is described in the GIA setup below.

\subsection{Federated learning setting}
We study a centralized, simulation based horizontal FL setting with full client participation in each communication round. The experiments consider two operational update types. In FedSGD, each client communicates a gradient computed from one local mini batch. In FedAvg, each client communicates a model delta after local training.

In our FedSGD settings, each client computes gradients on one local mini batch per communication round in the standard setting used throughout the main experiments. Concretely, for client \(c\) at round \(t\), the server distributes the current global parameters \(w_t\), the client evaluates the loss on its current local mini batch \(B_c^{(t)}\), and returns the corresponding gradient
\(
g_c^{(t)} = \nabla_w \mathcal{L}(w_t; B_c^{(t)})
\).
The server then forms the unweighted mean over participating clients,
\(
\bar g^{(t)} = \frac{1}{K}\sum_{c=1}^{K} g_c^{(t)}
\)
and applies a single global stochastic gradient descent (SGD) step
\(
w_{t+1} = w_t - \eta \bar g^{(t)}
\),
where \(\eta\) is the global learning rate. No server momentum or adaptive server optimizer is used in this protocol. In the default FedSGD setting, each client computes one mini batch gradient per communication round, so each observed client gradient corresponds to exactly one local client batch.

In our FedAvg settings, each client starts from the current global model \(w_t\) and performs local SGD on its own client data, producing an updated local model \(w_{t,c}^{\mathrm{local}}\). The client transmits the model delta
\(
\Delta_c^{(t)} = w_{t,c}^{\mathrm{local}} - w_t.
\).
The server aggregates these deltas by a weighted average,
\(
w_{t+1} = w_t + \sum_{c=1}^{K} \frac{m_c^{(t)}}{\sum_{j=1}^{K} m_j^{(t)}} \Delta_c^{(t)}
\),
where \(m_c^{(t)}\) is the number of examples processed locally by client \(c\) in that round. Thus, unlike FedSGD, the communicated update is a post-training model delta rather than a raw gradient.

FedAvg local computation is varied through the number of local epochs and the amount of client data processed before communication. In the full pass setting, each local epoch iterates once through the client dataloader, so increasing the number of local epochs increases the amount of optimization performed before aggregation. The framework can also restrict a communication round to a fixed number of local mini batches, which allows the communicated update to represent a controlled amount of local computation. In all FedAvg settings, local optimization uses plain SGD style updates without momentum.

In both settings, a fixed communication round budget is not sufficient for fair comparison across batch sizes or local computation settings. Hence, the number of examples processed per round depends on the local batch size and on the amount of local computation performed before aggregation. As a result, two runs with the same number of rounds can correspond to substantially different amounts of data exposure and substantially different stages of model training. Since leakage changes over training, comparing configurations at matched round counts would confound the effect of the factor of interest with differences in optimization progress.

To reduce this confounding, we control training by a client exposure budget rather than by a fixed number of rounds. Formally, let \(n_c^{\mathrm{eff}}\) denote the effective number of training examples available to client \(c\) after excluding incomplete final mini batches in the client dataloaders, i.e.
\(
n_c^{\mathrm{eff}} = \left\lfloor \frac{n_c}{B} \right\rfloor B
\),
where \(n_c\) is the raw client dataset size and \(B\) is the local batch size. If client \(c\) has processed \(s_c^{(t)}\) examples by round \(t\), we define its exposure as
\(
e_c^{(t)} = \frac{s_c^{(t)}}{n_c^{\mathrm{eff}}}
\).
We then track
\(
e_{\min}^{(t)} = \min_c e_c^{(t)}
\),
\(
e_{\mathrm{avg}}^{(t)} = \frac{1}{K}\sum_{c=1}^{K} e_c^{(t)}
\),
\(
e_{\max}^{(t)} = \max_c e_c^{(t)}
\).

Training is budgeted by a minimum exposure target \(E\). A run continues until the least exposed client reaches this target, i.e. until \(e_{\min}^{(t)} \geq E\). This design ensures that comparisons across batch sizes are made at approximately matched client level data exposure rather than at matched round counts.

Validation and attack evaluation are indexed by exposure. Attack points are defined in two ways. In the automatic schedule, we attack five communication round positions: the first round, approximately \(25\%\), \(50\%\), and \(75\%\) of the training trajectory, and the final round. The first automatic attack point is evaluated on the client update computed from the initialized global model before any server aggregation has occurred. Since the total round budget is derived from the minimum exposure budget, these automatically scheduled attack points usually correspond to approximately spaced exposure levels, but they are not explicit exposure milestones. Exposure scheduled attack points instead select the communication round whose client computation advances the least exposed client to a configured exposure milestone. The scheduling choice only determines when the attack is run. In both schedules, attack evaluation uses the client updates produced before server aggregation. These updates are per client gradients in FedSGD and per client model deltas in FedAvg.

Throughout the results, the initialized attack point refers to the client update attacked before any server aggregation, computed from the initial global model. The final attack point refers to the last attacked client update selected by the experiment's attack schedule. In exposure scheduled analyses this corresponds to the final exposure milestone, while in automatically scheduled analyses it corresponds to the final communication round.

Our exposure metric is intended as the federated analogue of an epoch count. In centralized training, one epoch means one full pass over the training set. In FL, however, clients may have different effective dataset sizes and may process different numbers of examples per round. We therefore measure progress in units of client local effective passes rather than server rounds.

\subsection{Datasets and preprocessing}
Our empirical study uses four tabular datasets covering binary classification, multiclass classification, and regression. This dataset suite is designed to test whether gradient inversion leakage patterns persist across task types while also validating the framework on a realistic clinical prediction problem using MIMIC-IV. Table~\ref{tab:datasets-summary} summarizes the datasets, prediction targets, scale, and experimental role of each dataset.

\begin{table}[H]
    \centering
    \small
    \caption{Summary of the tabular datasets used in the empirical study.}
    \label{tab:datasets-summary}
    \begin{adjustbox}{max width=\textwidth}
    \begin{tabular}{
          L{0.24\textwidth}
          L{0.2\textwidth}
          L{0.05\textwidth}
          L{0.18\textwidth}
          L{0.33\textwidth}}
    \hline
    \textbf{Dataset} & \textbf{Prediction setting} & \textbf{Rows} & \textbf{Features (num./cat.)} & \textbf{Role and main characteristics}\\
    \hline
    Adult~\cite{adult_dataset} & Binary classification & 32{,}561 & 14 (5/9) & U.S. Census income benchmark, target income above \$50K.\\
    \hline
    Private multiclass benchmark & Multiclass classification & 6{,}000 & 101 (54/47) & Three-class benchmark with class proportions of approximately 30.0\%, 30.9\%, and 39.1\%.\\
    \hline
    California Housing~\cite{scikit-learn,KELLEYPACE1997291} & Regression & 20{,}640 & 8 (8/0) & Continuous housing value prediction benchmark, target median house value.\\
    \hline
    MIMIC-IV cohort~\cite{PhysioNet-mimiciv-3.1} & Binary classification & 30{,}000 & 219 (211/8) & In-hospital mortality prediction at the admission level, with 2.16\% positive labels, subject-disjoint splits, and tier-3 features available at admission time.\\
    \hline
    \end{tabular}
    \end{adjustbox}
\end{table}

All datasets are processed through a common tabular pipeline before federated partitioning. We first separate targets from features, remove rows with missing targets, determine feature types from metadata for MIMIC-IV and from the training data for the other datasets, and apply minimal missing value handling by imputing numerical features with 0 and categorical features with the dataset specific missing token. We then create centralized train, validation, and test splits: benchmark datasets use 70\%/15\%/15\% splits, with stratification for classification and shuffled splits for regression, while MIMIC-IV uses externally generated subject-disjoint splits. The training split is divided across clients using independent and identically distributed (IID) partitions by default, with stratified folds for classification and shuffled folds for regression; Dirichlet-based non-IID partitions are used only in the heterogeneity experiments, with regression targets first discretized into quantile buckets for partitioning. Categorical encoding depends on the target model: MLP and ResNet use one-hot encoding, whereas FT-Transformer uses ordinal encoding with an explicit \texttt{\_\_UNK\_\_} category for unseen values. Numerical features are normalized using centralized training statistics for validation and test data, while each client training partition uses its own local statistics; these local statistics are not available to the attacker and are used only by the evaluator to invert normalization and compute reconstruction metrics. For MIMIC-IV, the provided positive class weight is also used in the binary loss to account for class imbalance.

\subsection{Model families and training objectives}
We evaluate three neural model families for tabular FL: a configurable multilayer perceptron (MLP), ResNet, and FT-Transformer. The default small MLP uses two hidden layers of width 32, LayerNorm, GELU activation, and no dropout. The MLP also serves as the model family for the architectural sensitivity analysis, where we vary hidden width, hidden depth, normalization, activation, and dropout. The architecture grid spans hidden widths \(\{32, 64, 128\}\), numbers of hidden layers \(\{1, 2, 3\}\), normalization layers \(\{\text{BatchNorm}, \text{LayerNorm}\}\), activation functions \(\{\text{ReLU}, \text{GELU}\}\), and dropout rates \(\{0.0, 0.1\}\).

To complement this configurable baseline, we include two established deep tabular architectures, ResNet and FT-Transformer, using the tabular model implementations from prior work \cite{gorishniy2021revisiting}. The ResNet baseline uses two residual blocks with block width 192, hidden multiplier 2.0, first dropout 0.15, and second dropout 0.0. For FT-Transformer, we use a reduced capacity variant rather than the full default configuration. This variant uses two transformer blocks with block width 128, four attention heads, attention dropout 0.1, feedforward network hidden multiplier \(4/3\), feedforward network dropout 0.1, and residual dropout 0.0. The smaller configuration keeps the repeated federated training and attack experiments practical while preserving the main architectural characteristics of FT-Transformer.

The input representation follows the model family. MLP and ResNet operate on one-hot encoded categorical features together with normalized numerical features, whereas FT-Transformer operates on ordinal encoded categorical inputs together with numerical features. This is not merely an implementation convenience. It ensures that each architecture is evaluated under the input representation it is designed to use, which is important both for utility and for attack realism.

Training objectives are matched to the prediction task. For binary classification, models are trained with binary cross-entropy with logits. For multiclass classification, we use multiclass cross-entropy, and for regression, we use mean squared error. For the imbalanced MIMIC-IV in-hospital mortality task, models are trained with binary cross-entropy with logits using a positive class weight of \(13\), as specified in the dataset metadata. This weights errors on positive mortality labels more strongly while leaving the attack pipeline otherwise unchanged.

\subsection{Gradient inversion attack setup}
\begin{figure}[H]
    \centering
    \includegraphics[width=0.90\textwidth]{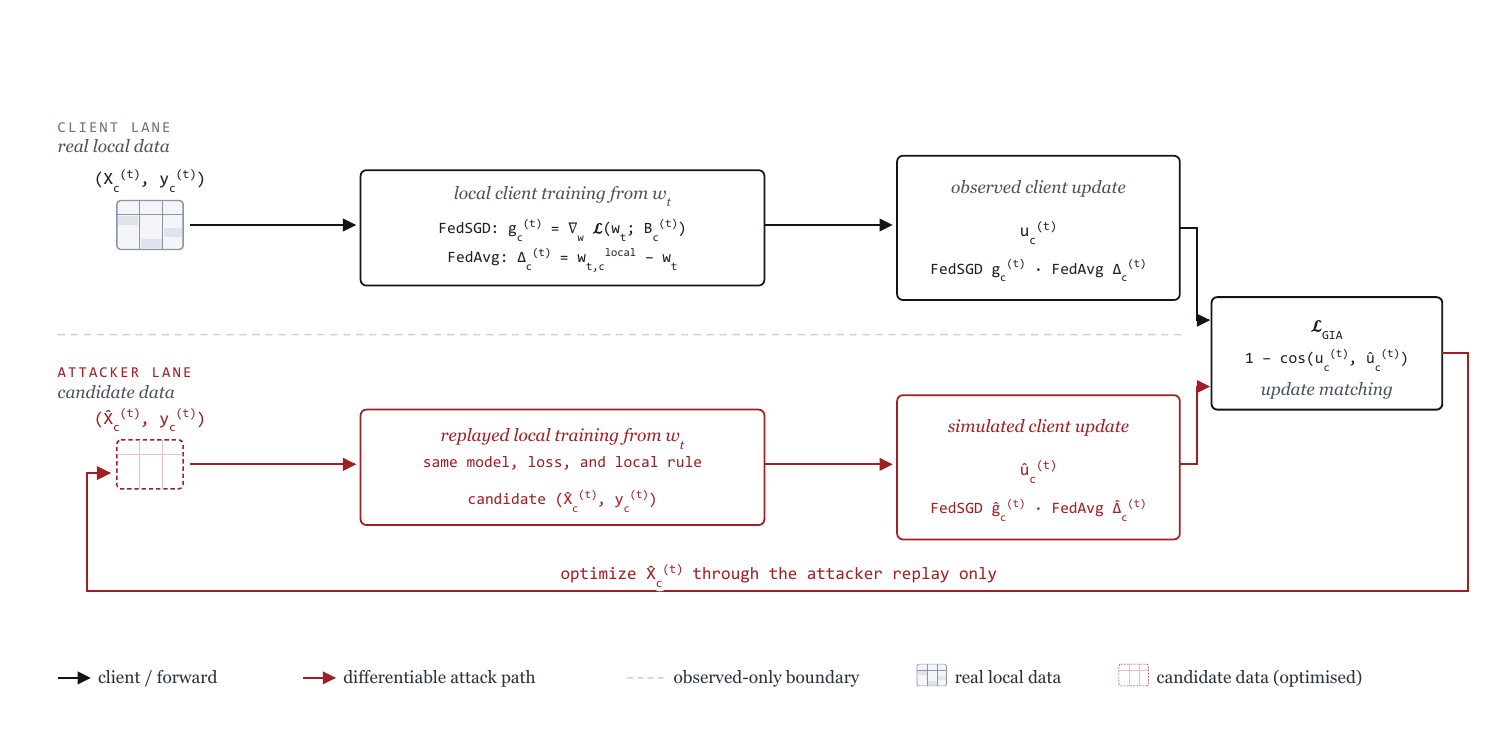}
    \caption[Federated update matching view of gradient inversion.]{The client and the attacker start from the same attacked round global model \(w_t\). The client performs local training on real data \((X,y)\) and produces an observed update \(u\), while the attacker applies the same local training rule to a candidate batch \((\hat{X},y)\) and obtains a simulated update \(\hat{u}\). In FedSGD, \(u\) is the per client gradient. In FedAvg, \(u\) is the client model delta after local training.}
    \label{fig:gradient-matching-attack-loop}
\end{figure}

We implement the attack pipeline on top of the open source LeakPro framework and adapt it to tabular FL. Concretely, our reconstruction pipeline builds on LeakPro's implementation of an optimization based gradient matching attack \cite{geiping2020inverting}, extending it with tabular preprocessing, tabular reconstruction metrics, and vectorized execution. The attack is always run by the server after observing the client update specified by the FL protocol. In FedSGD, this update is a per client batch gradient. In FedAvg, it is a per client model delta produced by local training. The attacker then optimizes a reconstructed batch so that the gradient or model delta induced by the reconstruction matches the observed client update as closely as possible. Figure \ref{fig:gradient-matching-attack-loop} presents a schematic view of the process.

Let \(u_c^{(t)}\) denote the observed client update after flattening the relevant gradient or model delta into a parameter vector, and let \(\hat{X}_c^{(t)}\) denote the reconstructed client batch. The attack minimizes a matching objective of the form
\[
  \mathcal{L}_{\mathrm{GIA}}(\hat{X}_c^{(t)}) = 1 - \cos\!\big(u_c^{(t)}, \hat{u}_c^{(t)}(\hat{X}_c^{(t)})\big)
\]
where \(\hat{u}_c^{(t)}(\hat{X}_c^{(t)})\) is the corresponding simulated gradient or model delta induced by the reconstruction under the same model, loss, local training rule, and number of local epochs as used by the attacked client. For FedSGD, \(\hat{u}_c^{(t)}(\hat{X}_c^{(t)})\) is a simulated gradient. For FedAvg, it is the model delta induced by replaying local training on the reconstructed batch. In practice, LeakPro computes this objective over the concatenated parameter space vector using cosine similarity.

For tabular data, the reconstruction variables are initialized as Gaussian noise in the model input space. When labels are assumed known, the true batch labels are copied into the attack setup. When labels are unknown, the attack uses dummy labels rather than the true batch labels, and only the input features are optimized. In the main experiments, labels are known to the attacker unless explicitly stated otherwise. The reconstruction variables are optimized with the Adam optimizer \cite{kingma2015adam} using an initial attack learning rate of \(0.06\), followed by linear learning rate decay. We do not replace reconstruction gradients by their elementwise signs, unlike the signed gradient variant described for image reconstruction by Geiping et al.~\cite{geiping2020inverting}. Unless noted otherwise, each attack is run for 10{,}000 optimization iterations and the iterate with the lowest attack loss is retained as the final reconstruction.

Client updates contain parameter gradients or model deltas, not nonparameter buffers such as BatchNorm running means or variances. For models with BatchNorm, client update computation and attack simulation use stored BatchNorm statistics rather than private batch statistics. This follows the more conservative setting discussed by Huang et al., where private batch statistics are not shared during federated training \cite{huang2021evaluating}. During reconstruction, the attacker optimizes in evaluation mode. This disables dropout in the simulated update and gives a more stable optimization objective, but it does not reveal the dropout masks used when a dropout enabled client update was produced \cite{scheliga2023dropout}.

A key implementation detail is that the reconstruction must be differentiable through the attacked model. For standard one-hot encoded tabular models such as the MLP and ResNet, the reconstructed inputs can be optimized directly in feature space. For FT-Transformer, however, categorical variables must pass through learned embeddings. We therefore use a model aware differentiable surrogate that maps reconstructed categorical variables into the representation consumed by the transformer during the forward pass. This ensures that the attack objective remains end-to-end differentiable for ordinal encoded categorical features rather than relying on a nondifferentiable decode and reembed procedure.

For FT-Transformer, we use two differentiable categorical parameterizations. The \emph{probability simplex} parameterization optimizes one expanded categorical block per feature and normalizes each block to the probability simplex before multiplying it by the learned categorical embedding matrix. The \emph{categorical logits} parameterization optimizes the same expanded categorical blocks as logits and maps each block to probabilities with a softmax before the embedding lookup. Unless otherwise stated, the categorical logits parameterization uses softmax temperature $\tau=1$ and initialization scale $5$. Tables and figures use the short labels \emph{Categorical logits} and \emph{Probability simplex}.

The attack framework supports two evaluation modes. In the standard attack point mode, the server attacks the client update observed at the selected attack point, with the corresponding client batch serving as ground truth for evaluation. In the fixed batch mode, the same client batch is attacked repeatedly across scheduled attack points, which separates changes caused by model training from changes caused by batch resampling. We use exposure scheduled attack points for fixed batch and FedAvg local computation experiments, where leakage must be compared at matched effective client exposure.

Relative to the original LeakPro codebase, our main modifications are practical rather than conceptual. We add a tabular specific data extension, support vectorized attack execution across multiple clients with matched batch shapes, and integrate reconstruction evaluation directly into the federated training loop. These changes make large scale attack experiments feasible while preserving the underlying optimization objective and threat model of the original gradient inversion framework.

\subsection{Evaluation protocol}
We evaluate both the predictive utility of the federated models and the privacy leakage of the corresponding GIAs. Because the study spans binary classification, multiclass classification, and regression, the exact utility metrics depend on the task objective. Privacy evaluation is based on feature level and record level reconstruction quality after aligning reconstructed and true client batches. Throughout, utility and privacy are evaluated at synchronized training states so that leakage can be interpreted relative to model quality at the same stage of training.

\subsubsection{Federated utility metrics}
For federated model utility, we report task appropriate validation and test metrics computed from the current global model. In binary classification, we record loss, accuracy, F1 score, receiver operating characteristic area under the curve (ROC-AUC), and precision-recall area under the curve (PR-AUC). Binary class predictions are obtained by thresholding the sigmoid output at 0.5 for threshold dependent metrics such as accuracy and F1, while ROC-AUC and PR-AUC are computed from the predicted probabilities. For the MIMIC-IV in-hospital mortality task, ROC-AUC is used as the primary utility metric in the main plots and tradeoff analyses, while PR-AUC remains informative because of class imbalance.

For multiclass classification, we report loss, accuracy, macro precision, macro recall, macro F1, and weighted F1. Macro metrics are included because they are less dominated by large classes and therefore provide a more balanced view of model quality under class imbalance. For regression, we report loss, mean squared error (MSE), mean absolute error (MAE), and \(R^2\). For plotting and privacy--utility analyses, the default validation utility metric is task dependent. We use validation ROC-AUC for binary classification, validation macro F1 for multiclass classification, and validation \(R^2\) for regression. Final held out test metrics are computed after restoring the best validation loss checkpoint.

\subsubsection{Privacy leakage metrics}
Our primary privacy metric is reconstruction accuracy, adapted from TabLeak \cite{vero2023tableak}. This metric evaluates each reconstructed row at the feature level and then averages over numerical and categorical attributes. For numerical feature \(j\), a reconstruction is counted as correct if it falls within \(\epsilon_j = 0.319\sigma_j\) of the ground truth value, where \(\sigma_j\) is the client local standard deviation. For categorical features, correctness requires exact recovery of the category. The per row TabLeak accuracy is then the fraction of all numerical and categorical features recovered correctly, and the aggregate reconstruction accuracy is the mean of this quantity across rows in the attacked batch.

To make the reconstruction behavior more interpretable, we also report reconstruction accuracies by feature type, which average only over numerical or categorical features. In addition, we compute exact match reconstruction rates. The exact match rate (EMR) counts the fraction of rows with perfect reconstruction.

Because raw reconstruction accuracy can be difficult to interpret without context, we compare the attacker against two simple baselines. The first is a client marginal prior baseline that reconstructs each numerical feature by the client local mean and each categorical feature by the most likely category under the client marginals. The second is a uniform random baseline that samples numerical values uniformly from the observed global training support and categorical values uniformly over valid categories, averaged over repeated Monte Carlo draws. We report both baseline accuracies and the attack gains over these baselines. These baselines are used as evaluation reference points rather than as additional information given to the attacker.

We interpret these baselines as reference points for the strength of privacy leakage. Reconstructions close to the uniform random baseline indicate little usable feature information beyond uninformed guessing. Reconstructions above random but below the client marginal prior baseline indicate that the update reveals some feature level or distributional structure, but does not outperform a simple client marginal reconstruction. Reconstructions that exceed the client marginal prior baseline provide stronger evidence of attack specific leakage beyond marginal feature recovery. High EMRs indicate row level recovery and therefore the strongest form of empirical privacy leakage measured in this study. These row level results are reported as auxiliary metrics.

Finally, we compute additional distributional diagnostics. These include the nearest neighbor distance from each reconstructed row to the true attacked batch and marginal likelihood statistics for reconstructed numerical and categorical values under the corresponding client distributions. These diagnostics are not used as the primary privacy outcome, but they help assess whether reconstructions preserve plausible client level feature structure beyond exact feature matches.

\subsubsection{Matching and reconstruction evaluation}
Gradient matching is permutation invariant with respect to the order of the rows inside an attacked batch. A reconstruction may contain the correct records in the wrong order and still produce nearly the same matched gradient or model delta. Direct row wise comparison without alignment would therefore underestimate reconstruction quality. To address this, we first align reconstructed and true rows using Hungarian matching \cite{kuhn1955hungarian} on the full encoded feature tensor with a row wise \(L_1\) cost. All reported privacy metrics are computed only after this assignment step.

Metric computation is performed in the original tabular feature space rather than in any internal attack parameterization. For numerical features, the evaluator transforms reconstructed values back from normalized client space to the original feature scale using the client local mean and standard deviation, which are not available to the attacker during reconstruction. For FT-Transformer, if the attack is optimized in a differentiable categorical surrogate space, the reconstruction is decoded back to ordinal input space before evaluation. This ensures that the reported privacy metrics correspond to recovered tabular records rather than to intermediate attack variables.

When labels are included in the attacked batch, they are retained in the attack artifacts for inspection. The main privacy metrics reported focus on feature reconstruction quality, since the central question is how much of the original client data can be inferred from the communicated client update.

\section{Experiments}
\label{sec:exp}
In this section, we outline the empirical evaluation of the proposed framework. We initially outline the experimental setup followed by the presentation of the obtained results.

\subsection{Setup}
We organize the empirical study around four questions. First, we test whether gradient inversion remains effective across tabular tasks, datasets, and model families. Second, we use exposure based training alignment to study how leakage evolves during FL training under comparable client data exposure. Third, we test whether the FT-Transformer embedded categorical representation changes reconstruction difficulty. Fourth, we vary MLP structural and module choices on MIMIC-IV to test whether ordinary architectural decisions amplify or suppress leakage. The experiments use a shared data processing, federated training, attack, and metric pipeline. Client batch size is the central aggregation factor and is evaluated first under FedSGD. Additional controlled analyses vary fixed versus resampled attacked batches, client partitioning, attacker label knowledge, attack budget, FedAvg local computation, MLP modules, and FT-Transformer attack parameterization. These analyses test whether the main leakage patterns persist under changes to the data, model, protocol, and attacker configuration.

Unless otherwise stated, FedSGD experiments use \(10\) clients, IID client partitioning, one mini batch gradient per client per round, global learning rate \(0.01\), and a minimum exposure budget of \(25\). Thus, the FedSGD setting does not perform a full local pass over each client dataset before communication. Each client contributes one gradient computed on the selected local mini batch. The FedSGD batch size comparisons use client batch sizes \(1,2,4,8,16,32,64,128,\) and \(256\). Fixed batch and MLP architecture experiments focus mainly on batch sizes \(8\) and \(32\), which provide representative intermediate settings where leakage remains measurable while aggregation effects are visible. Client heterogeneity experiments override the default IID partition with Dirichlet client splits. In addition, FedAvg experiments use \(3\) clients and vary the amount of local computation through the number of local epochs and the cap on client examples processed per round. These experiments use smaller batch sizes and local computation grids because each communicated update aggregates more local information than in FedSGD. The FedAvg local computation experiments use a minimum exposure budget of \(24\). This value is chosen because the local epoch grid contains \(1\), \(2\), and \(4\) local epochs. With \(4\) local epochs, one communication round advances client exposure by four effective local passes, so an exposure target of \(24\) aligns with all local epoch counts in the grid. An exposure target of \(25\) would force the four epoch runs to overshoot the target. Most FedSGD experiments use seeds \(7\), \(13\), and \(42\), while the FedAvg local computation experiments use seeds \(0\), \(1\), \(7\), \(13\), and \(42\).

These client counts are fixed to keep the empirical comparisons controlled and computationally tractable rather than to model a specific deployment scale. FedSGD uses \(10\) fully participating clients so that each round exposes multiple independent per client gradients while still allowing the batch size, dataset, and architecture comparisons to be repeated across seeds. FedAvg uses \(3\) clients because differentiating through local client training is substantially more expensive in memory and runtime. The learning rate is held fixed across comparable runs to avoid confounding privacy differences with model-specific optimizer tuning.

The default attack configuration uses LeakPro's optimization based gradient inversion with known labels, an initial attack learning rate of \(0.06\) with linear decay, and \(10{,}000\) optimization iterations. Label knowledge and attack budget are varied only in the corresponding sensitivity experiments. Most FedSGD experiments use five automatically selected attack points. Fixed batch FedSGD experiments use exposure scheduled attack points at exposure milestones \(0\), \(1\), \(5\), \(10\), and \(25\), where \(0\) denotes the first client update from the initial global model. FedAvg local computation experiments use exposure scheduled attacks at the initialized update and the final aligned exposure milestone \(24\). Thus, fixed batch and FedAvg local computation attacks are compared at matched effective client data exposure rather than at matched communication rounds.

The fixed batch experiments repeatedly attack the same selected client batch across exposure-scheduled attack points. This design separates changes caused by model training from changes caused by attacking different sampled batches at different rounds. Across all experiments, results are written as per attack, per round, and per run summaries. Reported tables give mean \(\pm\) standard deviation over seeds.

\subsection{Results}
\label{sec:results}
The results are organized around two questions: when gradient inversion becomes practically effective in tabular FL, and how model architecture changes reconstruction difficulty. We distinguish weakly aggregated settings, where the server observes updates from few client records, from settings where updates mix more records or pass through architectures with less direct input gradient relationships. Particular attention is given to FT-Transformer because its embedded categorical representation gives the attacker a different reconstruction problem than the one-hot MLP and ResNet baselines.

We first establish that gradient inversion remains feasible in federated tabular learning across multiple task types and datasets. The benchmark datasets span binary classification, multiclass classification, and regression, and serve as a controlled stage before the later MIMIC-IV analysis turns to a clinically motivated FL setting.

Table~\ref{tab:robustness-batch-size-trained-onehot} shows that gradient inversion remains feasible after training across binary classification, multiclass classification, and regression. Across the one-hot ResNet and small MLP models, smaller client batches generally produce higher reconstruction accuracy, while larger batches reduce it. This pattern is clearest on Adult and the private multiclass benchmark, where reconstruction falls substantially as batch size increases. California Housing is the main final attack point exception. At batch size \(1\), reconstruction is lower than at several larger batch sizes, with ResNet increasing from \(0.371\) at batch size \(1\) to \(0.748\) at batch size \(2\), and the small MLP increasing from \(0.462\) to \(0.815\). This suggests that the regression setting can change the relationship between aggregation and reconstruction after training, especially at the smallest batch size. However, this does not mean that regression batch size \(1\) updates are generally less vulnerable. At the initialized attack point, Appendix Table~\ref{tab:robustness-batch-size-initialized} shows strong reconstruction for California Housing at batch size \(1\), with \(0.992\) for ResNet and \(0.967\) for the small MLP. The initialized results also show stronger leakage before training across the benchmark settings, indicating that training attenuates but does not remove the privacy risk.

\begin{table}[H]
  \centering                                                                                
  \small
  \caption{Reconstruction accuracy at the final attack point across the benchmark datasets for ResNet and the small MLP under FedSGD batch size comparison. Each
  cell reports mean $\pm$ standard deviation across 3 seeds.}
  \label{tab:robustness-batch-size-trained-onehot}
  \resizebox{\textwidth}{!}{%
  \begin{tabular}{lcccccc}
  \hline
  \multicolumn{1}{l}{} & \multicolumn{2}{c}{\textbf{Adult}} & \multicolumn{2}{c}{\textbf{Private multiclass}} & \multicolumn{2}{c}{\textbf{California Housing}} \\
  \textbf{Batch size} & \textbf{ResNet} & \textbf{Small MLP} & \textbf{ResNet} & \textbf{Small MLP} & \textbf{ResNet} & \textbf{Small MLP} \\
  \hline
  1 & 0.945 $\pm$ 0.025 & 0.964 $\pm$ 0.062 & 0.774 $\pm$ 0.034 & 0.813 $\pm$ 0.080 & 0.371 $\pm$ 0.128 & 0.462 $\pm$ 0.121 \\
  2 & 0.780 $\pm$ 0.096 & 0.926 $\pm$ 0.017 & 0.543 $\pm$ 0.023 & 0.453 $\pm$ 0.099 & 0.748 $\pm$ 0.069 & 0.815 $\pm$ 0.112 \\
  4 & 0.676 $\pm$ 0.068 & 0.823 $\pm$ 0.031 & 0.455 $\pm$ 0.009 & 0.388 $\pm$ 0.008 & 0.690 $\pm$ 0.067 & 0.717 $\pm$ 0.080 \\
  8 & 0.573 $\pm$ 0.019 & 0.675 $\pm$ 0.030 & 0.433 $\pm$ 0.041 & 0.375 $\pm$ 0.012 & 0.558 $\pm$ 0.056 & 0.583 $\pm$ 0.053 \\
  16 & 0.444 $\pm$ 0.014 & 0.550 $\pm$ 0.011 & 0.403 $\pm$ 0.023 & 0.371 $\pm$ 0.007 & 0.474 $\pm$ 0.016 & 0.578 $\pm$ 0.032 \\
  32 & 0.403 $\pm$ 0.026 & 0.456 $\pm$ 0.002 & 0.376 $\pm$ 0.005 & 0.310 $\pm$ 0.007 & 0.417 $\pm$ 0.017 & 0.534 $\pm$ 0.020 \\
  64 & 0.352 $\pm$ 0.037 & 0.354 $\pm$ 0.005 & 0.356 $\pm$ 0.005 & 0.240 $\pm$ 0.002 & 0.402 $\pm$ 0.002 & 0.462 $\pm$ 0.011 \\
  128 & 0.366 $\pm$ 0.034 & 0.307 $\pm$ 0.005 & 0.338 $\pm$ 0.013 & 0.218 $\pm$ 0.000 & 0.401 $\pm$ 0.026 & 0.407 $\pm$ 0.005 \\
  256 & 0.482 $\pm$ 0.003 & 0.300 $\pm$ 0.002 & 0.301 $\pm$ 0.023 & 0.221 $\pm$ 0.001 & 0.396 $\pm$ 0.007 & 0.397 $\pm$ 0.004 \\
  \hline
  \end{tabular}
  }
\end{table}

Table~\ref{tab:robustness-batch-size-fttransformer} extends the benchmark to FT-Transformer. FT-Transformer is consistently harder to invert than the one-hot baselines, but it does not remove leakage. On Adult and California Housing, initialized attack point reconstruction exceeds \(0.70\) at batch size \(1\) and remains above \(0.35\) across the full batch size range. By the final attack point, reconstruction is generally lower than at initialization, although the decrease is less regular than for the one-hot baselines. Notably, the private multiclass benchmark behaves differently. It has \(101\) input features, compared with \(14\) for Adult and \(8\) for California Housing, and FT-Transformer starts from much lower initialized reconstruction on this dataset. Final attack point reconstruction is higher than initialized reconstruction at every batch size, with the largest increases at batch sizes \(1\) and \(2\), but remains below the one-hot baselines. This indicates that FT-Transformer changes the leakage profile, but should not be interpreted as a complete defense.

\begin{table}[H]
  \centering
  \footnotesize
  \caption{Reconstruction accuracy across the benchmark datasets for FT-Transformer under FedSGD batch size comparison. Each cell reports mean $\pm$ standard deviation across 3 seeds.}
  \label{tab:robustness-batch-size-fttransformer}
  \resizebox{\textwidth}{!}{%
  \begin{tabular}{lcccccc}
  \hline
  \multicolumn{1}{l}{} & \multicolumn{3}{c}{\textbf{Initialized attack point}} & \multicolumn{3}{c}{\textbf{Final attack point}} \\
  \textbf{Batch size} & \textbf{Adult} & \textbf{Private multiclass} & \textbf{California Housing} & \textbf{Adult} & \textbf{Private multiclass} & \textbf{California Housing} \\
  \hline
  1 & 0.721 $\pm$ 0.080 & 0.234 $\pm$ 0.032 & 0.721 $\pm$ 0.089 & 0.557 $\pm$ 0.075 & 0.320 $\pm$ 0.015 & 0.304 $\pm$ 0.051 \\
  2 & 0.495 $\pm$ 0.058 & 0.191 $\pm$ 0.021 & 0.496 $\pm$ 0.097 & 0.430 $\pm$ 0.021 & 0.295 $\pm$ 0.026 & 0.442 $\pm$ 0.080 \\
  4 & 0.424 $\pm$ 0.015 & 0.179 $\pm$ 0.006 & 0.384 $\pm$ 0.017 & 0.440 $\pm$ 0.024 & 0.231 $\pm$ 0.019 & 0.353 $\pm$ 0.011 \\
  8 & 0.373 $\pm$ 0.037 & 0.165 $\pm$ 0.006 & 0.414 $\pm$ 0.058 & 0.358 $\pm$ 0.048 & 0.217 $\pm$ 0.014 & 0.357 $\pm$ 0.059 \\
  16 & 0.374 $\pm$ 0.044 & 0.155 $\pm$ 0.006 & 0.424 $\pm$ 0.054 & 0.338 $\pm$ 0.025 & 0.206 $\pm$ 0.004 & 0.356 $\pm$ 0.012 \\
  32 & 0.377 $\pm$ 0.035 & 0.159 $\pm$ 0.007 & 0.434 $\pm$ 0.057 & 0.343 $\pm$ 0.011 & 0.188 $\pm$ 0.008 & 0.337 $\pm$ 0.031 \\
  64 & 0.383 $\pm$ 0.021 & 0.165 $\pm$ 0.010 & 0.485 $\pm$ 0.039 & 0.322 $\pm$ 0.021 & 0.183 $\pm$ 0.009 & 0.305 $\pm$ 0.015 \\
  128 & 0.396 $\pm$ 0.012 & 0.174 $\pm$ 0.012 & 0.503 $\pm$ 0.042 & 0.330 $\pm$ 0.010 & 0.198 $\pm$ 0.007 & 0.310 $\pm$ 0.017 \\
  256 & 0.412 $\pm$ 0.015 & 0.180 $\pm$ 0.010 & 0.537 $\pm$ 0.034 & 0.335 $\pm$ 0.014 & 0.190 $\pm$ 0.005 & 0.330 $\pm$ 0.025 \\
  \hline
  \end{tabular}
  }
\end{table}

We next analyze how leakage evolves during federated training on MIMIC-IV, using exposure aligned comparisons rather than communication round counts. Table~\ref{tab:batch-size-mimic-tableak-acc} shows that client batch size remains a strong privacy lever on MIMIC-IV, but the effect is model dependent. For ResNet and the small MLP, reconstruction decreases substantially as batch size grows. At batch size \(1\), both one-hot models are almost perfectly reconstructed at the initialized attack point, whereas at batch sizes \(128\) and \(256\) reconstruction is much lower at both attack points. Training also reduces leakage for the one-hot models, although it does not remove it. At the final attack point, batch size \(1\) still reaches \(0.669\) for ResNet and \(0.850\) for the small MLP. As before, FT-Transformer behaves differently. It is consistently harder to invert than the one-hot baselines, but its leakage is less monotonic in batch size. Reconstruction peaks at batch size \(16\), reaching \(0.426\) at the initialized attack point and \(0.462\) at the final attack point, before dropping at larger batch sizes. This indicates that MIMIC-IV leakage is shaped by both aggregation and architecture specific reconstruction behavior. The EMR results in Appendix Table~\ref{tab:batch-size-mimic-iv-strict-emr} further show that feature level reconstruction should not be equated with complete record recovery. At batch size \(1\), final attack point EMR reaches \(0.500\) for ResNet and \(0.700\) for the small MLP, but is \(0.000\) for FT-Transformer. The fixed client batch control in Appendix Table~\ref{tab:fixed-batch-size-mimic-tableak-acc} presents results with similar patterns. Holding the attacked batch fixed changes the exact reconstruction levels, but it does not alter the main conclusion that the batch size trend is not explained by unusually easy or unusually hard batch draws. FT-Transformer remains harder to invert than the one-hot models, and larger one-hot batches remain less vulnerable than smaller ones.

\begin{table}[H]
    \centering
    \small
    \caption{Reconstruction accuracy at the initialized attack point and final attack point for MIMIC-IV. Initialized attack point denotes the attack before any global model aggregation, and final attack point denotes the last attacked point under the exposure budget. Each cell reports mean $\pm$ standard deviation across 3 seeds.}
    \label{tab:batch-size-mimic-tableak-acc}
    \resizebox{\textwidth}{!}{%
    \begin{tabular}{lcccccc}
    \hline
    \multicolumn{1}{l}{} & \multicolumn{3}{c}{\textbf{Initialized attack point}} & \multicolumn{3}{c}{\textbf{Final attack point}} \\
    \textbf{Batch size} & \textbf{FT-Transformer} & \textbf{ResNet} & \textbf{Small MLP} & \textbf{FT-Transformer} & \textbf{ResNet} & \textbf{Small MLP} \\
    \hline
    1 & 0.254 $\pm$ 0.028 & 0.988 $\pm$ 0.010 & 0.989 $\pm$ 0.009 & 0.248 $\pm$ 0.028 & 0.669 $\pm$ 0.116 & 0.850 $\pm$ 0.038 \\
    2 & 0.300 $\pm$ 0.031 & 0.953 $\pm$ 0.026 & 0.970 $\pm$ 0.006 & 0.210 $\pm$ 0.107 & 0.635 $\pm$ 0.024 & 0.764 $\pm$ 0.085 \\
    4 & 0.343 $\pm$ 0.020 & 0.926 $\pm$ 0.053 & 0.962 $\pm$ 0.029 & 0.337 $\pm$ 0.096 & 0.607 $\pm$ 0.025 & 0.657 $\pm$ 0.010 \\
    8 & 0.357 $\pm$ 0.044 & 0.840 $\pm$ 0.021 & 0.901 $\pm$ 0.019 & 0.333 $\pm$ 0.186 & 0.703 $\pm$ 0.024 & 0.623 $\pm$ 0.016 \\
    16 & 0.426 $\pm$ 0.041 & 0.836 $\pm$ 0.065 & 0.822 $\pm$ 0.034 & 0.462 $\pm$ 0.230 & 0.736 $\pm$ 0.089 & 0.650 $\pm$ 0.037 \\
    32 & 0.384 $\pm$ 0.096 & 0.656 $\pm$ 0.079 & 0.624 $\pm$ 0.001 & 0.338 $\pm$ 0.100 & 0.504 $\pm$ 0.065 & 0.480 $\pm$ 0.044 \\
    64 & 0.248 $\pm$ 0.001 & 0.460 $\pm$ 0.097 & 0.471 $\pm$ 0.023 & 0.247 $\pm$ 0.001 & 0.332 $\pm$ 0.069 & 0.378 $\pm$ 0.058 \\
    128 & 0.363 $\pm$ 0.089 & 0.398 $\pm$ 0.071 & 0.358 $\pm$ 0.020 & 0.201 $\pm$ 0.048 & 0.212 $\pm$ 0.034 & 0.291 $\pm$ 0.015 \\
    256 & 0.396 $\pm$ 0.130 & 0.316 $\pm$ 0.031 & 0.302 $\pm$ 0.022 & 0.168 $\pm$ 0.048 & 0.192 $\pm$ 0.021 & 0.243 $\pm$ 0.016 \\
    \hline
    \end{tabular}
    }
\end{table}

The MIMIC-IV experiments require more careful interpretation than the benchmark datasets. MIMIC-IV contains many sparse clinical history indicators and repeated marginal patterns, so aggregate reconstruction accuracy can be inflated by correctly recovering common feature values. Appendix Table~\ref{tab:batch-size-mimic-iv-baseline-reference} reports the corresponding baselines: the client marginal prior remains near \(0.94\) across batch sizes, while uniform random reconstruction remains near \(0.04\)--\(0.05\). We therefore interpret the MIMIC-IV results primarily through relative comparisons across models and aggregation levels, together with exact row recovery, rather than as direct evidence that each reconstruction uniquely identifies a record.

\begin{figure}[H]
    \centering
    \includegraphics[width=0.99\textwidth]{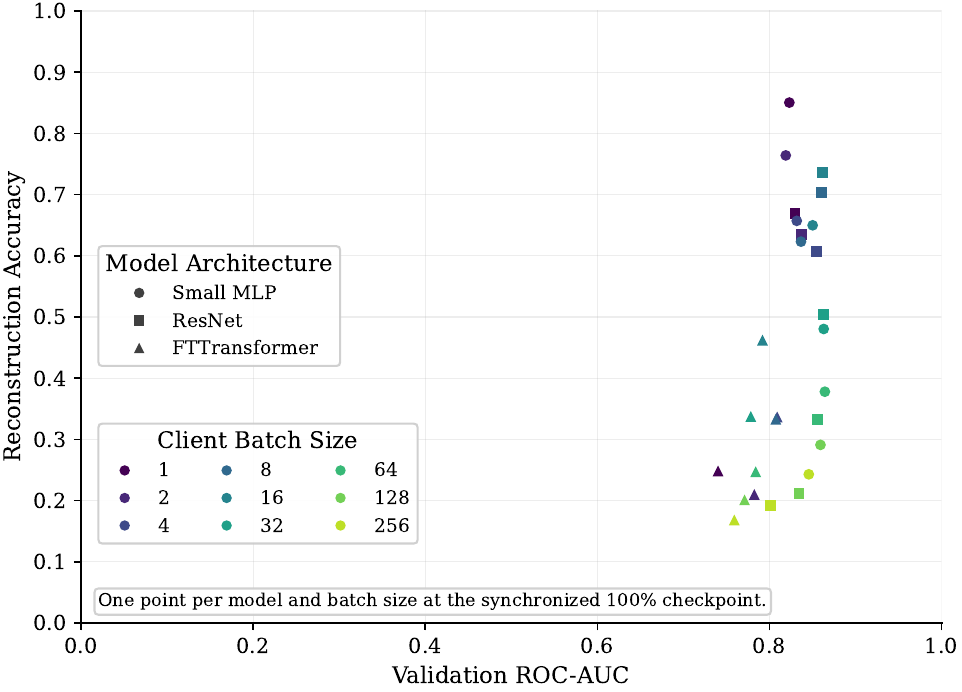}
    \caption{Privacy and utility tradeoff at the final attack point across client batch sizes on MIMIC-IV. Each marker represents a model architecture and each point corresponds to one batch size.}
    \label{fig:mimic-batch-cross-model-tradeoff}
\end{figure}

Figure~\ref{fig:mimic-batch-cross-model-tradeoff} shows the final attack point privacy and utility tradeoff across client batch sizes on MIMIC-IV, connecting the batch size leakage results to model utility. Utility is measured by validation ROC-AUC at the synchronized final attack point. For ResNet and the small MLP, increasing the client batch size substantially reduces leakage while maintaining or improving utility. ResNet improves from \(0.830\) ROC-AUC at batch size \(1\) to \(0.863\) at batch size \(32\), while reconstruction falls from \(0.669\) to \(0.504\). The small MLP shows the same direction, with ROC-AUC rising from \(0.823\) to \(0.863\) and reconstruction falling from \(0.850\) to \(0.480\). FT-Transformer has lower leakage overall, but its tradeoff curve is less smooth: utility improves at small batch sizes, then declines at larger batch sizes, while leakage peaks at intermediate batch sizes.

\begin{figure}[H]
      \centering

      \begin{subfigure}[t]{0.32\textwidth}
          \centering
          \includegraphics[width=\textwidth]{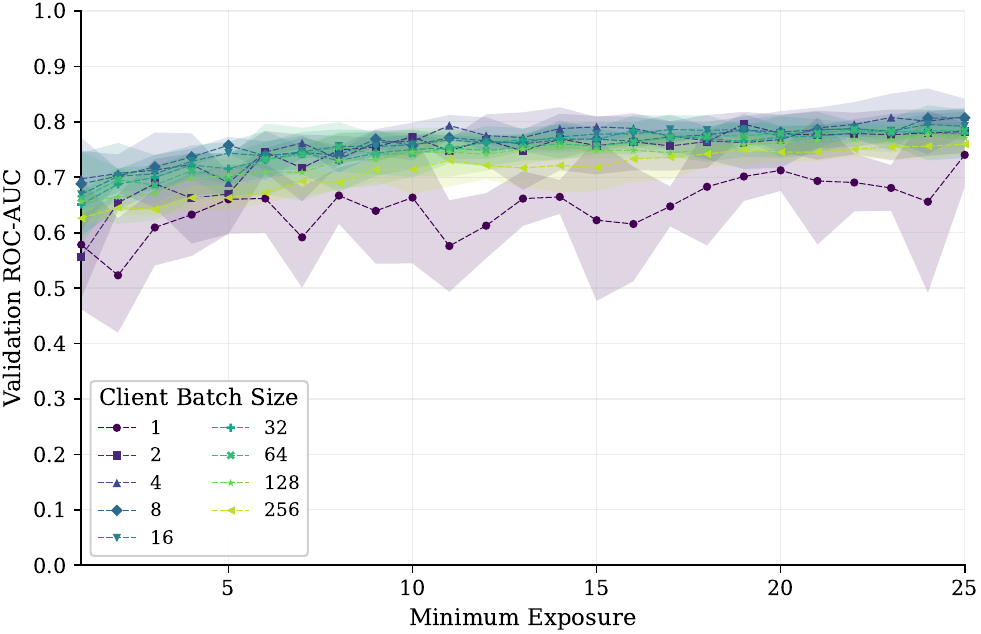}
          \caption{FT-Transformer}
      \end{subfigure}
      \hfill
      \begin{subfigure}[t]{0.32\textwidth}
          \centering
          \includegraphics[width=\textwidth]{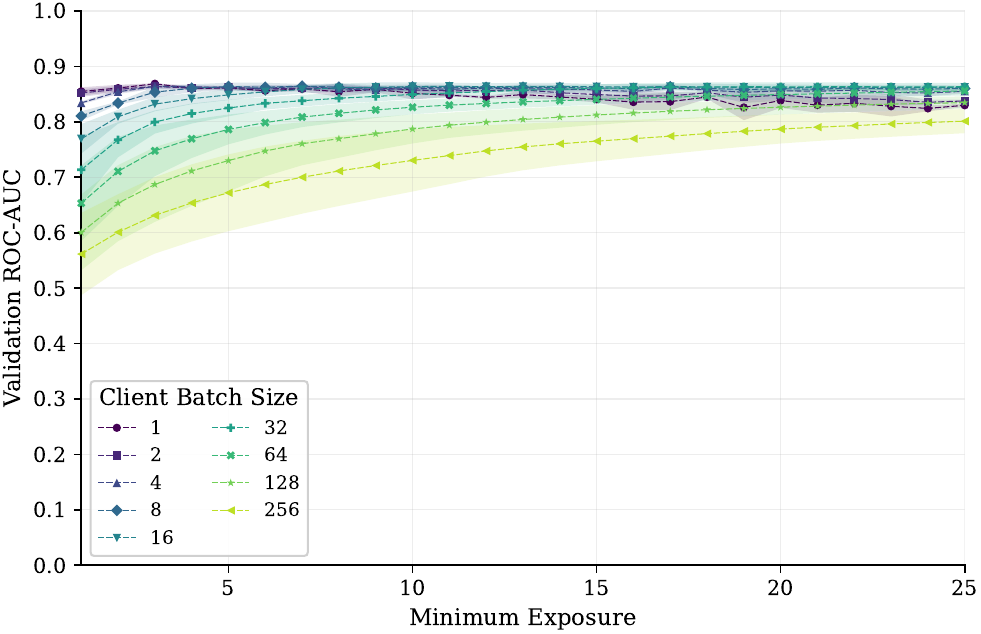}
          \caption{ResNet}
      \end{subfigure}
      \hfill
      \begin{subfigure}[t]{0.32\textwidth}
          \centering
          \includegraphics[width=\textwidth]{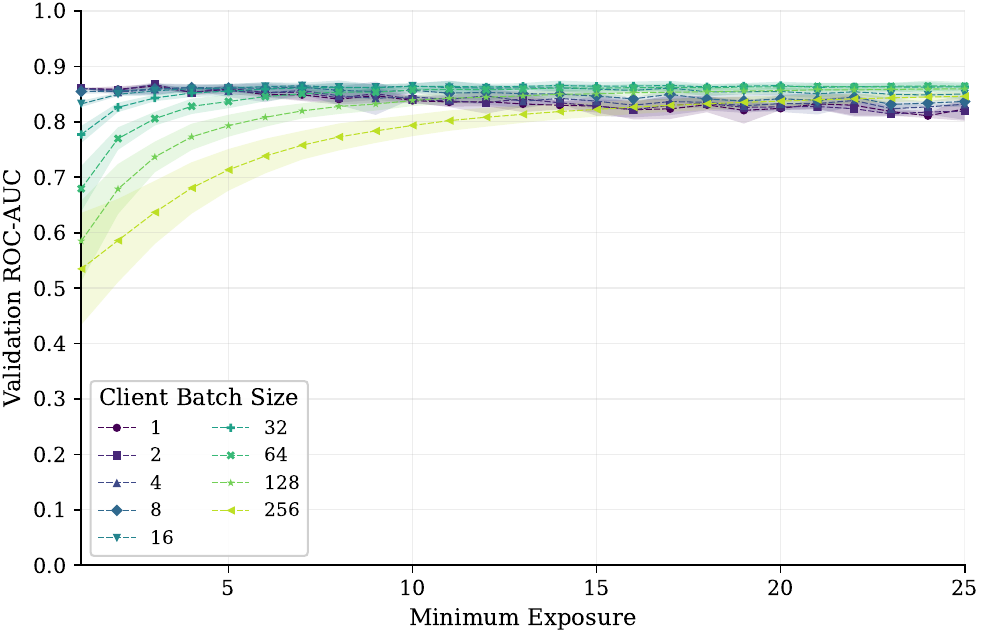}
          \caption{Small MLP}
      \end{subfigure}

      \vspace{0.5em}

      \begin{subfigure}[t]{0.32\textwidth}
          \centering
          \includegraphics[width=\textwidth]{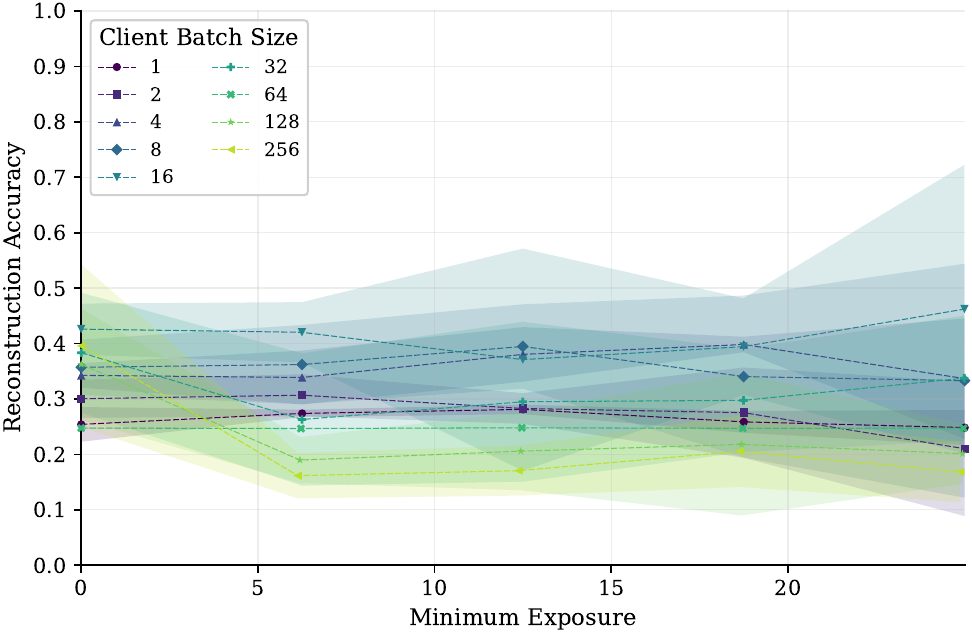}
          \caption{FT-Transformer}
      \end{subfigure}
      \hfill
      \begin{subfigure}[t]{0.32\textwidth}
          \centering
          \includegraphics[width=\textwidth]{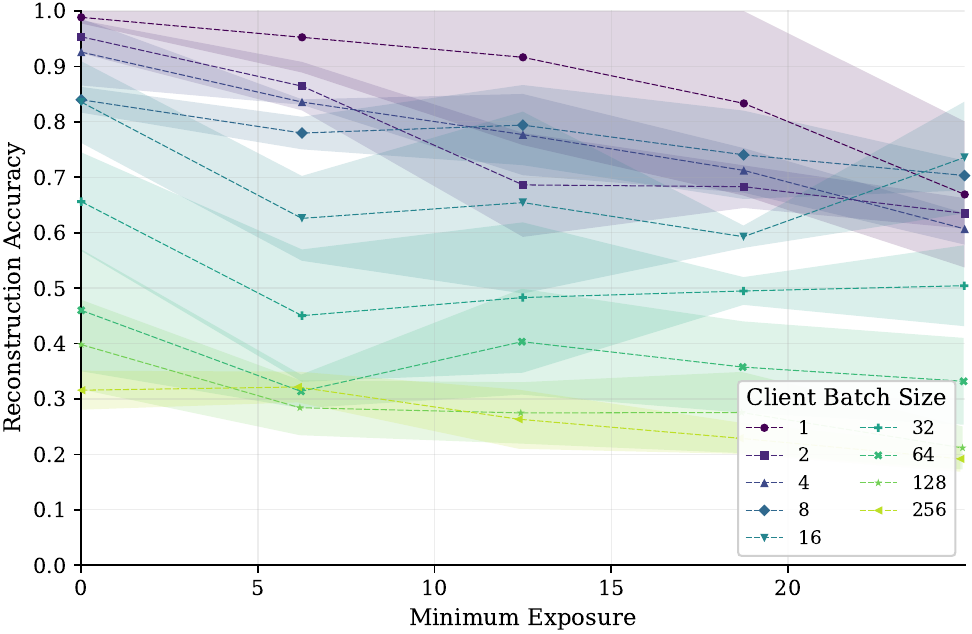}
          \caption{ResNet}
      \end{subfigure}
      \hfill
      \begin{subfigure}[t]{0.32\textwidth}
          \centering
          \includegraphics[width=\textwidth]{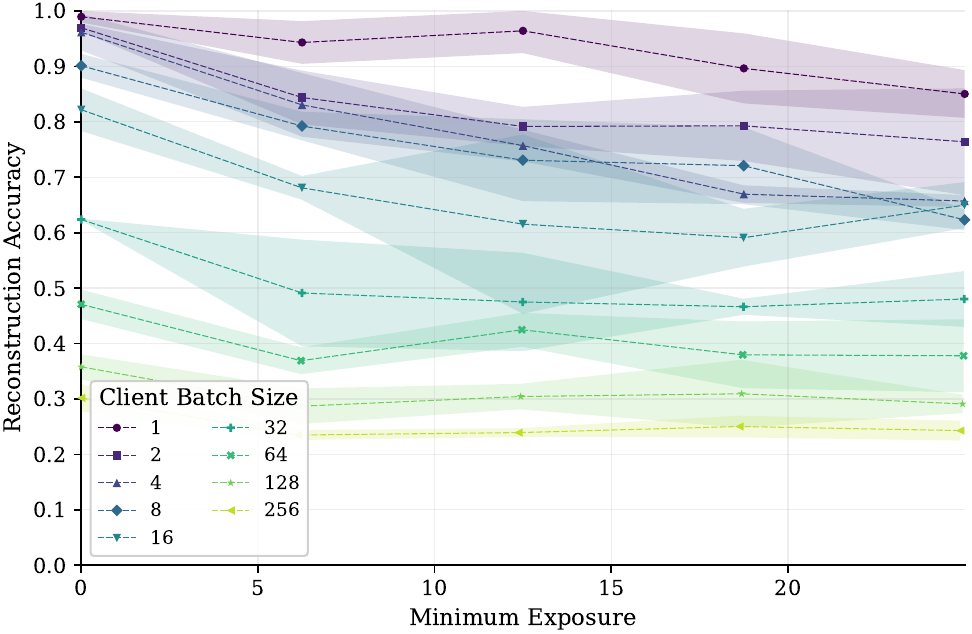}
          \caption{Small MLP}
      \end{subfigure}

      \caption{The top row shows validation utility over exposure, and the bottom row shows reconstruction accuracy over exposure for MIMIC-IV.}
      \label{fig:mimic-cross-model-batchsize-privacy-utility}
  \end{figure}

Figure~\ref{fig:mimic-cross-model-batchsize-privacy-utility} shows how utility and leakage evolve together over training exposure. For ResNet and the small MLP, larger client batches start from lower utility at early exposure, but improve steadily and ultimately match or exceed the smaller batch settings. Their reconstruction accuracy is also lower throughout much of training and typically decreases further as exposure increases. FT-Transformer again shows a less regular dynamic, with intermediate batch sizes remaining among the more vulnerable settings over much of training and the largest batches becoming clearly more private only at later exposure.

In summary, the exposure trajectories show that larger client batches often provide sustained privacy gains for the one-hot baselines without sacrificing final utility, whereas the same relationship is weaker and less stable for FT-Transformer.

The FedAvg experiment tests whether the MIMIC-IV leakage patterns persist when client updates are produced by local training trajectories rather than single mini batch gradients. In this setting, each client is restricted to a fixed local dataset, so the same records contribute to the attacked update at each attack point. The comparison therefore controls for changes in the attacked client record set across training. Here, \(n\) batches denotes how many mini batches are needed to process the fixed local dataset once, so larger values correspond to smaller local mini batches and more local optimizer steps per epoch, not to more distinct client examples.

In this context, Table~\ref{tab:fedavg-mimic-tableak-acc} shows that FedAvg model deltas remain reconstructable. The initialized FedAvg update is already highly exposed for the one-hot baselines, with ResNet ranging from \(0.683\) to \(0.894\) and the small MLP ranging from \(0.738\) to \(0.812\). FT-Transformer remains substantially lower, between \(0.262\) and \(0.335\), preserving the same architecture ordering observed under FedSGD. Additional local epochs do not act as a uniform privacy mechanism. At the initialized attack point, increasing the number of local epochs raises reconstruction for all three model families at each mini batch granularity, which indicates that repeated local optimization over the same small client record set can strengthen the reconstructive signal rather than hide it. Furthermore, the benchmark FedAvg results in Appendix~\ref{app:benchmark-fedavg} show the same broader pattern across Adult, the private multiclass benchmark, and California Housing. Reconstruction remains nontrivial when the server observes model deltas rather than one-step gradients, but FedAvg does not have a uniform privacy effect. Changing the number of local batches or local epochs can either reduce or increase reconstruction depending on the model and dataset. This supports the MIMIC-IV result that FedAvg leakage depends on how local computation is structured, not only on whether the observed update is a gradient or a model delta. On the \textit{Adult} dataset with 32 local examples per client, the small MLP follows the direction reported by TabLeak, where more local epochs reduce reconstruction accuracy under FedAvg \cite{vero2023tableak}. ResNet does not follow that pattern, since moving from 1 to 4 local epochs increases reconstruction across all three mini batch granularities. The MIMIC-IV results strengthen this point. With 16 local examples per client, additional local epochs usually preserve or increase leakage, especially for FT-Transformer and ResNet. Thus, the privacy effect of FedAvg local epochs is not universal. It depends on the architecture, dataset, local client dataset size, and attack formulation.

The same FedAvg comparison with 64 local examples per client is reported in Appendix~\ref{app:fedavg-local-computation}. Increasing the fixed client dataset size lowers reconstruction substantially for the one-hot baselines, but the architecture ordering remains and the effect of local epochs is still not uniformly protective. This suggests that the 16-example result is partly amplified by the small client dataset, while the fact that local epochs are not uniformly protective persists beyond that most exposed setting.

\begin{table}[H]
    \centering
    \small
    \caption{Reconstruction accuracy at the initialized attack point and final attack point for MIMIC-IV under FedAvg with 16 local examples per client. Initialized attack point denotes the first client update computed from the initial global model, and final attack point denotes the last attacked point under the exposure budget. The row variable \(n.\) batches denotes how many local mini batches are needed to process the local client data once. Each cell reports mean $\pm$ standard deviation across 5 seeds.}
    \label{tab:fedavg-mimic-tableak-acc}
    \resizebox{\textwidth}{!}{%
    \begin{tabular}{lcccccc}
    \hline
    \multicolumn{1}{l}{} & \multicolumn{3}{c}{\textbf{Initialized attack point}} & \multicolumn{3}{c}{\textbf{Final attack point}} \\
    \textbf{n. batches} & \textbf{FT-Transformer} & \textbf{ResNet} & \textbf{Small MLP} & \textbf{FT-Transformer} & \textbf{ResNet} & \textbf{Small MLP} \\
    \hline
    \multicolumn{7}{c}{1 local epoch} \\
    \hline
    1 & 0.304 $\pm$ 0.058 & 0.826 $\pm$ 0.060 & 0.744 $\pm$ 0.083 & 0.318 $\pm$ 0.069 & 0.845 $\pm$ 0.059 & 0.715 $\pm$ 0.051 \\
    2 & 0.279 $\pm$ 0.047 & 0.830 $\pm$ 0.078 & 0.738 $\pm$ 0.103 & 0.294 $\pm$ 0.067 & 0.769 $\pm$ 0.096 & 0.698 $\pm$ 0.070 \\
    4 & 0.262 $\pm$ 0.038 & 0.683 $\pm$ 0.119 & 0.756 $\pm$ 0.086 & 0.328 $\pm$ 0.079 & 0.689 $\pm$ 0.046 & 0.660 $\pm$ 0.076 \\
    \hline
    \multicolumn{7}{c}{2 local epochs} \\
    \hline
    1 & 0.309 $\pm$ 0.055 & 0.874 $\pm$ 0.079 & 0.763 $\pm$ 0.081 & 0.349 $\pm$ 0.075 & 0.872 $\pm$ 0.068 & 0.700 $\pm$ 0.087 \\
    2 & 0.301 $\pm$ 0.063 & 0.882 $\pm$ 0.055 & 0.780 $\pm$ 0.076 & 0.356 $\pm$ 0.062 & 0.820 $\pm$ 0.084 & 0.694 $\pm$ 0.056 \\
    4 & 0.291 $\pm$ 0.039 & 0.731 $\pm$ 0.119 & 0.768 $\pm$ 0.073 & 0.354 $\pm$ 0.122 & 0.709 $\pm$ 0.093 & 0.647 $\pm$ 0.076 \\
    \hline
    \multicolumn{7}{c}{4 local epochs} \\
    \hline
    1 & 0.335 $\pm$ 0.072 & 0.894 $\pm$ 0.064 & 0.773 $\pm$ 0.098 & 0.366 $\pm$ 0.084 & 0.866 $\pm$ 0.051 & 0.756 $\pm$ 0.061 \\
    2 & 0.318 $\pm$ 0.070 & 0.893 $\pm$ 0.052 & 0.812 $\pm$ 0.060 & 0.363 $\pm$ 0.097 & 0.822 $\pm$ 0.085 & 0.730 $\pm$ 0.081 \\
    4 & 0.302 $\pm$ 0.053 & 0.764 $\pm$ 0.120 & 0.796 $\pm$ 0.067 & 0.373 $\pm$ 0.105 & 0.746 $\pm$ 0.100 & 0.697 $\pm$ 0.060 \\
    \hline
    \end{tabular}
    }
\end{table}

Unlike the one-hot baselines, FT-Transformer does not expose categorical variables as direct input coordinates. Categorical features are ordinal encoded and passed through learned embeddings, so the attacker must optimize through a differentiable categorical surrogate. We therefore compare two categorical attack parameterizations on MIMIC-IV: categorical logits and probability simplex. The categorical logits parameterization is used as the default FT-Transformer attacker in the main MIMIC-IV batch size experiment. These controls are reported up to batch size \(128\), while the main MIMIC-IV batch size experiment also includes batch size \(256\). Table~\ref{tab:batch-size-mimic-tableak-acc-attack-paths} shows that the two categorical parameterizations lead to similar reconstruction levels and the same qualitative interpretation. The categorical logits parameterization is slightly stronger at several final attack points, including batch sizes \(1\), \(4\), \(8\), \(16\), and \(32\), while probability simplex is competitive or stronger in some settings, such as the final attack point at batch size \(2\) and the initialized attack point at batch size \(64\). The fixed batch control in Appendix Table~\ref{tab:fixed-batch-mimic-fttransformer-attack-paths} supports the same conclusion. Thus, the lower FT-Transformer leakage observed above is not explained by a single categorical relaxation choice.

\begin{table}[H]
  \centering
  \small
  \caption{Reconstruction accuracy at the initialized attack point and final attack point for MIMIC-IV under two FT-Transformer attack parameterizations. Initialized attack point denotes the attack before any global model aggregation, and final attack point denotes the last attacked point under the exposure budget. Each cell reports mean $\pm$ standard deviation across 3 seeds.}
  \label{tab:batch-size-mimic-tableak-acc-attack-paths}
  \begin{tabular}{lcccc}
  \hline
  \multicolumn{1}{l}{} & \multicolumn{2}{c}{\textbf{Initialized attack point}} & \multicolumn{2}{c}{\textbf{Final attack point}} \\
  \textbf{Batch size} & \textbf{Categorical logits} & \textbf{Probability simplex} & \textbf{Categorical logits} & \textbf{Probability simplex} \\
  \hline
  1 & 0.254 $\pm$ 0.028 & 0.251 $\pm$ 0.011 & 0.248 $\pm$ 0.028 & 0.167 $\pm$ 0.086 \\
  2 & 0.300 $\pm$ 0.031 & 0.286 $\pm$ 0.016 & 0.210 $\pm$ 0.107 & 0.290 $\pm$ 0.088 \\
  4 & 0.343 $\pm$ 0.020 & 0.333 $\pm$ 0.026 & 0.337 $\pm$ 0.096 & 0.306 $\pm$ 0.116 \\
  8 & 0.357 $\pm$ 0.044 & 0.355 $\pm$ 0.048 & 0.333 $\pm$ 0.186 & 0.330 $\pm$ 0.111 \\
  16 & 0.426 $\pm$ 0.041 & 0.407 $\pm$ 0.037 & 0.462 $\pm$ 0.230 & 0.413 $\pm$ 0.174 \\
  32 & 0.384 $\pm$ 0.096 & 0.375 $\pm$ 0.100 & 0.338 $\pm$ 0.100 & 0.328 $\pm$ 0.105 \\
  64 & 0.248 $\pm$ 0.001 & 0.368 $\pm$ 0.081 & 0.247 $\pm$ 0.001 & 0.236 $\pm$ 0.068 \\
  128 & 0.363 $\pm$ 0.089 & 0.359 $\pm$ 0.087 & 0.201 $\pm$ 0.048 & 0.202 $\pm$ 0.046 \\
  \hline
  \end{tabular}
\end{table}

\begin{figure}[H]
        \centering

        \begin{subfigure}[t]{0.48\textwidth}
            \centering
            \includegraphics[width=\textwidth]{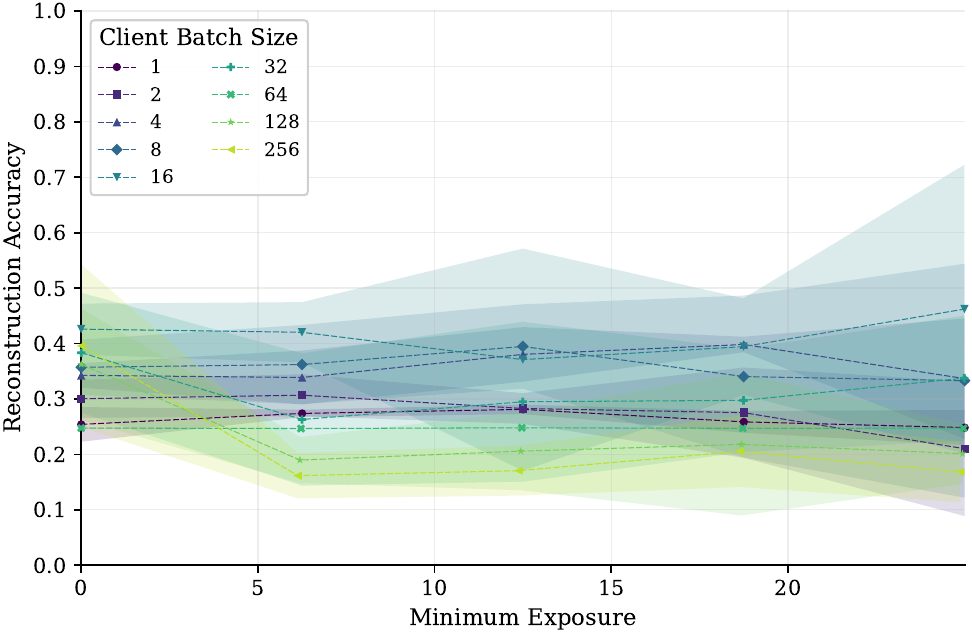}
            \caption{Categorical logits, dropout on}
        \end{subfigure}
        \hfill
        \begin{subfigure}[t]{0.48\textwidth}
            \centering
            \includegraphics[width=\textwidth]{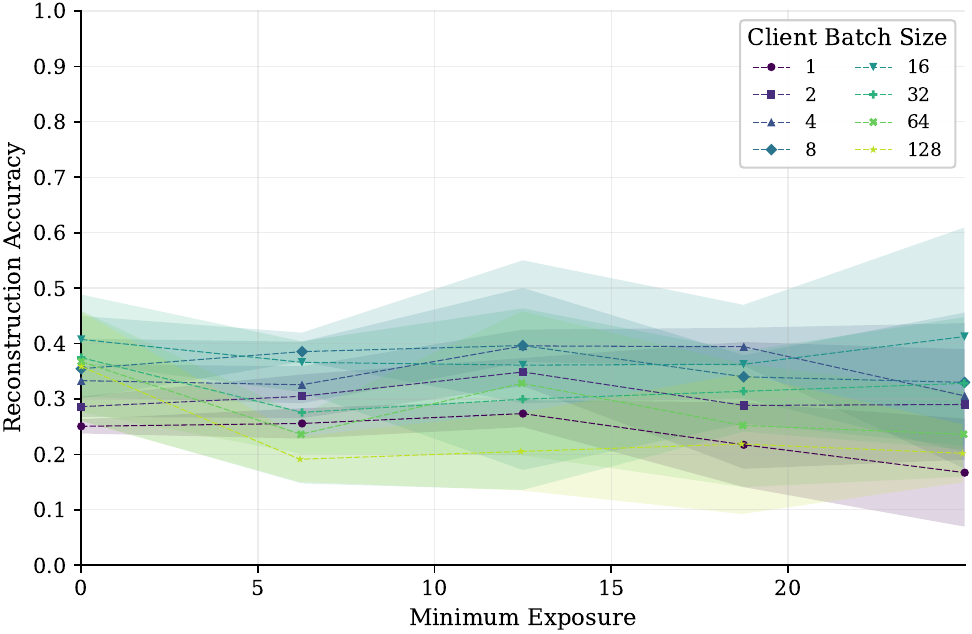}
            \caption{Probability simplex, dropout on}
        \end{subfigure}

        \vspace{0.5em}

        \begin{subfigure}[t]{0.48\textwidth}
            \centering
            \includegraphics[width=\textwidth]{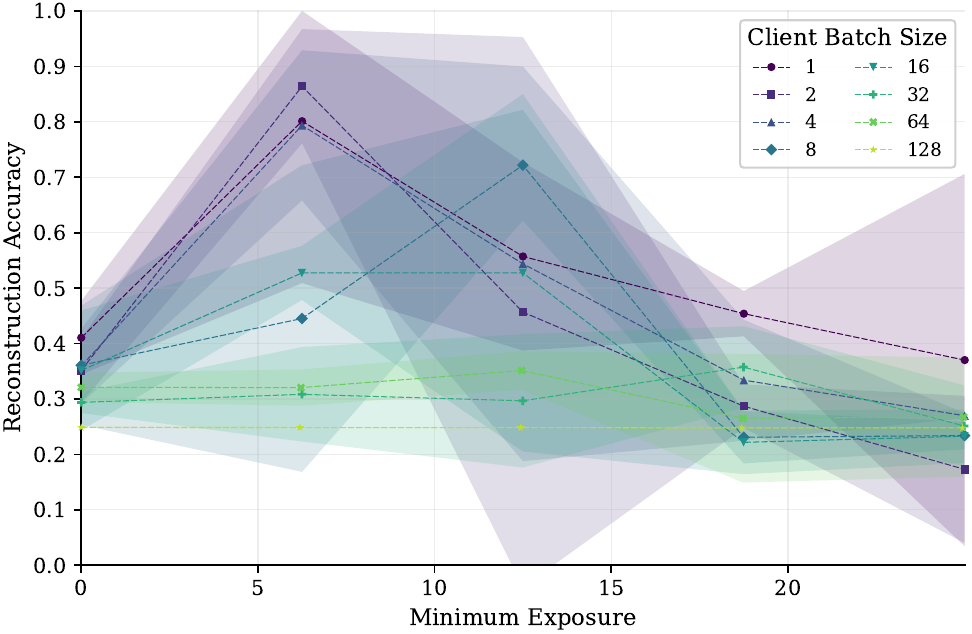}
            \caption{Categorical logits, dropout off}
        \end{subfigure}
        \hfill
        \begin{subfigure}[t]{0.48\textwidth}
            \centering
            \includegraphics[width=\textwidth]{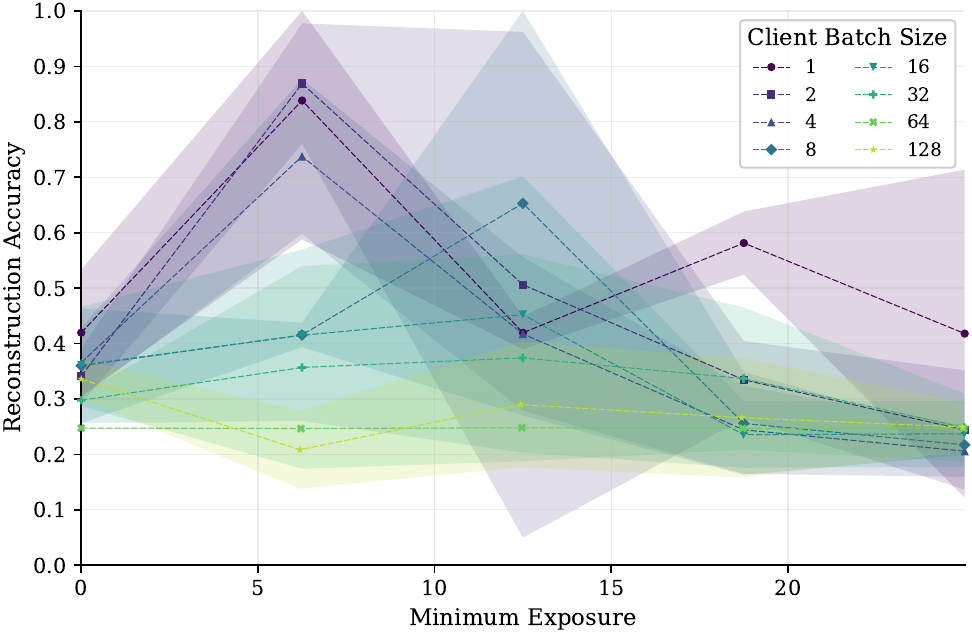}
            \caption{Probability simplex, dropout off}
        \end{subfigure}

        \caption{FT-Transformer reconstruction accuracy over client exposure on MIMIC-IV under the two categorical attack parameterizations, with and without dropout in the target model.}
        \label{fig:fttransformer-dropout-attack-exposure}
  \end{figure}

Dropout is a natural alternative explanation for the lower FT-Transformer leakage. Appendix Table~\ref{tab:batch-size-mimic-fttransformer-attack-paths-dropout0} reports the initialized and final attack point results when attention and feedforward dropout are disabled. Removing dropout increases reconstruction most clearly at batch size \(1\), but the effect is not systematic across batch sizes or attack points. Figure~\ref{fig:fttransformer-dropout-attack-exposure} therefore reports the full exposure trajectories to show whether dropout changes when leakage occurs during training. The exposure trajectories show that the dropout effect is strongly temporal. Without dropout, small batches produce a pronounced early training leakage spike. At batch size \(1\), reconstruction increases from \(0.411\) at the initialized attack point to \(0.801\) at the first post initialization attack point under categorical logits, and from \(0.420\) to \(0.838\) under probability simplex. With dropout enabled, the corresponding early spike is much weaker. By the final attack point, removing dropout does not produce uniformly higher reconstruction. Dropout therefore affects the timing and stability of FT-Transformer leakage, rather than providing a sufficient explanation for the architecture level gap between FT-Transformer and the one-hot baselines. The corresponding EMRs in Appendix Table~\ref{tab:batch-size-mimic-fttransformer-attack-paths-dropout0-emr} show that this aggregate leakage rarely translates into complete row recovery.

Leakage in tabular gradient inversion is not determined by the training protocol alone, but also by the architectural choices used in the target model. Because tabular model performance is known to be dataset dependent, architecture level leakage patterns should also be interpreted relative to the data setting \cite{gorishniy2021revisiting}. To isolate these effects within the MLP family, we evaluate a grid of standard NN modules and hyperparameters, including hidden width, hidden depth, normalization, activation, and dropout, while keeping the remaining training setup fixed. We first summarize the factor averaged effect of these choices on final attack point reconstruction accuracy, and then examine reconstruction trajectories over exposure to show how these effects interact across full configurations.

Tables~\ref{tab:torch-modules-mimic-structural} and~\ref{tab:torch-modules-mimic-modules} show clear factor averaged trends for MIMIC-IV. Structural choices still affect leakage through model capacity. Increasing hidden width increases reconstruction accuracy at both batch sizes, from 0.473 to 0.526 at batch size 8 and from 0.334 to 0.409 at batch size 32 when moving from width 32 to 128. Hidden depth is more dataset dependent. At batch size 8, deeper models are less vulnerable on average, with reconstruction accuracy decreasing from 0.540 for one hidden layer to 0.470 for three hidden layers. At batch size 32, the depth effect is weak. Module level choices also remain important. GELU is more vulnerable than ReLU at both batch sizes, and dropout reduces leakage on average. At batch size 8, switching from GELU to ReLU reduces the average reconstruction accuracy from 0.588 to 0.413, while adding dropout reduces it from 0.514 to 0.487. The same pattern holds at batch size 32, where ReLU reduces reconstruction accuracy from 0.422 to 0.321 and dropout reduces it from 0.398 to 0.345. LayerNorm is slightly less vulnerable than BatchNorm at both batch sizes, although the difference is smaller than the activation effect. These results indicate that architectural choices alone can shift leakage by a meaningful margin even when the training protocol is unchanged. A second consistent pattern is that the absolute leakage level is lower at the larger batch size. Larger client batches still reduce leakage overall, but within a fixed batch size setting, wider models and GELU activations remain systematically more vulnerable. 

\begin{table}[H]
    \centering
    \small
    \caption{Reconstruction accuracy at the final attack point for structural architectural factors on MIMIC-IV. Each cell reports mean $\pm$ standard deviation across the remaining configurations at the final attack point.}
    \label{tab:torch-modules-mimic-structural}
    \resizebox{\textwidth}{!}{%
    \begin{tabular}{lcccccc}
    \hline
    \multicolumn{1}{l}{} & \multicolumn{3}{c}{\textbf{Hidden Width}} & \multicolumn{3}{c}{\textbf{Hidden Layers}} \\
    \textbf{Batch size} & \textbf{32} & \textbf{64} & \textbf{128} & \textbf{1} & \textbf{2} & \textbf{3} \\
    \hline
    8 & 0.473 $\pm$ 0.161 & 0.503 $\pm$ 0.147 & 0.526 $\pm$ 0.137 & 0.540 $\pm$ 0.139 & 0.492 $\pm$ 0.147 & 0.470 $\pm$ 0.154 \\
    32 & 0.334 $\pm$ 0.109 & 0.372 $\pm$ 0.109 & 0.409 $\pm$ 0.099 & 0.369 $\pm$ 0.095 & 0.376 $\pm$ 0.113 & 0.369 $\pm$ 0.123 \\
    \hline
    \end{tabular}
    }
\end{table}

\begin{table}[H]
    \centering
    \small
    \caption{Reconstruction accuracy at the final attack point for module level architectural factors on MIMIC-IV. Each cell reports mean $\pm$ standard deviation across the remaining configurations at the final attack point.}
    \label{tab:torch-modules-mimic-modules}
    \resizebox{\textwidth}{!}{%
    \begin{tabular}{lcccccc}
    \hline
    \multicolumn{1}{l}{} & \multicolumn{2}{c}{\textbf{Normalization}} & \multicolumn{2}{c}{\textbf{Activation}} & \multicolumn{2}{c}{\textbf{Dropout}} \\
    \textbf{Batch size} & \textbf{BatchNorm} & \textbf{LayerNorm} & \textbf{ReLU} & \textbf{GELU} & \textbf{0.0} & \textbf{0.1} \\
    \hline
    8 & 0.554 $\pm$ 0.079 & 0.447 $\pm$ 0.180 & 0.413 $\pm$ 0.149 & 0.588 $\pm$ 0.081 & 0.514 $\pm$ 0.150 & 0.487 $\pm$ 0.146 \\
    32 & 0.380 $\pm$ 0.075 & 0.363 $\pm$ 0.136 & 0.321 $\pm$ 0.105 & 0.422 $\pm$ 0.088 & 0.398 $\pm$ 0.105 & 0.345 $\pm$ 0.108 \\
    \hline
    \end{tabular}
    }
\end{table}

\begin{figure}[H]
    \centering
    \includegraphics[width=0.9\textwidth]{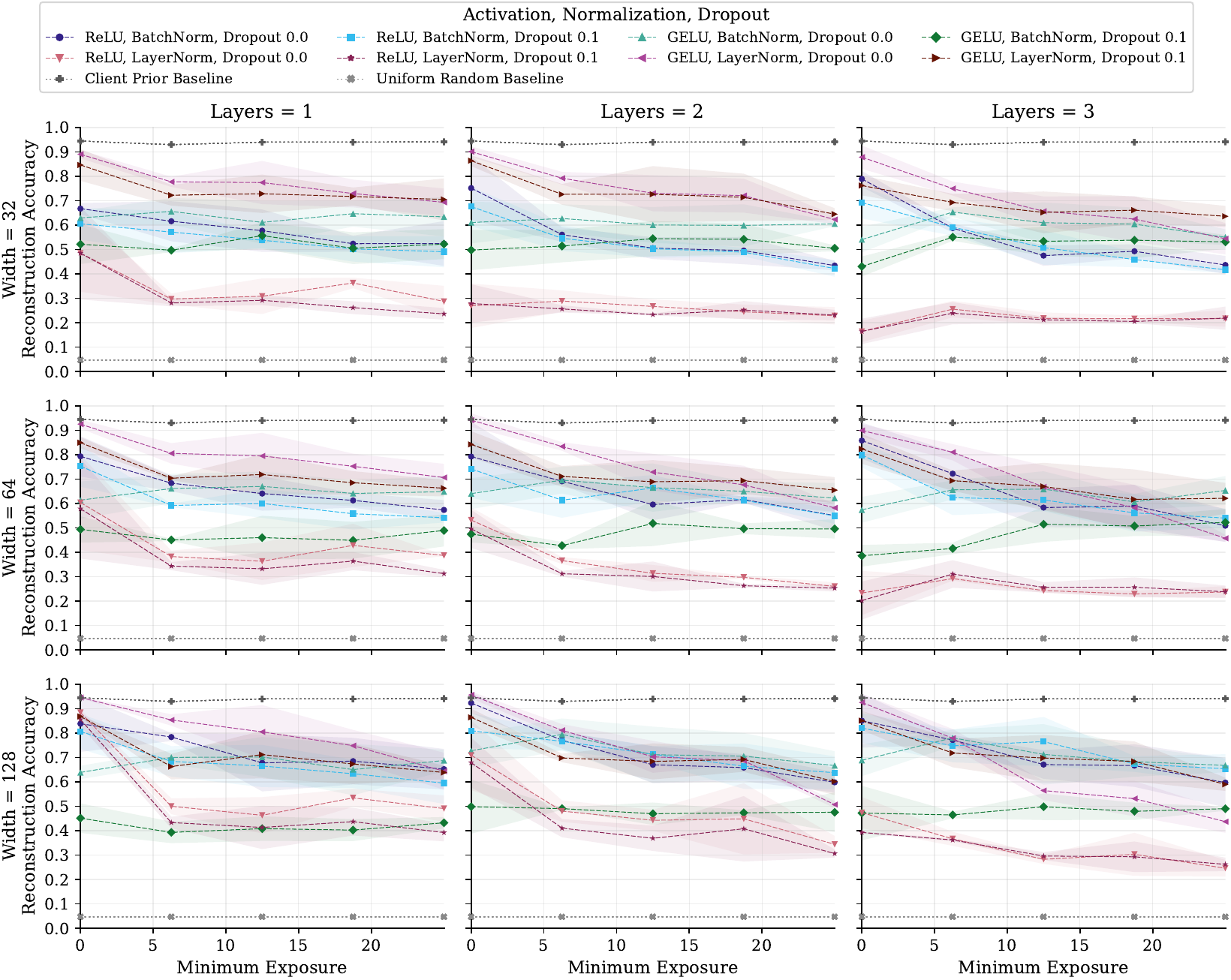}
    \caption{Reconstruction accuracy from FedSGD GIAs during training for the structural and module configurations with client batch size 8 on MIMIC-IV.}
    \label{fig:fedsgd_torch_modules_config_attack_grid__dataset_mimic_admission_tier3_binary_train__batch_8}
\end{figure}

The factor tables hide important interaction effects that become visible in the full configuration trajectories. As such, Figure~\ref{fig:fedsgd_torch_modules_config_attack_grid__dataset_mimic_admission_tier3_binary_train__batch_8} shows that the marginal patterns depend strongly on the surrounding structural configuration. Consistent with the factor averages, the lowest trajectory averages are concentrated in LayerNorm and ReLU settings. The lowest full configuration over the batch size \(8\) trajectory is width \(32\), three hidden layers, LayerNorm, ReLU, and dropout \(0.1\), with mean reconstruction accuracy \(0.208\) across attack points. However, the interaction view also reveals exceptions that the marginal tables hide. For width \(128\) with one hidden layer, BatchNorm with GELU and dropout \(0.1\) has the lowest trajectory average within that structural setting, despite GELU being more vulnerable on average. Across the full batch size \(8\) grid, trajectory averaged reconstruction ranges from \(0.208\) to \(0.799\), a difference of \(59.1\) percentage points. This shows that ordinary MLP design choices can substantially change gradient inversion leakage.

Table~\ref{tab:torch-modules-mimic-module-extremes-b32} reports the lowest and highest reconstruction module combinations at batch size \(32\), averaged over width and depth. The complete module ranking for both evaluated batch sizes is reported in Appendix Table~\ref{tab:torch-modules-mimic-module-ranking}. These rankings are consistent with the trajectory analysis. At both batch sizes, the lowest average reconstruction accuracy is achieved by LayerNorm with ReLU and dropout \(0.1\), followed by LayerNorm with ReLU and no dropout. The highest reconstruction settings are GELU based, with LayerNorm plus GELU producing the most vulnerable combinations. The final attack point rankings therefore support the same conclusion as the full trajectories. Activation is the clearest module level leakage driver, while width, depth, normalization, and dropout determine the final leakage level through their interaction in the complete architecture. In addition, the benchmark architecture grid results in Appendix~\ref{app:benchmark-mlp-architecture} show a related pattern across Adult, the private multiclass benchmark, and California Housing. In all three benchmark datasets, LayerNorm with GELU activation and no dropout is the highest reconstruction module combination at both evaluated batch sizes. MIMIC-IV is slightly less uniform, but its highest reconstruction settings are also GELU based. The utility rankings in Appendix Tables~\ref{tab:torch-modules-mimic-top-utility-configs-b8} and~\ref{tab:torch-modules-mimic-top-utility-configs-b32} show no clear utility advantage for the leakier MIMIC-IV configurations. At batch size \(32\), the best validation ROC-AUC is achieved by LayerNorm with ReLU and dropout \(0.1\), which is also the lowest reconstruction module combination. Thus, in this grid, lower leakage does not require a clear utility sacrifice.

\begin{table}[H]
    \centering
    \small
    \caption{Lowest and highest reconstruction module combinations on MIMIC-IV for batch size \(32\). Reconstruction accuracy is averaged over width and depth and reported at the final attack point.}
    \label{tab:torch-modules-mimic-module-extremes-b32}
    \begin{tabular}{ccccc}
    \hline
    \textbf{Rank} & \textbf{Normalization} & \textbf{Activation} & \textbf{Dropout} & \textbf{Recon. Acc.} \\
    \hline
    \multicolumn{5}{c}{Lowest reconstruction accuracy} \\
    \hline
    1 & LayerNorm & ReLU & 0.1 & 0.215 $\pm$ 0.048 \\
    2 & LayerNorm & ReLU & 0.0 & 0.259 $\pm$ 0.064 \\
    3 & BatchNorm & GELU & 0.1 & 0.294 $\pm$ 0.016 \\
    \hline
    \multicolumn{5}{c}{Highest reconstruction accuracy} \\
    \hline
    1 & LayerNorm & GELU & 0.0 & 0.500 $\pm$ 0.024 \\
    2 & LayerNorm & GELU & 0.1 & 0.478 $\pm$ 0.028 \\
    3 & BatchNorm & GELU & 0.0 & 0.417 $\pm$ 0.059 \\
    \hline
    \end{tabular}
\end{table}

Additional sensitivity analyses for client heterogeneity, attack budget, and label unknown attacks are reported in Appendix~\ref{app:mimic-sensitivity}.

\section{Discussion}
\label{sec:discussion}
In this study, we profile privacy preservation in tabular FL under GIAs across a wide range of configurations. The results show that tabular FL can be practically privacy preserving against complete record reconstruction when the architecture and FL configuration are chosen carefully. At the same time, GIAs remain feasible across binary classification, multiclass classification, and regression. Leakage is highest when the observed update is weakly aggregated, especially for very small FedSGD client batches, direct one-hot input representations, and strong attacker assumptions such as known labels. In general, reconstruction decreases as client batch size increases and as the model representation becomes less directly tied to individual input features. The clearest example is FT-Transformer on MIMIC-IV. EMR is often zero, even when dropout is disabled. In the dropout disabled ablation, nonzero recovery appears only in the most exposed setting, categorical logits with batch size \(1\). Nonzero aggregate feature recovery therefore does not necessarily imply complete record recovery.

Larger client batches often reduce reconstruction, but the effect depends on the model representation and does not by itself make the observed update private. As shown in Table~\ref{tab:robustness-batch-size-trained-onehot}, Table~\ref{tab:robustness-batch-size-fttransformer}, and Table~\ref{tab:batch-size-mimic-tableak-acc}, the strongest leakage appears when the observed update is weakly aggregated, especially for small FedSGD client batches. At the same time, the same tables show that increasing the batch size reduces reconstruction but does not eliminate it, and Table~\ref{tab:fedavg-mimic-tableak-acc} further shows that FedAvg model deltas remain reconstructable even when the server observes a local training trajectory rather than a one-step gradient. FedSGD is therefore used as the main controlled leakage setting because each observed update has a direct interpretation as the gradient of one client batch. FedAvg is closer to many deployment settings, but it is less controlled as an inversion target because each model delta reflects a local training trajectory, optimizer dynamics, and potentially many costly unrolled steps. This pattern agrees with prior evidence that collaborative updates may expose unintended client information and that aggregation can still preserve recoverable input structure \cite{huang2021evaluating,dimitrov2022data,vero2023tableak}. The practical meaning is therefore that larger batches and local training should be viewed as leakage reduction mechanisms, not as privacy guarantees. This is also visible in Figure~\ref{fig:mimic-batch-cross-model-tradeoff} and Figure~\ref{fig:mimic-cross-model-batchsize-privacy-utility}, where larger batches often reduce reconstruction while maintaining or improving utility for the one-hot baselines. The sensitivity analyses further show that client heterogeneity, attack budget, and label knowledge modulate measured leakage, but do not overturn the broader pattern that aggregation and model representation are the most stable determinants.

The results also show that feature level reconstruction and complete record recovery should be interpreted separately. Table~\ref{tab:batch-size-mimic-tableak-acc} shows substantial feature level reconstruction on MIMIC-IV, whereas Appendix Table~\ref{tab:batch-size-mimic-iv-strict-emr} shows that EMR can be much lower, especially for FT-Transformer. This gap is important because sparse clinical tabular data may leak sensitive client level structure even when a full row is not perfectly recovered. The baseline comparisons point in the same direction: privacy risk is not limited to reconstructing a complete row, but also includes recovering sensitive feature structure or client specific distributions. Thus, the meaning of the MIMIC-IV results is not simply that some records are or are not recovered; rather, the combination of Table~\ref{tab:batch-size-mimic-tableak-acc}, Appendix Table~\ref{tab:batch-size-mimic-iv-strict-emr}, and the baseline comparisons in the appendix shows that tabular leakage must be evaluated at multiple levels: feature recovery, exact row recovery, and baseline relative gain.

Finally, the architecture level results show that privacy leakage is shaped by the representation through which gradients are produced. Table~\ref{tab:robustness-batch-size-fttransformer} and Table~\ref{tab:batch-size-mimic-tableak-acc} show that FT-Transformer is generally harder to invert than the one-hot MLP and ResNet baselines, while Table~\ref{tab:batch-size-mimic-tableak-acc-attack-paths} and Figure~\ref{fig:fttransformer-dropout-attack-exposure} show that this effect is not explained solely by one categorical attack parameterization or by dropout. A plausible mechanism is that FT-Transformer leakage depends on the state of its learned categorical embedding representation. The exposure trajectories in Figure~\ref{fig:fttransformer-dropout-attack-exposure} are consistent with this interpretation. At initialization, the categorical embeddings and downstream transformer layers are not yet aligned with the task. After training, the learned representation may make the update more informative about category choices and feature interactions in the attacked batch. This effect competes with the usual reduction in gradient signal as training progresses, which helps explain why training stage effects are architecture and dataset dependent. This supports the view that embeddings and transformer style feature processing change the relationship between input features and observed gradients. They should not be interpreted as a complete defense, since transformer and attention based models have also been shown to leak through gradients in other modalities \cite{hatamizadeh2022gradvit,balunovic2022lamp}. The MLP grid leads to the same broader conclusion: Table~\ref{tab:torch-modules-mimic-structural}, Table~\ref{tab:torch-modules-mimic-modules}, Figure~\ref{fig:fedsgd_torch_modules_config_attack_grid__dataset_mimic_admission_tier3_binary_train__batch_8}, and Table~\ref{tab:torch-modules-mimic-module-extremes-b32} show that width, activation, normalization, and dropout can shift leakage substantially without necessarily producing a utility advantage. Hence, the main implication is that privacy-aware tabular FL design should evaluate the full model update pipeline before deployment rather than relying on architectural complexity, FedAvg, or predictive utility as proxies for privacy.

This study has several limitations. First, the threat model is intentionally narrow. We study an honest-but-curious server that follows the FL protocol and reconstructs from individual client updates. We do not evaluate servers that manipulate training, malicious or colluding clients, secure aggregation, or differential privacy. These settings change either how client updates are generated or what update is available to the attacker. The results should therefore be read as empirical leakage under the stated threat model. Second, the attack family is fixed. Our reconstruction pipeline is based on the open source LeakPro implementation of the optimization based GIA. Other attack families, including analytical inversion, generative priors, or attacks designed specifically for embedded categorical representations, may produce different leakage patterns. Third, the attacker's assumption used in the main experiments is strong. The label known setting provides an upper bound estimate of feature leakage, and our label unknown sensitivity analysis weakens this assumption only partially, since the attacker still knows the model architecture, training pipeline, feature types, column positions, and feature semantics. The empirical scope is also bounded by the chosen datasets, feature sets, and architectures; in particular, the architecture comparison covers MLP, ResNet, and FT-Transformer, but not other tabular architectures such as TabNet or TabTransformer. Fourth, the federated configuration is controlled to isolate leakage mechanisms. FedAvg local computation is restricted to plain SGD style updates without momentum, server side adaptive optimization, or update post processing. These choices make the experiments comparable across protocols, but limit how directly the conclusions transfer to larger production FL systems with more clients, heterogeneous compute, and additional optimizer machinery. Finally, the reconstruction metrics measure recoverability after row matching. They are well suited for comparing attacks and architectures, but they do not by themselves determine how severe a partial reconstruction is in privacy terms. Thus, as a future work, the architecture and aggregation findings should be tested together with formal, system level, and update level defenses. In addition, the threat model and attack family can also be expanded, including cases where adversaries do change the training process, including servers that manipulate the global model, modify the loss, or select clients to amplify reconstruction. Furthermore, other tabular NN architectures, including TabNet \cite{arik2021tabnet} and TabTransformer \cite{huang2020tabtransformer} encode feature interactions in different ways. Each of these models defines a different differentiable input gradient relationship, and the resulting reconstructability could differ significantly. A systematic comparison that includes these architectures, together with controlled embedding dimension and attention depth ablations, would help clarify which architectural primitives most strongly modulate gradient inversion risk in tabular settings.

Taken jointly, our results indicate that tabular FL privacy depends on the interaction between aggregation, architecture, data exposure, and dataset structure. Larger batches, embedded categorical representations, dropout, and favorable MLP module choices can reduce measured reconstructability under the studied threat model, but they do not replace formal defenses such as secure aggregation or differential privacy when formal privacy guarantees are required. Scholars and practitioners should take these findings into account when developing tabular FL models and deploying them in production environments.

\section*{Declarations}

\subsection*{Conflicts of interest/competing interests}
The authors declare that they have no competing interests.

\subsection*{Code and data availability}
The source code used for this study is publicly available at \url{https://github.com/scaleoutsystems/tabular-GIA/}. The Adult and California Housing datasets are publicly available from their sources. The MIMIC-IV v3.1 cohort is governed by the PhysioNet credentialed access process; researchers who have completed the required training and data use agreement can obtain MIMIC-IV from PhysioNet. The private multiclass benchmark used as an intermediate evaluation dataset cannot be shared because it contains sensitive data.

\subsection*{Acknowledgments}
The authors thank Fazeleh Hoseini and Mattias {\AA}kesson from AI Sweden and Scaleout Systems for helping facilitate the project and for providing helpful feedback during the work. 

\subsection*{Funding}
This work was supported by Vinnova — Sweden's Innovation Agency — through the Advanced Digitalization Research and Innovation Programme (grant no. 2025-03066). The funder had no role in the study design, data collection, analysis, interpretation, manuscript preparation, or decision to submit the article for publication.

\subsection*{Declaration of generative AI and AI-assisted technologies in the manuscript preparation process}
The authors used AI-based tools for language editing, code assistance, and manuscript preparation support. All content was manually reviewed and edited by the authors, who take full responsibility for the final manuscript.

\subsection*{Author contributions}
Ivo {\"O}sterberg Nilsson: Conceptualization, Methodology, Software, Formal analysis, Investigation, Writing - Original Draft, Writing - review \& editing, Visualization, Data Curation, Validation. \\
Maximilian Birr Engvall: Conceptualization, Methodology, Software, Formal analysis, Investigation, Writing - Original Draft, Visualization, Data Curation, Validation. \\
Viktor Valadi: Methodology, Validation, Resources, Writing - review \& editing, Supervision, Project administration. \\
Teddy Lazebnik: Validation, Resources, Data Curation, Writing - Review \& Editing, Supervision. \\

\bibliography{biblio}
\bibliographystyle{unsrt}
 
\appendix

\section{Benchmark Robustness and Sensitivity Analyses}
\label{appendix}

The benchmark robustness results support the main tabular FL analysis by extending the main attack point summary tables with exposure trajectories, exact row recovery summaries, fixed batch controls, client heterogeneity controls, attack budget controls, label unknown attacks, FedAvg results, and additional MLP architecture results. These results test whether the main conclusions persist across datasets, model families, attacker assumptions, and update protocols.

All reconstruction trajectory figures in this appendix report the five attack points sampled during FL training, with the horizontal axis showing minimum client exposure.

\subsection{Leakage across client batch sizes}
\label{app:benchmark-batch-size}
Table~\ref{tab:robustness-batch-size-initialized} shows that the two one-hot baselines leak substantial information before the first global aggregation. At batch sizes 1 and 2, ResNet and the small MLP often reach near perfect reconstruction, especially on Adult and the private multiclass benchmark. California Housing also shows strong small batch leakage, although the decline with increasing batch size is less monotonic than for the two classification tasks. Across datasets, the dominant pattern is local aggregation: increasing the client batch size generally reduces reconstruction accuracy before any global model update has occurred. This establishes that federated tabular inversion is not confined to a single prediction objective, and that client batch size is already a major privacy lever at the initialized attack point. The corresponding benchmark baseline values are reported in Appendix Table~\ref{tab:batch-size-benchmark-baseline-reference}. They provide the dataset specific reference levels for interpreting when reconstruction exceeds the client marginal or uniform random baselines.

\begin{table}[H]
  \centering
  \small
  \caption{Reconstruction accuracy at the initialized attack point across the benchmark datasets for ResNet and the small MLP under the FedSGD batch size comparison. Each cell reports mean $\pm$ standard deviation across 3 seeds.}
  \label{tab:robustness-batch-size-initialized}
  \resizebox{\textwidth}{!}{%
  \begin{tabular}{lcccccc}
  \hline
  \multicolumn{1}{l}{} & \multicolumn{2}{c}{\textbf{Adult}} & \multicolumn{2}{c}{\textbf{Private multiclass}} & \multicolumn{2}{c}{\textbf{California Housing}} \\
  \textbf{Batch size} & \textbf{ResNet} & \textbf{Small MLP} & \textbf{ResNet} & \textbf{Small MLP} & \textbf{ResNet} & \textbf{Small MLP} \\
  \hline
  1   & 0.995 $\pm$ 0.004 & 1.000 $\pm$ 0.000 & 0.980 $\pm$ 0.019 & 0.961 $\pm$ 0.035 & 0.992 $\pm$ 0.014 & 0.967 $\pm$ 0.058 \\
  2   & 0.949 $\pm$ 0.013 & 0.980 $\pm$ 0.018 & 0.983 $\pm$ 0.013 & 0.872 $\pm$ 0.129 & 0.781 $\pm$ 0.063 & 0.892 $\pm$ 0.067 \\
  4   & 0.869 $\pm$ 0.037 & 0.970 $\pm$ 0.019 & 0.944 $\pm$ 0.032 & 0.679 $\pm$ 0.063 & 0.662 $\pm$ 0.054 & 0.792 $\pm$ 0.057 \\
  8   & 0.788 $\pm$ 0.016 & 0.903 $\pm$ 0.033 & 0.797 $\pm$ 0.024 & 0.554 $\pm$ 0.056 & 0.537 $\pm$ 0.016 & 0.729 $\pm$ 0.023 \\
  16  & 0.713 $\pm$ 0.018 & 0.748 $\pm$ 0.017 & 0.624 $\pm$ 0.016 & 0.456 $\pm$ 0.007 & 0.503 $\pm$ 0.025 & 0.663 $\pm$ 0.037 \\
  32  & 0.589 $\pm$ 0.011 & 0.482 $\pm$ 0.010 & 0.443 $\pm$ 0.008 & 0.322 $\pm$ 0.005 & 0.494 $\pm$ 0.013 & 0.609 $\pm$ 0.019 \\
  64  & 0.553 $\pm$ 0.007 & 0.358 $\pm$ 0.006 & 0.338 $\pm$ 0.010 & 0.248 $\pm$ 0.003 & 0.507 $\pm$ 0.016 & 0.601 $\pm$ 0.026 \\
  128 & 0.533 $\pm$ 0.007 & 0.326 $\pm$ 0.008 & 0.286 $\pm$ 0.009 & 0.225 $\pm$ 0.002 & 0.515 $\pm$ 0.009 & 0.605 $\pm$ 0.035 \\
  256 & 0.527 $\pm$ 0.012 & 0.324 $\pm$ 0.009 & 0.261 $\pm$ 0.010 & 0.225 $\pm$ 0.000 & 0.564 $\pm$ 0.013 & 0.624 $\pm$ 0.030 \\
  \hline
  \end{tabular}
  }
\end{table}

\begin{figure}[H]
      \centering

      \begin{subfigure}[t]{0.32\textwidth}
          \centering
          \includegraphics[width=\textwidth]{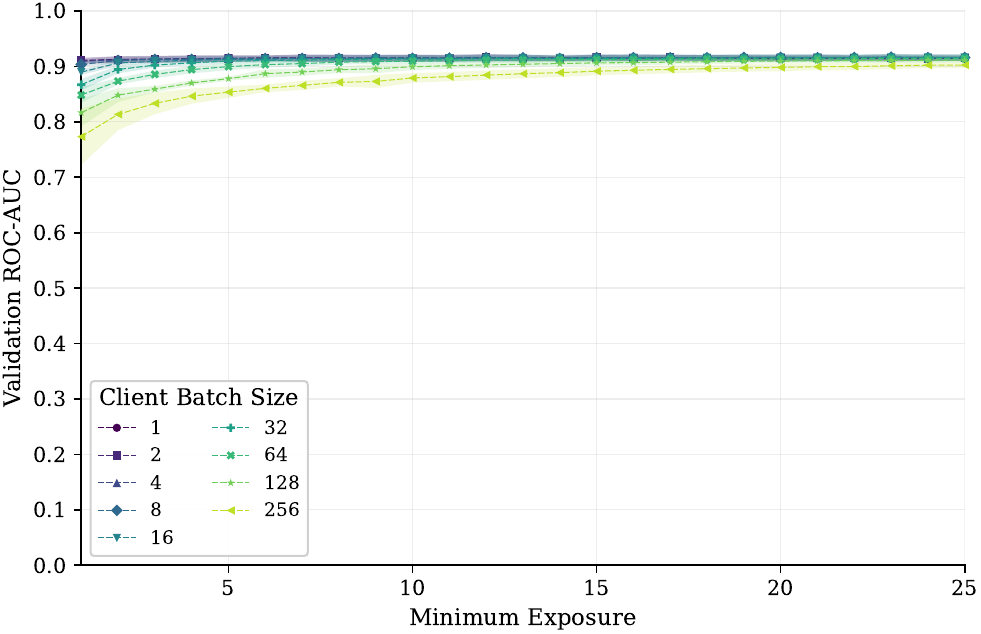}
          \caption{FT-Transformer}
      \end{subfigure}
      \hfill
      \begin{subfigure}[t]{0.32\textwidth}
          \centering
          \includegraphics[width=\textwidth]{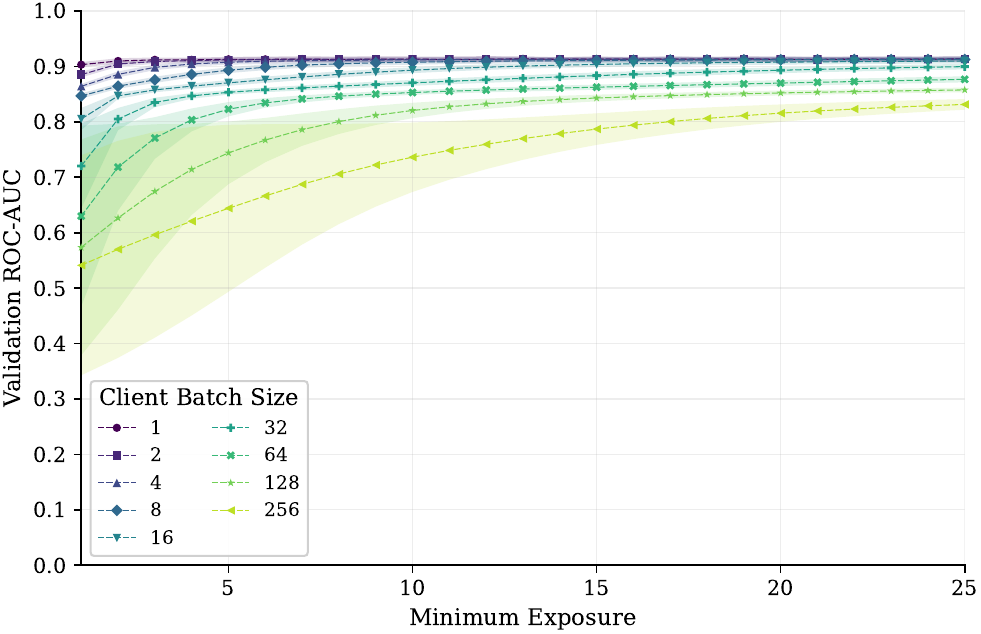}
          \caption{ResNet}
      \end{subfigure}
      \hfill
      \begin{subfigure}[t]{0.32\textwidth}
          \centering
          \includegraphics[width=\textwidth]{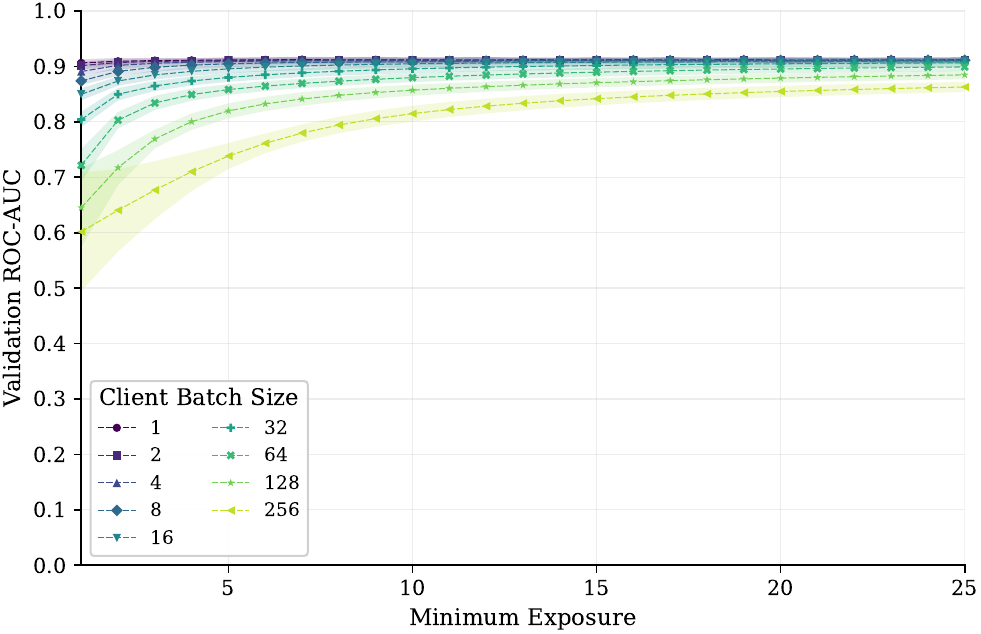}
          \caption{Small MLP}
      \end{subfigure}

      \vspace{0.5em}

      \begin{subfigure}[t]{0.32\textwidth}
          \centering
          \includegraphics[width=\textwidth]{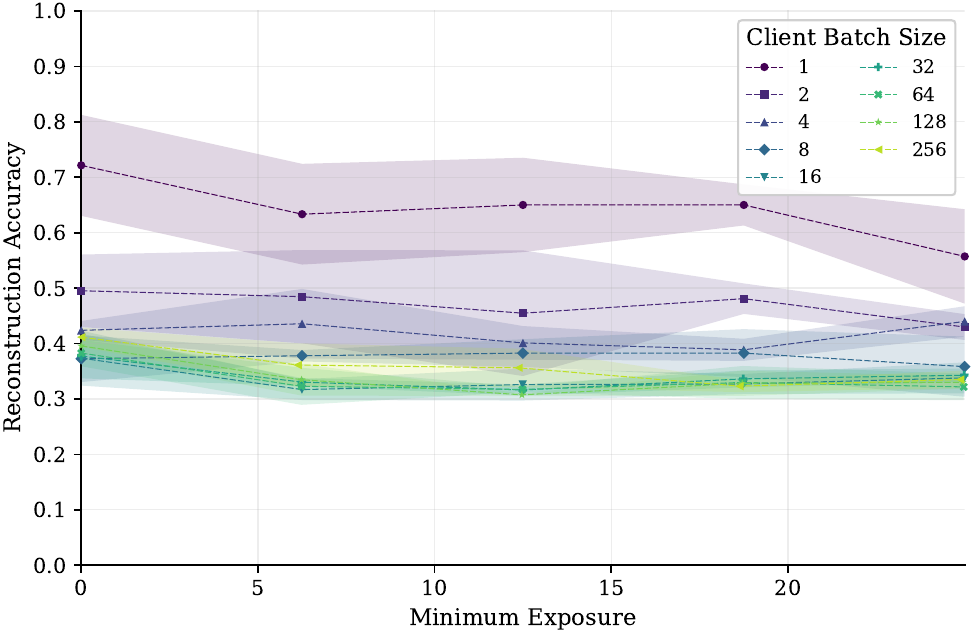}
          \caption{FT-Transformer}
      \end{subfigure}
      \hfill
      \begin{subfigure}[t]{0.32\textwidth}
          \centering
          \includegraphics[width=\textwidth]{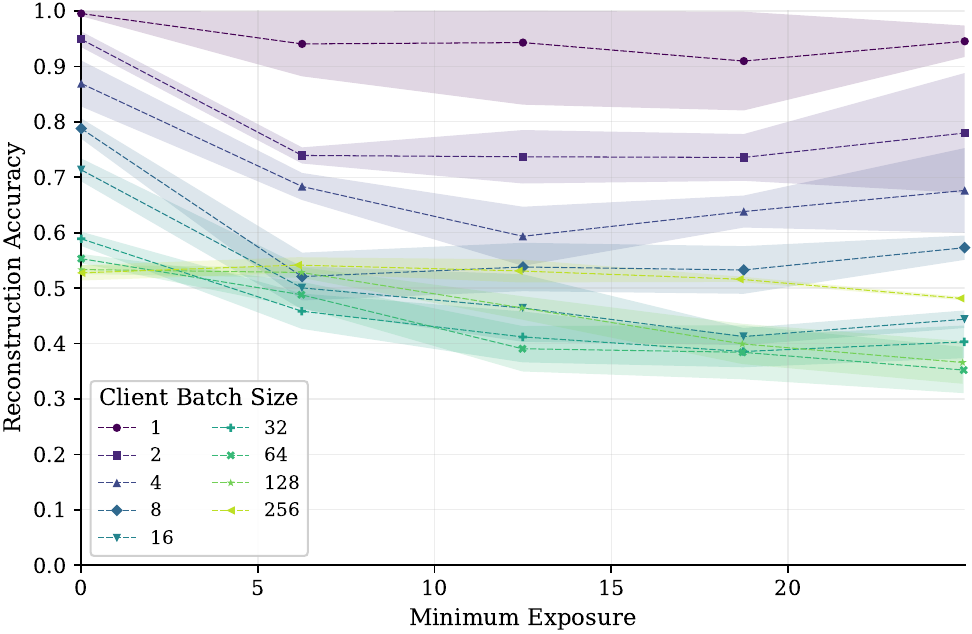}
          \caption{ResNet}
      \end{subfigure}
      \hfill
      \begin{subfigure}[t]{0.32\textwidth}
          \centering
          \includegraphics[width=\textwidth]{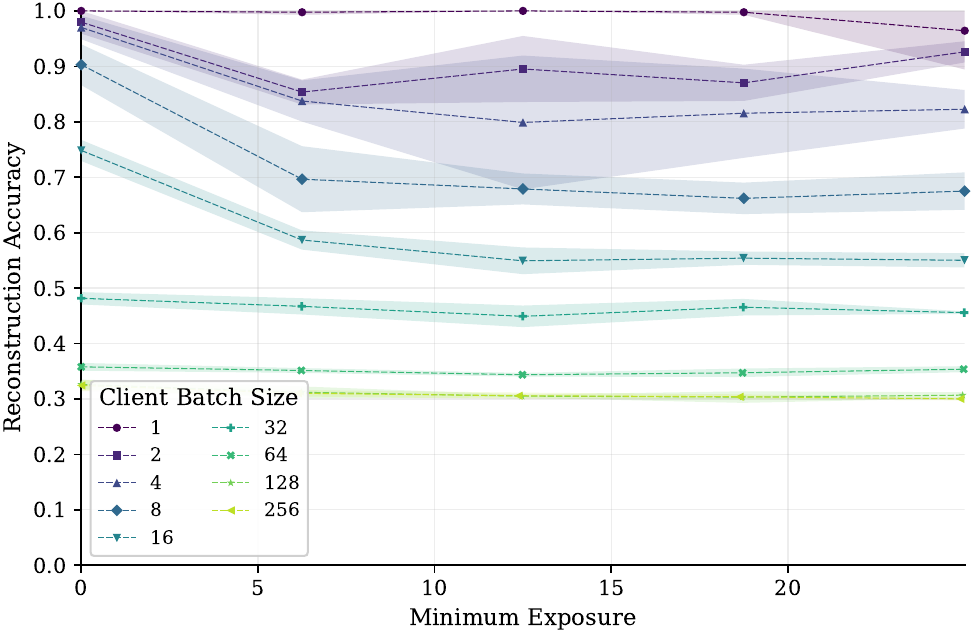}
          \caption{Small MLP}
      \end{subfigure}

      \caption{The top row shows validation utility over exposure, and the bottom row shows reconstruction accuracy over exposure for Adult.}
      \label{fig:adult-cross-model-batchsize-privacy-utility}
\end{figure}

\begin{figure}[H]
      \centering

      \begin{subfigure}[t]{0.32\textwidth}
          \centering
          \includegraphics[width=\textwidth]{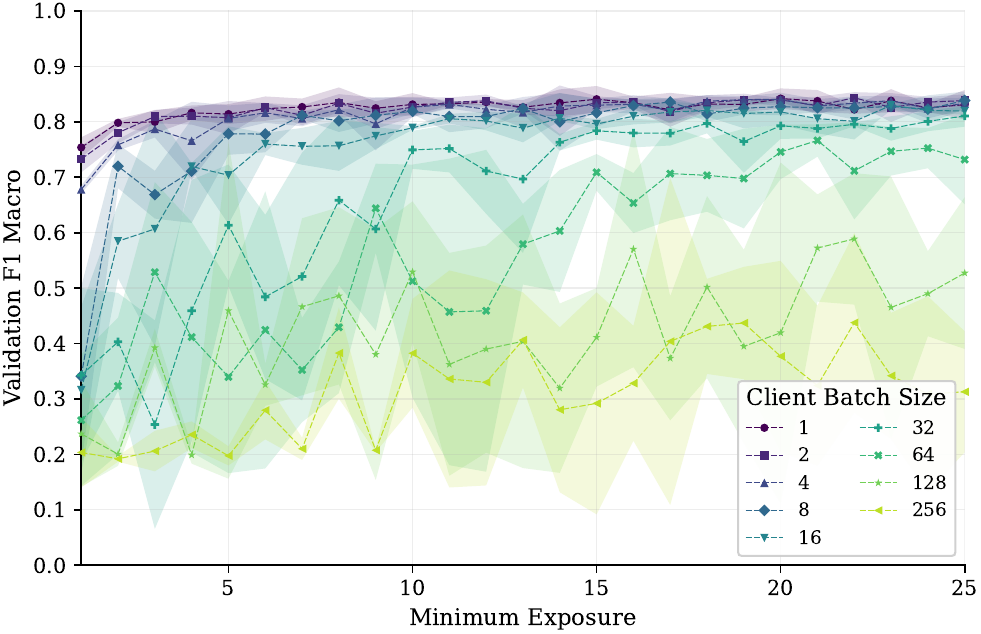}
          \caption{FT-Transformer}
      \end{subfigure}
      \hfill
      \begin{subfigure}[t]{0.32\textwidth}
          \centering
          \includegraphics[width=\textwidth]{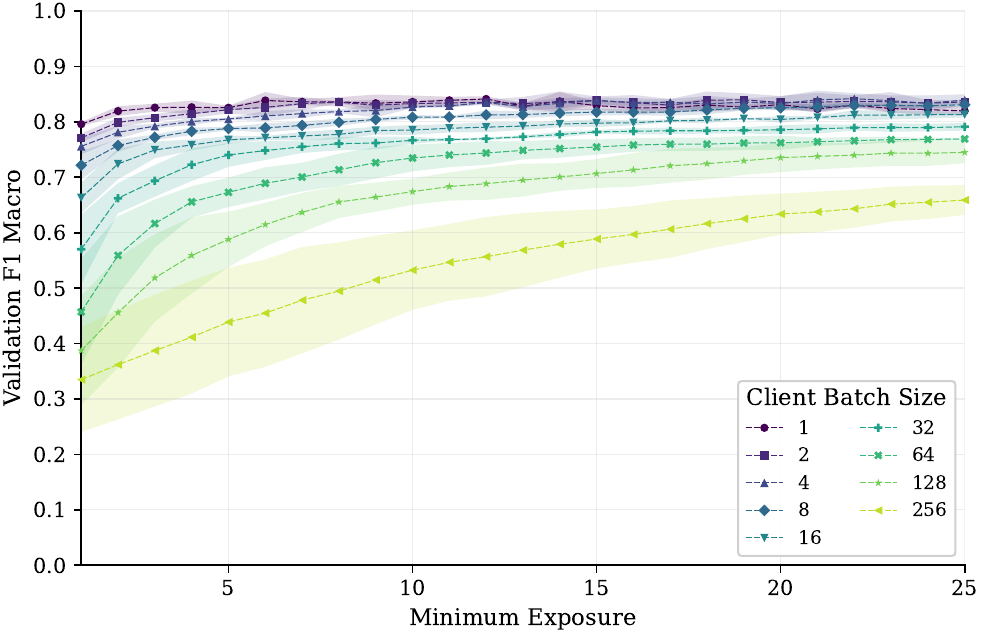}
          \caption{ResNet}
      \end{subfigure}
      \hfill
      \begin{subfigure}[t]{0.32\textwidth}
          \centering
          \includegraphics[width=\textwidth]{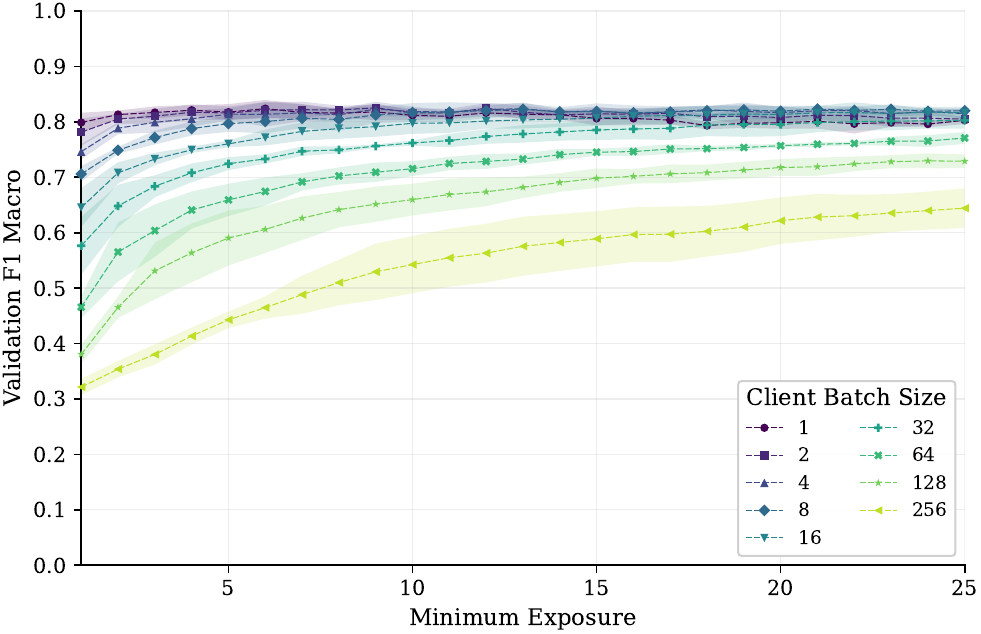}
          \caption{Small MLP}
      \end{subfigure}

      \vspace{0.5em}

      \begin{subfigure}[t]{0.32\textwidth}
          \centering
          \includegraphics[width=\textwidth]{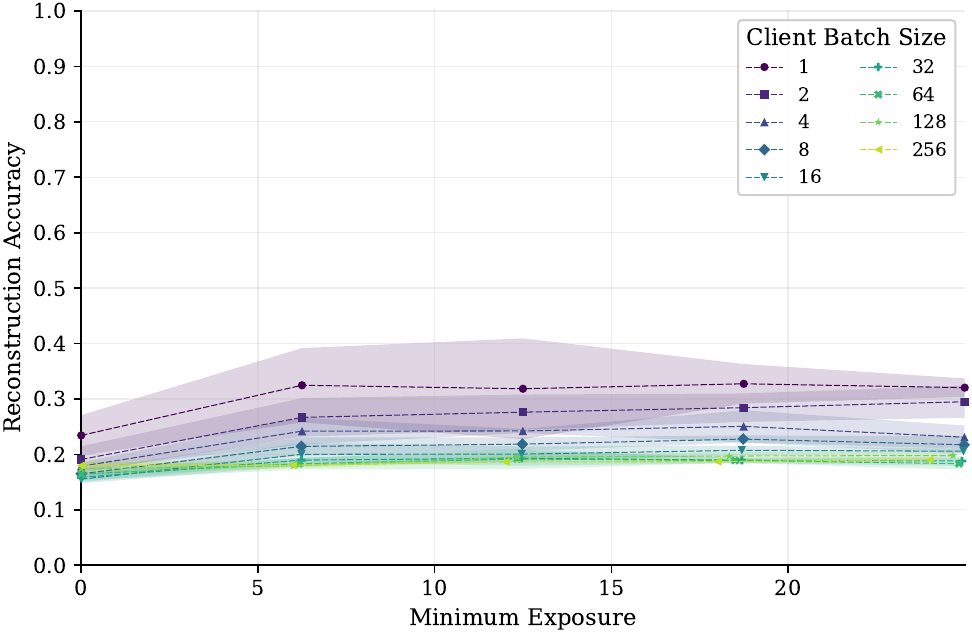}
          \caption{FT-Transformer}
      \end{subfigure}
      \hfill
      \begin{subfigure}[t]{0.32\textwidth}
          \centering
          \includegraphics[width=\textwidth]{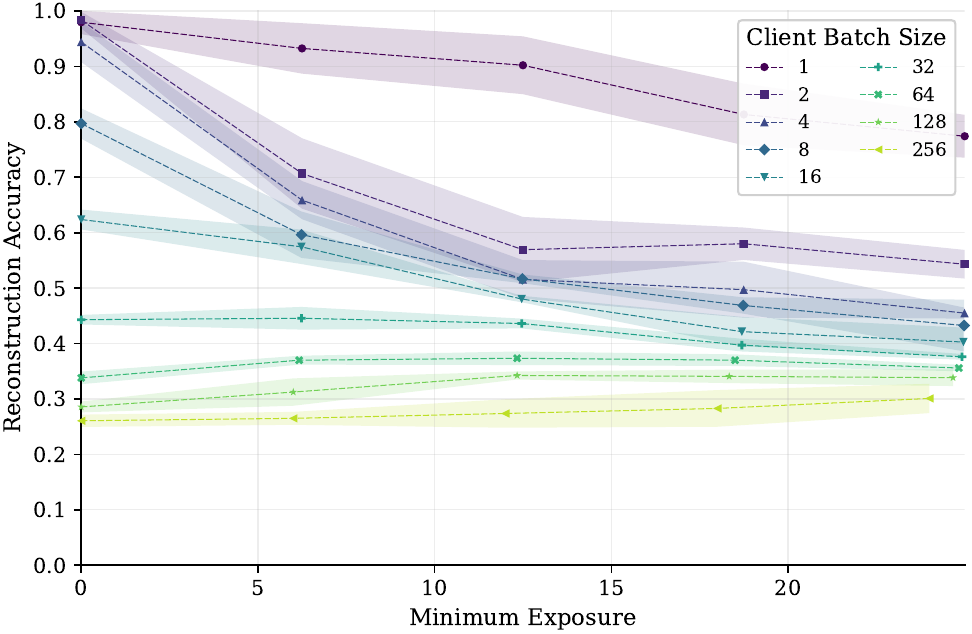}
          \caption{ResNet}
      \end{subfigure}
      \hfill
      \begin{subfigure}[t]{0.32\textwidth}
          \centering
          \includegraphics[width=\textwidth]{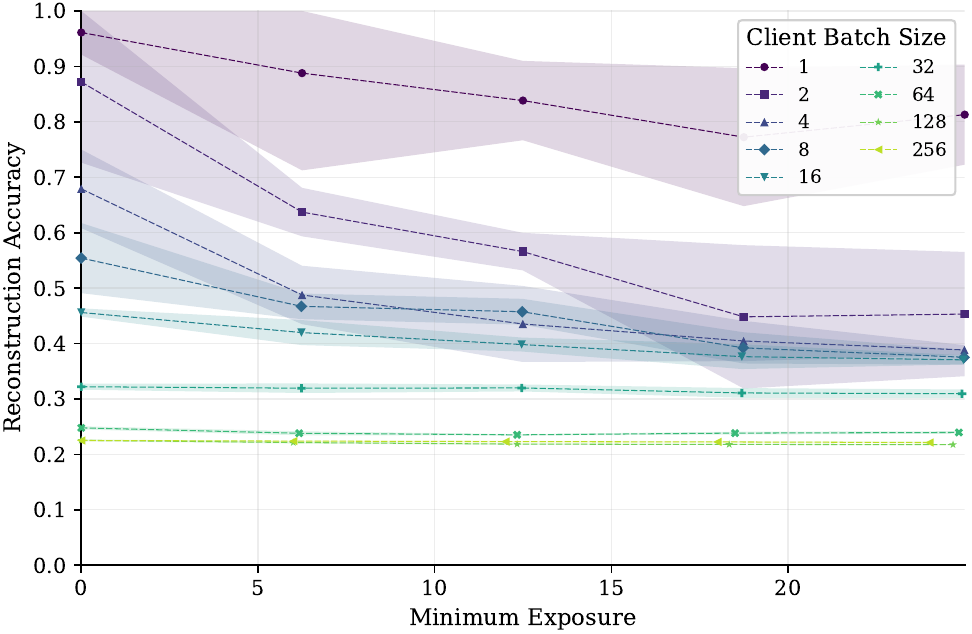}
          \caption{Small MLP}
      \end{subfigure}

      \caption{The top row shows validation utility over exposure, and the bottom row shows reconstruction accuracy over exposure for the private multiclass benchmark.}
      \label{fig:pandemic-cross-model-batchsize-privacy-utility}
\end{figure}

\begin{figure}[H]
      \centering

      \begin{subfigure}[t]{0.32\textwidth}
          \centering
          \includegraphics[width=\textwidth]{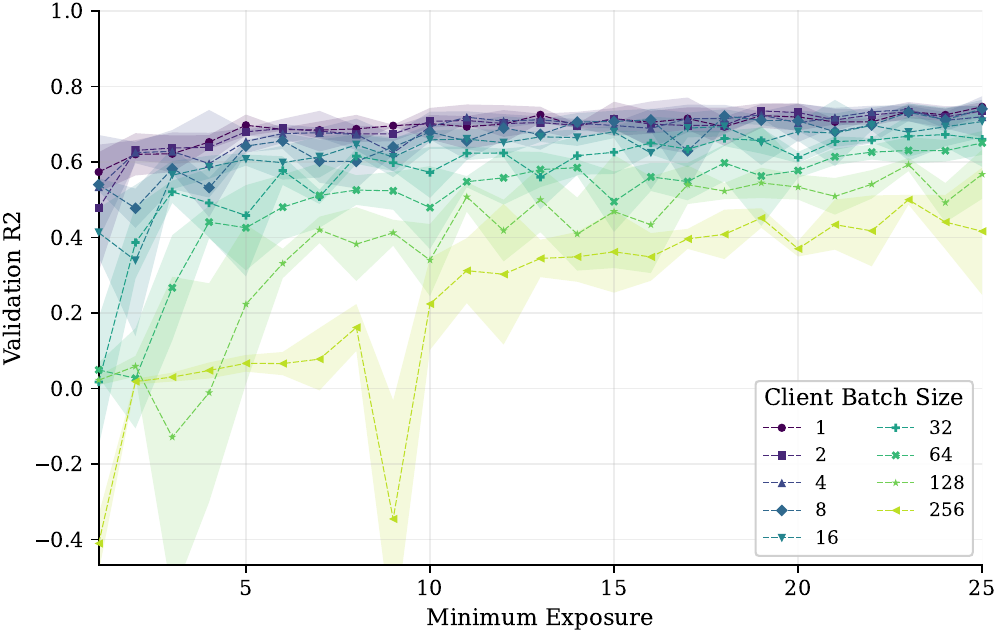}
          \caption{FT-Transformer}
      \end{subfigure}
      \hfill
      \begin{subfigure}[t]{0.32\textwidth}
          \centering
          \includegraphics[width=\textwidth]{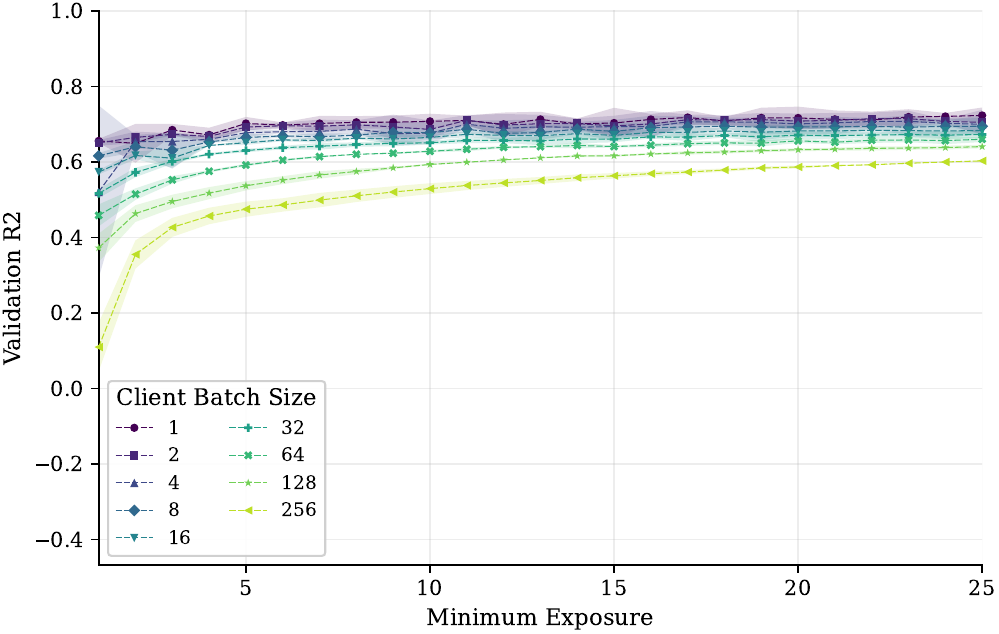}
          \caption{ResNet}
      \end{subfigure}
      \hfill
      \begin{subfigure}[t]{0.32\textwidth}
          \centering
          \includegraphics[width=\textwidth]{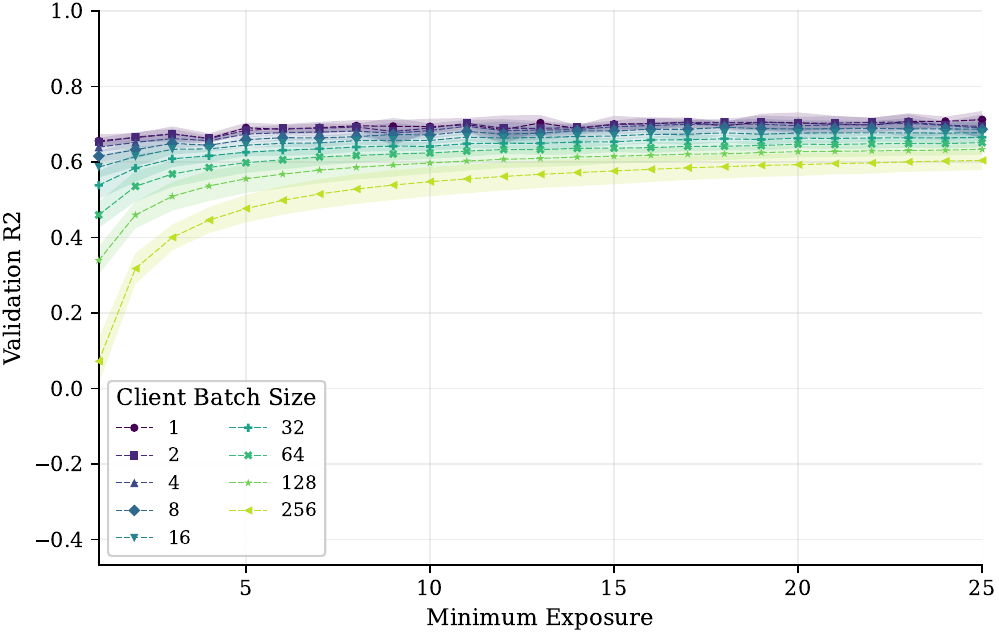}
          \caption{Small MLP}
      \end{subfigure}

      \vspace{0.5em}

      \begin{subfigure}[t]{0.32\textwidth}
          \centering
          \includegraphics[width=\textwidth]{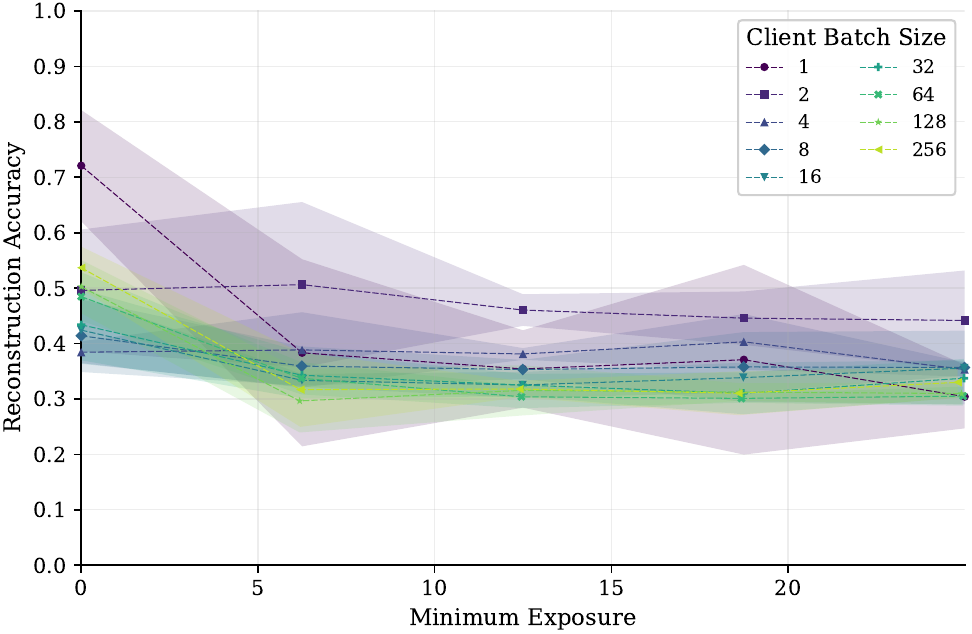}
          \caption{FT-Transformer}
      \end{subfigure}
      \hfill
      \begin{subfigure}[t]{0.32\textwidth}
          \centering
          \includegraphics[width=\textwidth]{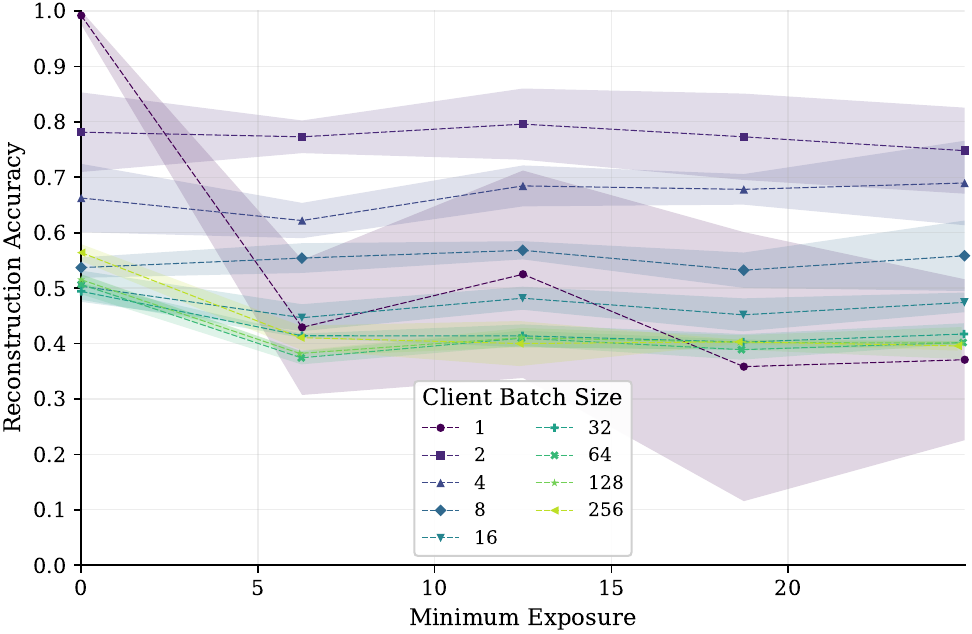}
          \caption{ResNet}
      \end{subfigure}
      \hfill
      \begin{subfigure}[t]{0.32\textwidth}
          \centering
          \includegraphics[width=\textwidth]{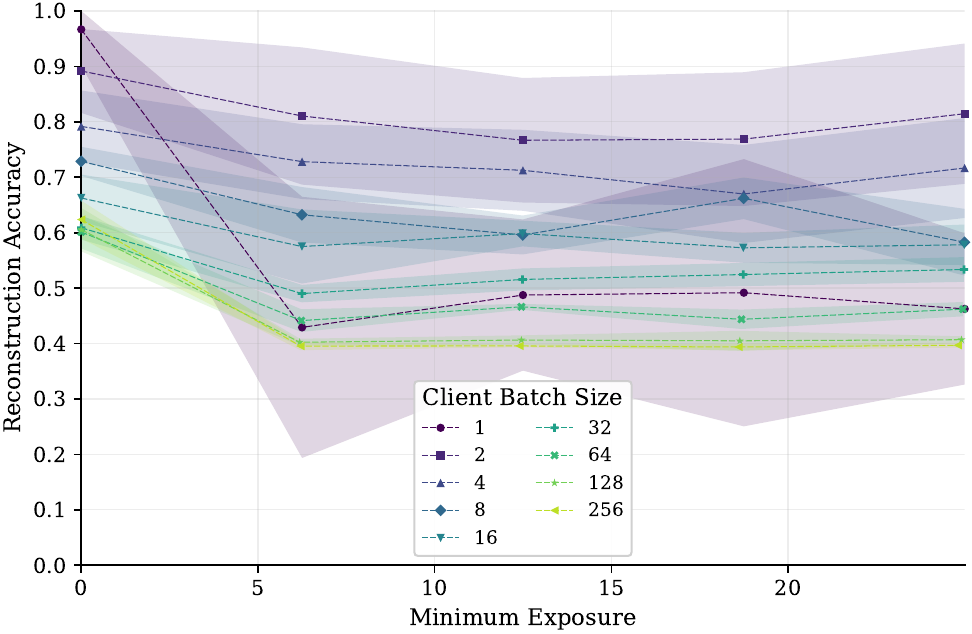}
          \caption{Small MLP}
      \end{subfigure}

      \caption{The top row shows validation utility over exposure, and the bottom row shows reconstruction accuracy over exposure for California Housing.}
      \label{fig:cali-cross-model-batchsize-privacy-utility}
\end{figure}

\begin{figure}[H]
    \centering
    \includegraphics[width=0.9\textwidth]{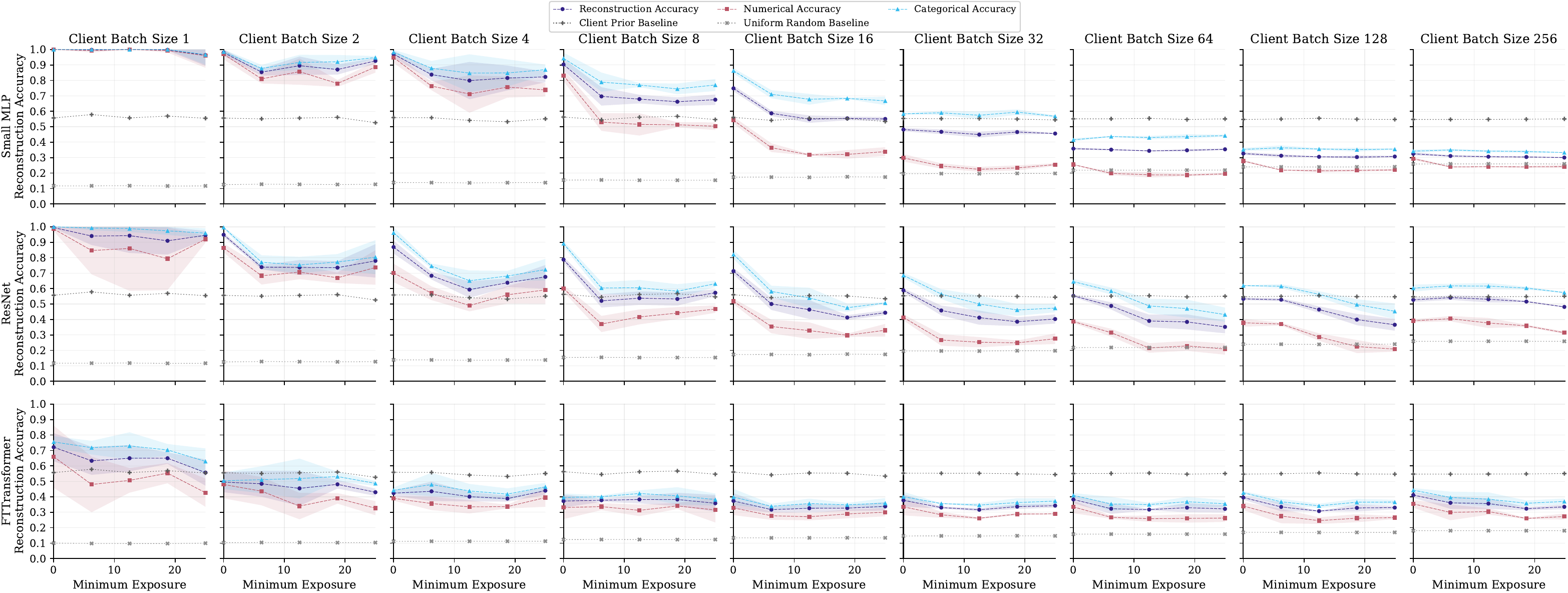}
    \caption{Reconstruction trajectories across model families and client batch sizes on Adult.}
    \label{fig:adult-all-models-all-batchsize-trajectories}
\end{figure}

\begin{figure}[H]
    \centering
    \includegraphics[width=0.9\textwidth]{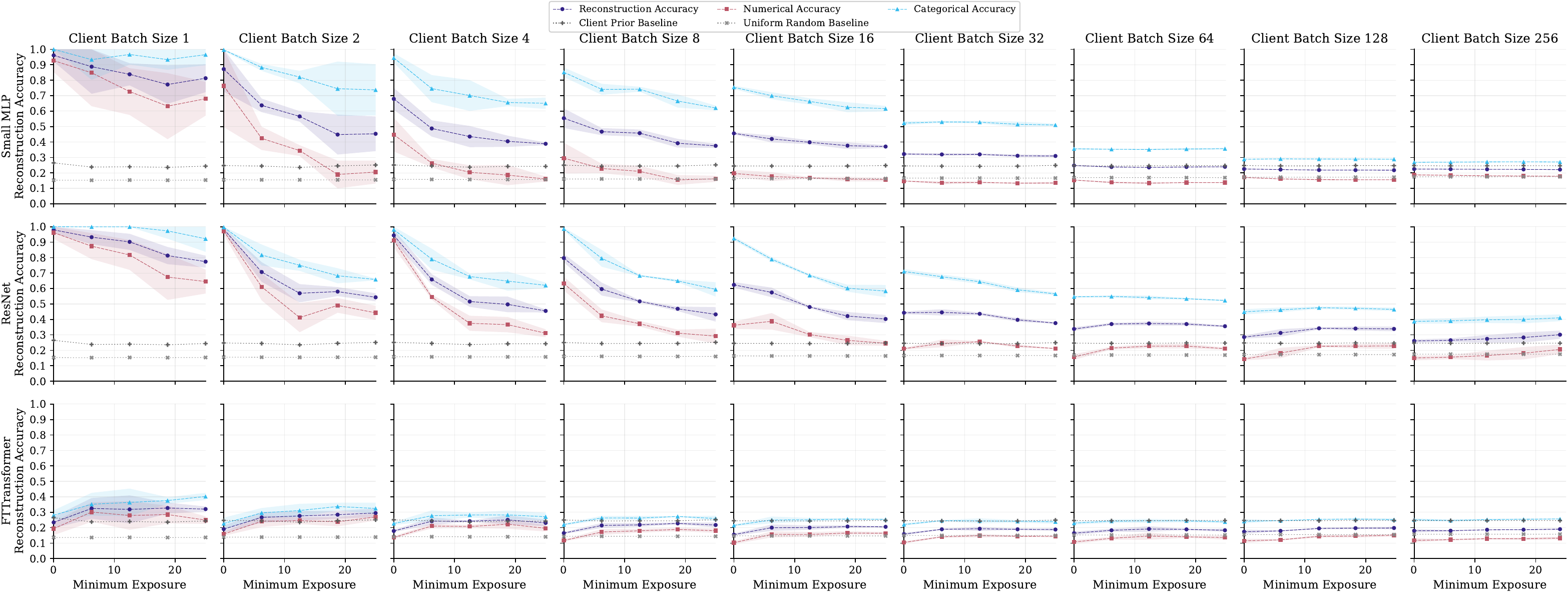}
    \caption{Reconstruction trajectories across model families and client batch sizes on the private multiclass benchmark.}
    \label{fig:pandemic-all-models-all-batchsize-trajectories}
\end{figure}

\begin{figure}[H]
    \centering
    \includegraphics[width=0.9\textwidth]{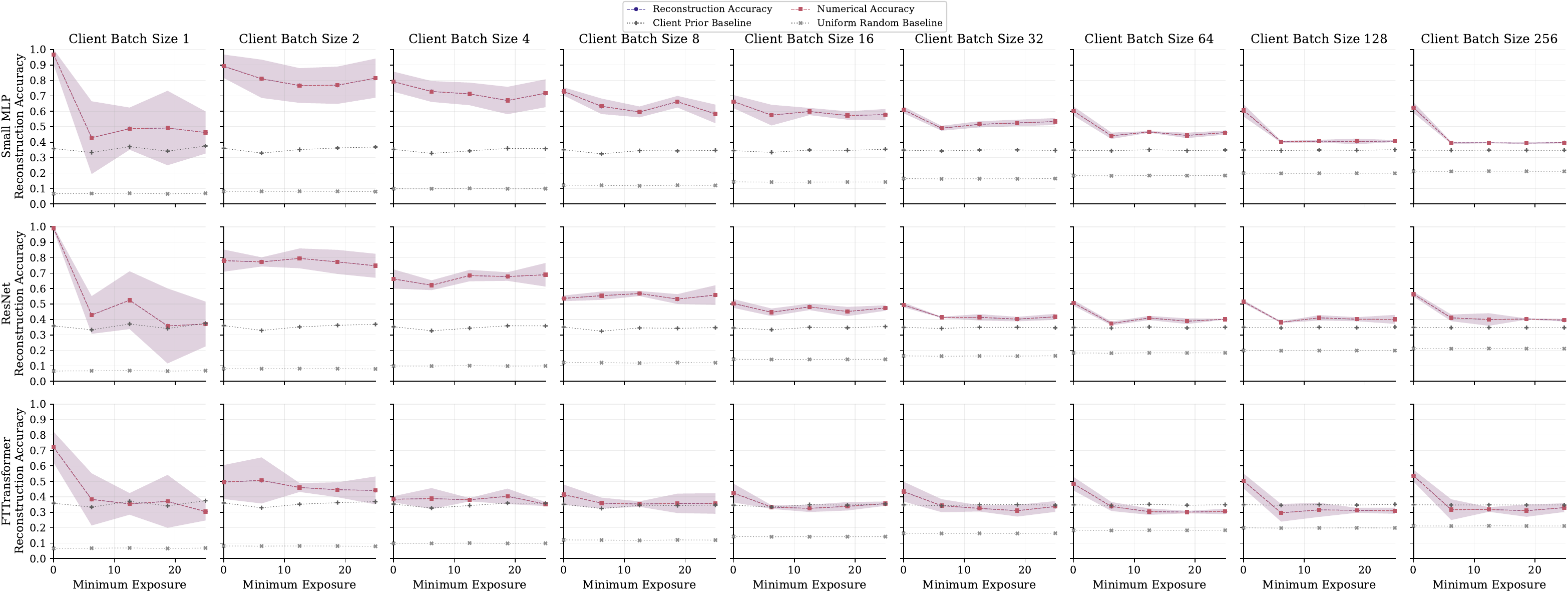}
    \caption{Reconstruction trajectories across model families and client batch sizes on California Housing.}
    \label{fig:cali-all-models-all-batchsize-trajectories}
\end{figure}

Figures~\ref{fig:adult-cross-model-batchsize-privacy-utility}--\ref{fig:cali-cross-model-batchsize-privacy-utility} report utility and reconstruction jointly, while Figures~\ref{fig:adult-all-models-all-batchsize-trajectories}--\ref{fig:cali-all-models-all-batchsize-trajectories} isolate reconstruction accuracy across model families and client batch sizes. These trajectories support the attack point summary tables in the main text by showing that the batch size effect is present over training rather than only at the selected attack points.

Table~\ref{tab:batch-size-benchmark-baseline-reference} reports the client marginal prior and uniform random reconstruction reference values for the same benchmark FedSGD batch size experiments. These baselines are not additional attacker information. They provide evaluation reference points for interpreting aggregate reconstruction accuracy across datasets. The table shows that marginal recoverability differs by dataset, which helps separate attack specific leakage from reconstruction that is already explained by simple client local feature marginals.

\begin{table}[H]
  \centering
  \small
  \caption{Client marginal prior and uniform random reconstruction baselines for the benchmark FedSGD batch size experiments. Values are computed on the same model runs and attack points as the main reconstruction experiments and reported as mean $\pm$ standard deviation. Prior denotes client marginal prior reconstruction accuracy, and Random denotes uniform random reconstruction accuracy.}
  \label{tab:batch-size-benchmark-baseline-reference}
  \resizebox{\textwidth}{!}{%
  \begin{tabular}{lcccccc}
  \hline
  \multicolumn{1}{l}{} & \multicolumn{2}{c}{\textbf{Adult}} & \multicolumn{2}{c}{\textbf{Private multiclass}} & \multicolumn{2}{c}{\textbf{California Housing}} \\
  \textbf{Batch size} & \textbf{Prior} & \textbf{Random} & \textbf{Prior} & \textbf{Random} & \textbf{Prior} & \textbf{Random} \\
  \hline
  1 & 0.563 $\pm$ 0.056 & 0.111 $\pm$ 0.009 & 0.244 $\pm$ 0.020 & 0.148 $\pm$ 0.008 & 0.356 $\pm$ 0.045 & 0.067 $\pm$ 0.003 \\
  2 & 0.550 $\pm$ 0.022 & 0.119 $\pm$ 0.011 & 0.245 $\pm$ 0.013 & 0.150 $\pm$ 0.008 & 0.355 $\pm$ 0.037 & 0.081 $\pm$ 0.002 \\
  4 & 0.548 $\pm$ 0.020 & 0.129 $\pm$ 0.012 & 0.244 $\pm$ 0.011 & 0.152 $\pm$ 0.008 & 0.348 $\pm$ 0.022 & 0.099 $\pm$ 0.002 \\
  8 & 0.557 $\pm$ 0.016 & 0.144 $\pm$ 0.015 & 0.247 $\pm$ 0.005 & 0.155 $\pm$ 0.008 & 0.342 $\pm$ 0.018 & 0.120 $\pm$ 0.002 \\
  16 & 0.549 $\pm$ 0.013 & 0.161 $\pm$ 0.019 & 0.245 $\pm$ 0.004 & 0.158 $\pm$ 0.008 & 0.346 $\pm$ 0.017 & 0.142 $\pm$ 0.002 \\
  32 & 0.550 $\pm$ 0.008 & 0.180 $\pm$ 0.024 & 0.245 $\pm$ 0.004 & 0.161 $\pm$ 0.008 & 0.347 $\pm$ 0.013 & 0.163 $\pm$ 0.003 \\
  64 & 0.551 $\pm$ 0.004 & 0.198 $\pm$ 0.029 & 0.246 $\pm$ 0.002 & 0.164 $\pm$ 0.008 & 0.348 $\pm$ 0.007 & 0.183 $\pm$ 0.002 \\
  128 & 0.549 $\pm$ 0.004 & 0.216 $\pm$ 0.033 & 0.247 $\pm$ 0.002 & 0.167 $\pm$ 0.008 & 0.349 $\pm$ 0.007 & 0.199 $\pm$ 0.002 \\
  256 & 0.548 $\pm$ 0.003 & 0.233 $\pm$ 0.037 & 0.246 $\pm$ 0.001 & 0.169 $\pm$ 0.008 & 0.348 $\pm$ 0.006 & 0.212 $\pm$ 0.001 \\
  \hline
  \end{tabular}
  }
\end{table}

Tables~\ref{tab:batch-size-adult-strict-emr}--\ref{tab:batch-size-california-housing-strict-emr} report EMRs for the same FedSGD batch size experiments. These row level results separate complete record recovery from feature level reconstruction. Exact row recovery is concentrated in the smallest one-hot settings and falls rapidly as client batch size increases. FT-Transformer produces almost no exact row recovery on Adult and the private multiclass benchmark, and only very low rates on California Housing. Thus, the benchmark results show that aggregate reconstruction corresponds to full record recovery mainly in highly exposed one-hot settings.

\begin{table}[H]
  \centering
  \small
  \caption{EMR at the initialized attack point and final attack point for Adult. Each cell reports mean $\pm$ standard deviation across 3 seeds.}
  \label{tab:batch-size-adult-strict-emr}
  \resizebox{\textwidth}{!}{%
  \begin{tabular}{lcccccc}
  \hline
  \multicolumn{1}{l}{} & \multicolumn{3}{c}{\textbf{Initialized attack point}} & \multicolumn{3}{c}{\textbf{Final attack point}} \\
  \textbf{Batch size} & \textbf{FT-Transformer} & \textbf{ResNet} & \textbf{Small MLP} & \textbf{FT-Transformer} & \textbf{ResNet} & \textbf{Small MLP} \\
  \hline
  1 & 0.033 $\pm$ 0.058 & 0.933 $\pm$ 0.058 & 1.000 $\pm$ 0.000 & 0.000 $\pm$ 0.000 & 0.867 $\pm$ 0.058 & 0.933 $\pm$ 0.115 \\
  2 & 0.000 $\pm$ 0.000 & 0.567 $\pm$ 0.104 & 0.933 $\pm$ 0.058 & 0.017 $\pm$ 0.029 & 0.583 $\pm$ 0.126 & 0.767 $\pm$ 0.029 \\
  4 & 0.000 $\pm$ 0.000 & 0.275 $\pm$ 0.132 & 0.908 $\pm$ 0.029 & 0.017 $\pm$ 0.014 & 0.317 $\pm$ 0.058 & 0.533 $\pm$ 0.072 \\
  8 & 0.000 $\pm$ 0.000 & 0.192 $\pm$ 0.085 & 0.679 $\pm$ 0.069 & 0.000 $\pm$ 0.000 & 0.158 $\pm$ 0.019 & 0.225 $\pm$ 0.025 \\
  16 & 0.000 $\pm$ 0.000 & 0.067 $\pm$ 0.031 & 0.229 $\pm$ 0.042 & 0.000 $\pm$ 0.000 & 0.050 $\pm$ 0.019 & 0.069 $\pm$ 0.041 \\
  32 & 0.000 $\pm$ 0.000 & 0.010 $\pm$ 0.002 & 0.009 $\pm$ 0.008 & 0.000 $\pm$ 0.000 & 0.014 $\pm$ 0.008 & 0.018 $\pm$ 0.013 \\
  64 & 0.000 $\pm$ 0.000 & 0.002 $\pm$ 0.000 & 0.000 $\pm$ 0.000 & 0.000 $\pm$ 0.000 & 0.001 $\pm$ 0.001 & 0.001 $\pm$ 0.001 \\
  128 & 0.000 $\pm$ 0.000 & 0.001 $\pm$ 0.001 & 0.000 $\pm$ 0.000 & 0.000 $\pm$ 0.000 & 0.000 $\pm$ 0.000 & 0.000 $\pm$ 0.000 \\
  256 & 0.000 $\pm$ 0.000 & 0.001 $\pm$ 0.001 & 0.000 $\pm$ 0.000 & 0.000 $\pm$ 0.000 & 0.000 $\pm$ 0.000 & 0.000 $\pm$ 0.000 \\
  \hline
  \end{tabular}
  }
\end{table}

\begin{table}[H]
  \centering
  \small
  \caption{EMR at the initialized attack point and final attack point for the private multiclass benchmark. Each cell reports mean $\pm$ standard deviation across 3 seeds.}
  \label{tab:batch-size-private-multiclass-benchmark-strict-emr}
  \resizebox{\textwidth}{!}{%
  \begin{tabular}{lcccccc}
  \hline
  \multicolumn{1}{l}{} & \multicolumn{3}{c}{\textbf{Initialized attack point}} & \multicolumn{3}{c}{\textbf{Final attack point}} \\
  \textbf{Batch size} & \textbf{FT-Transformer} & \textbf{ResNet} & \textbf{Small MLP} & \textbf{FT-Transformer} & \textbf{ResNet} & \textbf{Small MLP} \\
  \hline
  1 & 0.000 $\pm$ 0.000 & 0.867 $\pm$ 0.058 & 0.900 $\pm$ 0.100 & 0.000 $\pm$ 0.000 & 0.567 $\pm$ 0.058 & 0.533 $\pm$ 0.115 \\
  2 & 0.000 $\pm$ 0.000 & 0.850 $\pm$ 0.050 & 0.600 $\pm$ 0.328 & 0.000 $\pm$ 0.000 & 0.267 $\pm$ 0.076 & 0.100 $\pm$ 0.000 \\
  4 & 0.000 $\pm$ 0.000 & 0.667 $\pm$ 0.146 & 0.175 $\pm$ 0.115 & 0.000 $\pm$ 0.000 & 0.067 $\pm$ 0.029 & 0.033 $\pm$ 0.029 \\
  8 & 0.000 $\pm$ 0.000 & 0.083 $\pm$ 0.047 & 0.037 $\pm$ 0.037 & 0.000 $\pm$ 0.000 & 0.050 $\pm$ 0.025 & 0.013 $\pm$ 0.013 \\
  16 & 0.000 $\pm$ 0.000 & 0.000 $\pm$ 0.000 & 0.000 $\pm$ 0.000 & 0.000 $\pm$ 0.000 & 0.010 $\pm$ 0.004 & 0.000 $\pm$ 0.000 \\
  32 & 0.000 $\pm$ 0.000 & 0.000 $\pm$ 0.000 & 0.000 $\pm$ 0.000 & 0.000 $\pm$ 0.000 & 0.002 $\pm$ 0.002 & 0.000 $\pm$ 0.000 \\
  64 & 0.000 $\pm$ 0.000 & 0.000 $\pm$ 0.000 & 0.000 $\pm$ 0.000 & 0.000 $\pm$ 0.000 & 0.000 $\pm$ 0.000 & 0.000 $\pm$ 0.000 \\
  128 & 0.000 $\pm$ 0.000 & 0.000 $\pm$ 0.000 & 0.000 $\pm$ 0.000 & 0.000 $\pm$ 0.000 & 0.000 $\pm$ 0.000 & 0.000 $\pm$ 0.000 \\
  256 & 0.000 $\pm$ 0.000 & 0.000 $\pm$ 0.000 & 0.000 $\pm$ 0.000 & 0.000 $\pm$ 0.000 & 0.000 $\pm$ 0.000 & 0.000 $\pm$ 0.000 \\
  \hline
  \end{tabular}
  }
\end{table}

\begin{table}[H]
  \centering
  \small
  \caption{EMR at the initialized attack point and final attack point for California Housing. Each cell reports mean $\pm$ standard deviation across 3 seeds.}
  \label{tab:batch-size-california-housing-strict-emr}
  \resizebox{\textwidth}{!}{%
  \begin{tabular}{lcccccc}
  \hline
  \multicolumn{1}{l}{} & \multicolumn{3}{c}{\textbf{Initialized attack point}} & \multicolumn{3}{c}{\textbf{Final attack point}} \\
  \textbf{Batch size} & \textbf{FT-Transformer} & \textbf{ResNet} & \textbf{Small MLP} & \textbf{FT-Transformer} & \textbf{ResNet} & \textbf{Small MLP} \\
  \hline
  1 & 0.067 $\pm$ 0.115 & 0.967 $\pm$ 0.058 & 0.967 $\pm$ 0.058 & 0.033 $\pm$ 0.058 & 0.233 $\pm$ 0.153 & 0.267 $\pm$ 0.115 \\
  2 & 0.017 $\pm$ 0.029 & 0.517 $\pm$ 0.126 & 0.750 $\pm$ 0.132 & 0.067 $\pm$ 0.058 & 0.617 $\pm$ 0.076 & 0.750 $\pm$ 0.132 \\
  4 & 0.008 $\pm$ 0.014 & 0.308 $\pm$ 0.052 & 0.617 $\pm$ 0.072 & 0.008 $\pm$ 0.014 & 0.425 $\pm$ 0.087 & 0.475 $\pm$ 0.164 \\
  8 & 0.004 $\pm$ 0.007 & 0.087 $\pm$ 0.037 & 0.396 $\pm$ 0.026 & 0.033 $\pm$ 0.007 & 0.212 $\pm$ 0.033 & 0.250 $\pm$ 0.078 \\
  16 & 0.006 $\pm$ 0.000 & 0.060 $\pm$ 0.013 & 0.296 $\pm$ 0.022 & 0.006 $\pm$ 0.011 & 0.100 $\pm$ 0.023 & 0.212 $\pm$ 0.041 \\
  32 & 0.001 $\pm$ 0.002 & 0.043 $\pm$ 0.012 & 0.207 $\pm$ 0.031 & 0.002 $\pm$ 0.002 & 0.041 $\pm$ 0.008 & 0.157 $\pm$ 0.021 \\
  64 & 0.009 $\pm$ 0.007 & 0.039 $\pm$ 0.003 & 0.146 $\pm$ 0.021 & 0.002 $\pm$ 0.002 & 0.013 $\pm$ 0.002 & 0.083 $\pm$ 0.010 \\
  128 & 0.010 $\pm$ 0.003 & 0.033 $\pm$ 0.012 & 0.097 $\pm$ 0.018 & 0.002 $\pm$ 0.001 & 0.007 $\pm$ 0.002 & 0.034 $\pm$ 0.004 \\
  256 & 0.012 $\pm$ 0.006 & 0.038 $\pm$ 0.008 & 0.083 $\pm$ 0.011 & 0.002 $\pm$ 0.001 & 0.002 $\pm$ 0.001 & 0.012 $\pm$ 0.005 \\
  \hline
  \end{tabular}
  }
\end{table}

\subsection{Fixed client batches across training}
\label{app:benchmark-fixed-client-batches}
\begin{figure}[H]
      \centering

      \begin{subfigure}[t]{0.48\textwidth}
          \centering
          \includegraphics[width=\textwidth]{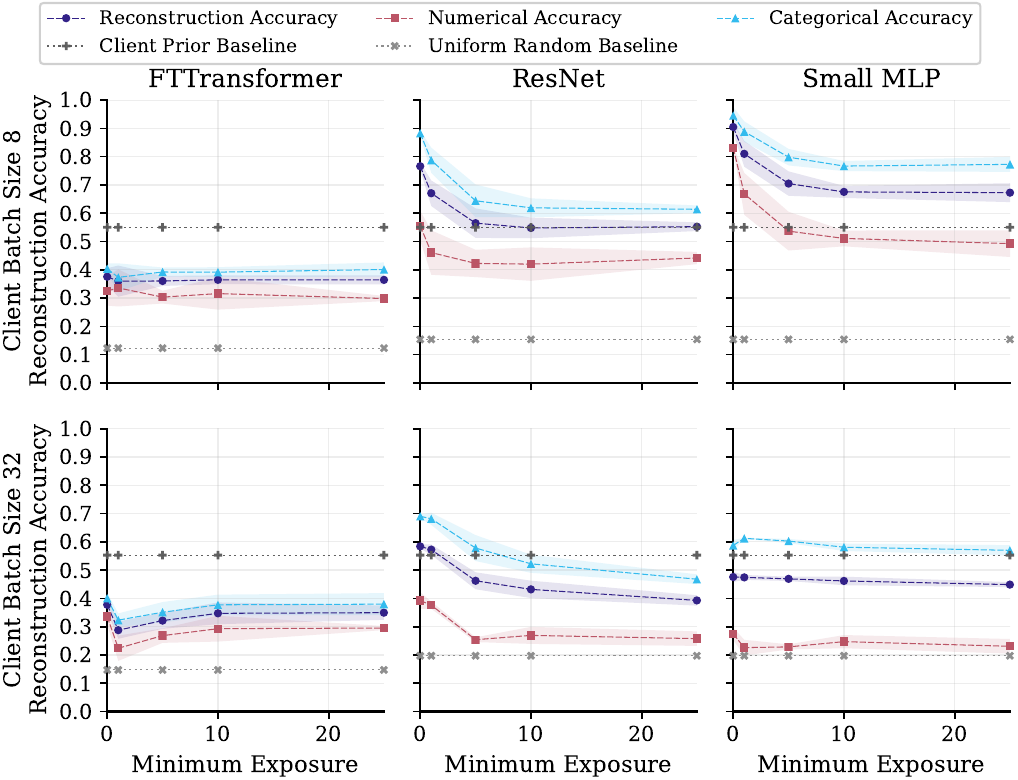}
          \caption{Adult}
      \end{subfigure}
      \hfill
      \begin{subfigure}[t]{0.48\textwidth}
          \centering
          \includegraphics[width=\textwidth]{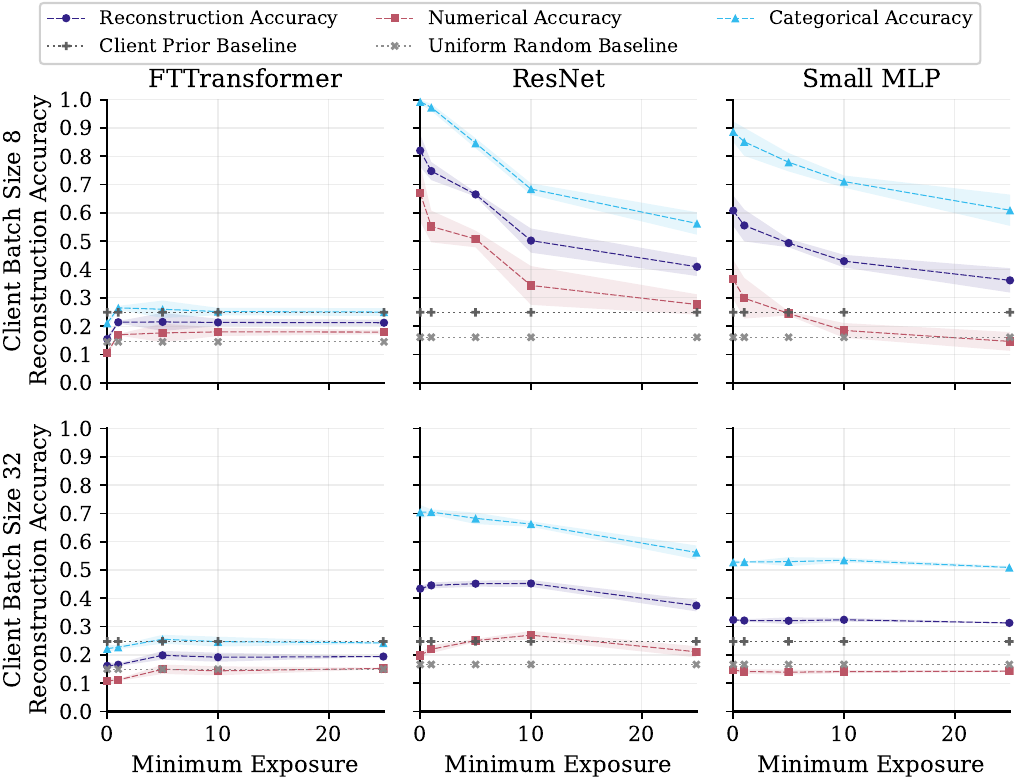}
          \caption{Private multiclass}
      \end{subfigure}

      \vspace{0.5em}

      \begin{subfigure}[t]{0.48\textwidth}
          \centering
          \includegraphics[width=\textwidth]{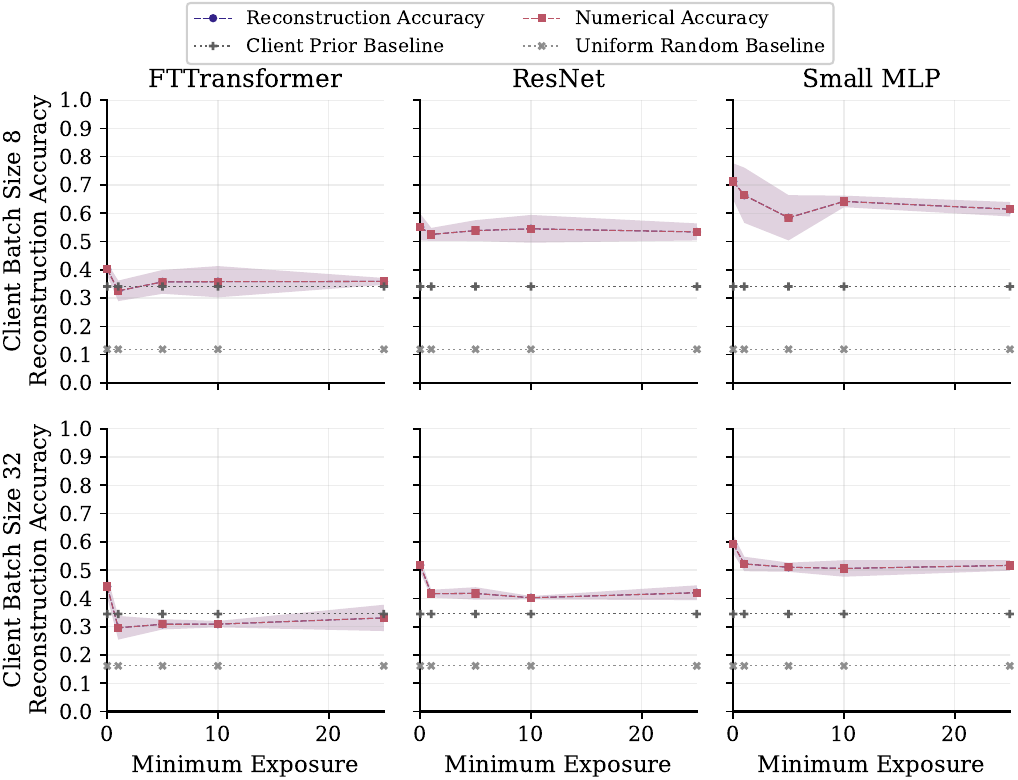}
          \caption{California Housing}
      \end{subfigure}

      \caption{Reconstruction accuracy over FL training using fixed batches per client for client batch sizes 8 and 32 across all three models and the three benchmark datasets.}
      \label{fig:benchmark-fixed-batch-all-models}
\end{figure}

Tables~\ref{tab:fixed-batch-adult-tableak-acc}--\ref{tab:fixed-batch-california-housing-tableak-acc} and Figure~\ref{fig:benchmark-fixed-batch-all-models} report fixed batch controls for the benchmark datasets. These experiments attack the same selected client batches at the initialized attack point and final attack point. This separates changes caused by model training from changes caused by attacking different sampled batches.

\begin{table}[H]
  \centering
  \small
  \caption{Reconstruction accuracy at the initialized attack point and final attack point for Adult. Initialized attack point denotes the attack before any global model aggregation, and final attack point denotes the last attacked point under the exposure budget. Each cell reports mean $\pm$ standard deviation across 3 seeds.}
  \label{tab:fixed-batch-adult-tableak-acc}
  \resizebox{\textwidth}{!}{%
  \begin{tabular}{lcccccc}
  \hline
  \multicolumn{1}{l}{} & \multicolumn{3}{c}{\textbf{Initialized attack point}} & \multicolumn{3}{c}{\textbf{Final attack point}} \\
  \textbf{Batch size} & \textbf{FT-Transformer} & \textbf{ResNet} & \textbf{Small MLP} & \textbf{FT-Transformer} & \textbf{ResNet} & \textbf{Small MLP} \\
  \hline
  8 & 0.376 $\pm$ 0.026 & 0.766 $\pm$ 0.012 & 0.905 $\pm$ 0.018 & 0.364 $\pm$ 0.014 & 0.552 $\pm$ 0.014 & 0.673 $\pm$ 0.029 \\
  32 & 0.378 $\pm$ 0.020 & 0.584 $\pm$ 0.007 & 0.476 $\pm$ 0.016 & 0.349 $\pm$ 0.022 & 0.393 $\pm$ 0.016 & 0.449 $\pm$ 0.008 \\
  \hline
  \end{tabular}
  }
\end{table}

\begin{table}[H]
  \centering
  \small
  \caption{Reconstruction accuracy at the initialized attack point and final attack point for the private multiclass benchmark. Initialized attack point denotes the attack before any global model aggregation, and final attack point denotes the last attacked point under the exposure budget. Each cell reports mean $\pm$ standard deviation across 3
  seeds.}
  \label{tab:fixed-batch-private-multiclass-tableak-acc}
  \resizebox{\textwidth}{!}{%
  \begin{tabular}{lcccccc}
  \hline
  \multicolumn{1}{l}{} & \multicolumn{3}{c}{\textbf{Initialized attack point}} & \multicolumn{3}{c}{\textbf{Final attack point}} \\
  \textbf{Batch size} & \textbf{FT-Transformer} & \textbf{ResNet} & \textbf{Small MLP} & \textbf{FT-Transformer} & \textbf{ResNet} & \textbf{Small MLP} \\
  \hline
  8 & 0.154 $\pm$ 0.011 & 0.820 $\pm$ 0.051 & 0.608 $\pm$ 0.047 & 0.212 $\pm$ 0.009 & 0.410 $\pm$ 0.029 & 0.362 $\pm$ 0.038 \\
  32 & 0.161 $\pm$ 0.012 & 0.434 $\pm$ 0.008 & 0.324 $\pm$ 0.003 & 0.194 $\pm$ 0.001 & 0.374 $\pm$ 0.018 & 0.313 $\pm$ 0.002 \\
  \hline
  \end{tabular}
  }
\end{table}

\begin{table}[H]
  \centering
  \small
  \caption{Reconstruction accuracy at the initialized attack point and final attack point for California Housing. Initialized attack point denotes the attack before any global model aggregation, and final attack point denotes the last attacked point under the exposure budget. Each cell reports mean $\pm$ standard deviation across 3 seeds.}
  \label{tab:fixed-batch-california-housing-tableak-acc}
  \resizebox{\textwidth}{!}{%
  \begin{tabular}{lcccccc}
  \hline
  \multicolumn{1}{l}{} & \multicolumn{3}{c}{\textbf{Initialized attack point}} & \multicolumn{3}{c}{\textbf{Final attack point}} \\
  \textbf{Batch size} & \textbf{FT-Transformer} & \textbf{ResNet} & \textbf{Small MLP} & \textbf{FT-Transformer} & \textbf{ResNet} & \textbf{Small MLP} \\
  \hline
  8 & 0.402 $\pm$ 0.020 & 0.551 $\pm$ 0.042 & 0.712 $\pm$ 0.057 & 0.359 $\pm$ 0.011 & 0.534 $\pm$ 0.027 & 0.614 $\pm$ 0.023 \\
  32 & 0.442 $\pm$ 0.033 & 0.518 $\pm$ 0.022 & 0.592 $\pm$ 0.029 & 0.331 $\pm$ 0.041 & 0.420 $\pm$ 0.023 & 0.517 $\pm$ 0.016 \\
  \hline
  \end{tabular}
  }
\end{table}

\subsection{Non-IID client partitions}
\label{app:benchmark-noniid}
\begin{figure}[H]
    \centering
    \includegraphics[
        width=0.95\textwidth,
        height=0.78\textheight,
        keepaspectratio
    ]{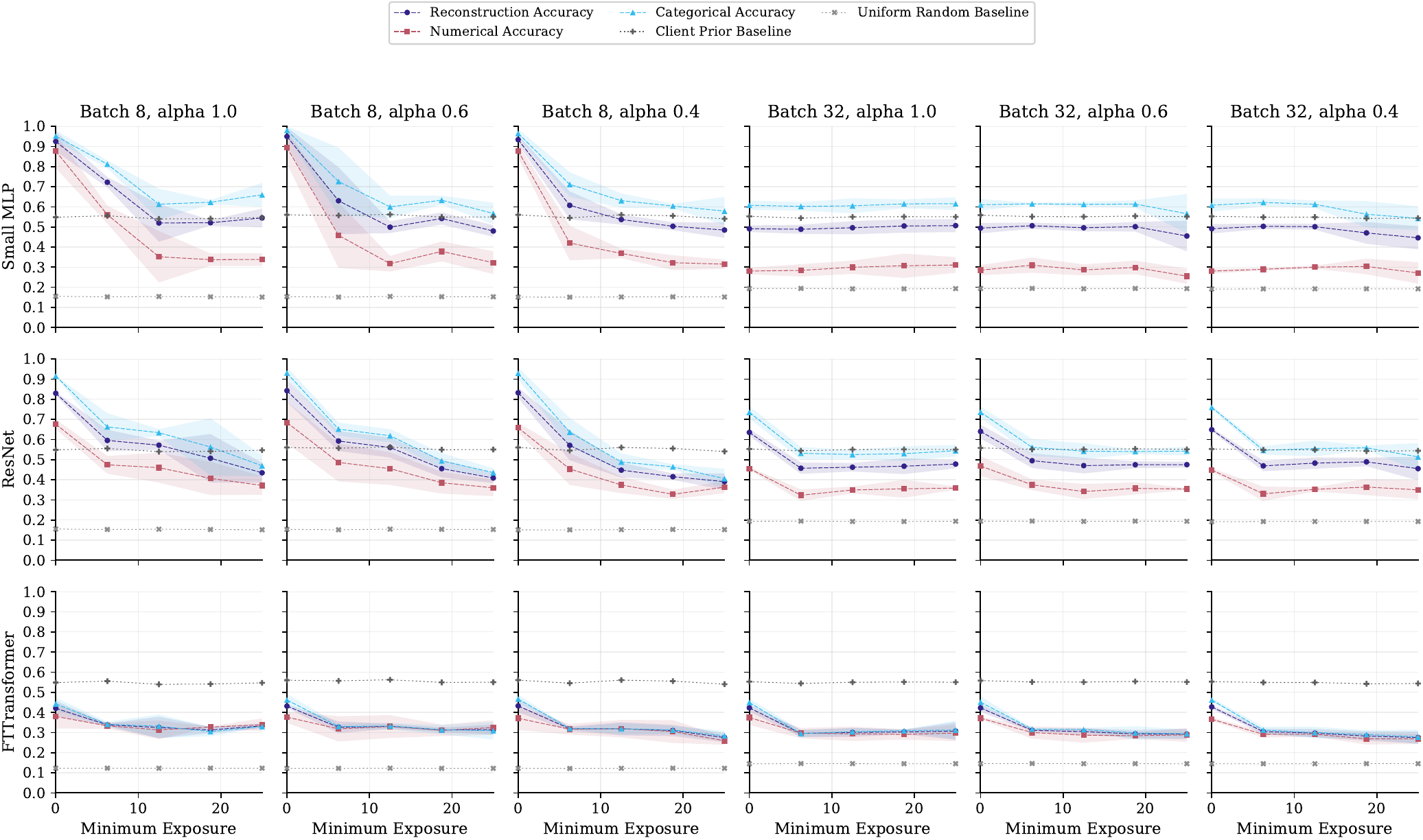}
    \caption{Non-IID robustness analysis on Adult.}
    \label{fig:appendix-benchmark-robustness-noniid-adult}
\end{figure}

\begin{figure}[H]
    \centering
    \includegraphics[
        width=0.95\textwidth,
        height=0.78\textheight,
        keepaspectratio
    ]{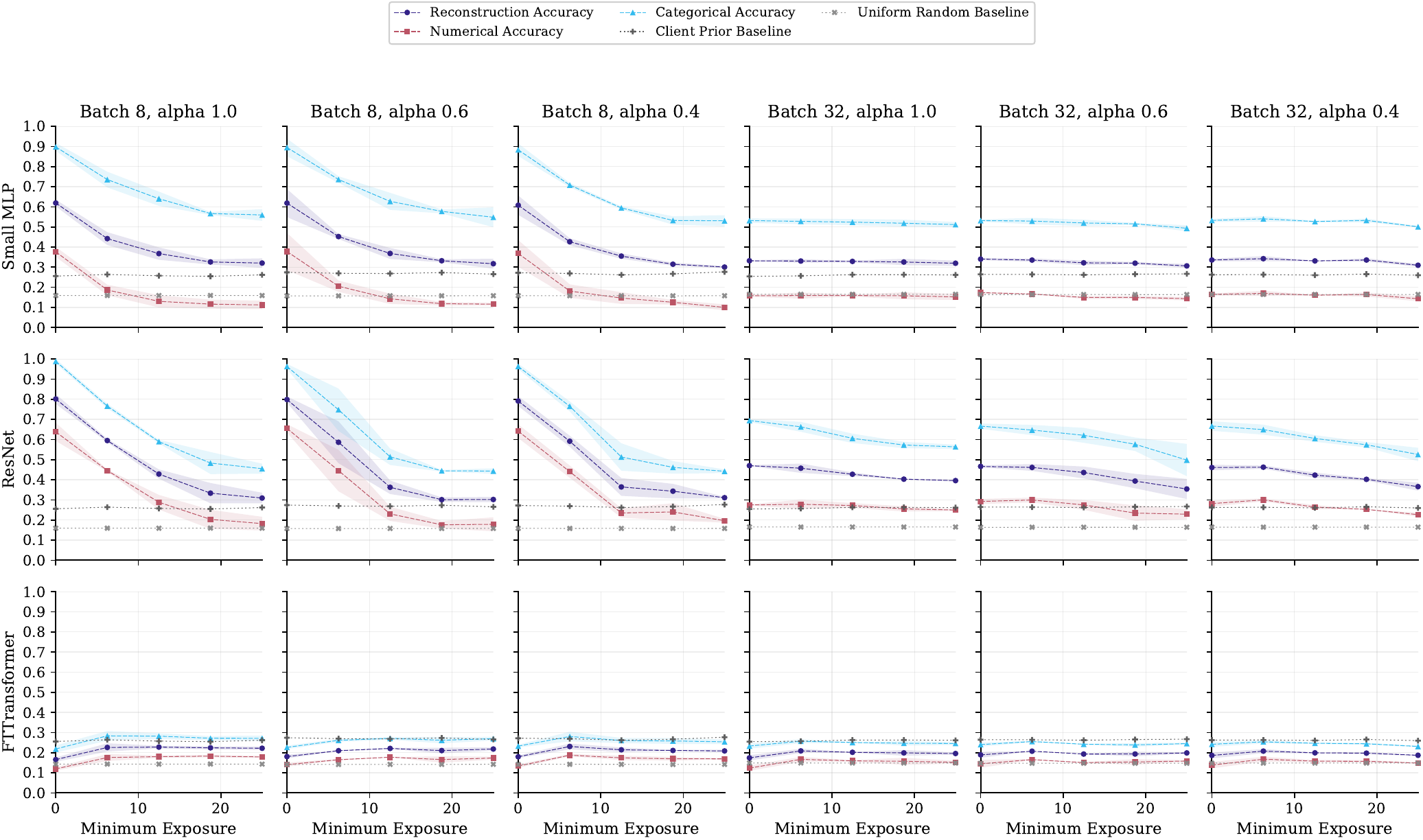}
    \caption{Non-IID robustness analysis on the private multiclass benchmark.}
    \label{fig:appendix-benchmark-robustness-noniid-private-multiclass}
\end{figure}

\begin{figure}[H]
    \centering
    \includegraphics[
        width=0.95\textwidth,
        height=0.78\textheight,
        keepaspectratio
    ]{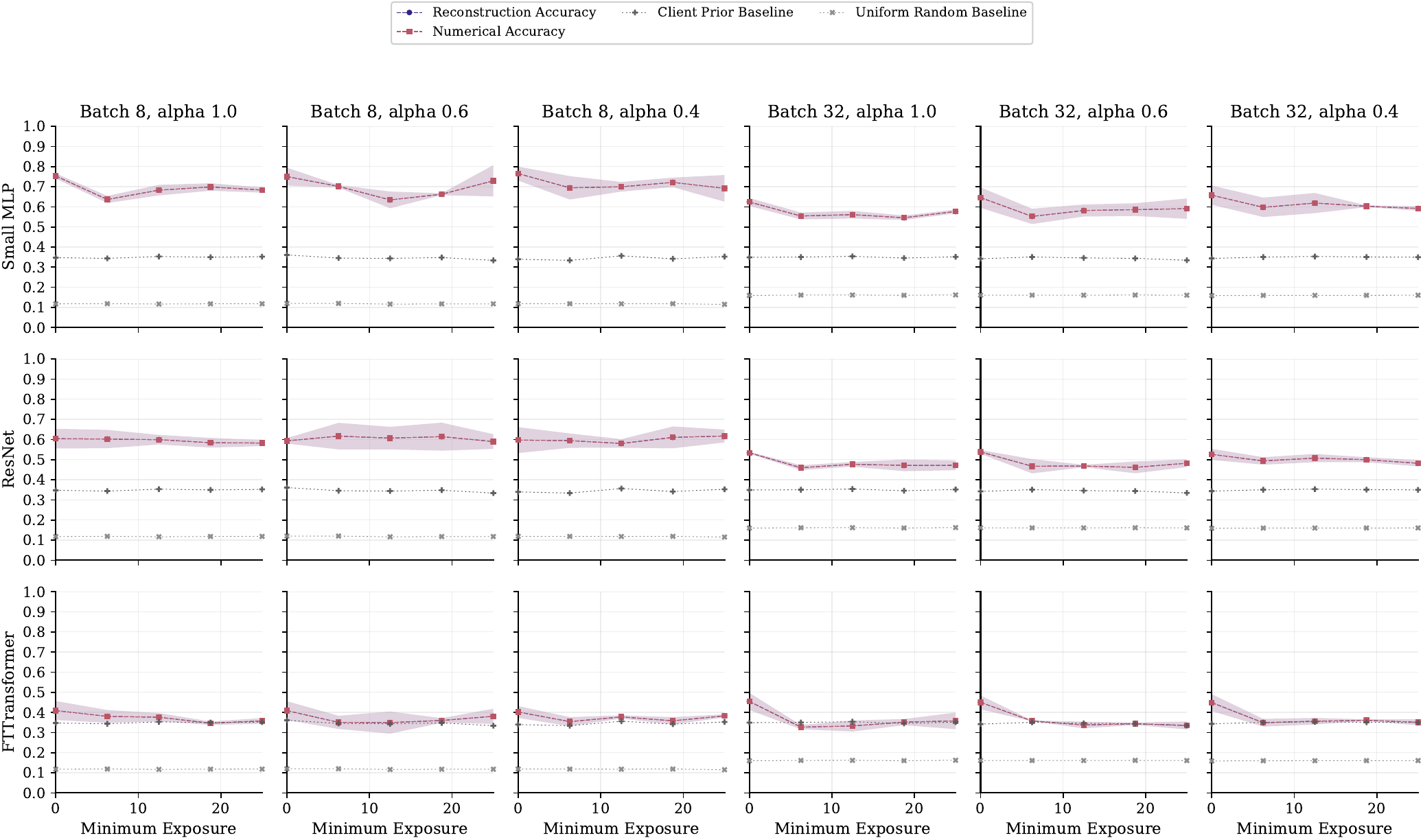}
    \caption{Non-IID robustness analysis on California Housing.}
    \label{fig:appendix-benchmark-robustness-noniid-california}
\end{figure}

Table~\ref{tab:appendix-robustness-noniid-b8-init-final} and Figures~\ref{fig:appendix-benchmark-robustness-noniid-adult}--\ref{fig:appendix-benchmark-robustness-noniid-california} report the non-IID benchmark controls. The Dirichlet concentration changes client composition while keeping the attack and evaluation pipeline fixed. These results test whether the benchmark leakage trends persist when client local data distributions are less evenly distributed.

\begin{figure}[H]
    \centering
    \includegraphics[
        width=0.95\textwidth,
        height=0.72\textheight,
        keepaspectratio
    ]{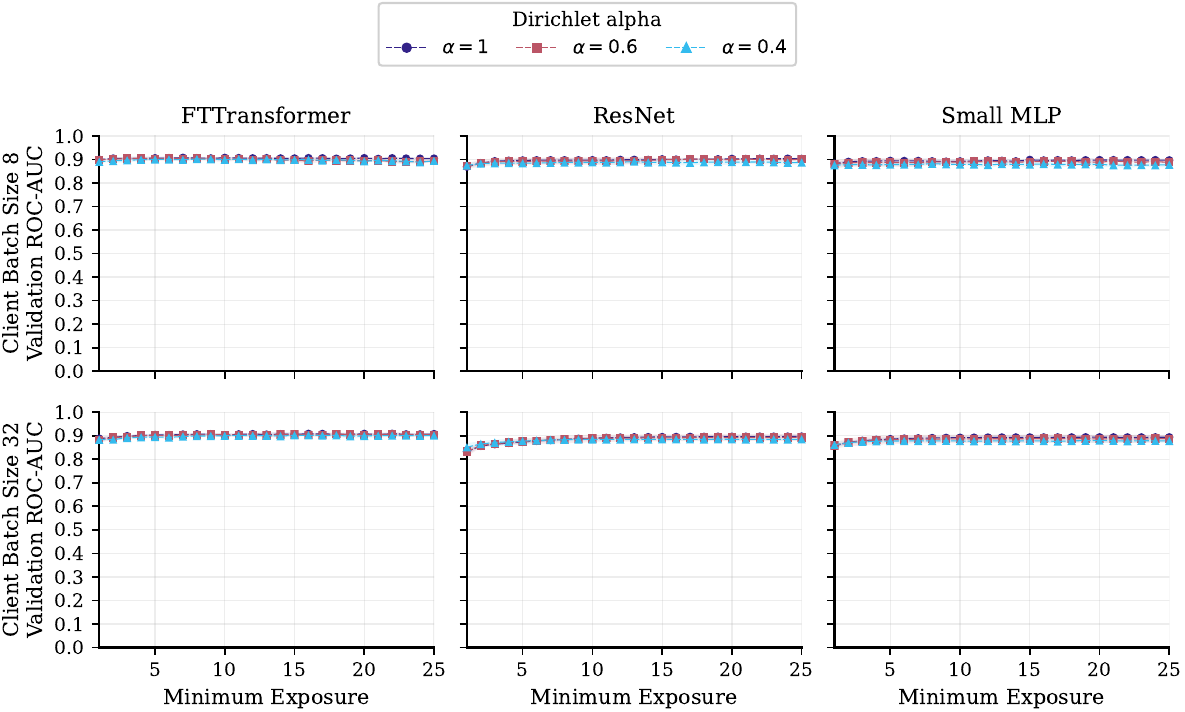}
    \caption{Validation utility under non-IID client partitions on Adult. The figure uses the same Dirichlet client partitions as the reconstruction robustness analysis.}
    \label{fig:appendix-noniid-utility-adult}
\end{figure}

\begin{figure}[H]
    \centering
    \includegraphics[
        width=0.95\textwidth,
        height=0.72\textheight,
        keepaspectratio
    ]{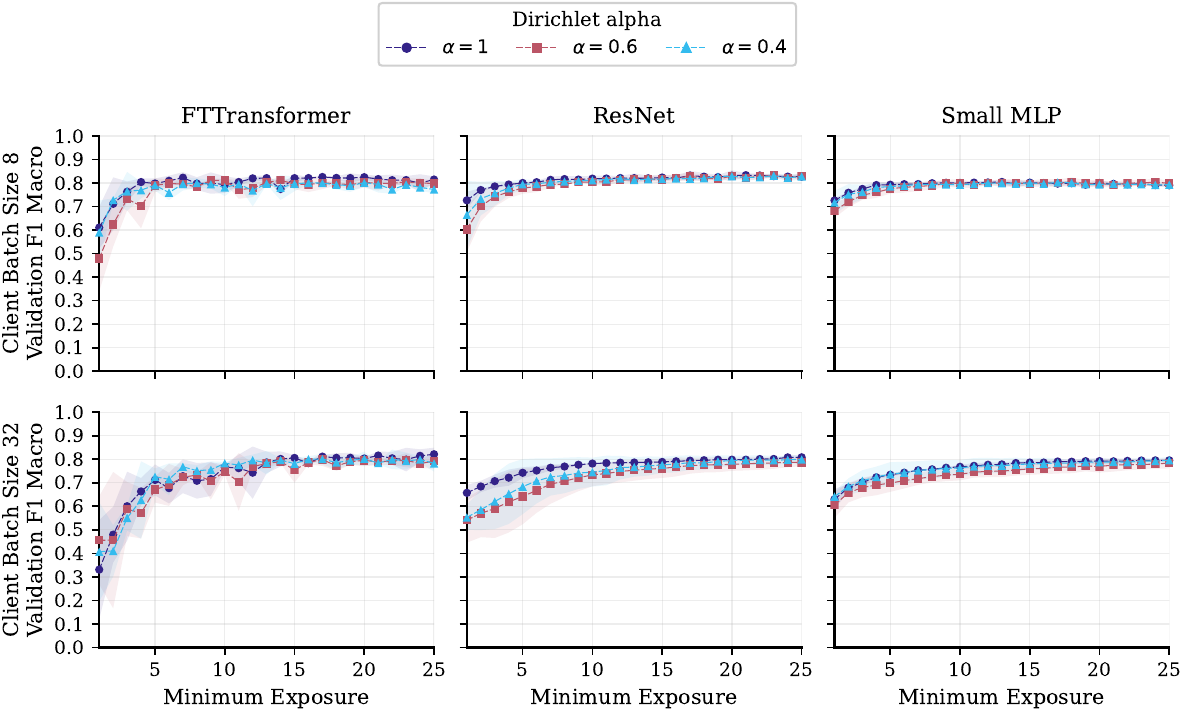}
    \caption{Validation utility under non-IID client partitions on the private multiclass benchmark. The figure uses the same Dirichlet client partitions as the reconstruction robustness analysis.}
    \label{fig:appendix-noniid-utility-private-multiclass}
\end{figure}

\begin{figure}[H]
    \centering
    \includegraphics[
        width=0.95\textwidth,
        height=0.72\textheight,
        keepaspectratio
    ]{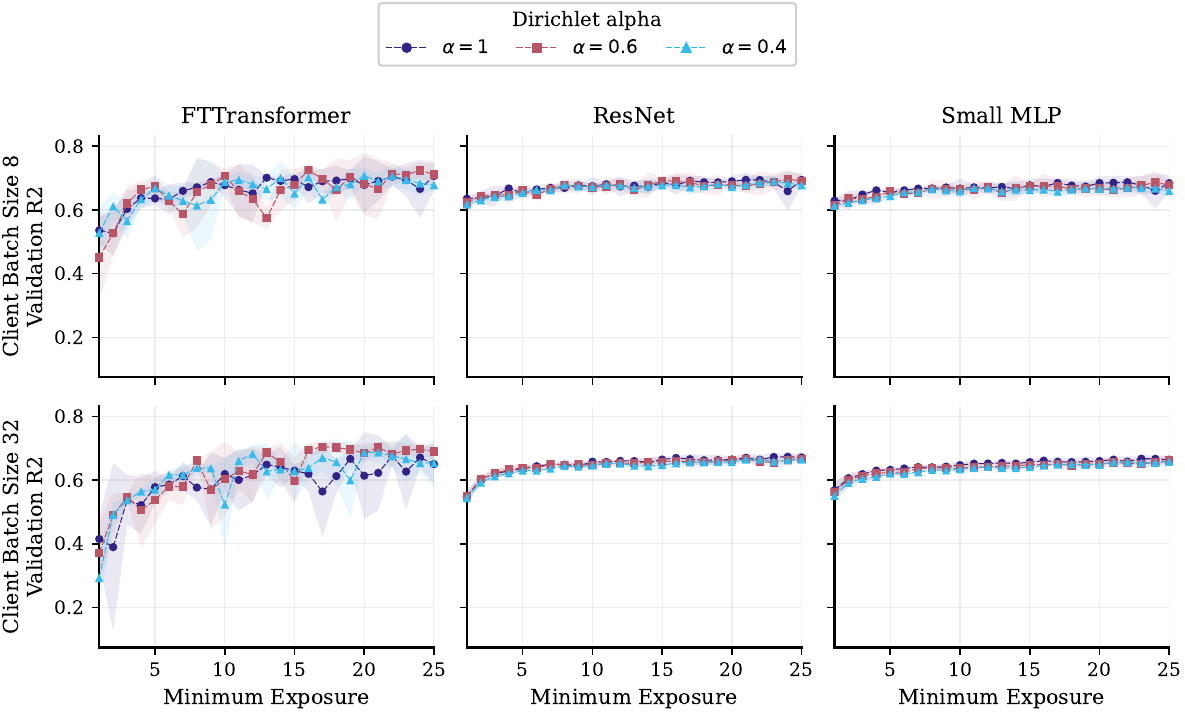}
    \caption{Validation utility under non-IID client partitions on California Housing. The figure uses the same Dirichlet client partitions as the reconstruction robustness analysis.}
    \label{fig:appendix-noniid-utility-california}
\end{figure}

Figures~\ref{fig:appendix-noniid-utility-adult}--\ref{fig:appendix-noniid-utility-california} report validation utility under the same non-IID client partitions. These curves verify that the privacy comparisons are not separated from the corresponding predictive behavior of the trained models.

\begin{table}[H]
    \centering
    \small
    \caption{Reconstruction accuracy under non-IID client partitions at batch size 8. Each cell reports mean $\pm$ standard deviation across 3 seeds.}
    \label{tab:appendix-robustness-noniid-b8-init-final}
    \resizebox{\textwidth}{!}{%
    \begin{tabular}{lccccccccc}
    \hline
     & \multicolumn{3}{c}{\textbf{Adult}} & \multicolumn{3}{c}{\textbf{Private multiclass}} & \multicolumn{3}{c}{\textbf{California Housing}} \\
    $\alpha$ & \textbf{FT-Transformer} & \textbf{ResNet} & \textbf{Small MLP} & \textbf{FT-Transformer} & \textbf{ResNet} & \textbf{Small MLP} & \textbf{FT-Transformer} & \textbf{ResNet} & \textbf{Small MLP} \\
    \hline
    \multicolumn{10}{c}{Initialized attack point} \\
    \hline
    $\alpha=0.4$ & 0.432 $\pm$ 0.028 & 0.832 $\pm$ 0.023 & 0.934 $\pm$ 0.019 & 0.180 $\pm$ 0.006 & 0.791 $\pm$ 0.022 & 0.608 $\pm$ 0.042 & 0.402 $\pm$ 0.026 & 0.597 $\pm$ 0.057 & 0.765 $\pm$ 0.031 \\
    $\alpha=0.6$ & 0.431 $\pm$ 0.008 & 0.842 $\pm$ 0.049 & 0.949 $\pm$ 0.040 & 0.181 $\pm$ 0.008 & 0.798 $\pm$ 0.014 & 0.618 $\pm$ 0.062 & 0.409 $\pm$ 0.041 & 0.593 $\pm$ 0.012 & 0.750 $\pm$ 0.039 \\
    $\alpha=1.0$ & 0.420 $\pm$ 0.033 & 0.829 $\pm$ 0.006 & 0.925 $\pm$ 0.043 & 0.165 $\pm$ 0.009 & 0.801 $\pm$ 0.022 & 0.619 $\pm$ 0.014 & 0.409 $\pm$ 0.042 & 0.604 $\pm$ 0.043 & 0.753 $\pm$ 0.011 \\
    \hline
    \multicolumn{10}{c}{Final attack point} \\
    \hline
    $\alpha=0.4$ & 0.275 $\pm$ 0.020 & 0.390 $\pm$ 0.036 & 0.485 $\pm$ 0.034 & 0.209 $\pm$ 0.006 & 0.311 $\pm$ 0.004 & 0.301 $\pm$ 0.005 & 0.383 $\pm$ 0.007 & 0.617 $\pm$ 0.028 & 0.692 $\pm$ 0.058 \\
    $\alpha=0.6$ & 0.315 $\pm$ 0.022 & 0.409 $\pm$ 0.022 & 0.480 $\pm$ 0.025 & 0.218 $\pm$ 0.007 & 0.302 $\pm$ 0.012 & 0.317 $\pm$ 0.022 & 0.381 $\pm$ 0.033 & 0.590 $\pm$ 0.032 & 0.729 $\pm$ 0.069 \\
    $\alpha=1.0$ & 0.333 $\pm$ 0.012 & 0.435 $\pm$ 0.044 & 0.545 $\pm$ 0.041 & 0.222 $\pm$ 0.007 & 0.309 $\pm$ 0.023 & 0.320 $\pm$ 0.021 & 0.358 $\pm$ 0.012 & 0.582 $\pm$ 0.013 & 0.683 $\pm$ 0.011 \\
    \hline
    \end{tabular}
    }
\end{table}

Table~\ref{tab:robustness-noniid-b32-init-final} shows that client heterogeneity changes leakage, but the direction and size of the effect are dataset dependent. At the initialized attack point, reconstruction is still primarily governed by architecture and aggregation: FT-Transformer remains less vulnerable than the one-hot baselines on Adult and the private multiclass benchmark, while California Housing shows a smaller separation between model families. The effect of $\alpha$ is modest at this stage. At the final attack point, Adult is the clearest case in which the split closest to IID leaks more across all three model families. For example, the small MLP increases from 0.446 at $\alpha=0.4$ to 0.507 at $\alpha=1.0$, and FT-Transformer increases from 0.275 to 0.307. The private multiclass benchmark remains comparatively stable for FT-Transformer and the small MLP, while ResNet leaks more at $\alpha=1.0$ than under the more heterogeneous splits. California Housing does not show a consistent monotonic relationship with $\alpha$. Overall, client heterogeneity modifies the client gradient or model update through class, feature, or target bucket skew, but architecture and aggregation remain more stable predictors of reconstruction than the Dirichlet concentration alone. The corresponding batch size 8 results are reported in Table~\ref{tab:appendix-robustness-noniid-b8-init-final}.

\begin{table}[H]
    \centering
    \caption{Reconstruction accuracy under non-IID client partitions at batch size 32. Each cell reports mean $\pm$ standard deviation across 3 seeds.}
    \label{tab:robustness-noniid-b32-init-final}
    \resizebox{\textwidth}{!}{%
    \begin{tabular}{lccccccccc}
    \hline
     & \multicolumn{3}{c}{\textbf{Adult}} & \multicolumn{3}{c}{\textbf{Private multiclass}} & \multicolumn{3}{c}{\textbf{California Housing}} \\
    $\alpha$ & \textbf{FT-Transformer} & \textbf{ResNet} & \textbf{Small MLP} & \textbf{FT-Transformer} & \textbf{ResNet} & \textbf{Small MLP} & \textbf{FT-Transformer} & \textbf{ResNet} & \textbf{Small MLP} \\
    \hline
    \multicolumn{10}{c}{Initialized attack point} \\
    \hline
    $\alpha=0.4$ & 0.427 $\pm$ 0.009 & 0.649 $\pm$ 0.007 & 0.491 $\pm$ 0.021 & 0.186 $\pm$ 0.013 & 0.461 $\pm$ 0.012 & 0.336 $\pm$ 0.006 & 0.449 $\pm$ 0.037 & 0.526 $\pm$ 0.024 & 0.658 $\pm$ 0.042 \\
    $\alpha=0.6$ & 0.422 $\pm$ 0.017 & 0.640 $\pm$ 0.030 & 0.494 $\pm$ 0.023 & 0.189 $\pm$ 0.013 & 0.466 $\pm$ 0.004 & 0.340 $\pm$ 0.004 & 0.450 $\pm$ 0.030 & 0.538 $\pm$ 0.008 & 0.646 $\pm$ 0.045 \\
    $\alpha=1.0$ & 0.423 $\pm$ 0.016 & 0.635 $\pm$ 0.015 & 0.491 $\pm$ 0.014 & 0.174 $\pm$ 0.011 & 0.470 $\pm$ 0.002 & 0.332 $\pm$ 0.003 & 0.455 $\pm$ 0.038 & 0.534 $\pm$ 0.004 & 0.624 $\pm$ 0.017 \\
    \hline
    \multicolumn{10}{c}{Final attack point} \\
    \hline
    $\alpha=0.4$ & 0.275 $\pm$ 0.026 & 0.455 $\pm$ 0.051 & 0.446 $\pm$ 0.051 & 0.187 $\pm$ 0.002 & 0.366 $\pm$ 0.018 & 0.310 $\pm$ 0.006 & 0.351 $\pm$ 0.013 & 0.482 $\pm$ 0.014 & 0.592 $\pm$ 0.010 \\
    $\alpha=0.6$ & 0.292 $\pm$ 0.023 & 0.475 $\pm$ 0.013 & 0.455 $\pm$ 0.067 & 0.199 $\pm$ 0.003 & 0.354 $\pm$ 0.044 & 0.306 $\pm$ 0.008 & 0.335 $\pm$ 0.017 & 0.482 $\pm$ 0.017 & 0.591 $\pm$ 0.045 \\
    $\alpha=1.0$ & 0.307 $\pm$ 0.039 & 0.478 $\pm$ 0.019 & 0.507 $\pm$ 0.029 & 0.196 $\pm$ 0.005 & 0.396 $\pm$ 0.004 & 0.320 $\pm$ 0.012 & 0.358 $\pm$ 0.036 & 0.472 $\pm$ 0.021 & 0.577 $\pm$ 0.008 \\
    \hline
    \end{tabular}
    }
\end{table}

Figures~\ref{fig:appendix-benchmark-robustness-noniid-adult}--\ref{fig:appendix-benchmark-robustness-noniid-california}
report the non-IID robustness analysis on the benchmark datasets. The curves compare FedSGD gradient inversion
reconstruction accuracy across Dirichlet client partition settings and client batch sizes for each evaluated model
family. The client marginal prior and uniform random baselines provide reference leakage levels for distribution aware and
distribution free reconstruction, respectively.

\subsection{Attack optimization budget}
\label{app:benchmark-attack-budget}
\begin{figure}[H]
    \centering
    \includegraphics[
        width=0.95\textwidth,
        height=0.78\textheight,
        keepaspectratio
    ]{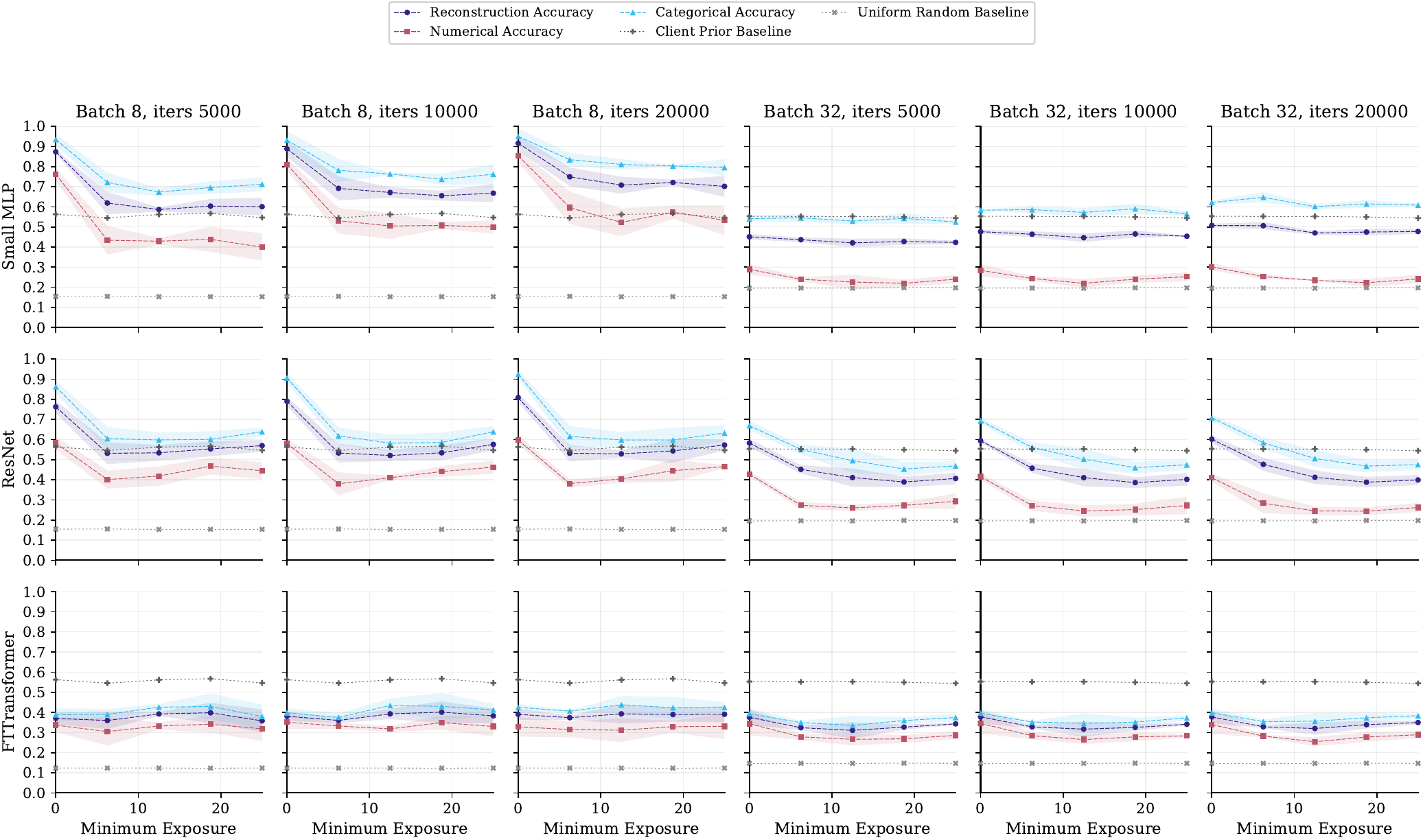}
    \caption{Attack budget robustness analysis on Adult.}
    \label{fig:appendix-benchmark-robustness-attack-budget-adult}
\end{figure}

\begin{figure}[H]
    \centering
    \includegraphics[
        width=0.95\textwidth,
        height=0.78\textheight,
        keepaspectratio
    ]{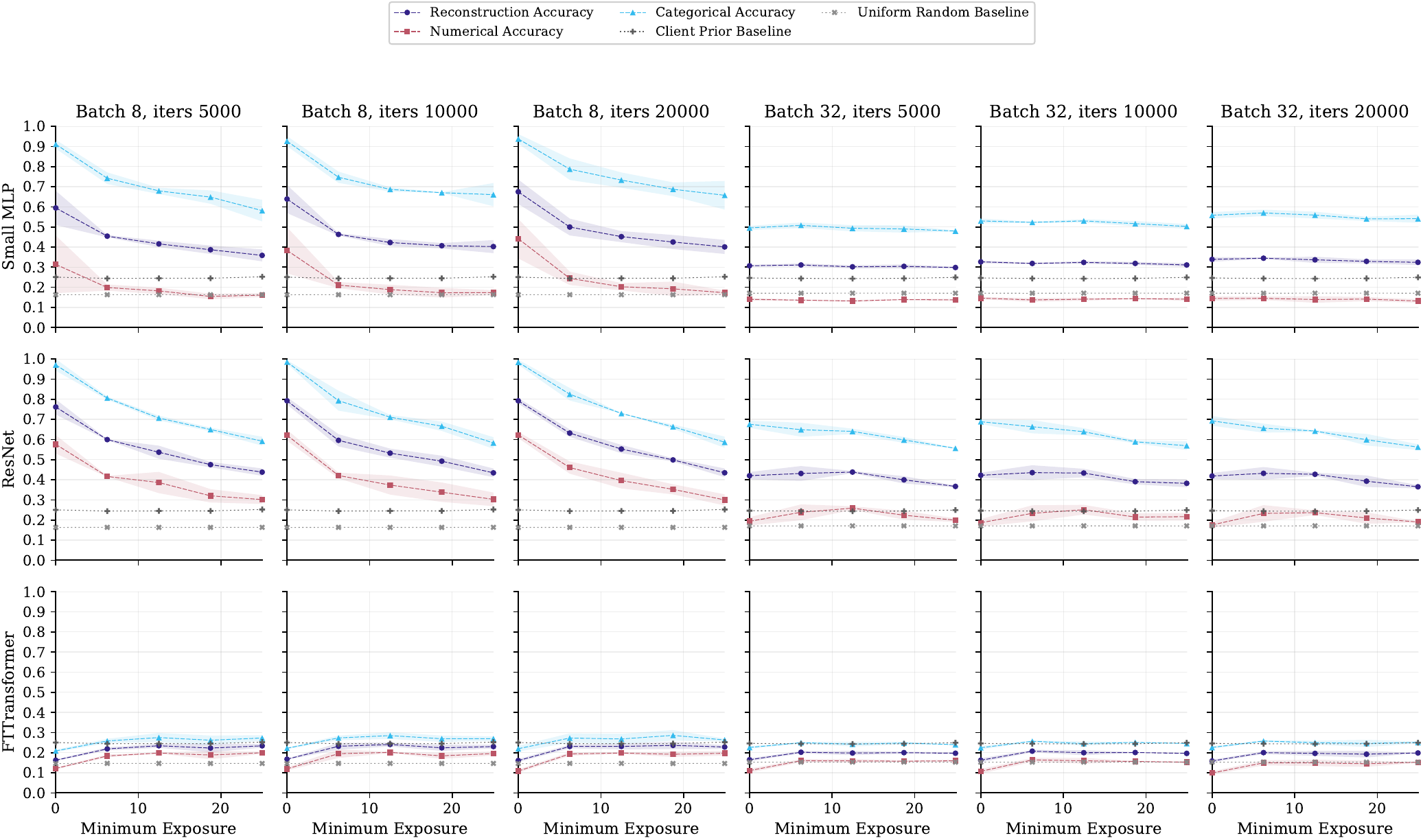}
    \caption{Attack budget robustness analysis on the private multiclass benchmark.}
    \label{fig:appendix-benchmark-robustness-attack-budget-private-multiclass}
\end{figure}

\begin{figure}[H]
    \centering
    \includegraphics[
        width=0.95\textwidth,
        height=0.78\textheight,
        keepaspectratio
    ]{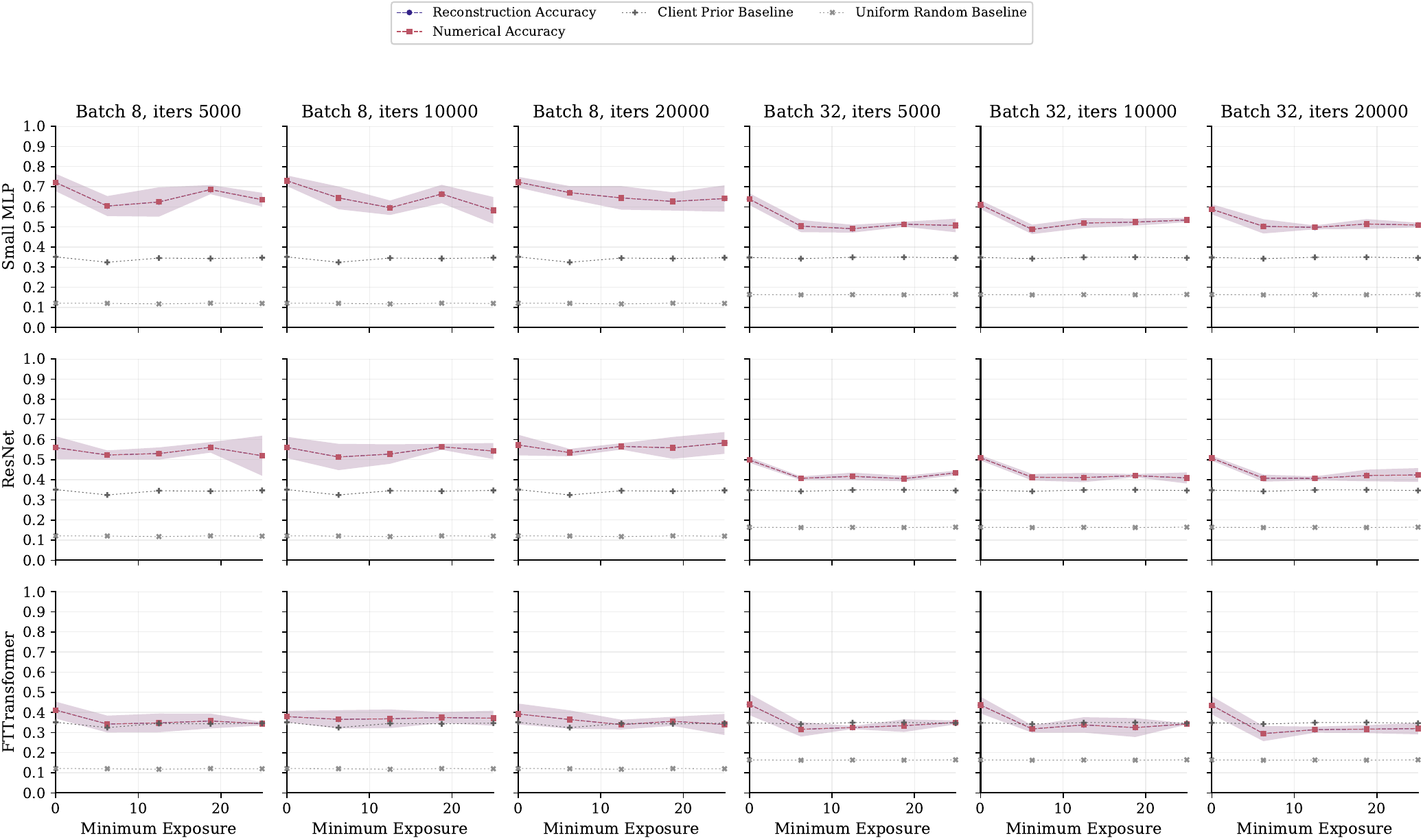}
    \caption{Attack budget robustness analysis on California Housing.}
    \label{fig:appendix-benchmark-robustness-attack-budget-california}
\end{figure}

Table~\ref{tab:appendix-robustness-attack-budget-b8-init-final} and
Figures~\ref{fig:appendix-benchmark-robustness-attack-budget-adult}--\ref{fig:appendix-benchmark-robustness-attack-budget-california}
evaluate whether benchmark reconstruction depends strongly on the number of attack optimization iterations. The FL configuration is held fixed while the offline attack budget is varied. These controls test whether the reported attack point conclusions are artifacts of an insufficiently optimized attack.

\begin{table}[H]
    \centering
    \caption{Reconstruction accuracy under attack budget variation at batch size 8. Each cell reports mean $\pm$ standard deviation across 3 seeds.}
    \label{tab:appendix-robustness-attack-budget-b8-init-final}
    \resizebox{\textwidth}{!}{%
    \begin{tabular}{lccccccccc}
    \hline
     & \multicolumn{3}{c}{\textbf{Adult}} & \multicolumn{3}{c}{\textbf{Private multiclass}} & \multicolumn{3}{c}{\textbf{California Housing}} \\
    \textbf{Attack iterations} & \textbf{FT-Transformer} & \textbf{ResNet} & \textbf{Small MLP} & \textbf{FT-Transformer} & \textbf{ResNet} & \textbf{Small MLP} & \textbf{FT-Transformer} & \textbf{ResNet} & \textbf{Small MLP} \\
    \hline
    \multicolumn{10}{c}{Initialized attack point} \\
    \hline
    5{,}000  & 0.370 $\pm$ 0.026 & 0.762 $\pm$ 0.027 & 0.873 $\pm$ 0.007 & 0.162 $\pm$ 0.004 & 0.761 $\pm$ 0.033 & 0.594 $\pm$ 0.075 & 0.411 $\pm$ 0.037 & 0.559 $\pm$ 0.050 & 0.721 $\pm$ 0.038 \\
    10{,}000 & 0.381 $\pm$ 0.021 & 0.790 $\pm$ 0.020 & 0.888 $\pm$ 0.034 & 0.168 $\pm$ 0.002 & 0.792 $\pm$ 0.016 & 0.639 $\pm$ 0.061 & 0.379 $\pm$ 0.025 & 0.560 $\pm$ 0.047 & 0.729 $\pm$ 0.023 \\
    20{,}000 & 0.390 $\pm$ 0.024 & 0.808 $\pm$ 0.020 & 0.915 $\pm$ 0.032 & 0.161 $\pm$ 0.007 & 0.792 $\pm$ 0.012 & 0.675 $\pm$ 0.054 & 0.392 $\pm$ 0.046 & 0.572 $\pm$ 0.045 & 0.722 $\pm$ 0.024 \\
    \hline
    \multicolumn{10}{c}{Final attack point} \\
    \hline
    5{,}000  & 0.358 $\pm$ 0.050 & 0.569 $\pm$ 0.025 & 0.601 $\pm$ 0.039 & 0.234 $\pm$ 0.008 & 0.438 $\pm$ 0.016 & 0.359 $\pm$ 0.026 & 0.344 $\pm$ 0.008 & 0.519 $\pm$ 0.088 & 0.635 $\pm$ 0.030 \\
    10{,}000 & 0.383 $\pm$ 0.032 & 0.576 $\pm$ 0.026 & 0.668 $\pm$ 0.039 & 0.230 $\pm$ 0.008 & 0.434 $\pm$ 0.019 & 0.402 $\pm$ 0.029 & 0.371 $\pm$ 0.033 & 0.543 $\pm$ 0.035 & 0.582 $\pm$ 0.059 \\
    20{,}000 & 0.391 $\pm$ 0.035 & 0.573 $\pm$ 0.024 & 0.701 $\pm$ 0.046 & 0.228 $\pm$ 0.008 & 0.435 $\pm$ 0.019 & 0.401 $\pm$ 0.031 & 0.340 $\pm$ 0.046 & 0.583 $\pm$ 0.048 & 0.641 $\pm$ 0.058 \\
    \hline
    \end{tabular}
    }
\end{table}

Table~\ref{tab:robustness-attack-budget-b32-init-final} evaluates whether the benchmark conclusions depend on the number of attack optimization iterations. At batch size 32, increasing the attack budget from 5{,}000 to 20{,}000 iterations produces only moderate and model dependent changes. The clearest increase is observed for the small MLP on Adult, where final attack point reconstruction increases from 0.423 to 0.477. FT-Transformer is comparatively stable on Adult and the private multiclass benchmark, while California Housing is not monotonic with respect to the attack budget. The same batch size 8 sensitivity is reported in Table~\ref{tab:appendix-robustness-attack-budget-b8-init-final}.

\begin{table}[H]
    \centering
    \small
    \caption{Reconstruction accuracy under attack budget variation at batch size 32. Each cell reports mean $\pm$ standard deviation across 3 seeds.}
    \label{tab:robustness-attack-budget-b32-init-final}
    \resizebox{\textwidth}{!}{%
    \begin{tabular}{lccccccccc}
    \hline
     & \multicolumn{3}{c}{\textbf{Adult}} & \multicolumn{3}{c}{\textbf{Private multiclass}} & \multicolumn{3}{c}{\textbf{California Housing}} \\
    \textbf{Attack iterations} & \textbf{FT-Transformer} & \textbf{ResNet} & \textbf{Small MLP} & \textbf{FT-Transformer} & \textbf{ResNet} & \textbf{Small MLP} & \textbf{FT-Transformer} & \textbf{ResNet} & \textbf{Small MLP} \\
    \hline
    \multicolumn{10}{c}{Initialized attack point} \\
    \hline
    5{,}000  & 0.376 $\pm$ 0.033 & 0.583 $\pm$ 0.018 & 0.451 $\pm$ 0.012 & 0.165 $\pm$ 0.007 & 0.420 $\pm$ 0.016 & 0.307 $\pm$ 0.004 & 0.440 $\pm$ 0.047 & 0.498 $\pm$ 0.012 & 0.638 $\pm$ 0.026 \\
    10{,}000 & 0.378 $\pm$ 0.028 & 0.594 $\pm$ 0.007 & 0.477 $\pm$ 0.009 & 0.162 $\pm$ 0.009 & 0.422 $\pm$ 0.011 & 0.326 $\pm$ 0.006 & 0.436 $\pm$ 0.037 & 0.509 $\pm$ 0.011 & 0.610 $\pm$ 0.019 \\
    20{,}000 & 0.377 $\pm$ 0.028 & 0.601 $\pm$ 0.015 & 0.507 $\pm$ 0.008 & 0.159 $\pm$ 0.006 & 0.418 $\pm$ 0.014 & 0.339 $\pm$ 0.010 & 0.435 $\pm$ 0.040 & 0.507 $\pm$ 0.011 & 0.588 $\pm$ 0.022 \\
    \hline
    \multicolumn{10}{c}{Final attack point} \\
    \hline
    5{,}000  & 0.343 $\pm$ 0.005 & 0.406 $\pm$ 0.025 & 0.423 $\pm$ 0.008 & 0.197 $\pm$ 0.004 & 0.367 $\pm$ 0.004 & 0.298 $\pm$ 0.004 & 0.350 $\pm$ 0.008 & 0.434 $\pm$ 0.010 & 0.507 $\pm$ 0.030 \\
    10{,}000 & 0.341 $\pm$ 0.005 & 0.402 $\pm$ 0.028 & 0.454 $\pm$ 0.002 & 0.196 $\pm$ 0.002 & 0.382 $\pm$ 0.017 & 0.311 $\pm$ 0.008 & 0.343 $\pm$ 0.003 & 0.409 $\pm$ 0.024 & 0.534 $\pm$ 0.009 \\
    20{,}000 & 0.350 $\pm$ 0.008 & 0.399 $\pm$ 0.022 & 0.477 $\pm$ 0.011 & 0.199 $\pm$ 0.001 & 0.365 $\pm$ 0.007 & 0.324 $\pm$ 0.012 & 0.320 $\pm$ 0.025 & 0.424 $\pm$ 0.030 & 0.510 $\pm$ 0.010 \\
    \hline
    \end{tabular}
    }
\end{table}

Overall, the attack budget sensitivity does not indicate that the reported benchmark trends are an artifact of a
single optimization budget. Larger budgets can improve some reconstructions, but they do not consistently increase leakage across datasets and architectures, nor do they change the broader ordering of model families observed in the benchmark experiments. The attack budget should therefore be interpreted as a robustness check rather than as the primary privacy lever.

Figures~\ref{fig:appendix-benchmark-robustness-attack-budget-adult}--\ref{fig:appendix-benchmark-robustness-attack-budget-california} evaluate whether benchmark reconstruction depends strongly on the number of attack optimization iterations. The FL configuration is held fixed while the offline attack budget is varied. These controls test whether the reported attack point conclusions are artifacts of an insufficiently optimized attack.

\subsection{Leakage without true batch labels}
\label{app:benchmark-label-unknown}
\begin{figure}[H]
    \centering
    \begin{subfigure}[t]{0.98\textwidth}
        \centering
        \includegraphics[width=\textwidth]{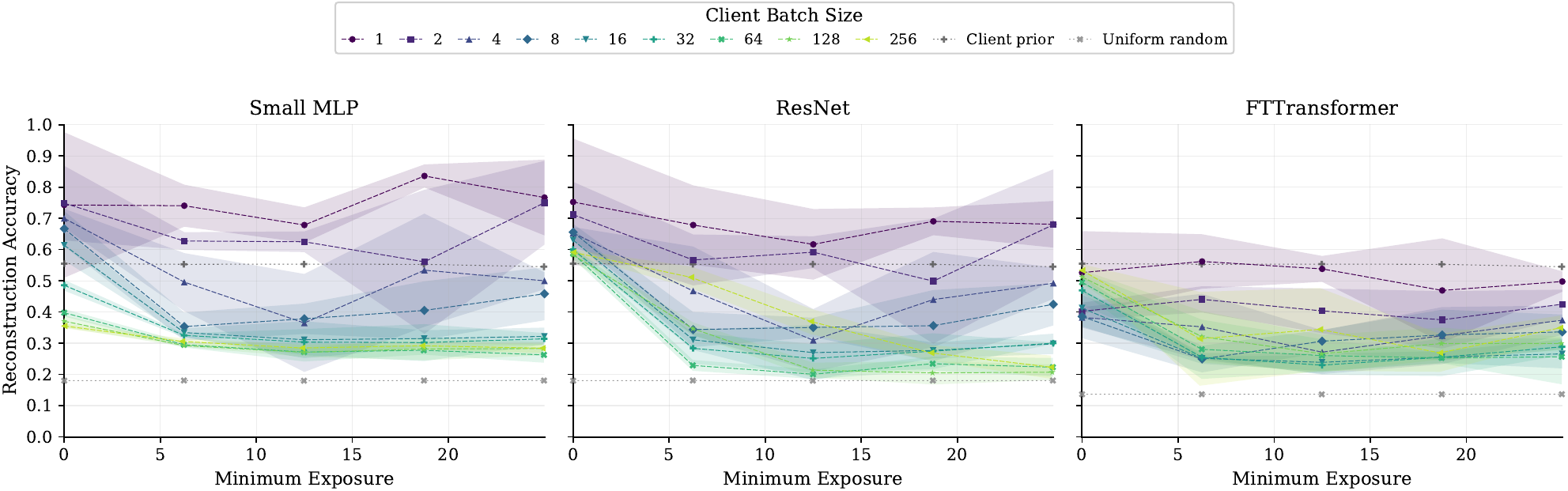}
        \caption{Adult}
    \end{subfigure}

    \vspace{0.75em}

    \begin{subfigure}[t]{0.98\textwidth}
        \centering
        \includegraphics[width=\textwidth]{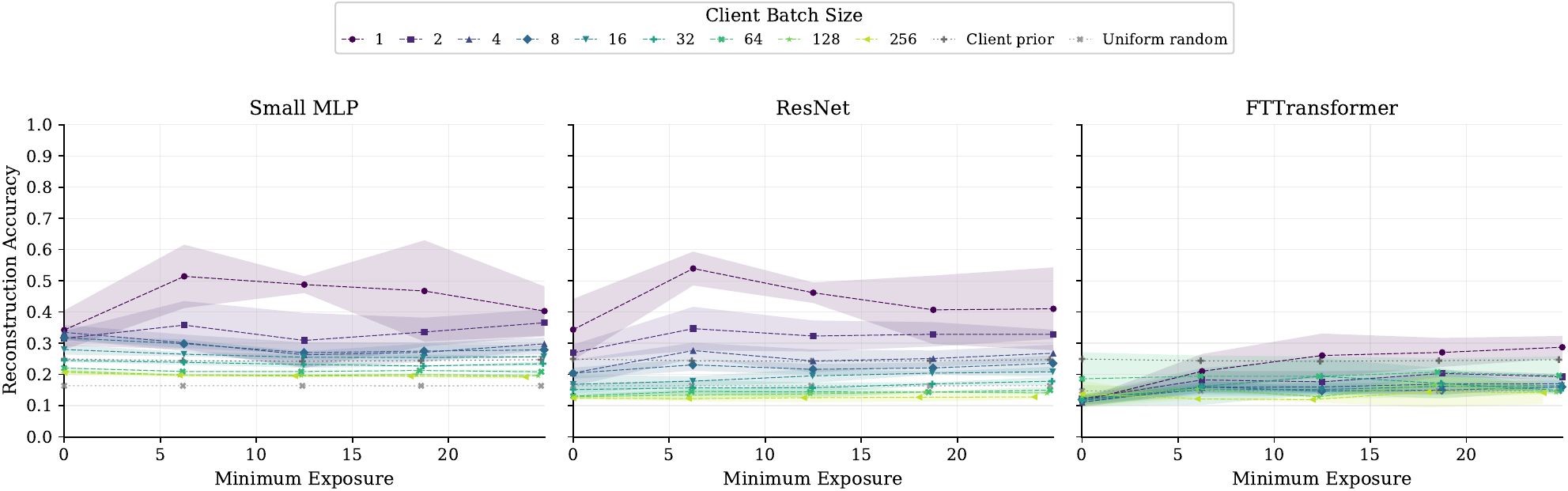}
        \caption{Private multiclass}
    \end{subfigure}

    \vspace{0.75em}

    \begin{subfigure}[t]{0.98\textwidth}
        \centering
        \includegraphics[width=\textwidth]{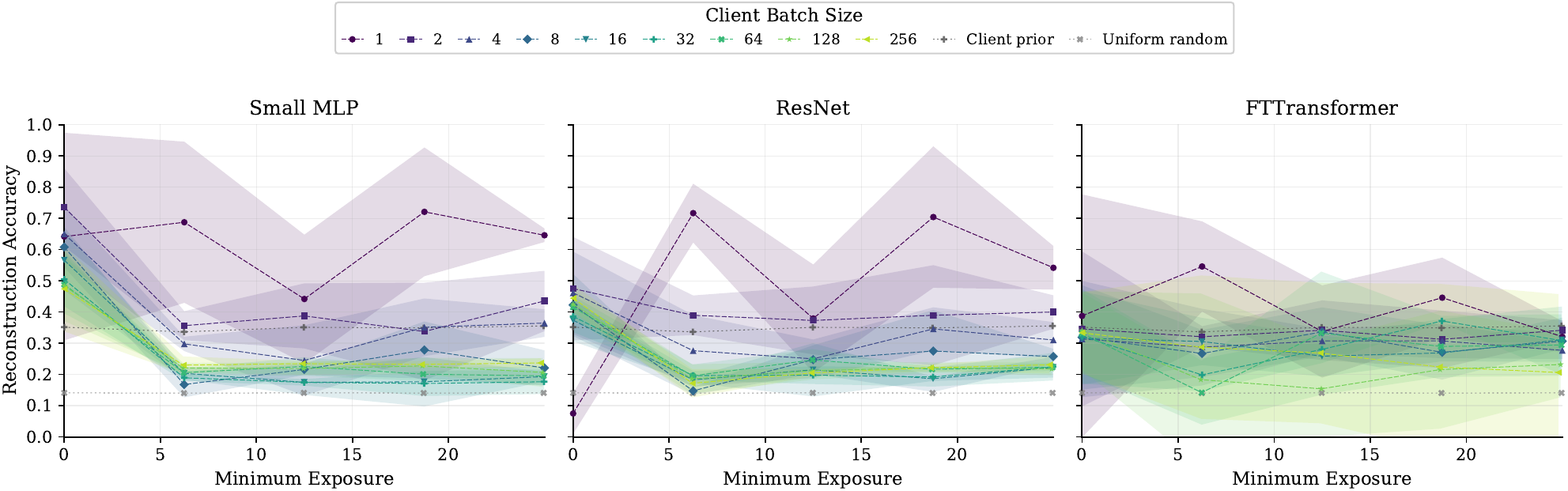}
        \caption{California Housing}
    \end{subfigure}
    \caption{Label unknown robustness analysis on the benchmark datasets. The curves report reconstruction accuracy over exposure when the attacker is not given the true labels or regression targets. Comparing these trajectories with the client marginal prior and uniform random baselines shows how much feature information remains in the observed update under the weaker target knowledge assumption.}
    \label{fig:appendix-benchmark-robustness-label-unknown}
\end{figure}

Table~\ref{tab:robustness-label-unknown-final} and Figure~\ref{fig:appendix-benchmark-robustness-label-unknown} report benchmark attacks without access to the true batch labels or regression targets. This setting removes target information from the attacker while preserving the same observed client updates and model families. The comparison tests whether the benchmark conclusions rely on the stronger label known assumption used by default.

The label unknown benchmark final attack point results are reported in Table~\ref{tab:robustness-label-unknown-final}, with label unknown trajectories shown in Figure~\ref{fig:appendix-benchmark-robustness-label-unknown}; the corresponding label known California Housing trajectories are shown in Figure~\ref{fig:cali-all-models-all-batchsize-trajectories}. These results generally support the batch size conclusion under the weaker attacker assumption, but California Housing reveals a regression specific label knowledge effect rather than a simple reduction in attacker strength. Under label known attacks, batch size \(1\) reconstruction drops during training for all three model families, from \(0.721\) to \(0.304\) for FT-Transformer, from \(0.992\) to \(0.371\) for ResNet, and from \(0.967\) to \(0.463\) for the small MLP. In the label unknown trajectories, the direction reverses, with initially weak reconstruction recovering during training, most clearly for ResNet. This suggests that regression targets interact with the gradient matching objective through the MSE residual, so label knowledge should be interpreted more carefully for regression than for classification.

\begin{table}[H]
    \centering
    \small
    \caption{Reconstruction accuracy at the final attack point for label unknown FedSGD attacks on the benchmark datasets. The attacker does not receive the true labels. Each cell reports mean $\pm$ standard deviation across 3 seeds.}
    \label{tab:robustness-label-unknown-final}
    \resizebox{\textwidth}{!}{%
    \begin{tabular}{lcccccc}
    \hline
     & \multicolumn{2}{c}{\textbf{Adult}} & \multicolumn{2}{c}{\textbf{Private multiclass}} & \multicolumn{2}{c}{\textbf{California Housing}} \\
    \textbf{Batch size} & \textbf{ResNet} & \textbf{Small MLP} & \textbf{ResNet} & \textbf{Small MLP} & \textbf{ResNet} & \textbf{Small MLP} \\
    \hline
    1   & 0.684 $\pm$ 0.058 & 0.753 $\pm$ 0.073 & 0.432 $\pm$ 0.031 & 0.443 $\pm$ 0.071 & 0.483 $\pm$ 0.044 & 0.628 $\pm$ 0.128 \\
    2   & 0.610 $\pm$ 0.015 & 0.662 $\pm$ 0.019 & 0.319 $\pm$ 0.005 & 0.337 $\pm$ 0.009 & 0.405 $\pm$ 0.039 & 0.450 $\pm$ 0.023 \\
    4   & 0.472 $\pm$ 0.028 & 0.519 $\pm$ 0.001 & 0.249 $\pm$ 0.016 & 0.294 $\pm$ 0.015 & 0.329 $\pm$ 0.035 & 0.382 $\pm$ 0.029 \\
    8   & 0.426 $\pm$ 0.015 & 0.452 $\pm$ 0.011 & 0.222 $\pm$ 0.003 & 0.288 $\pm$ 0.009 & 0.270 $\pm$ 0.015 & 0.298 $\pm$ 0.011 \\
    16  & 0.357 $\pm$ 0.027 & 0.379 $\pm$ 0.013 & 0.191 $\pm$ 0.010 & 0.262 $\pm$ 0.002 & 0.236 $\pm$ 0.023 & 0.263 $\pm$ 0.011 \\
    32  & 0.342 $\pm$ 0.005 & 0.346 $\pm$ 0.009 & 0.163 $\pm$ 0.002 & 0.235 $\pm$ 0.004 & 0.239 $\pm$ 0.015 & 0.243 $\pm$ 0.025 \\
    64  & 0.295 $\pm$ 0.007 & 0.301 $\pm$ 0.004 & 0.143 $\pm$ 0.002 & 0.212 $\pm$ 0.002 & 0.259 $\pm$ 0.005 & 0.263 $\pm$ 0.022 \\
    128 & 0.311 $\pm$ 0.012 & 0.300 $\pm$ 0.010 & 0.137 $\pm$ 0.006 & 0.200 $\pm$ 0.001 & 0.252 $\pm$ 0.008 & 0.271 $\pm$ 0.016 \\
    256 & 0.393 $\pm$ 0.023 & 0.304 $\pm$ 0.003 & 0.125 $\pm$ 0.007 & 0.197 $\pm$ 0.004 & 0.255 $\pm$ 0.010 & 0.281 $\pm$ 0.016 \\
    \hline
    \end{tabular}
    }
\end{table}

The final attack point table shows that the main batch size conclusion persists without true labels or targets. Averaged over the two one-hot models, Adult drops from \(0.719\) at batch size \(1\) to \(0.344\) at batch size \(32\), California Housing drops from \(0.556\) to \(0.241\), and the private multiclass benchmark drops from \(0.438\) to \(0.199\). Thus, the central conclusion is not conditional on the strongest label known attacker. Even when labels or targets are unavailable, small client batches remain the most exposed regime.

Figure~\ref{fig:appendix-benchmark-robustness-label-unknown} shows the full label unknown trajectories. The California Housing curves reveal a regression specific label knowledge effect that is not visible from final attack point averages alone. In the label known batch size experiment, batch size \(1\) reconstruction drops during training for all model families, despite being the least aggregated setting. In the label unknown trajectories, reconstruction instead starts low and recovers during training, with the clearest reversal appearing for ResNet. This behavior is most pronounced in the continuous target setting. Under MSE, the observed gradient depends directly on the prediction residual, and training changes the scale and informativeness of that residual. Removing the true target therefore changes the residual structure of the simulated gradient rather than only weakening the attacker.

\subsection{Leakage from FedAvg model updates}
\label{app:benchmark-fedavg}
Tables~\ref{tab:fedavg-adult-tableak-acc-initialized}--\ref{tab:fedavg-california-tableak-acc-trained} report benchmark FedAvg reconstruction accuracy with 32 local examples per client. These controls test whether leakage remains measurable when the server observes model deltas rather than one-step gradients. They complement the FedSGD benchmark analysis and the MIMIC-IV FedAvg results.

\begin{table}[H]
  \centering
  \small
  \caption{Reconstruction accuracy at the initialized attack point for Adult. Initialized attack point denotes the attack before any global model aggregation, and final attack point denotes the last attacked point under the exposure budget. Each cell reports mean $\pm$ standard deviation across 3 seeds.}
  \label{tab:fedavg-adult-tableak-acc-initialized}
  \resizebox{\textwidth}{!}{%
  \begin{tabular}{lcccccc}
  \hline
  \multicolumn{1}{l}{} & \multicolumn{3}{c}{\textbf{1 local epoch}} & \multicolumn{3}{c}{\textbf{4 local epochs}} \\
  \textbf{n. batches} & \textbf{FT-Transformer} & \textbf{ResNet} & \textbf{Small MLP} & \textbf{FT-Transformer} & \textbf{ResNet} & \textbf{Small MLP} \\
  \hline
  1 & 0.383 $\pm$ 0.015 & 0.605 $\pm$ 0.023 & 0.492 $\pm$ 0.024 & 0.399 $\pm$ 0.019 & 0.631 $\pm$ 0.019 & 0.470 $\pm$ 0.022 \\
  2 & 0.345 $\pm$ 0.030 & 0.602 $\pm$ 0.022 & 0.479 $\pm$ 0.016 & 0.374 $\pm$ 0.030 & 0.625 $\pm$ 0.012 & 0.474 $\pm$ 0.033 \\
  4 & 0.321 $\pm$ 0.022 & 0.592 $\pm$ 0.018 & 0.492 $\pm$ 0.010 & 0.353 $\pm$ 0.015 & 0.614 $\pm$ 0.027 & 0.481 $\pm$ 0.022 \\
  \hline
  \end{tabular}
  }
\end{table}

\begin{table}[H]
  \centering
  \small
  \caption{Reconstruction accuracy at the final attack point for Adult. Initialized attack point denotes the attack before any global model aggregation, and final attack point denotes the last attacked point under the exposure budget. Each cell reports mean $\pm$ standard deviation across 3 seeds.}
  \label{tab:fedavg-adult-tableak-acc-trained}
  \resizebox{\textwidth}{!}{%
  \begin{tabular}{lcccccc}
  \hline
  \multicolumn{1}{l}{} & \multicolumn{3}{c}{\textbf{1 local epoch}} & \multicolumn{3}{c}{\textbf{4 local epochs}} \\
  \textbf{n. batches} & \textbf{FT-Transformer} & \textbf{ResNet} & \textbf{Small MLP} & \textbf{FT-Transformer} & \textbf{ResNet} & \textbf{Small MLP} \\
  \hline
  1 & 0.325 $\pm$ 0.013 & 0.585 $\pm$ 0.026 & 0.492 $\pm$ 0.016 & 0.356 $\pm$ 0.028 & 0.615 $\pm$ 0.022 & 0.489 $\pm$ 0.018 \\
  2 & 0.316 $\pm$ 0.015 & 0.588 $\pm$ 0.032 & 0.486 $\pm$ 0.021 & 0.360 $\pm$ 0.017 & 0.618 $\pm$ 0.046 & 0.460 $\pm$ 0.022 \\
  4 & 0.310 $\pm$ 0.031 & 0.564 $\pm$ 0.047 & 0.480 $\pm$ 0.027 & 0.351 $\pm$ 0.016 & 0.616 $\pm$ 0.056 & 0.463 $\pm$ 0.029 \\
  \hline
  \end{tabular}
  }
\end{table}

\begin{table}[H]
  \centering
  \small
  \caption{Reconstruction accuracy at the initialized attack point for the private multiclass benchmark. Initialized attack point denotes the attack before any global model aggregation, and final attack point denotes the last attacked point under the exposure budget. Each cell reports mean $\pm$ standard deviation across 3 seeds.}
  \label{tab:fedavg-pandemic-tableak-acc-initialized}
  \resizebox{\textwidth}{!}{%
  \begin{tabular}{lcccccc}
  \hline
  \multicolumn{1}{l}{} & \multicolumn{3}{c}{\textbf{1 local epoch}} & \multicolumn{3}{c}{\textbf{4 local epochs}} \\
  \textbf{n. batches} & \textbf{FT-Transformer} & \textbf{ResNet} & \textbf{Small MLP} & \textbf{FT-Transformer} & \textbf{ResNet} & \textbf{Small MLP} \\
  \hline
  1 & 0.156 $\pm$ 0.007 & 0.436 $\pm$ 0.022 & 0.325 $\pm$ 0.017 & 0.180 $\pm$ 0.004 & 0.454 $\pm$ 0.012 & 0.325 $\pm$ 0.014 \\
  2 & 0.148 $\pm$ 0.008 & 0.417 $\pm$ 0.026 & 0.324 $\pm$ 0.008 & 0.175 $\pm$ 0.008 & 0.439 $\pm$ 0.013 & 0.329 $\pm$ 0.014 \\
  4 & 0.143 $\pm$ 0.010 & 0.417 $\pm$ 0.025 & 0.328 $\pm$ 0.011 & 0.163 $\pm$ 0.008 & 0.427 $\pm$ 0.017 & 0.331 $\pm$ 0.016 \\
  \hline
  \end{tabular}
  }
\end{table}

\begin{table}[H]
  \centering
  \small
  \caption{Reconstruction accuracy at the final attack point for the private multiclass benchmark. Initialized attack point denotes the attack before any global model aggregation, and final attack point denotes the last attacked point under the exposure budget. Each cell reports mean $\pm$ standard deviation across 3 seeds.}
  \label{tab:fedavg-pandemic-tableak-acc-trained}
  \resizebox{\textwidth}{!}{%
  \begin{tabular}{lcccccc}
  \hline
  \multicolumn{1}{l}{} & \multicolumn{3}{c}{\textbf{1 local epoch}} & \multicolumn{3}{c}{\textbf{4 local epochs}} \\
  \textbf{n. batches} & \textbf{FT-Transformer} & \textbf{ResNet} & \textbf{Small MLP} & \textbf{FT-Transformer} & \textbf{ResNet} & \textbf{Small MLP} \\
  \hline
  1 & 0.177 $\pm$ 0.016 & 0.431 $\pm$ 0.025 & 0.315 $\pm$ 0.014 & 0.202 $\pm$ 0.012 & 0.464 $\pm$ 0.032 & 0.310 $\pm$ 0.010 \\
  2 & 0.179 $\pm$ 0.007 & 0.414 $\pm$ 0.027 & 0.311 $\pm$ 0.009 & 0.198 $\pm$ 0.008 & 0.452 $\pm$ 0.026 & 0.297 $\pm$ 0.010 \\
  4 & 0.178 $\pm$ 0.012 & 0.397 $\pm$ 0.046 & 0.309 $\pm$ 0.010 & 0.212 $\pm$ 0.012 & 0.446 $\pm$ 0.022 & 0.307 $\pm$ 0.010 \\
  \hline
  \end{tabular}
  }
\end{table}

\begin{table}[H]
  \centering
  \small
  \caption{Reconstruction accuracy at the initialized attack point for California Housing. Initialized attack point denotes the attack before any global model aggregation, and final attack point denotes the last attacked point under the exposure budget. Each cell reports mean $\pm$ standard deviation across 3 seeds.}
  \label{tab:fedavg-california-tableak-acc-initialized}
  \resizebox{\textwidth}{!}{%
  \begin{tabular}{lcccccc}
  \hline
  \multicolumn{1}{l}{} & \multicolumn{3}{c}{\textbf{1 local epoch}} & \multicolumn{3}{c}{\textbf{4 local epochs}} \\
  \textbf{n. batches} & \textbf{FT-Transformer} & \textbf{ResNet} & \textbf{Small MLP} & \textbf{FT-Transformer} & \textbf{ResNet} & \textbf{Small MLP} \\
  \hline
  1 & 0.381 $\pm$ 0.034 & 0.460 $\pm$ 0.022 & 0.561 $\pm$ 0.043 & 0.394 $\pm$ 0.025 & 0.532 $\pm$ 0.030 & 0.477 $\pm$ 0.044 \\
  2 & 0.271 $\pm$ 0.023 & 0.445 $\pm$ 0.023 & 0.463 $\pm$ 0.039 & 0.285 $\pm$ 0.036 & 0.497 $\pm$ 0.011 & 0.468 $\pm$ 0.051 \\
  4 & 0.232 $\pm$ 0.033 & 0.414 $\pm$ 0.010 & 0.431 $\pm$ 0.048 & 0.230 $\pm$ 0.031 & 0.454 $\pm$ 0.031 & 0.429 $\pm$ 0.037 \\
  \hline
  \end{tabular}
  }
\end{table}

\begin{table}[H]
  \centering
  \small
  \caption{Reconstruction accuracy at the final attack point for California Housing. Initialized attack point denotes the attack before any global model aggregation, and final attack point denotes the last attacked point under the exposure budget. Each cell reports mean $\pm$ standard deviation across 3 seeds.}
  \label{tab:fedavg-california-tableak-acc-trained}
  \resizebox{\textwidth}{!}{%
  \begin{tabular}{lcccccc}
  \hline
  \multicolumn{1}{l}{} & \multicolumn{3}{c}{\textbf{1 local epoch}} & \multicolumn{3}{c}{\textbf{4 local epochs}} \\
  \textbf{n. batches} & \textbf{FT-Transformer} & \textbf{ResNet} & \textbf{Small MLP} & \textbf{FT-Transformer} & \textbf{ResNet} & \textbf{Small MLP} \\
  \hline
  1 & 0.222 $\pm$ 0.037 & 0.421 $\pm$ 0.027 & 0.499 $\pm$ 0.044 & 0.233 $\pm$ 0.047 & 0.471 $\pm$ 0.024 & 0.483 $\pm$ 0.038 \\
  2 & 0.265 $\pm$ 0.052 & 0.388 $\pm$ 0.030 & 0.483 $\pm$ 0.023 & 0.280 $\pm$ 0.016 & 0.436 $\pm$ 0.022 & 0.439 $\pm$ 0.028 \\
  4 & 0.223 $\pm$ 0.033 & 0.368 $\pm$ 0.025 & 0.472 $\pm$ 0.043 & 0.260 $\pm$ 0.027 & 0.420 $\pm$ 0.024 & 0.438 $\pm$ 0.019 \\
  \hline
  \end{tabular}
  }
\end{table}

\subsection{MLP architecture effects on reconstruction}
\label{app:benchmark-mlp-architecture}
\begin{figure}[H]
      \centering

      \begin{subfigure}[t]{0.48\textwidth}
          \centering
          \includegraphics[width=\textwidth]{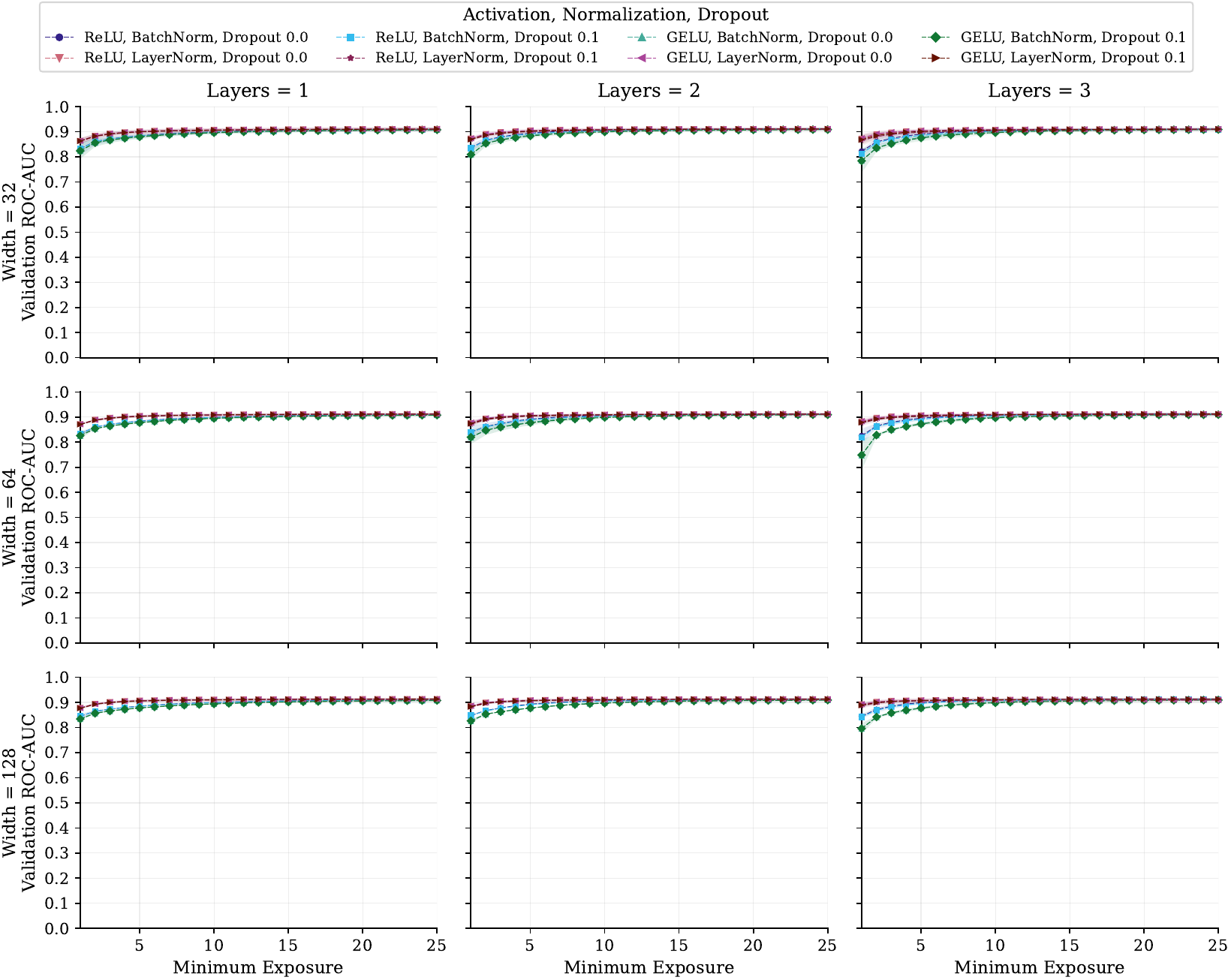}
          \caption{Utility curve ROC-AUC}
      \end{subfigure}
      \hfill
      \begin{subfigure}[t]{0.48\textwidth}
          \centering
          \includegraphics[width=\textwidth]{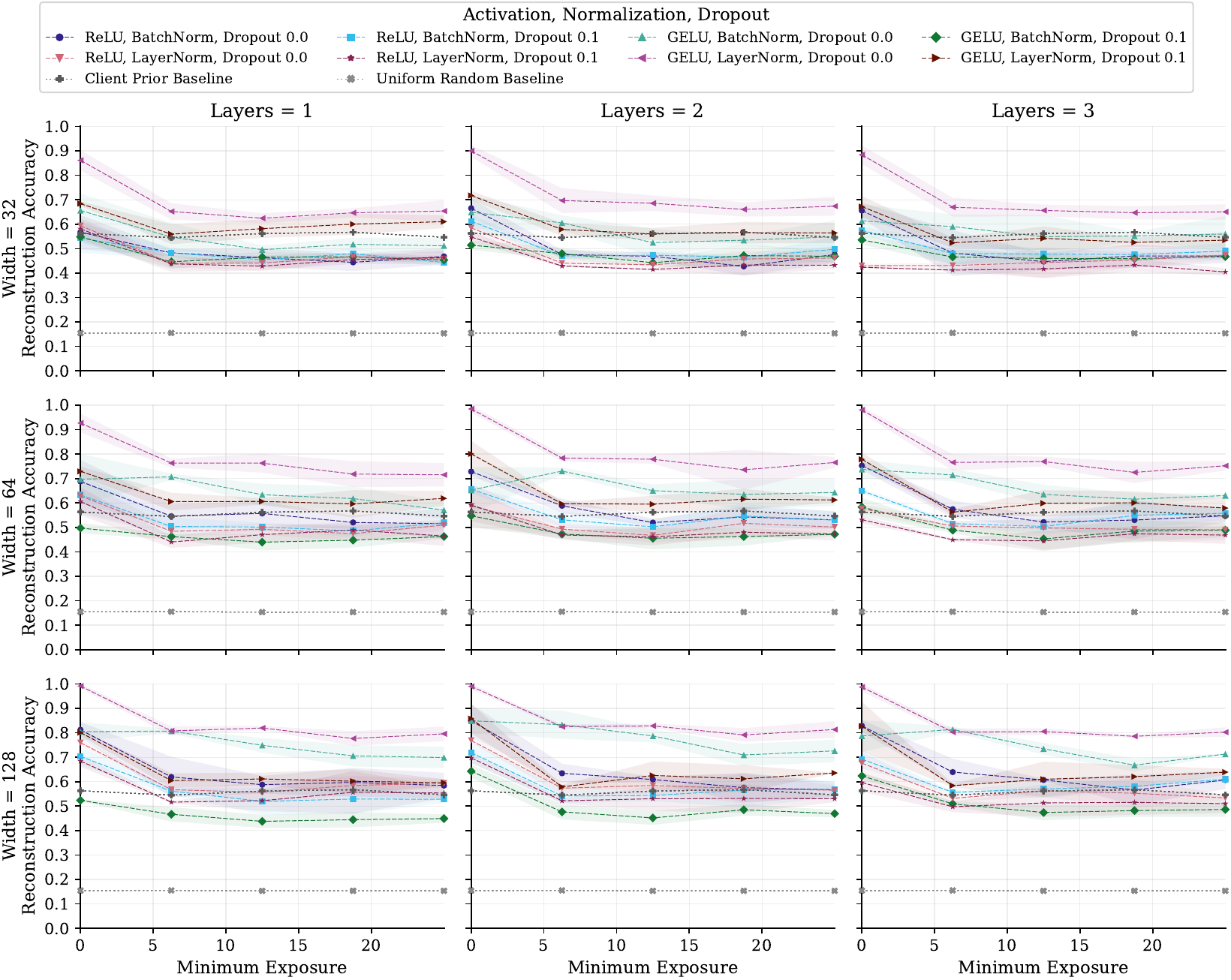}
          \caption{Reconstruction accuracy}
      \end{subfigure}

      \vspace{0.5em}

      \begin{subfigure}[t]{0.48\textwidth}
          \centering
          \includegraphics[width=\textwidth]{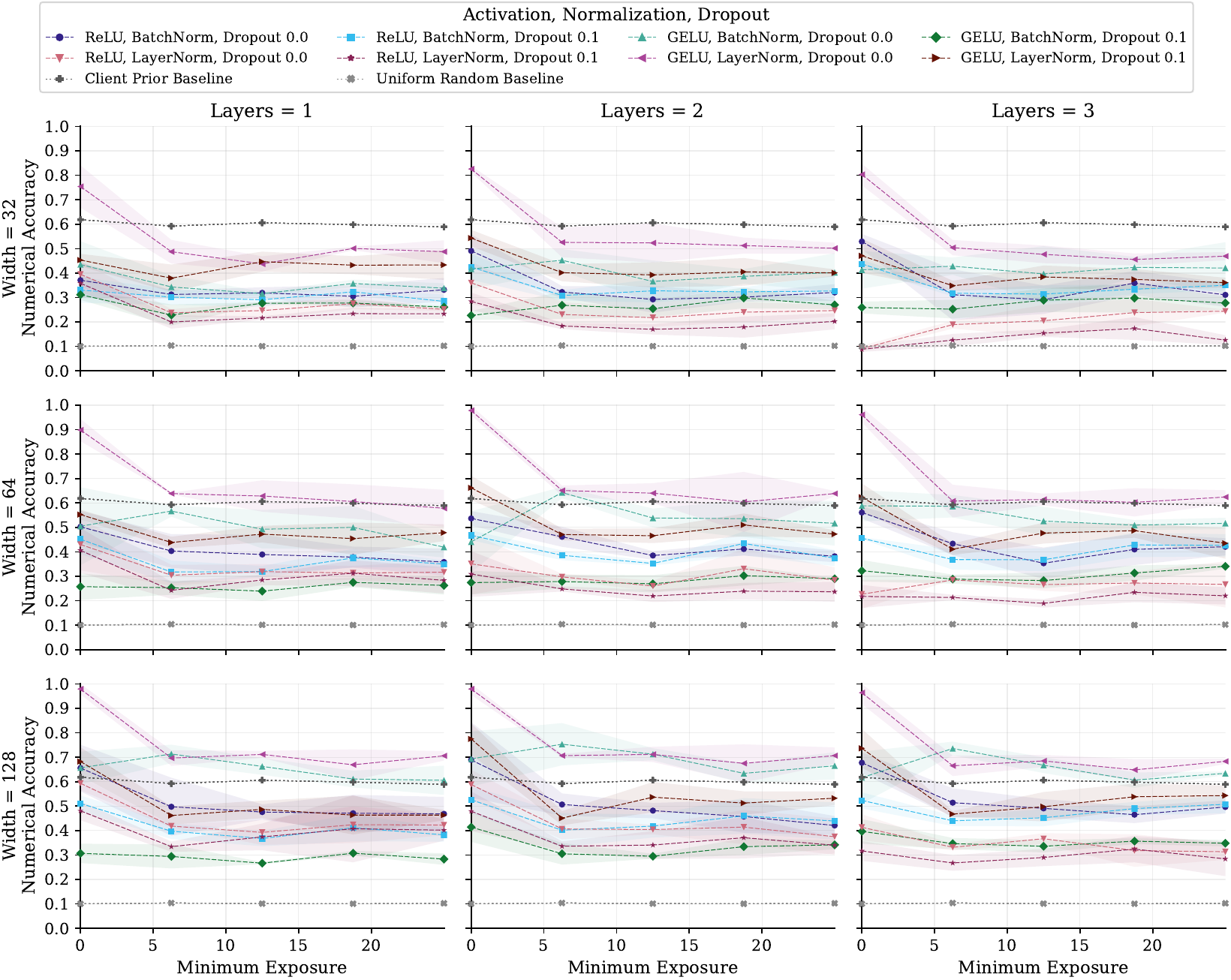}
          \caption{Numerical accuracy}
      \end{subfigure}
      \hfill
      \begin{subfigure}[t]{0.48\textwidth}
          \centering
          \includegraphics[width=\textwidth]{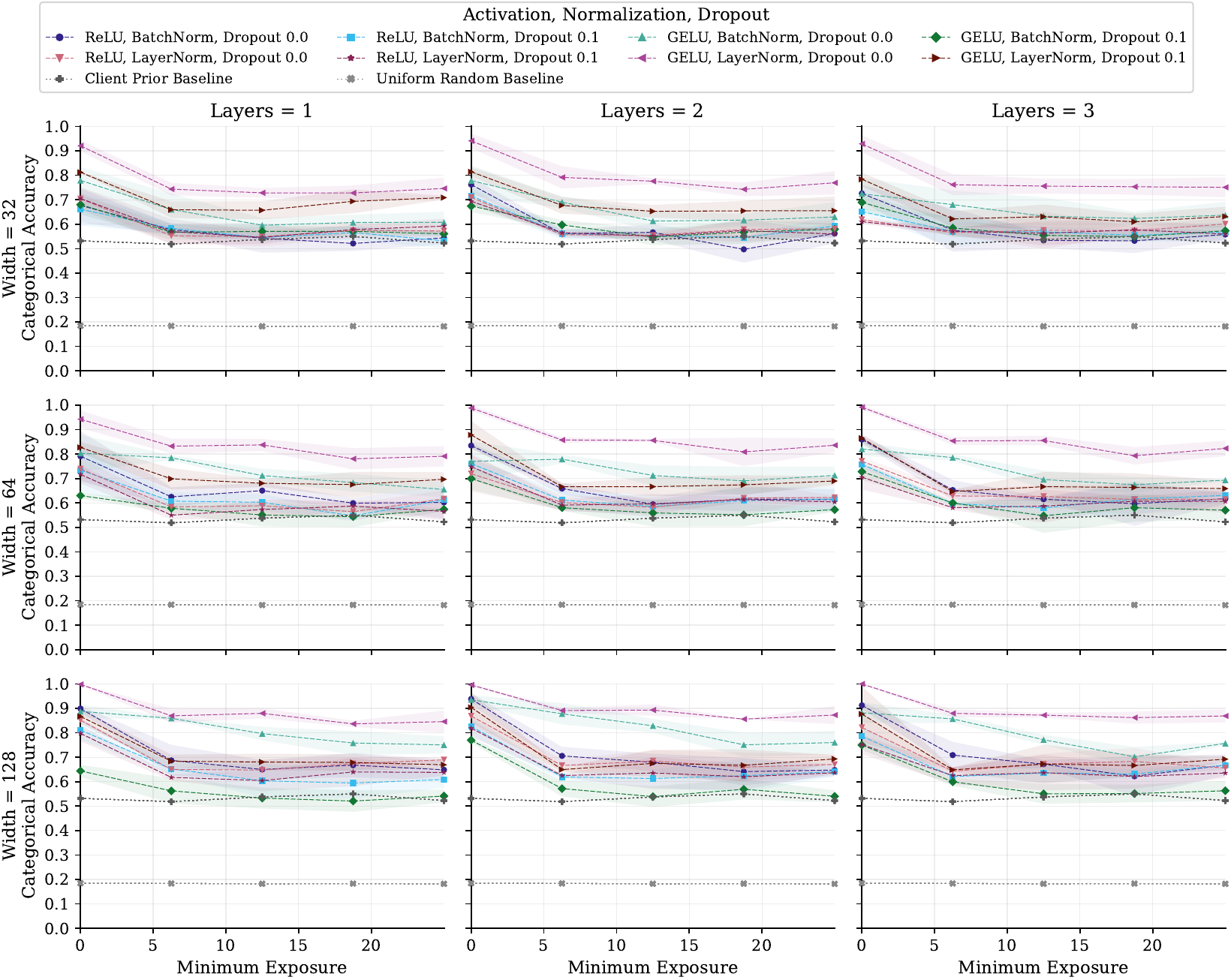}
          \caption{Categorical accuracy}
      \end{subfigure}

      \caption{Utility and inversion accuracy metrics over FL training for the MLP architecture grid using client batch size 8 on Adult.}
      \label{fig:adult-torch-modules-privacy-utility-b8}
\end{figure}

\begin{figure}[H]
      \centering

      \begin{subfigure}[t]{0.48\textwidth}
          \centering
          \includegraphics[width=\textwidth]{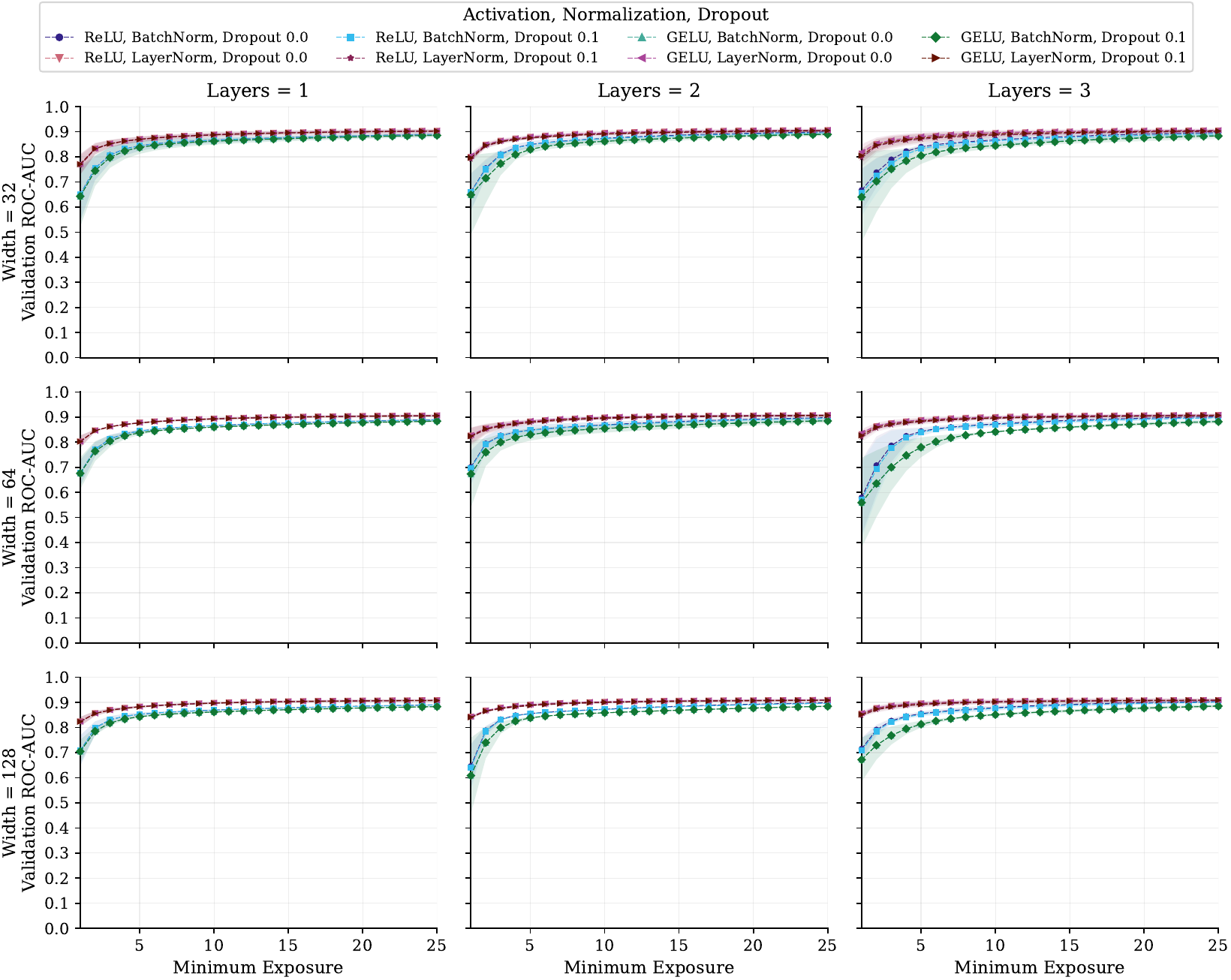}
          \caption{Utility curve ROC-AUC}
      \end{subfigure}
      \hfill
      \begin{subfigure}[t]{0.48\textwidth}
          \centering
          \includegraphics[width=\textwidth]{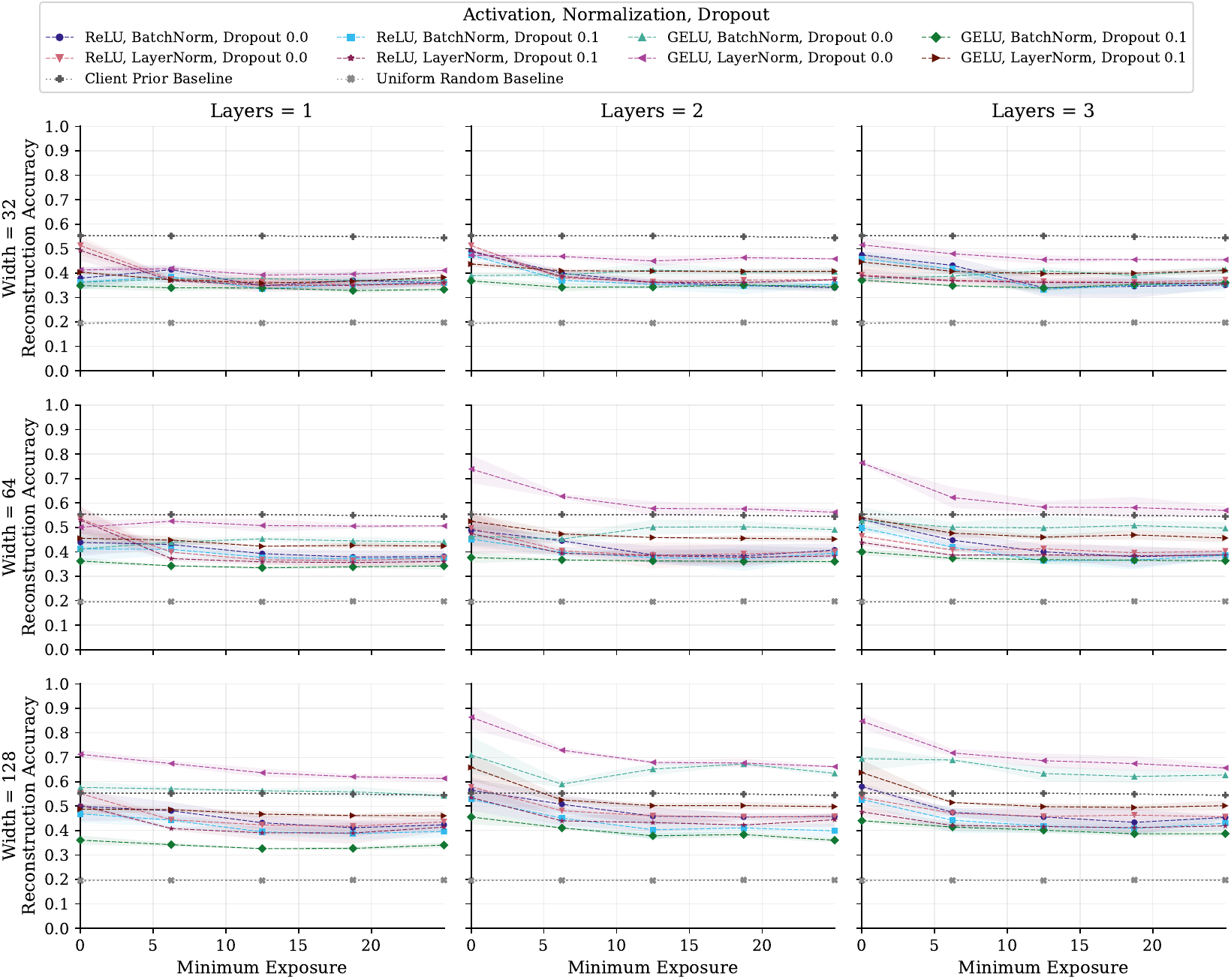}
          \caption{Reconstruction accuracy}
      \end{subfigure}

      \vspace{0.5em}

      \begin{subfigure}[t]{0.48\textwidth}
          \centering
          \includegraphics[width=\textwidth]{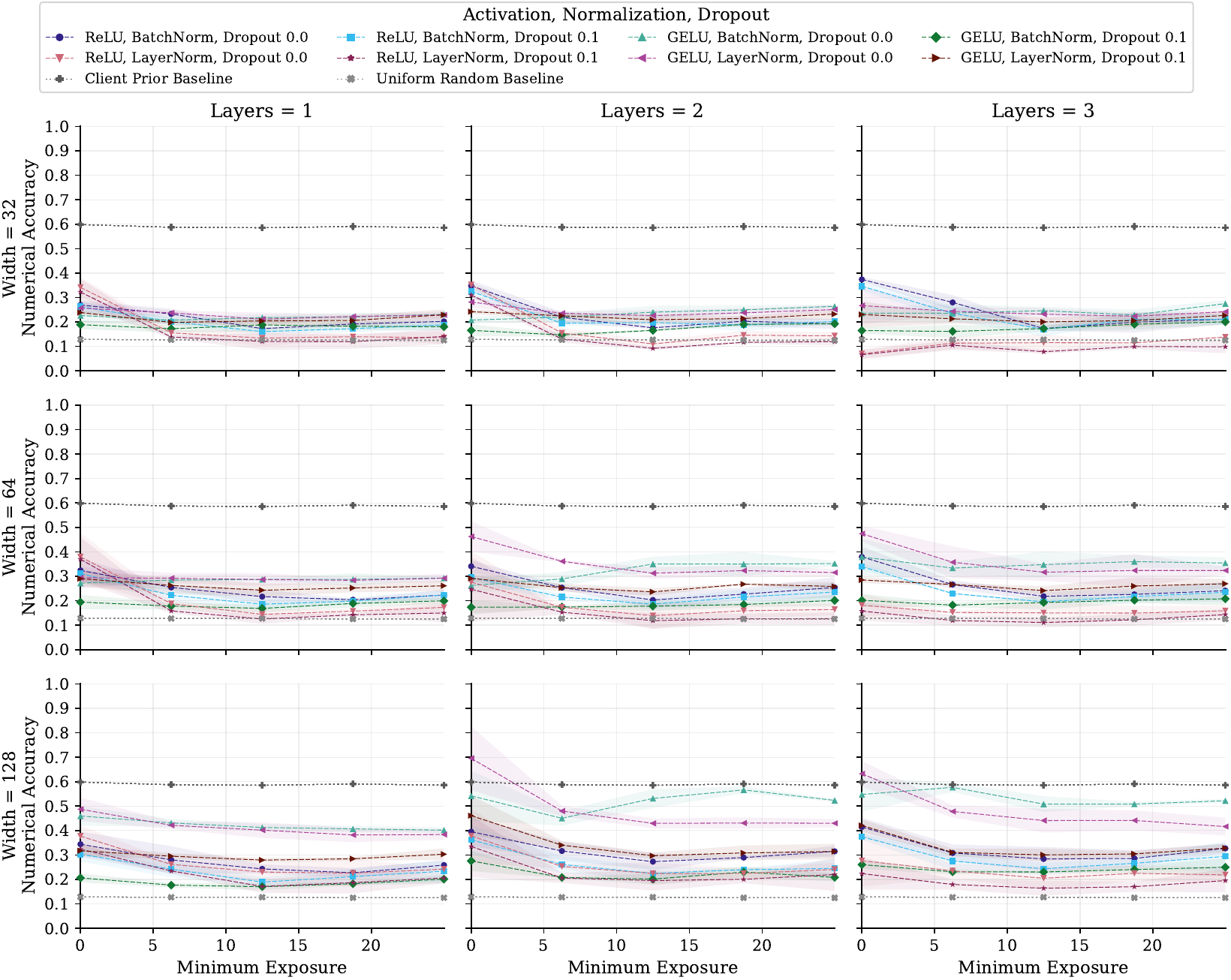}
          \caption{Numerical accuracy}
      \end{subfigure}
      \hfill
      \begin{subfigure}[t]{0.48\textwidth}
          \centering
          \includegraphics[width=\textwidth]{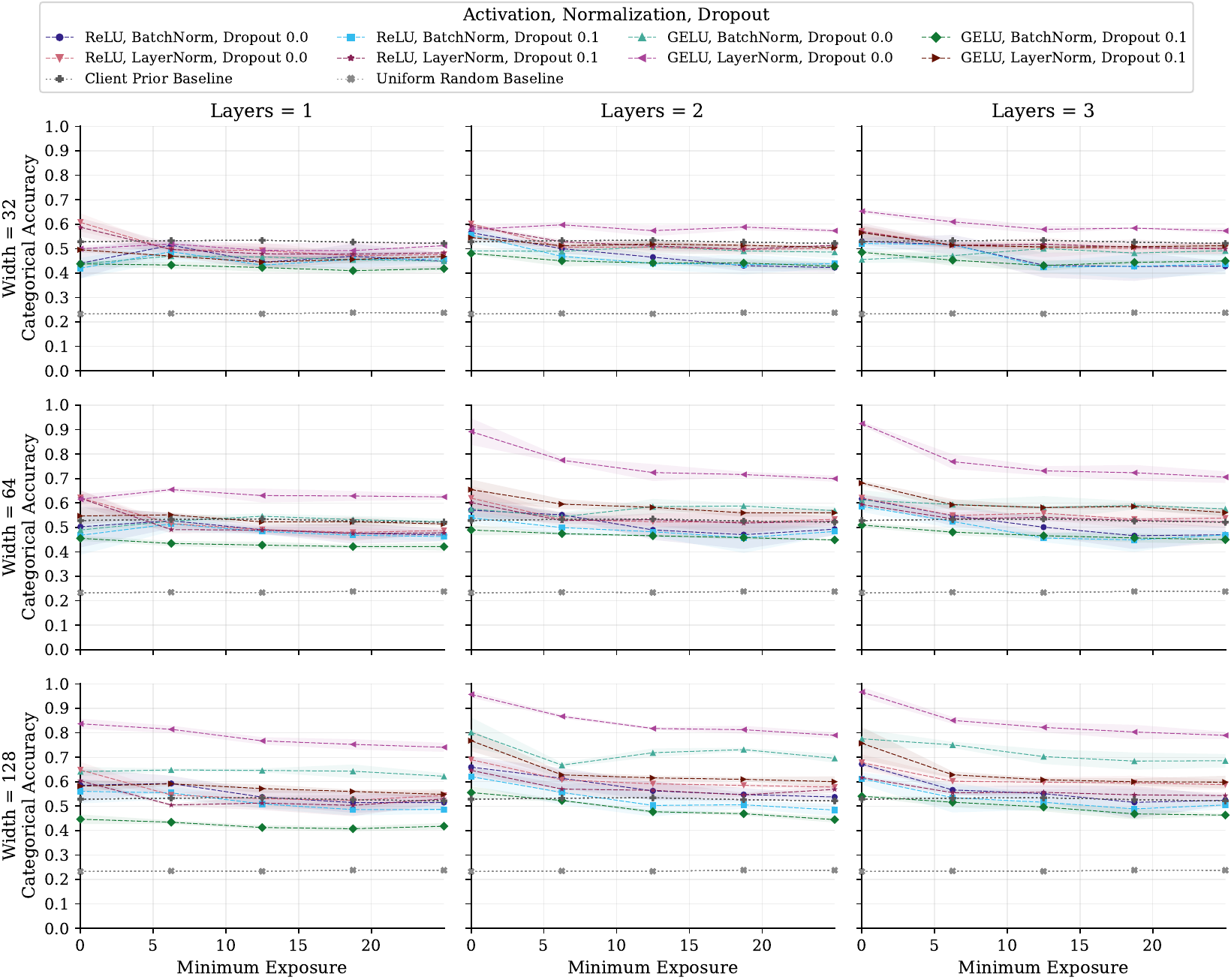}
          \caption{Categorical accuracy}
      \end{subfigure}

      \caption{Utility and inversion accuracy metrics over FL training for the MLP architecture grid using client batch size 32 on Adult.}
      \label{fig:adult-torch-modules-privacy-utility-b32}
\end{figure}

\begin{figure}[H]
      \centering

      \begin{subfigure}[t]{0.48\textwidth}
          \centering
          \includegraphics[width=\textwidth]{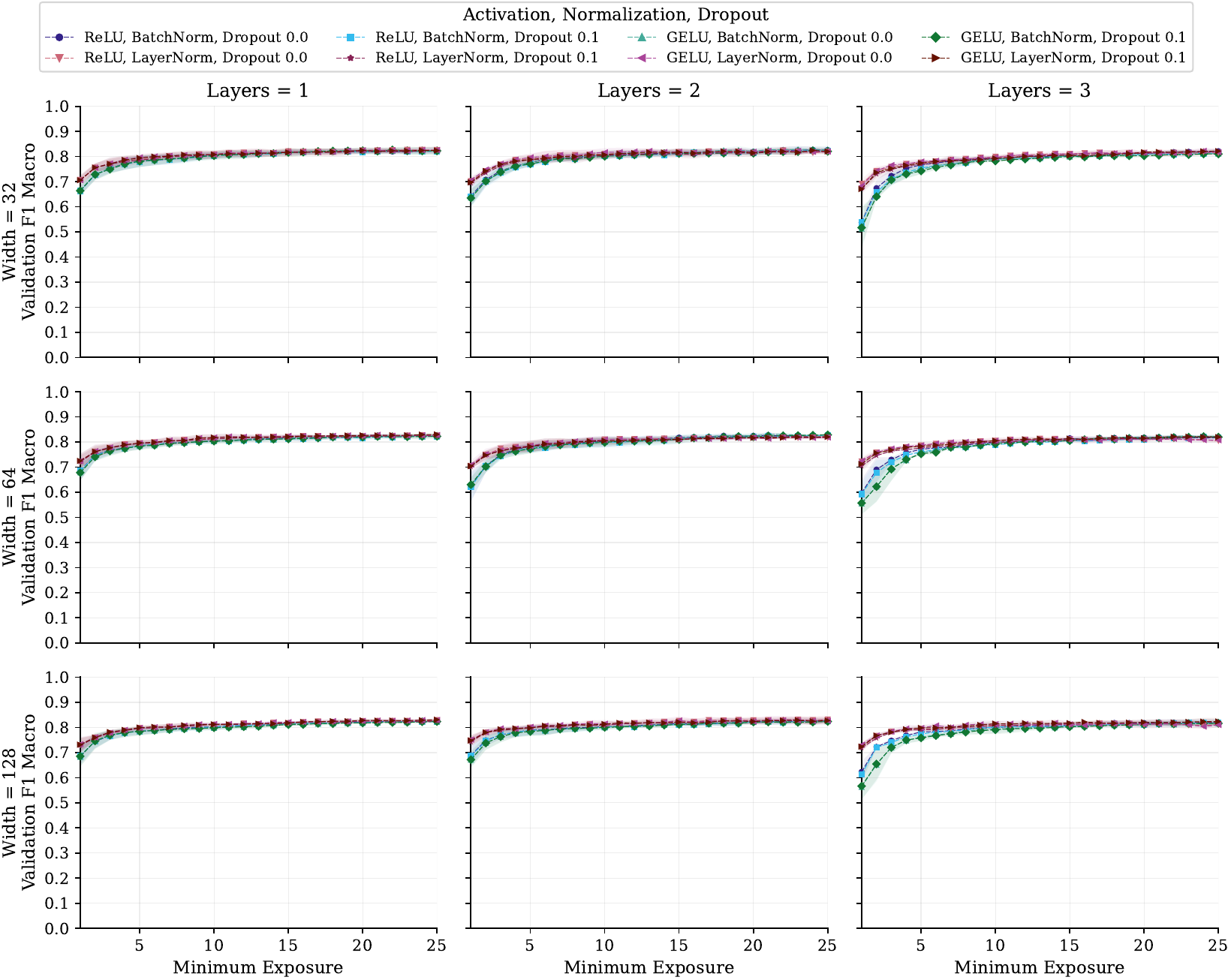}
          \caption{Utility curve Macro F1}
      \end{subfigure}
      \hfill
      \begin{subfigure}[t]{0.48\textwidth}
          \centering
          \includegraphics[width=\textwidth]{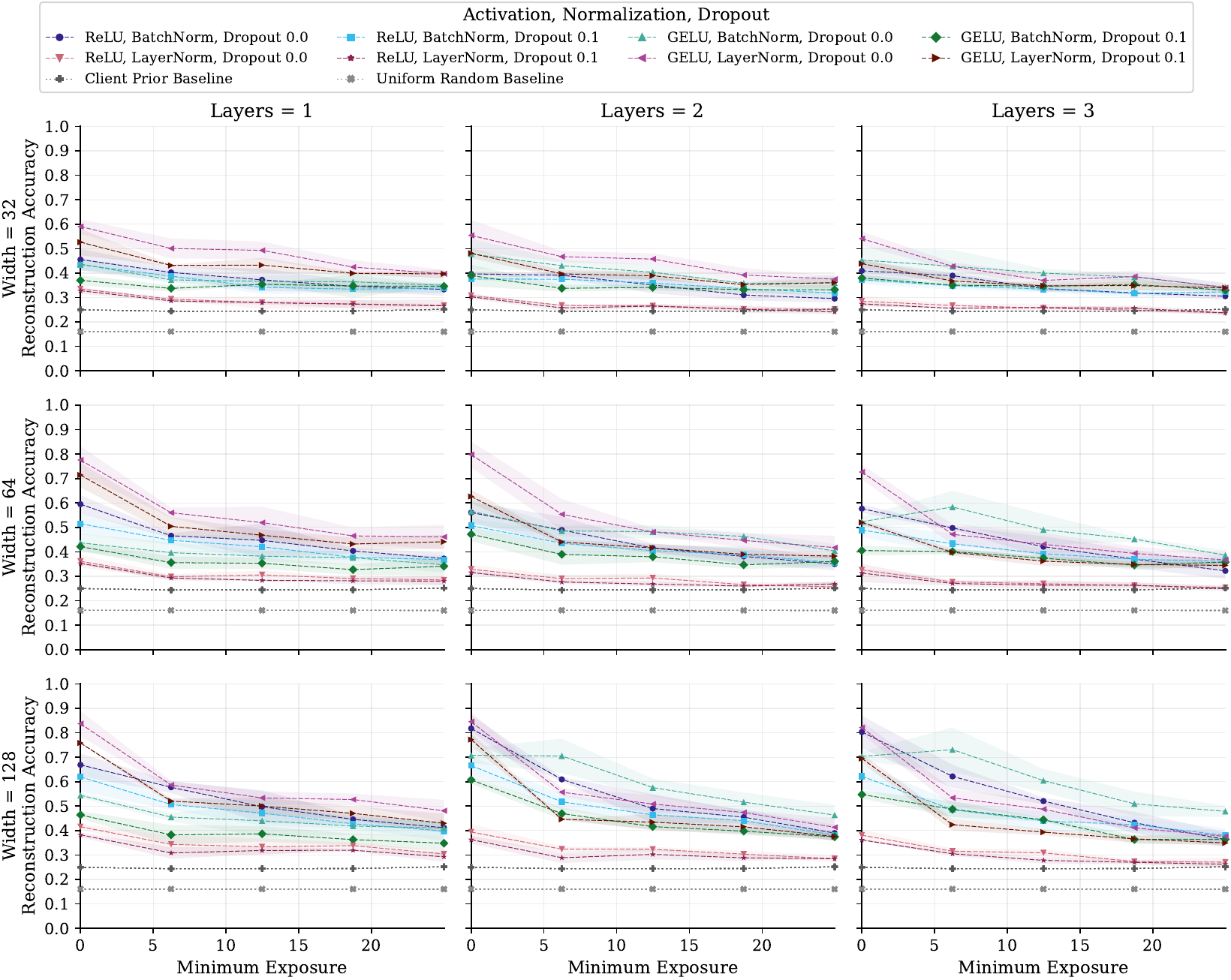}
          \caption{Reconstruction accuracy}
      \end{subfigure}

      \vspace{0.5em}

      \begin{subfigure}[t]{0.48\textwidth}
          \centering
          \includegraphics[width=\textwidth]{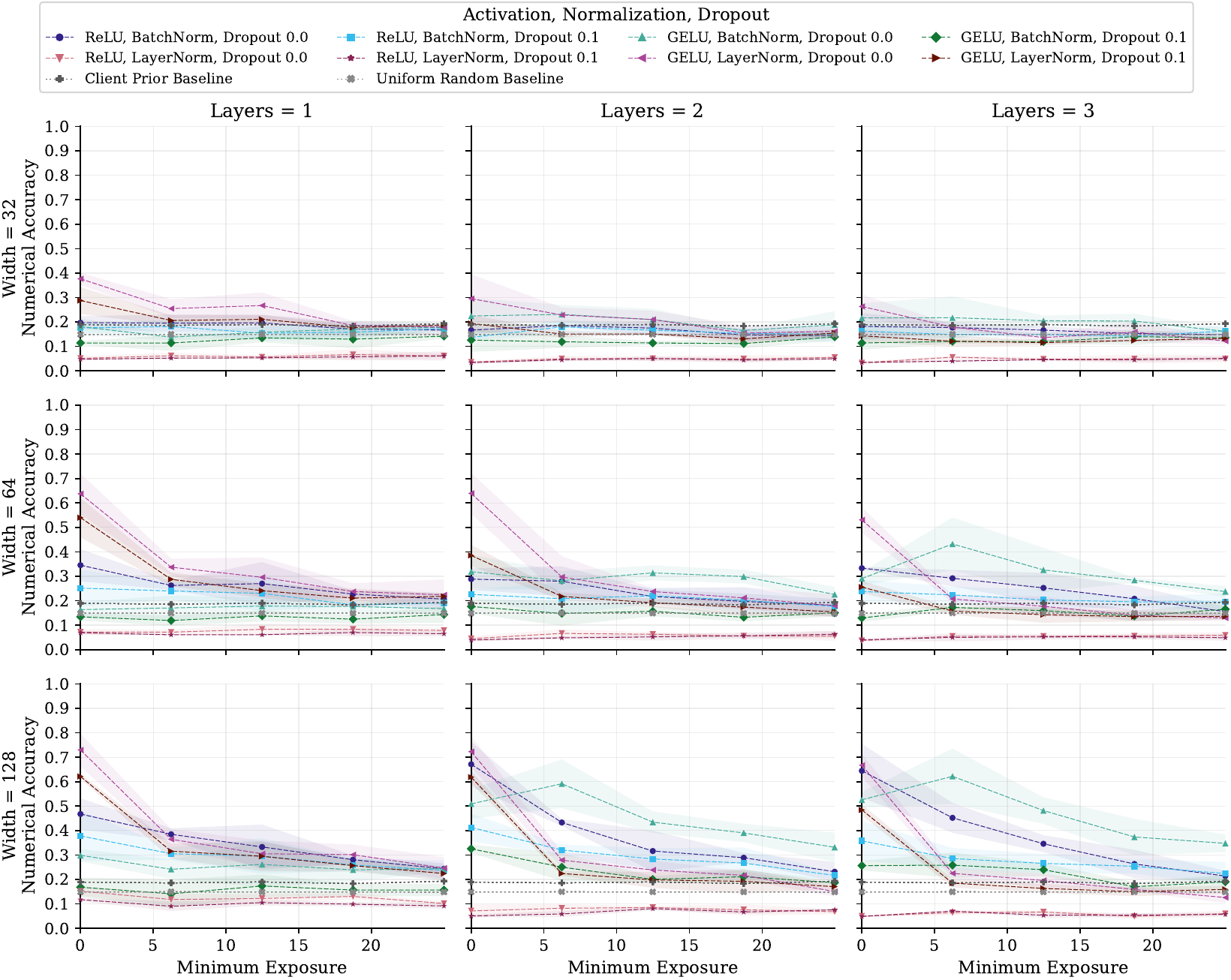}
          \caption{Numerical accuracy}
      \end{subfigure}
      \hfill
      \begin{subfigure}[t]{0.48\textwidth}
          \centering
          \includegraphics[width=\textwidth]{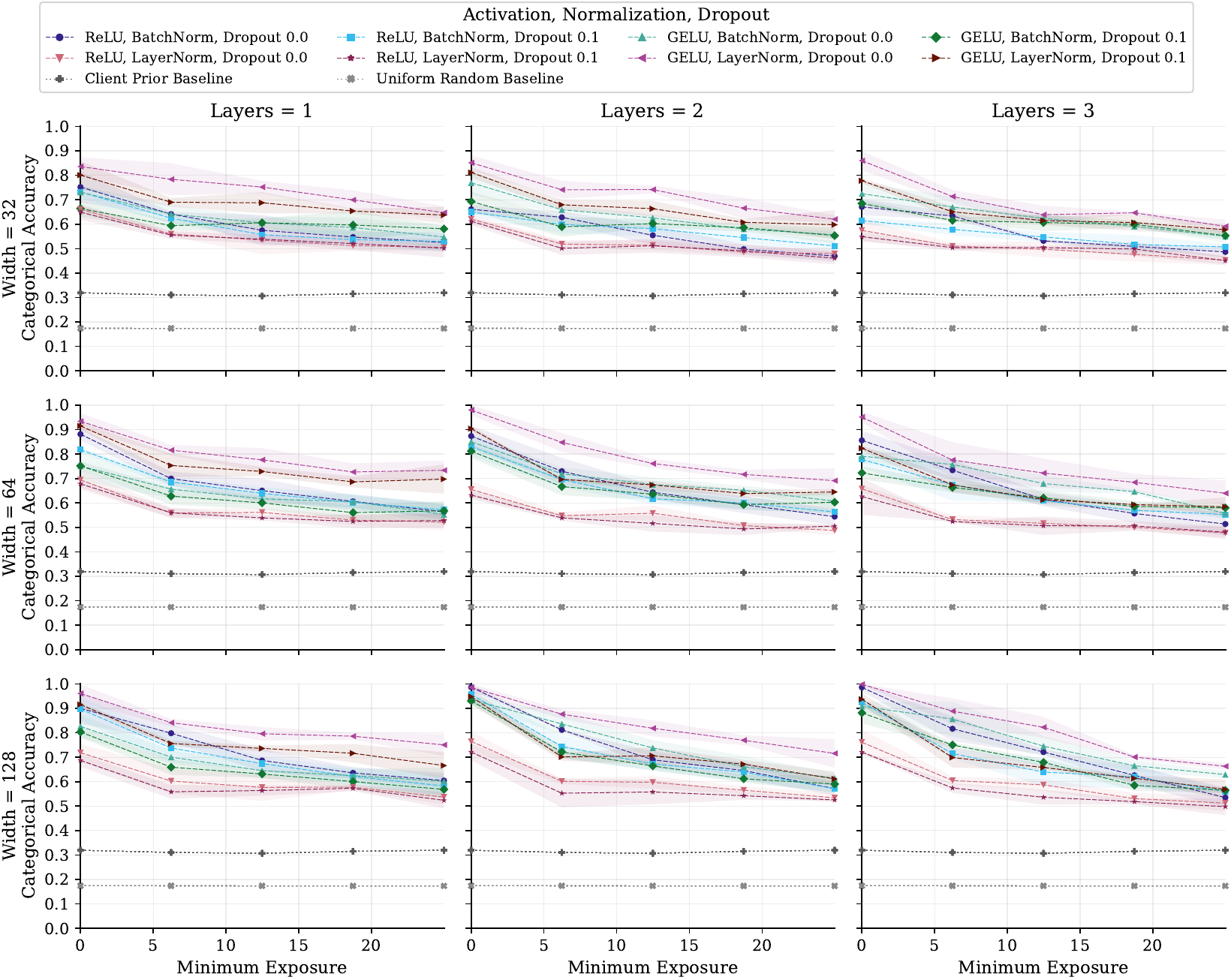}
          \caption{Categorical accuracy}
      \end{subfigure}

      \caption{Utility and inversion accuracy metrics over FL training for the MLP architecture grid using client batch size 8 on the private multiclass benchmark.}
      \label{fig:pandemic-torch-modules-privacy-utility-b8}
\end{figure}

\begin{figure}[H]
      \centering

      \begin{subfigure}[t]{0.48\textwidth}
          \centering
          \includegraphics[width=\textwidth]{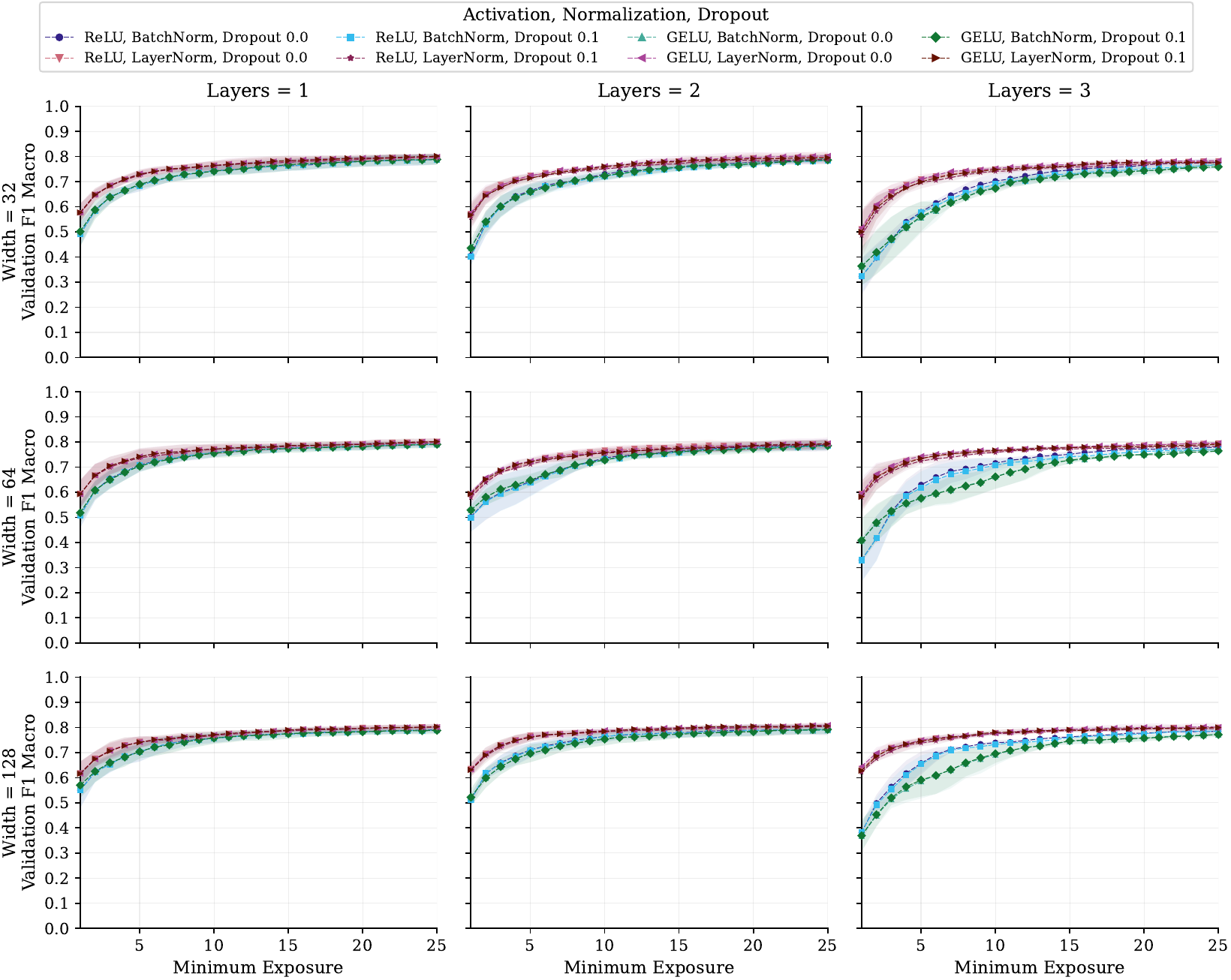}
          \caption{Utility curve Macro F1}
      \end{subfigure}
      \hfill
      \begin{subfigure}[t]{0.48\textwidth}
          \centering
          \includegraphics[width=\textwidth]{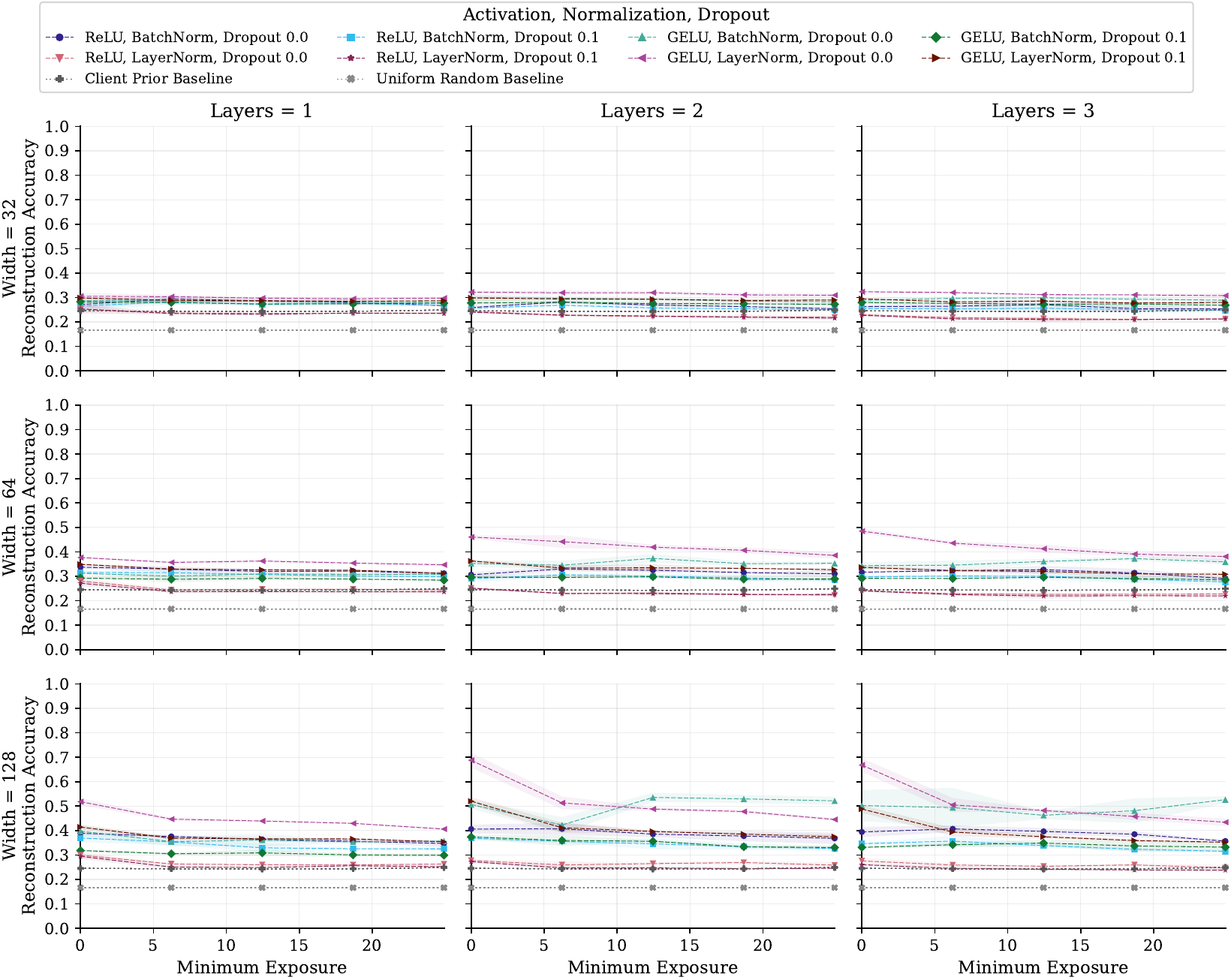}
          \caption{Reconstruction accuracy}
      \end{subfigure}

      \vspace{0.5em}

      \begin{subfigure}[t]{0.48\textwidth}
          \centering
          \includegraphics[width=\textwidth]{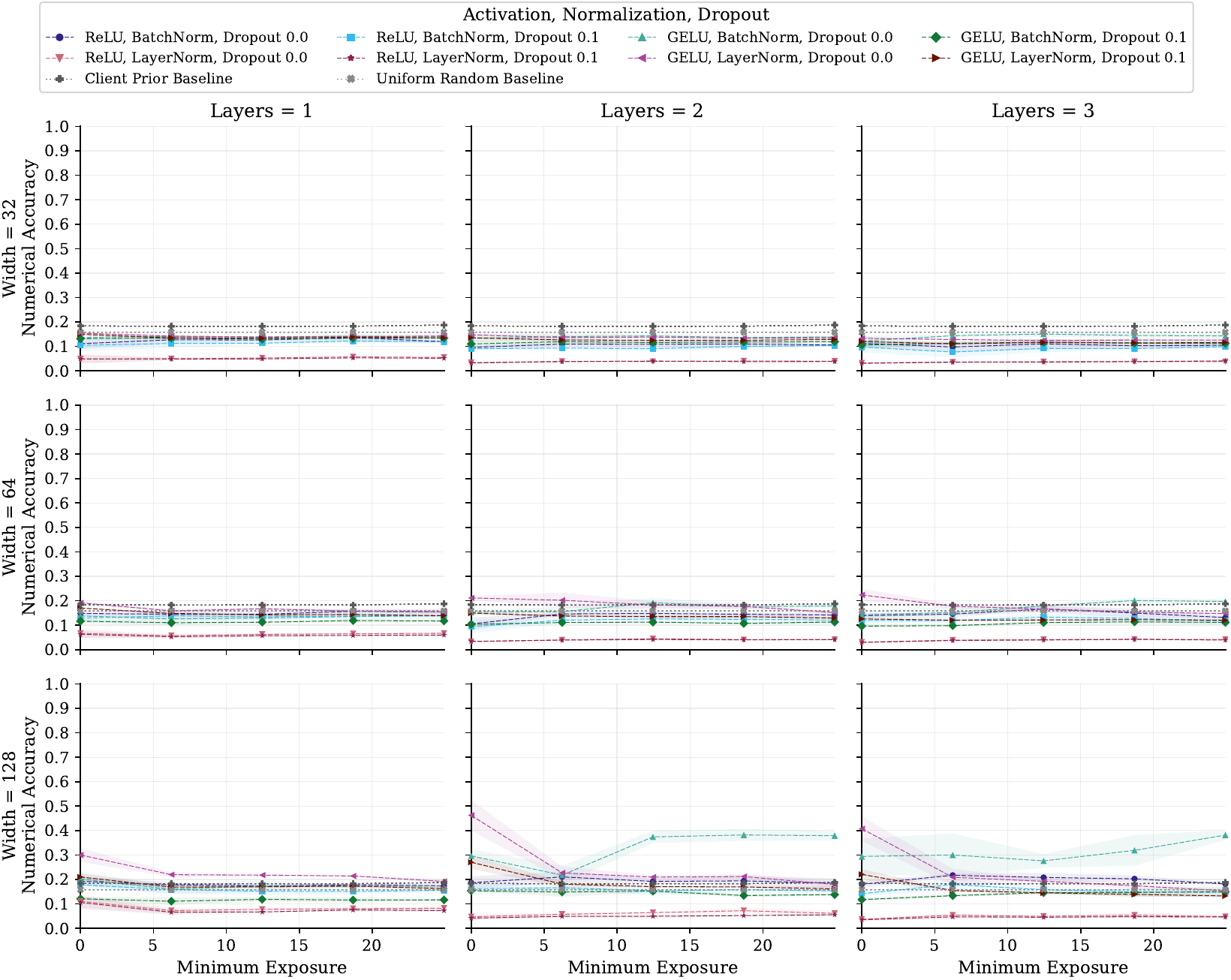}
          \caption{Numerical accuracy}
      \end{subfigure}
      \hfill
      \begin{subfigure}[t]{0.48\textwidth}
          \centering
          \includegraphics[width=\textwidth]{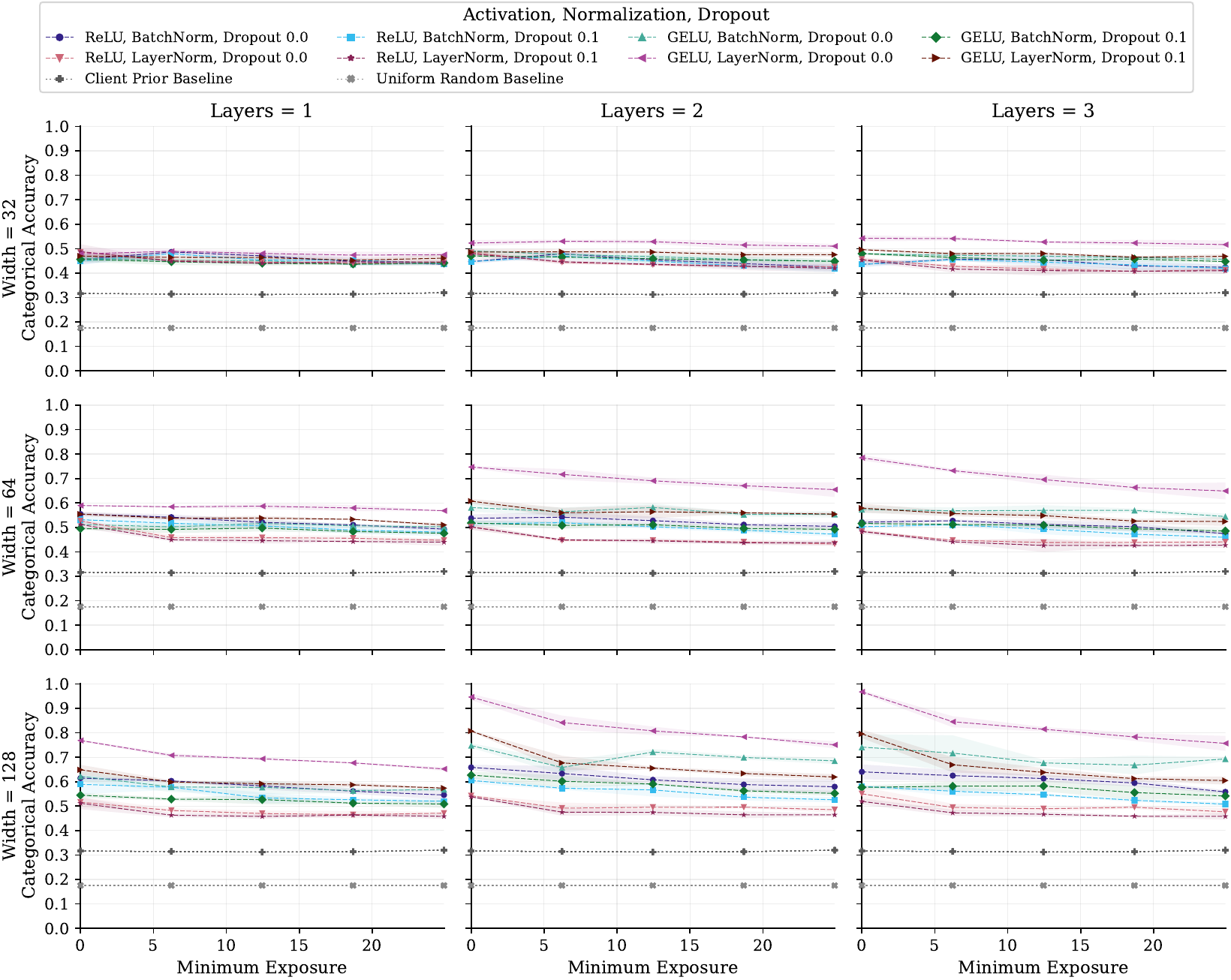}
          \caption{Categorical accuracy}
      \end{subfigure}

      \caption{Utility and inversion accuracy metrics over FL training for the MLP architecture grid using client batch size 32 on the private multiclass benchmark.}
      \label{fig:pandemic-torch-modules-privacy-utility-b32}
\end{figure}

\begin{figure}[H]
      \centering

      \begin{subfigure}[t]{0.48\textwidth}
          \centering
          \includegraphics[width=\textwidth]{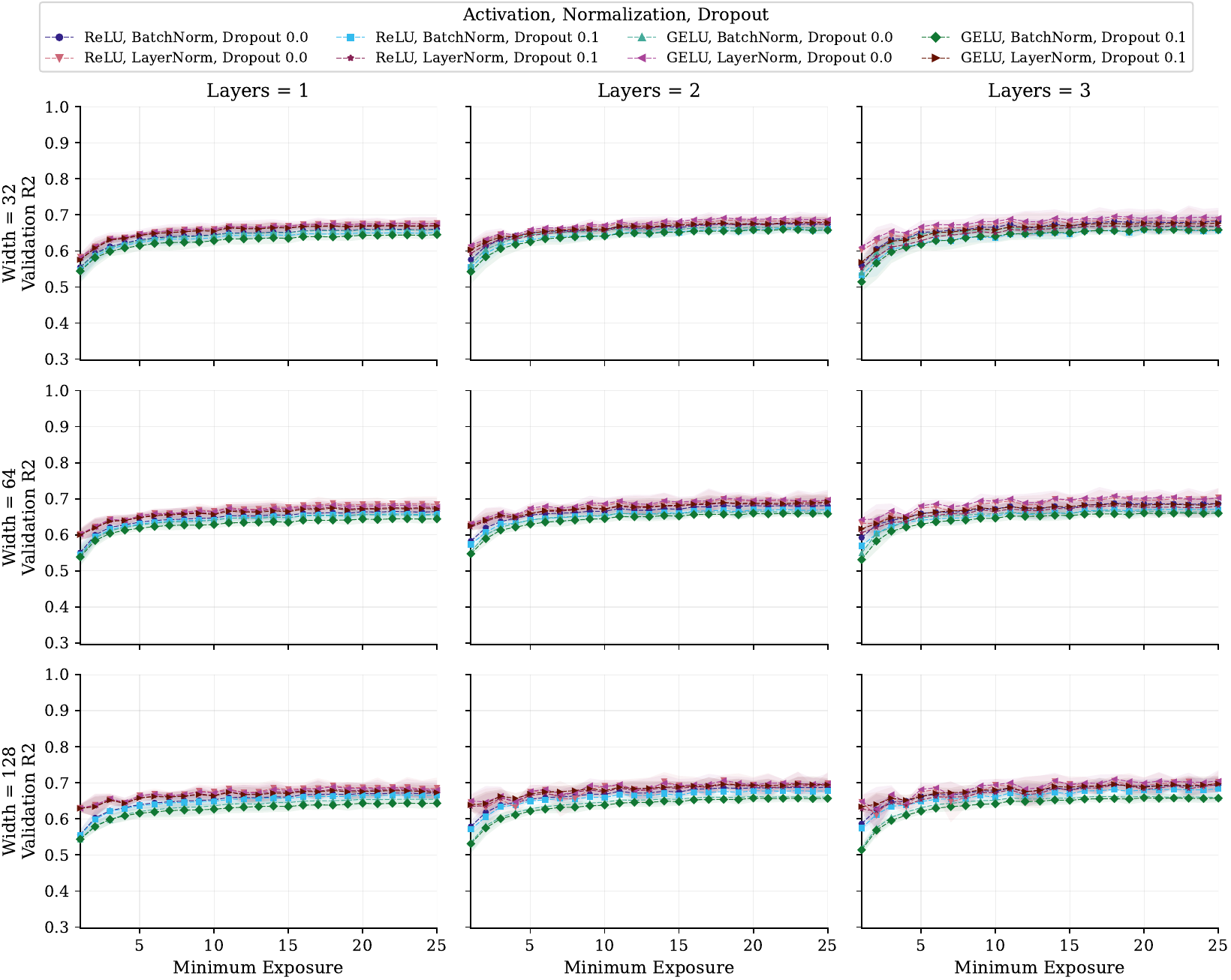}
          \caption{Utility curve $R^2$ batch size 8}
      \end{subfigure}
      \hfill
      \begin{subfigure}[t]{0.48\textwidth}
          \centering
          \includegraphics[width=\textwidth]{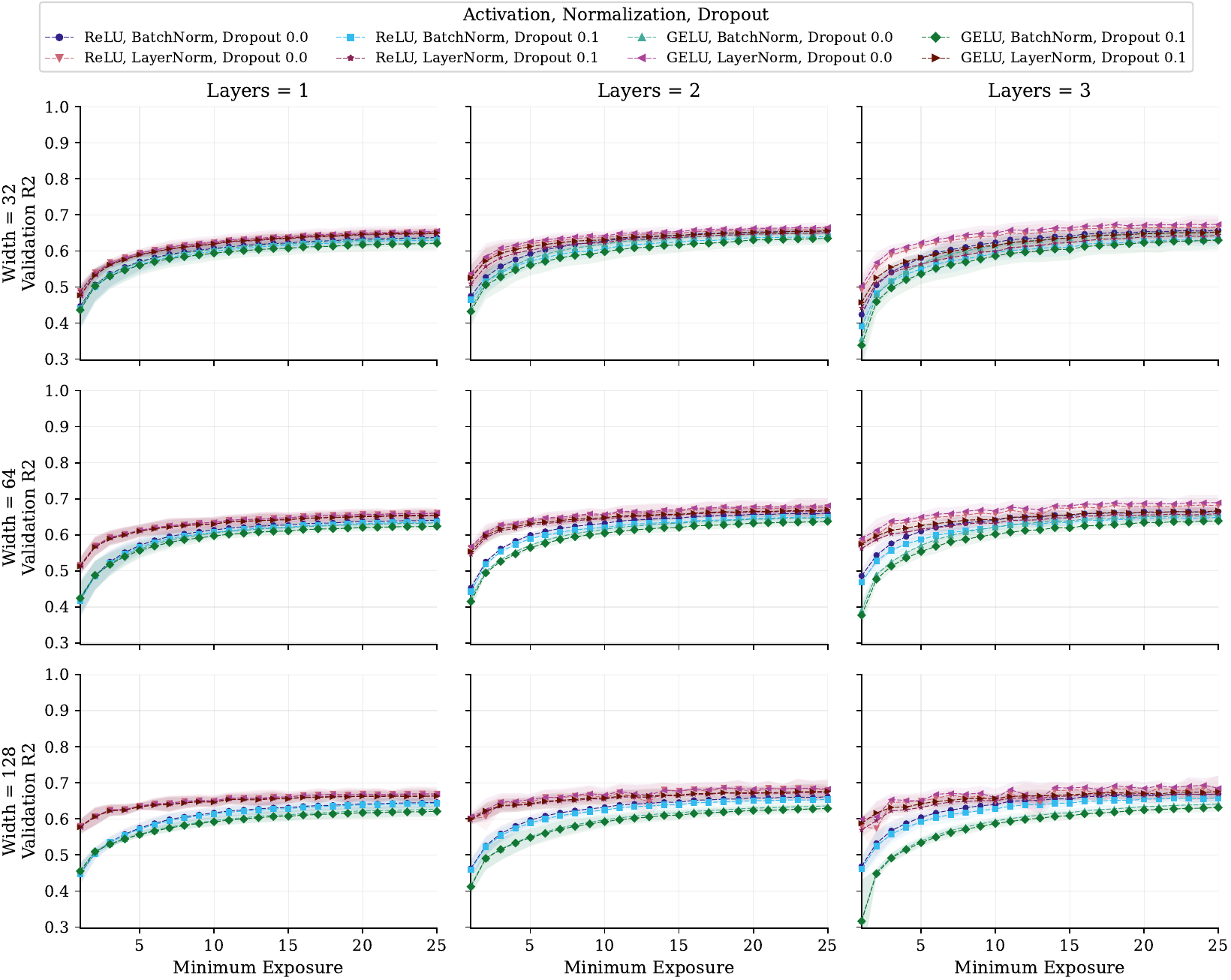}
          \caption{Utility curve $R^2$ batch size 32}
      \end{subfigure}

      \vspace{0.5em}

      \begin{subfigure}[t]{0.48\textwidth}
          \centering
          \includegraphics[width=\textwidth]{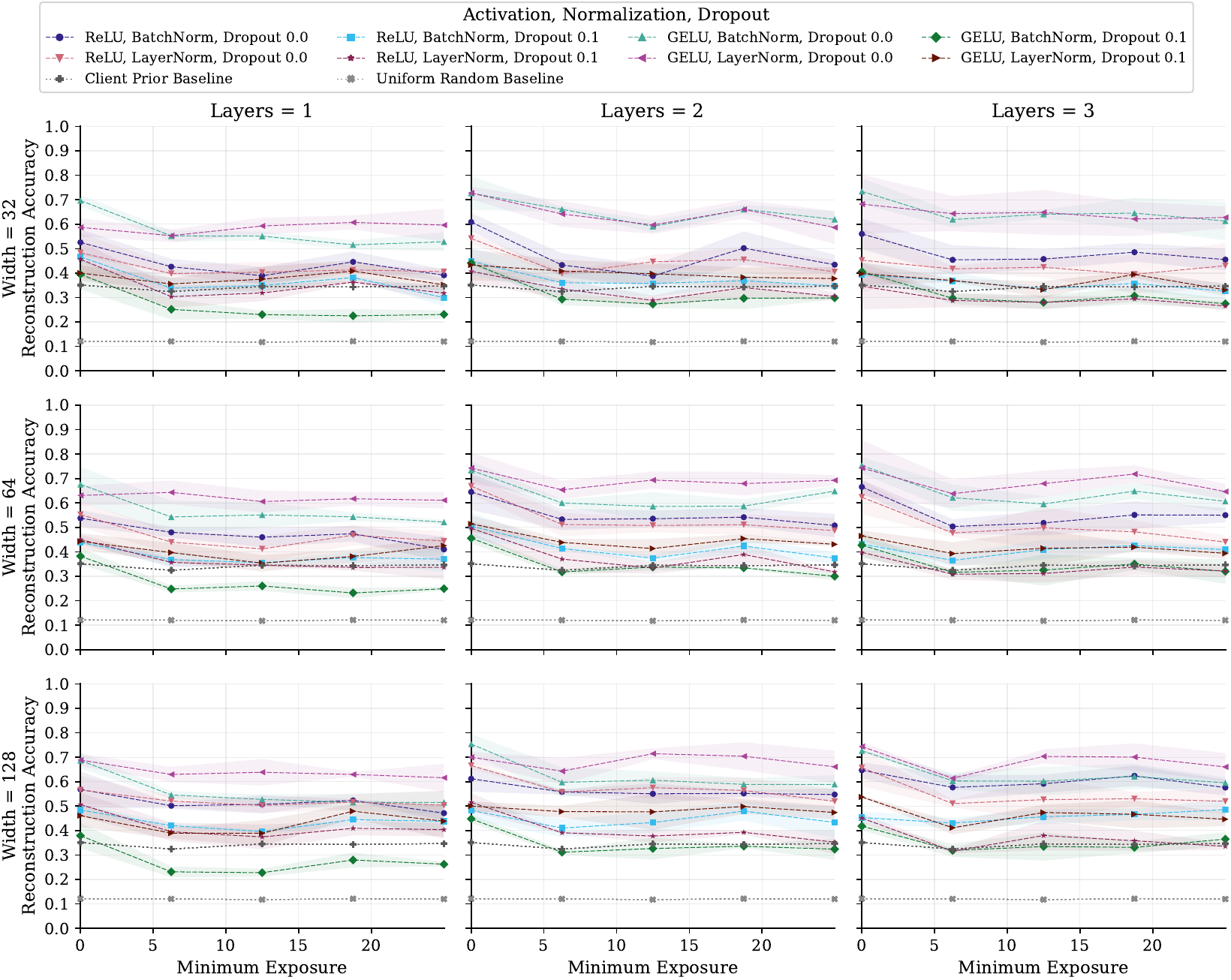}
          \caption{Reconstruction accuracy batch size 8}
      \end{subfigure}
      \hfill
      \begin{subfigure}[t]{0.48\textwidth}
          \centering
          \includegraphics[width=\textwidth]{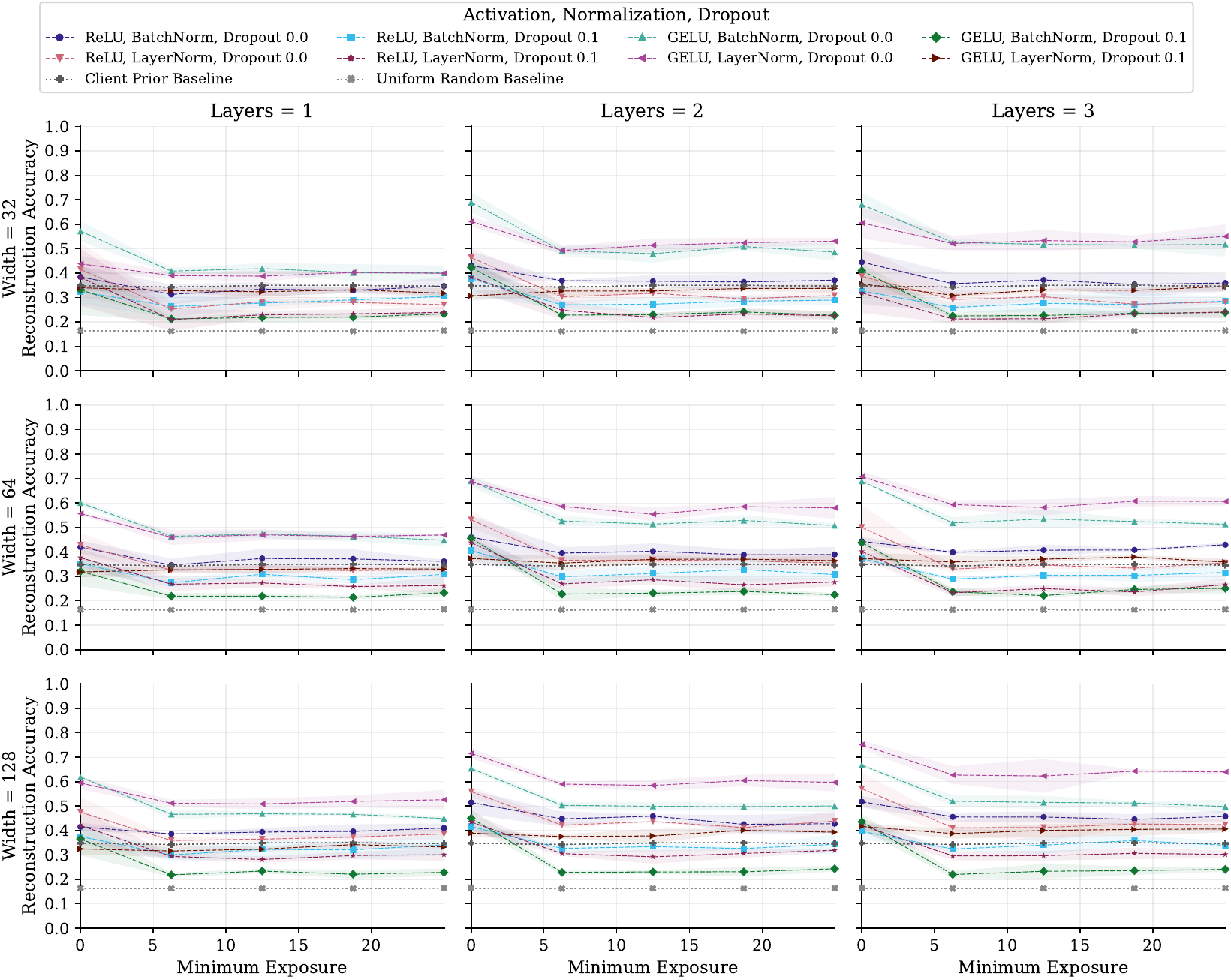}
          \caption{Reconstruction accuracy batch size 32}
      \end{subfigure}

      \caption{Utility and reconstruction accuracy over FL training for the MLP architecture grid using client batch sizes 8 and 32 on California Housing; reconstruction accuracy equals numerical accuracy because the dataset contains only numerical input features.}
      \label{fig:cali-torch-modules-privacy-utility}
\end{figure}

The benchmark MLP architecture grid covers Adult, the private multiclass benchmark, and California Housing. Tables~\ref{tab:torch-modules-adult-structural}, \ref{tab:torch-modules-pandemic-structural}, and~\ref{tab:torch-modules-california-structural} summarize the marginal effects of hidden width and hidden depth. Tables~\ref{tab:torch-modules-adult-modules}, \ref{tab:torch-modules-pandemic-modules}, and~\ref{tab:torch-modules-california-modules} summarize the marginal effects of normalization, activation, and dropout. Tables~\ref{tab:torch-modules-adult-module-ranking}, \ref{tab:torch-modules-pandemic-module-ranking}, and~\ref{tab:torch-modules-california-module-ranking} rank the complete module combinations from lowest to highest final attack point reconstruction accuracy after averaging over width and depth. Table~\ref{tab:torch-modules-benchmark-top-utility-configs-b8} reports the highest utility configurations at client batch size \(8\). Figures~\ref{fig:adult-torch-modules-privacy-utility-b8} and~\ref{fig:adult-torch-modules-privacy-utility-b32} report the Adult trajectories, Figures~\ref{fig:pandemic-torch-modules-privacy-utility-b8} and~\ref{fig:pandemic-torch-modules-privacy-utility-b32} report the private multiclass benchmark trajectories, and Figure~\ref{fig:cali-torch-modules-privacy-utility} reports the California Housing trajectories. These results support the architecture analysis by testing whether module level leakage and utility patterns also appear across the benchmark tasks.

\begin{table}[H]
  \centering
  \small
  \caption{Reconstruction accuracy at the final attack point for structural architectural factors on Adult. Each cell reports mean $\pm$ standard deviation across the remaining configurations at the final attack point.}
  \label{tab:torch-modules-adult-structural}
  \resizebox{\textwidth}{!}{%
  \begin{tabular}{lcccccc}
  \hline
  \multicolumn{1}{l}{} & \multicolumn{3}{c}{\textbf{Hidden Width}} & \multicolumn{3}{c}{\textbf{Hidden Layers}} \\
  \textbf{Batch size} & \textbf{32} & \textbf{64} & \textbf{128} & \textbf{1} & \textbf{2} & \textbf{3} \\
  \hline
  8 & 0.510 $\pm$ 0.074 & 0.559 $\pm$ 0.090 & 0.607 $\pm$ 0.103 & 0.552 $\pm$ 0.095 & 0.564 $\pm$ 0.102 & 0.561 $\pm$ 0.099 \\
  32 & 0.378 $\pm$ 0.034 & 0.421 $\pm$ 0.063 & 0.478 $\pm$ 0.095 & 0.407 $\pm$ 0.068 & 0.434 $\pm$ 0.085 & 0.436 $\pm$ 0.083 \\
  \hline
  \end{tabular}
  }
\end{table}

\begin{table}[H]
  \centering
  \small
  \caption{Reconstruction accuracy at the final attack point for structural architectural factors on the private multiclass benchmark. Each cell reports mean $\pm$ standard deviation across the remaining configurations at the final attack point.}
  \label{tab:torch-modules-pandemic-structural}
  \resizebox{\textwidth}{!}{%
  \begin{tabular}{lcccccc}
  \hline
  \multicolumn{1}{l}{} & \multicolumn{3}{c}{\textbf{Hidden Width}} & \multicolumn{3}{c}{\textbf{Hidden Layers}} \\
  \textbf{Batch size} & \textbf{32} & \textbf{64} & \textbf{128} & \textbf{1} & \textbf{2} & \textbf{3} \\
  \hline
  8 & 0.320 $\pm$ 0.048 & 0.347 $\pm$ 0.058 & 0.370 $\pm$ 0.063 & 0.361 $\pm$ 0.061 & 0.346 $\pm$ 0.059 & 0.331 $\pm$ 0.057 \\
  32 & 0.263 $\pm$ 0.029 & 0.295 $\pm$ 0.049 & 0.345 $\pm$ 0.080 & 0.295 $\pm$ 0.044 & 0.307 $\pm$ 0.075 & 0.301 $\pm$ 0.074 \\
  \hline
  \end{tabular}
  }
\end{table}

\begin{table}[H]
  \centering
  \small
  \caption{Reconstruction accuracy at the final attack point for structural architectural factors on California Housing. Each cell reports mean $\pm$ standard deviation across the remaining configurations at the final attack point.}
  \label{tab:torch-modules-california-structural}
  \resizebox{\textwidth}{!}{%
  \begin{tabular}{lcccccc}
  \hline
  \multicolumn{1}{l}{} & \multicolumn{3}{c}{\textbf{Hidden Width}} & \multicolumn{3}{c}{\textbf{Hidden Layers}} \\
  \textbf{Batch size} & \textbf{32} & \textbf{64} & \textbf{128} & \textbf{1} & \textbf{2} & \textbf{3} \\
  \hline
  8 & 0.409 $\pm$ 0.124 & 0.451 $\pm$ 0.124 & 0.481 $\pm$ 0.107 & 0.422 $\pm$ 0.110 & 0.460 $\pm$ 0.125 & 0.458 $\pm$ 0.127 \\
  32 & 0.338 $\pm$ 0.099 & 0.368 $\pm$ 0.106 & 0.398 $\pm$ 0.106 & 0.343 $\pm$ 0.081 & 0.377 $\pm$ 0.111 & 0.384 $\pm$ 0.120 \\
  \hline
  \end{tabular}
  }
\end{table}

\begin{table}[H]
  \centering
  \small
  \caption{Reconstruction accuracy at the final attack point for module level architectural factors on Adult. Each cell reports mean $\pm$ standard deviation across the remaining configurations at the final attack point.}
  \label{tab:torch-modules-adult-modules}
  \resizebox{\textwidth}{!}{%
  \begin{tabular}{lcccccc}
  \hline
  \multicolumn{1}{l}{} & \multicolumn{2}{c}{\textbf{Normalization}} & \multicolumn{2}{c}{\textbf{Activation}} & \multicolumn{2}{c}{\textbf{Dropout}} \\
  \textbf{Batch size} & \textbf{BatchNorm} & \textbf{LayerNorm} & \textbf{ReLU} & \textbf{GELU} & \textbf{0.0} & \textbf{0.1} \\
  \hline
  8 & 0.537 $\pm$ 0.076 & 0.581 $\pm$ 0.112 & 0.511 $\pm$ 0.051 & 0.606 $\pm$ 0.109 & 0.599 $\pm$ 0.108 & 0.518 $\pm$ 0.064 \\
  32 & 0.406 $\pm$ 0.073 & 0.445 $\pm$ 0.081 & 0.393 $\pm$ 0.034 & 0.458 $\pm$ 0.097 & 0.459 $\pm$ 0.093 & 0.393 $\pm$ 0.043 \\
  \hline
  \end{tabular}
  }
\end{table}

\begin{table}[H]
  \centering
  \small
  \caption{Reconstruction accuracy at the final attack point for module level architectural factors on the private multiclass benchmark. Each cell reports mean $\pm$ standard deviation across the remaining configurations at the final attack point.}
  \label{tab:torch-modules-pandemic-modules}
  \resizebox{\textwidth}{!}{%
  \begin{tabular}{lcccccc}
  \hline
  \multicolumn{1}{l}{} & \multicolumn{2}{c}{\textbf{Normalization}} & \multicolumn{2}{c}{\textbf{Activation}} & \multicolumn{2}{c}{\textbf{Dropout}} \\
  \textbf{Batch size} & \textbf{BatchNorm} & \textbf{LayerNorm} & \textbf{ReLU} & \textbf{GELU} & \textbf{0.0} & \textbf{0.1} \\
  \hline
  8 & 0.363 $\pm$ 0.038 & 0.329 $\pm$ 0.071 & 0.310 $\pm$ 0.052 & 0.382 $\pm$ 0.043 & 0.354 $\pm$ 0.067 & 0.338 $\pm$ 0.051 \\
  32 & 0.313 $\pm$ 0.062 & 0.289 $\pm$ 0.067 & 0.266 $\pm$ 0.043 & 0.336 $\pm$ 0.065 & 0.319 $\pm$ 0.079 & 0.283 $\pm$ 0.041 \\
  \hline
  \end{tabular}
  }
\end{table}

\begin{table}[H]
  \centering
  \small
  \caption{Reconstruction accuracy at the final attack point for module level architectural factors on California Housing. Each cell reports mean $\pm$ standard deviation across the remaining configurations at the final attack point.}
  \label{tab:torch-modules-california-modules}
  \resizebox{\textwidth}{!}{%
  \begin{tabular}{lcccccc}
  \hline
  \multicolumn{1}{l}{} & \multicolumn{2}{c}{\textbf{Normalization}} & \multicolumn{2}{c}{\textbf{Activation}} & \multicolumn{2}{c}{\textbf{Dropout}} \\
  \textbf{Batch size} & \textbf{BatchNorm} & \textbf{LayerNorm} & \textbf{ReLU} & \textbf{GELU} & \textbf{0.0} & \textbf{0.1} \\
  \hline
  8 & 0.436 $\pm$ 0.122 & 0.458 $\pm$ 0.120 & 0.415 $\pm$ 0.081 & 0.479 $\pm$ 0.145 & 0.540 $\pm$ 0.086 & 0.354 $\pm$ 0.065 \\
  32 & 0.356 $\pm$ 0.097 & 0.380 $\pm$ 0.114 & 0.332 $\pm$ 0.060 & 0.403 $\pm$ 0.128 & 0.442 $\pm$ 0.093 & 0.294 $\pm$ 0.051 \\
  \hline
  \end{tabular}
  }
\end{table}

Adult favors wide LayerNorm configurations, mostly with GELU. The private multiclass benchmark has several nearly tied configurations, including both ReLU and GELU. California Housing favors dropout disabled configurations, which differs from the MIMIC-IV utility ranking where the top configurations use dropout \(0.1\). This is consistent with the broader tabular learning observation that architectural performance is dataset dependent. The benchmark utility table should therefore be read together with the reconstruction rankings rather than as evidence for a universal best MLP architecture.

\begin{table}[H]
  \centering
  \small
  \caption{Complete ranking of module combinations by final attack point reconstruction accuracy on Adult for batch sizes 8 and 32. Rows are ordered from lowest to highest reconstruction accuracy. Reconstruction accuracy is averaged over width and depth and reported at the final attack point.}
  \label{tab:torch-modules-adult-module-ranking}
  \begin{tabular}{ccccc}
  \hline
  \textbf{Rank} & \textbf{Normalization} & \textbf{Activation} & \textbf{Dropout} & \textbf{Recon. Acc.} \\
  \hline
  \multicolumn{5}{c}{Batch size 8} \\
  \hline
  1 & BatchNorm & GELU & 0.1 & 0.469 $\pm$ 0.013 \\
  2 & LayerNorm & ReLU & 0.1 & 0.478 $\pm$ 0.047 \\
  3 & LayerNorm & ReLU & 0.0 & 0.510 $\pm$ 0.047 \\
  4 & BatchNorm & ReLU & 0.1 & 0.527 $\pm$ 0.049 \\
  5 & BatchNorm & ReLU & 0.0 & 0.529 $\pm$ 0.051 \\
  6 & LayerNorm & GELU & 0.1 & 0.599 $\pm$ 0.034 \\
  7 & BatchNorm & GELU & 0.0 & 0.622 $\pm$ 0.079 \\
  8 & LayerNorm & GELU & 0.0 & 0.736 $\pm$ 0.065 \\
  \hline
  \multicolumn{5}{c}{Batch size 32} \\
  \hline
  1 & BatchNorm & GELU & 0.1 & 0.355 $\pm$ 0.016 \\
  2 & BatchNorm & ReLU & 0.1 & 0.384 $\pm$ 0.024 \\
  3 & LayerNorm & ReLU & 0.1 & 0.389 $\pm$ 0.031 \\
  4 & BatchNorm & ReLU & 0.0 & 0.396 $\pm$ 0.043 \\
  5 & LayerNorm & ReLU & 0.0 & 0.403 $\pm$ 0.039 \\
  6 & LayerNorm & GELU & 0.1 & 0.444 $\pm$ 0.041 \\
  7 & BatchNorm & GELU & 0.0 & 0.492 $\pm$ 0.095 \\
  8 & LayerNorm & GELU & 0.0 & 0.544 $\pm$ 0.091 \\
  \hline
  \end{tabular}
\end{table}

\begin{table}[H]
  \centering
  \small
  \caption{Complete ranking of module combinations by final attack point reconstruction accuracy on the private multiclass benchmark for batch sizes 8 and 32. Rows are ordered from lowest to highest reconstruction accuracy. Reconstruction accuracy is averaged over width and depth and reported at the final attack point.}
  \label{tab:torch-modules-pandemic-module-ranking}
  \begin{tabular}{ccccc}
  \hline
  \textbf{Rank} & \textbf{Normalization} & \textbf{Activation} & \textbf{Dropout} & \textbf{Recon. Acc.} \\
  \hline
  \multicolumn{5}{c}{Batch size 8} \\
  \hline
  1 & LayerNorm & ReLU & 0.1 & 0.265 $\pm$ 0.019 \\
  2 & LayerNorm & ReLU & 0.0 & 0.268 $\pm$ 0.021 \\
  3 & BatchNorm & ReLU & 0.0 & 0.349 $\pm$ 0.039 \\
  4 & BatchNorm & GELU & 0.1 & 0.351 $\pm$ 0.015 \\
  5 & BatchNorm & ReLU & 0.1 & 0.358 $\pm$ 0.027 \\
  6 & LayerNorm & GELU & 0.1 & 0.380 $\pm$ 0.037 \\
  7 & BatchNorm & GELU & 0.0 & 0.393 $\pm$ 0.050 \\
  8 & LayerNorm & GELU & 0.0 & 0.403 $\pm$ 0.046 \\
  \hline
  \multicolumn{5}{c}{Batch size 32} \\
  \hline
  1 & LayerNorm & ReLU & 0.1 & 0.231 $\pm$ 0.014 \\
  2 & LayerNorm & ReLU & 0.0 & 0.237 $\pm$ 0.018 \\
  3 & BatchNorm & ReLU & 0.1 & 0.288 $\pm$ 0.030 \\
  4 & BatchNorm & GELU & 0.1 & 0.293 $\pm$ 0.024 \\
  5 & BatchNorm & ReLU & 0.0 & 0.307 $\pm$ 0.044 \\
  6 & LayerNorm & GELU & 0.1 & 0.320 $\pm$ 0.034 \\
  7 & BatchNorm & GELU & 0.0 & 0.364 $\pm$ 0.096 \\
  8 & LayerNorm & GELU & 0.0 & 0.368 $\pm$ 0.056 \\
  \hline
  \end{tabular}
\end{table}

\begin{table}[H]
  \centering
  \small
  \caption{Complete ranking of module combinations by final attack point reconstruction accuracy on California Housing for batch sizes 8 and 32. Rows are ordered from lowest to highest reconstruction accuracy. Reconstruction accuracy is averaged over width and depth and reported at the final attack point.}
  \label{tab:torch-modules-california-module-ranking}
  \begin{tabular}{ccccc}
  \hline
  \textbf{Rank} & \textbf{Normalization} & \textbf{Activation} & \textbf{Dropout} & \textbf{Recon. Acc.} \\
  \hline
  \multicolumn{5}{c}{Batch size 8} \\
  \hline
  1 & BatchNorm & GELU & 0.1 & 0.292 $\pm$ 0.042 \\
  2 & LayerNorm & ReLU & 0.1 & 0.328 $\pm$ 0.037 \\
  3 & BatchNorm & ReLU & 0.1 & 0.387 $\pm$ 0.060 \\
  4 & LayerNorm & GELU & 0.1 & 0.408 $\pm$ 0.047 \\
  5 & LayerNorm & ReLU & 0.0 & 0.462 $\pm$ 0.046 \\
  6 & BatchNorm & ReLU & 0.0 & 0.483 $\pm$ 0.066 \\
  7 & BatchNorm & GELU & 0.0 & 0.582 $\pm$ 0.048 \\
  8 & LayerNorm & GELU & 0.0 & 0.633 $\pm$ 0.035 \\
  \hline
  \multicolumn{5}{c}{Batch size 32} \\
  \hline
  1 & BatchNorm & GELU & 0.1 & 0.236 $\pm$ 0.009 \\
  2 & LayerNorm & ReLU & 0.1 & 0.270 $\pm$ 0.032 \\
  3 & BatchNorm & ReLU & 0.1 & 0.315 $\pm$ 0.022 \\
  4 & LayerNorm & ReLU & 0.0 & 0.350 $\pm$ 0.059 \\
  5 & LayerNorm & GELU & 0.1 & 0.354 $\pm$ 0.030 \\
  6 & BatchNorm & ReLU & 0.0 & 0.395 $\pm$ 0.038 \\
  7 & BatchNorm & GELU & 0.0 & 0.480 $\pm$ 0.039 \\
  8 & LayerNorm & GELU & 0.0 & 0.544 $\pm$ 0.074 \\
  \hline
  \end{tabular}
\end{table}

The benchmark rankings show a consistent high leakage module pattern. Across Adult, the private multiclass benchmark, and California Housing, the highest reconstruction configuration at both batch sizes is LayerNorm with GELU activation and no dropout. This supports the main MIMIC-IV architecture analysis by showing that the factor averaged trends are driven by module interactions rather than isolated choices alone. The benchmark results also reinforce that GELU based MLP configurations tend to be more vulnerable than ReLU based configurations, while the effect of normalization and dropout depends on the surrounding module combination.

\begin{table}[H]
    \centering
    \footnotesize
    \caption{Top raw architectural configurations on the benchmark datasets ranked by best validation utility for client batch size \(8\). Adult is
  ranked by validation ROC-AUC, the private multiclass benchmark by validation macro F1, and California Housing by validation \(R^2\). Scores are
  reported as mean \(\pm\) standard deviation across seeds at the best validation checkpoint for each configuration.}
    \label{tab:torch-modules-benchmark-top-utility-configs-b8}
    \begin{tabular}{ccccccc}
    \hline
    \textbf{Rank} & \textbf{Width} & \textbf{Layers} & \textbf{Normalization} & \textbf{Activation} & \textbf{Dropout} & Validation utility \\
    \hline
    \multicolumn{7}{c}{\textbf{Adult}} \\
    \hline
    1 & 128 & 1 & LayerNorm & GELU & 0.1 & 0.913 $\pm$ 0.005 \\
    2 & 128 & 1 & LayerNorm & GELU & 0.0 & 0.913 $\pm$ 0.005 \\
    3 & 128 & 2 & LayerNorm & GELU & 0.1 & 0.913 $\pm$ 0.005 \\
    4 & 128 & 3 & LayerNorm & GELU & 0.1 & 0.913 $\pm$ 0.005 \\
    5 & 128 & 1 & LayerNorm & ReLU & 0.1 & 0.913 $\pm$ 0.004 \\
    \hline
    \multicolumn{7}{c}{\textbf{Private multiclass}} \\
    \hline
    1 & 128 & 2 & LayerNorm & ReLU & 0.0 & 0.831 $\pm$ 0.014 \\
    2 & 128 & 1 & LayerNorm & GELU & 0.1 & 0.831 $\pm$ 0.005 \\
    3 & 64 & 2 & BatchNorm & GELU & 0.0 & 0.830 $\pm$ 0.005 \\
    4 & 128 & 2 & LayerNorm & ReLU & 0.1 & 0.830 $\pm$ 0.009 \\
    5 & 64 & 1 & LayerNorm & GELU & 0.1 & 0.830 $\pm$ 0.002 \\
    \hline
    \multicolumn{7}{c}{\textbf{California Housing}} \\
    \hline
    1 & 128 & 3 & LayerNorm & GELU & 0.0 & 0.710 $\pm$ 0.012 \\
    2 & 64 & 3 & LayerNorm & GELU & 0.0 & 0.708 $\pm$ 0.013 \\
    3 & 128 & 2 & LayerNorm & ReLU & 0.0 & 0.706 $\pm$ 0.014 \\
    4 & 128 & 2 & LayerNorm & GELU & 0.0 & 0.706 $\pm$ 0.011 \\
    5 & 128 & 3 & LayerNorm & ReLU & 0.0 & 0.705 $\pm$ 0.014 \\
    \hline
    \end{tabular}
\end{table}

\section{MIMIC-IV Sensitivity Analyses}
\label{app:mimic-sensitivity}
The MIMIC-IV sensitivity results support the main clinical benchmark analysis with full exposure trajectories, exact row recovery summaries, fixed batch trajectories, architecture grid supplements, a larger local data FedAvg control, and attack budget results.

\subsection{Leakage across client batch sizes on MIMIC-IV}
\label{app:mimic-batch-size}
Table~\ref{tab:fixed-batch-size-mimic-tableak-acc} reports the fixed client batch control for MIMIC-IV. These attacks reuse the same selected client batch across exposure scheduled attack points, separating training stage effects from variation in the sampled attacked batch.
\begin{table}[H]
    \centering
    \small
    \caption{Reconstruction accuracy at the initialized attack point and final attack point for MIMIC-IV on the same client batch. Initialized attack point denotes the attack before any global model aggregation, and final attack point denotes the last attacked point under the exposure budget. Each cell reports mean $\pm$ standard deviation across 3 seeds.}
    \label{tab:fixed-batch-size-mimic-tableak-acc}
    \resizebox{\textwidth}{!}{%
    \begin{tabular}{lcccccc}
    \hline
    \multicolumn{1}{l}{} & \multicolumn{3}{c}{\textbf{Initialized attack point}} & \multicolumn{3}{c}{\textbf{Final attack point}} \\
    \textbf{Batch size} & \textbf{FT-Transformer} & \textbf{ResNet} & \textbf{Small MLP} & \textbf{FT-Transformer} & \textbf{ResNet} & \textbf{Small MLP} \\
    \hline
    8 & 0.341 $\pm$ 0.020 & 0.785 $\pm$ 0.042 & 0.881 $\pm$ 0.024 & 0.367 $\pm$ 0.111 & 0.705 $\pm$ 0.040 & 0.642 $\pm$ 0.009 \\
    32 & 0.330 $\pm$ 0.009 & 0.605 $\pm$ 0.012 & 0.598 $\pm$ 0.049 & 0.365 $\pm$ 0.059 & 0.549 $\pm$ 0.084 & 0.471 $\pm$ 0.062 \\
    \hline
    \end{tabular}
    }
\end{table}

\begin{figure}[H]
    \centering
    \includegraphics[width=0.9\textwidth]{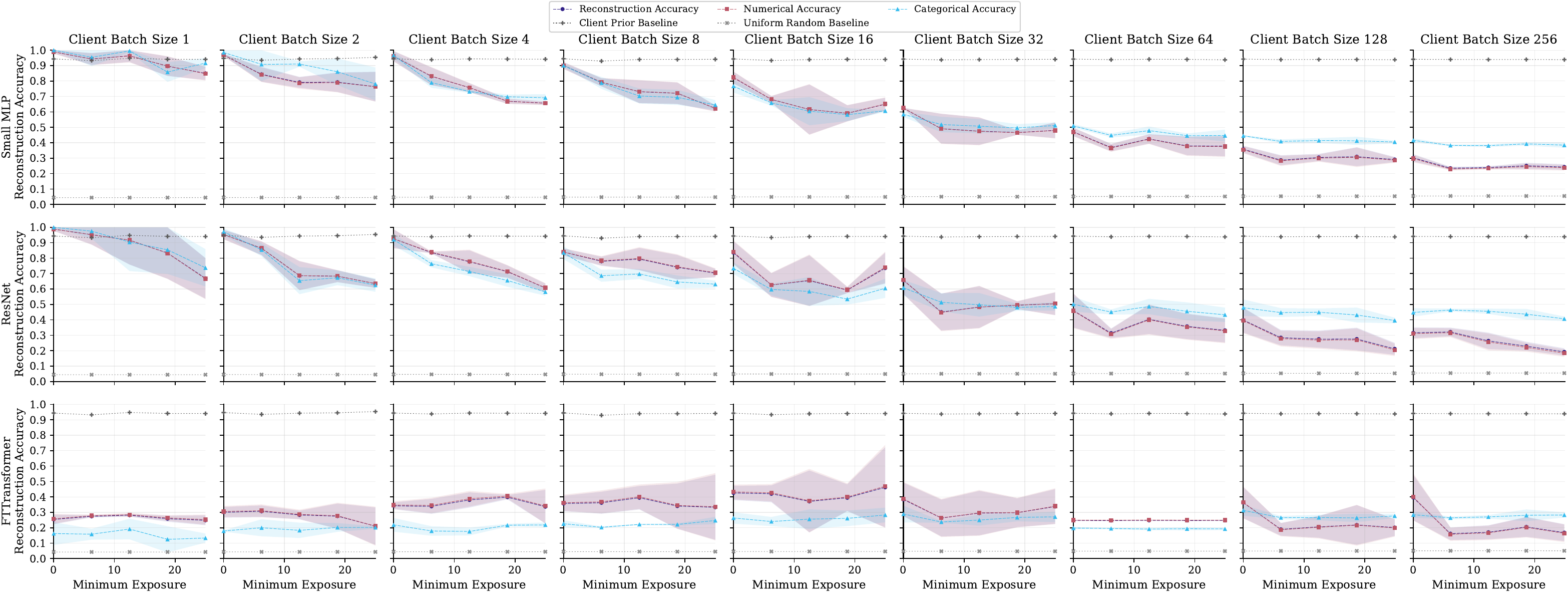}
    \caption{Reconstruction trajectories across model families and client batch sizes on MIMIC-IV.}
    \label{fig:mimic-all-models-all-batchsizes-num-cat-trajectories}
\end{figure}

\begin{figure}[H]
    \centering
    \includegraphics[width=0.7\textwidth]{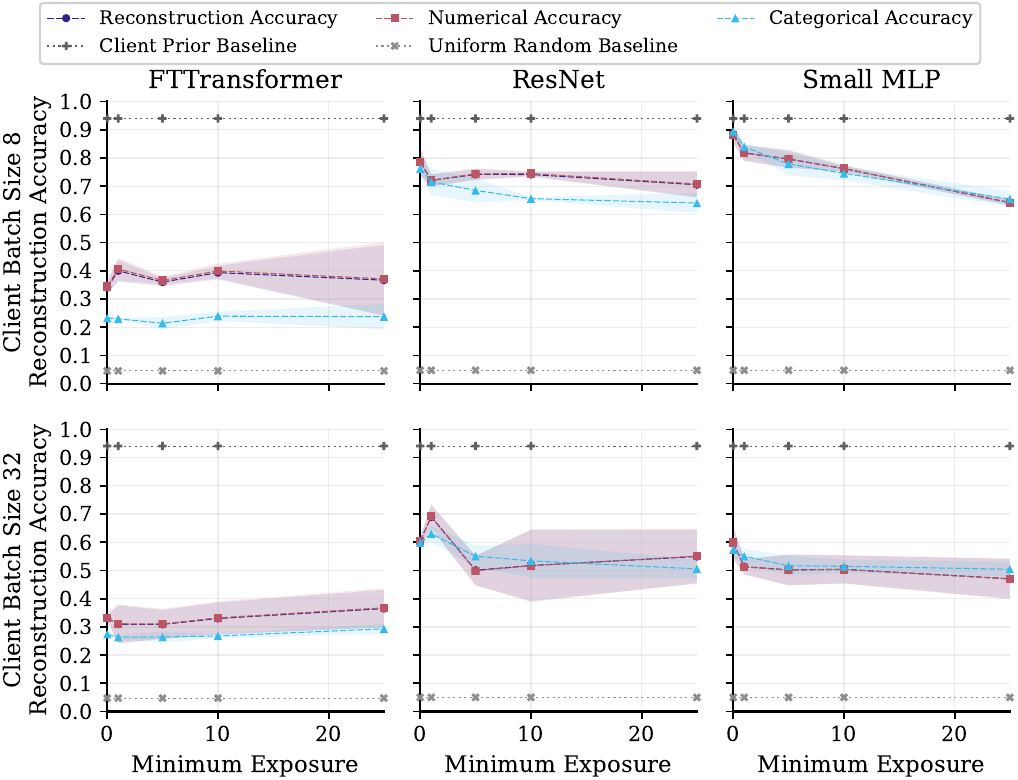}
    \caption{Reconstruction trajectories for fixed client batches across model families and client batch sizes on MIMIC-IV.}
    \label{fig:mimic-fixed-batch-all-models-num-cat-trajectories}
\end{figure}

Figures~\ref{fig:mimic-all-models-all-batchsizes-num-cat-trajectories} and~\ref{fig:mimic-fixed-batch-all-models-num-cat-trajectories} report MIMIC-IV reconstruction metrics over the five attack points used in the trajectory analyses. The first figure uses newly sampled client batches at each attack point, while the second reuses fixed client batches. Together, they show how aggregate, numerical, and categorical reconstruction vary with client exposure across model families and batch sizes.

Table~\ref{tab:batch-size-mimic-iv-baseline-reference} reports the baseline reference values for the same MIMIC-IV FedSGD batch size experiment. The client marginal prior baseline is high across all batch sizes, reflecting repeated and sparse clinical history indicators in the cohort. This means that aggregate reconstruction accuracy must be interpreted together with attack gains and EMRs. Uniform random reconstruction remains low, so values above random but below the client marginal prior indicate partial feature information rather than strong record recovery.

\begin{table}[H]
  \centering
  \small
  \caption{Client marginal prior and uniform random reconstruction baselines for MIMIC-IV in the FedSGD batch size experiment. Values are computed on the same model runs and attack points as the main reconstruction experiment and reported as mean $\pm$ standard deviation.}
  \label{tab:batch-size-mimic-iv-baseline-reference}
  \begin{tabular}{lcc}
  \hline
  \textbf{Batch size} & \textbf{Client marginal prior recon. acc.} & \textbf{Uniform random recon. acc.} \\
  \hline
  1 & 0.941 $\pm$ 0.011 & 0.044 $\pm$ 0.001 \\
  2 & 0.945 $\pm$ 0.008 & 0.044 $\pm$ 0.001 \\
  4 & 0.942 $\pm$ 0.005 & 0.046 $\pm$ 0.001 \\
  8 & 0.939 $\pm$ 0.006 & 0.047 $\pm$ 0.001 \\
  16 & 0.939 $\pm$ 0.004 & 0.048 $\pm$ 0.001 \\
  32 & 0.940 $\pm$ 0.003 & 0.049 $\pm$ 0.001 \\
  64 & 0.940 $\pm$ 0.002 & 0.051 $\pm$ 0.001 \\
  128 & 0.940 $\pm$ 0.002 & 0.052 $\pm$ 0.002 \\
  256 & 0.940 $\pm$ 0.001 & 0.053 $\pm$ 0.002 \\
  \hline
  \end{tabular}
\end{table}

Table~\ref{tab:batch-size-mimic-iv-strict-emr} reports EMRs for the MIMIC-IV FedSGD batch size experiment. These values distinguish complete row recovery from feature level reconstruction. Exact row recovery is substantial for the one-hot models at small batch sizes, but falls quickly as local aggregation increases. FT-Transformer shows no exact row recovery across the reported MIMIC-IV settings, despite nonzero feature level reconstruction accuracy.

\begin{table}[H]
  \centering
  \small
  \caption{EMR at the initialized attack point and final attack point for MIMIC-IV. Each cell reports mean $\pm$ standard deviation across 3 seeds.}
  \label{tab:batch-size-mimic-iv-strict-emr}
  \resizebox{\textwidth}{!}{%
  \begin{tabular}{lcccccc}
  \hline
  \multicolumn{1}{l}{} & \multicolumn{3}{c}{\textbf{Initialized attack point}} & \multicolumn{3}{c}{\textbf{Final attack point}} \\
  \textbf{Batch size} & \textbf{FT-Transformer} & \textbf{ResNet} & \textbf{Small MLP} & \textbf{FT-Transformer} & \textbf{ResNet} & \textbf{Small MLP} \\
  \hline
  1 & 0.000 $\pm$ 0.000 & 0.900 $\pm$ 0.000 & 0.933 $\pm$ 0.058 & 0.000 $\pm$ 0.000 & 0.500 $\pm$ 0.100 & 0.700 $\pm$ 0.100 \\
  2 & 0.000 $\pm$ 0.000 & 0.533 $\pm$ 0.153 & 0.817 $\pm$ 0.029 & 0.000 $\pm$ 0.000 & 0.450 $\pm$ 0.050 & 0.417 $\pm$ 0.076 \\
  4 & 0.000 $\pm$ 0.000 & 0.250 $\pm$ 0.152 & 0.800 $\pm$ 0.066 & 0.000 $\pm$ 0.000 & 0.267 $\pm$ 0.088 & 0.233 $\pm$ 0.080 \\
  8 & 0.000 $\pm$ 0.000 & 0.083 $\pm$ 0.059 & 0.342 $\pm$ 0.044 & 0.000 $\pm$ 0.000 & 0.196 $\pm$ 0.019 & 0.104 $\pm$ 0.019 \\
  16 & 0.000 $\pm$ 0.000 & 0.006 $\pm$ 0.006 & 0.044 $\pm$ 0.029 & 0.000 $\pm$ 0.000 & 0.069 $\pm$ 0.029 & 0.056 $\pm$ 0.013 \\
  32 & 0.000 $\pm$ 0.000 & 0.000 $\pm$ 0.000 & 0.001 $\pm$ 0.002 & 0.000 $\pm$ 0.000 & 0.009 $\pm$ 0.003 & 0.021 $\pm$ 0.005 \\
  64 & 0.000 $\pm$ 0.000 & 0.000 $\pm$ 0.000 & 0.002 $\pm$ 0.002 & 0.000 $\pm$ 0.000 & 0.002 $\pm$ 0.002 & 0.006 $\pm$ 0.001 \\
  128 & 0.000 $\pm$ 0.000 & 0.000 $\pm$ 0.000 & 0.000 $\pm$ 0.000 & 0.000 $\pm$ 0.000 & 0.000 $\pm$ 0.000 & 0.001 $\pm$ 0.001 \\
  256 & 0.000 $\pm$ 0.000 & 0.000 $\pm$ 0.000 & 0.000 $\pm$ 0.000 & 0.000 $\pm$ 0.000 & 0.000 $\pm$ 0.000 & 0.000 $\pm$ 0.000 \\
  \hline
  \end{tabular}
  }
\end{table}

Table~\ref{tab:batch-size-mimic-fttransformer-attack-paths-dropout0-emr} reports the corresponding EMRs for the dropout disabled FT-Transformer runs. Complete row recovery remains absent for probability simplex at all reported batch sizes and is observed for categorical logits only in the most exposed batch size \(1\) setting.

\begin{table}[H]
    \centering
    \small
    \caption{EMR at the initialized attack point and final attack point for MIMIC-IV under two FT-Transformer attack parameterizations with attention and feedforward dropout set to zero in the target model. Each cell reports mean $\pm$ standard deviation across 3 seeds.}
    \label{tab:batch-size-mimic-fttransformer-attack-paths-dropout0-emr}
    \begin{tabular}{lcccc}
    \hline
    \multicolumn{1}{l}{} & \multicolumn{2}{c}{\textbf{Initialized attack point}} & \multicolumn{2}{c}{\textbf{Final attack point}} \\
    \textbf{Batch size} & \textbf{Categorical logits} & \textbf{Probability simplex} & \textbf{Categorical logits} & \textbf{Probability simplex} \\
    \hline
    1   & 0.033 $\pm$ 0.058 & 0.000 $\pm$ 0.000 & 0.167 $\pm$ 0.289 & 0.000 $\pm$ 0.000 \\
    2   & 0.000 $\pm$ 0.000 & 0.000 $\pm$ 0.000 & 0.000 $\pm$ 0.000 & 0.000 $\pm$ 0.000 \\
    4   & 0.000 $\pm$ 0.000 & 0.000 $\pm$ 0.000 & 0.000 $\pm$ 0.000 & 0.000 $\pm$ 0.000 \\
    8   & 0.000 $\pm$ 0.000 & 0.000 $\pm$ 0.000 & 0.000 $\pm$ 0.000 & 0.000 $\pm$ 0.000 \\
    16  & 0.000 $\pm$ 0.000 & 0.000 $\pm$ 0.000 & 0.000 $\pm$ 0.000 & 0.000 $\pm$ 0.000 \\
    32  & 0.000 $\pm$ 0.000 & 0.000 $\pm$ 0.000 & 0.000 $\pm$ 0.000 & 0.000 $\pm$ 0.000 \\
    64  & 0.000 $\pm$ 0.000 & 0.000 $\pm$ 0.000 & 0.000 $\pm$ 0.000 & 0.000 $\pm$ 0.000 \\
    128 & 0.000 $\pm$ 0.000 & 0.000 $\pm$ 0.000 & 0.000 $\pm$ 0.000 & 0.000 $\pm$ 0.000 \\
    \hline
    \end{tabular}
\end{table}

\subsection{MIMIC-IV MLP architecture sensitivity}
\begin{figure}[H]
      \centering

      \begin{subfigure}[t]{0.48\textwidth}
          \centering
          \includegraphics[width=\textwidth]{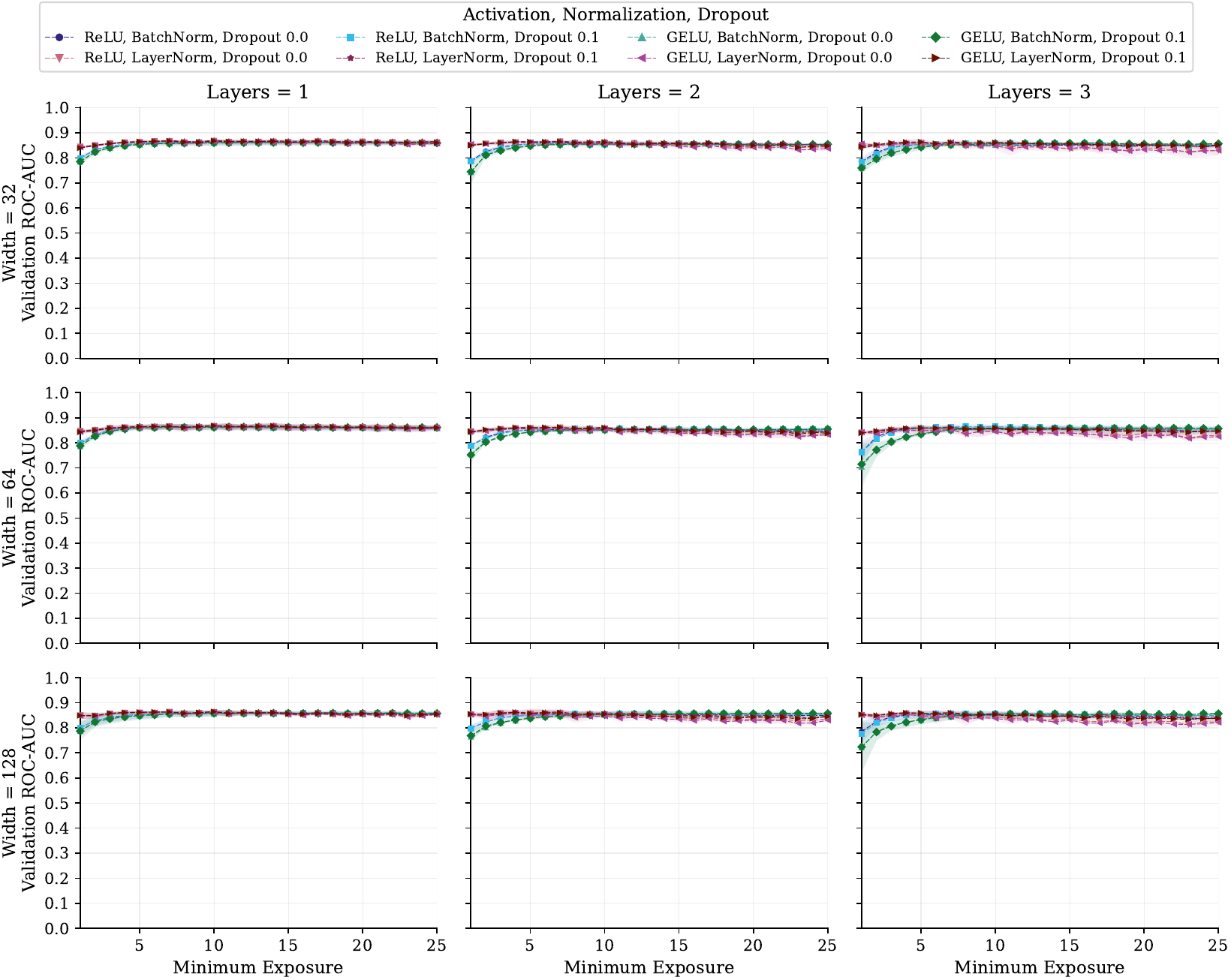}
          \caption{Utility curve ROC-AUC}
      \end{subfigure}
      \hfill
      \begin{subfigure}[t]{0.48\textwidth}
          \centering
          \includegraphics[width=\textwidth]{plots/fesgd/mimic/torch_modules/fedsgd_torch_modules_config_attack_grid__dataset_mimic_admission_tier3_binary_train__batch_8.pdf}
          \caption{Reconstruction accuracy}
      \end{subfigure}

      \vspace{0.5em}

      \begin{subfigure}[t]{0.48\textwidth}
          \centering
          \includegraphics[width=\textwidth]{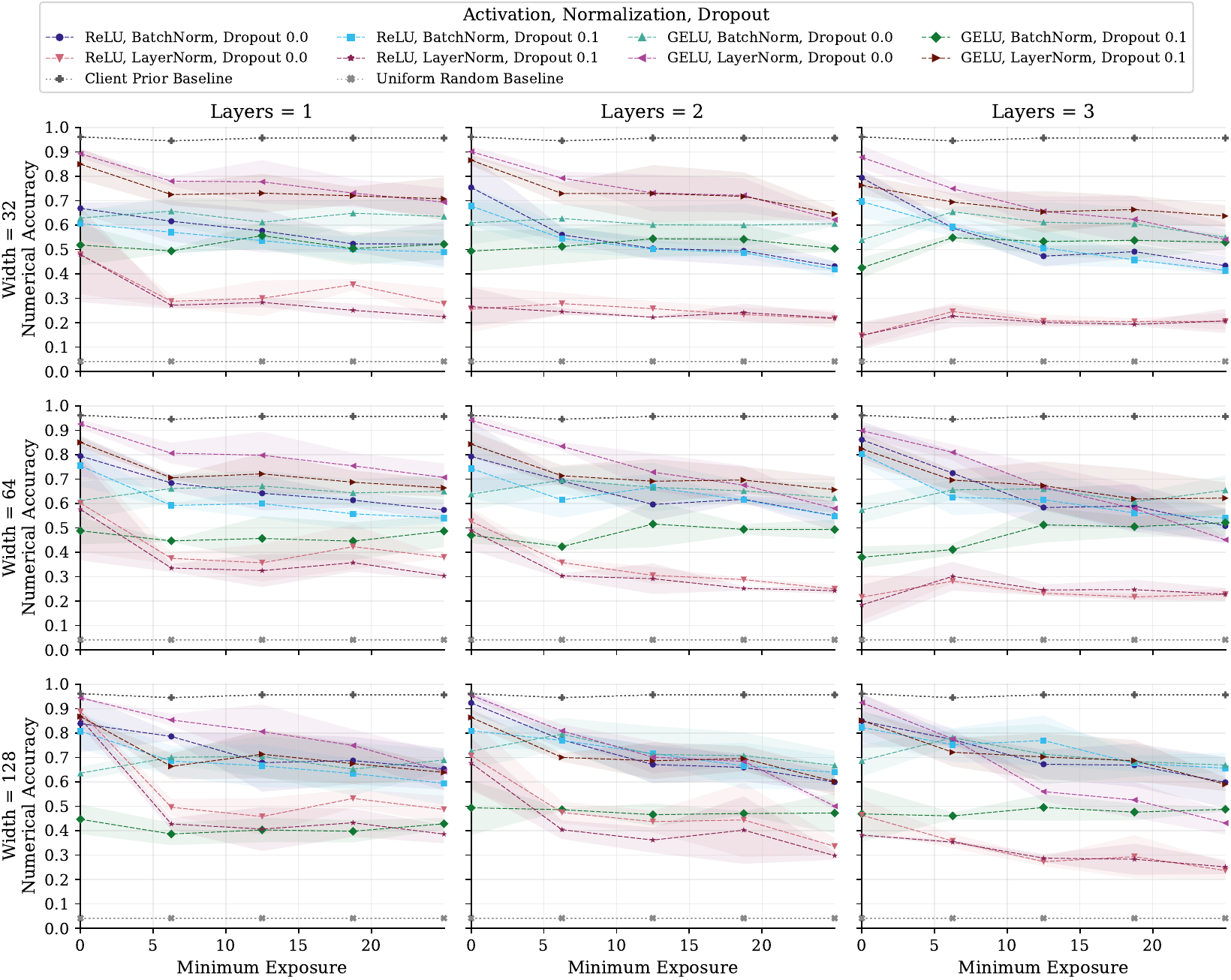}
          \caption{Numerical accuracy}
      \end{subfigure}
      \hfill
      \begin{subfigure}[t]{0.48\textwidth}
          \centering
          \includegraphics[width=\textwidth]{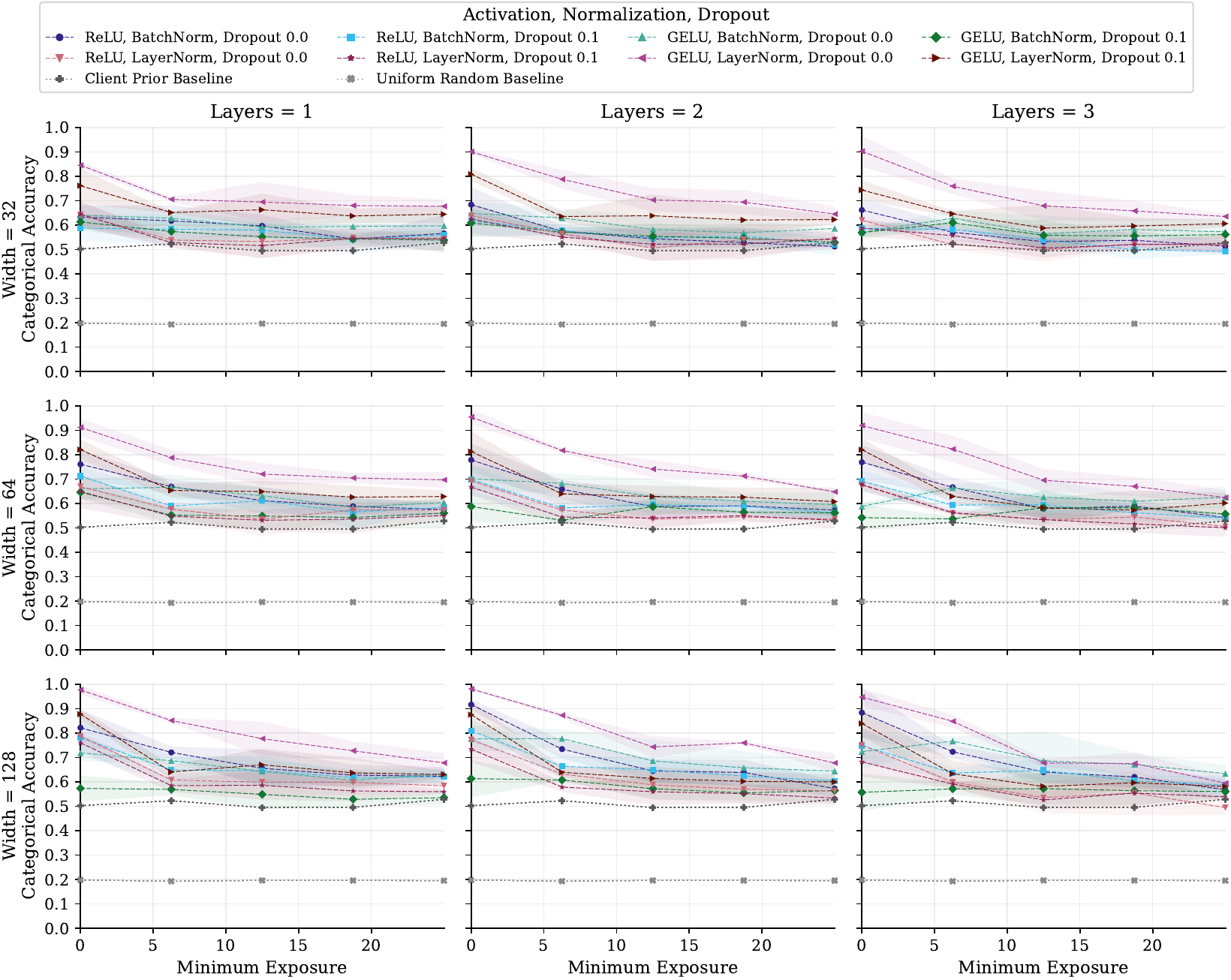}
          \caption{Categorical accuracy}
      \end{subfigure}

      \caption{Utility and inversion accuracy metrics over FL training for the MLP architecture grid using client batch size 8 on MIMIC-IV.}
      \label{fig:mimic-torch-modules-privacy-utility-b8}
\end{figure}

\begin{figure}[H]
      \centering

      \begin{subfigure}[t]{0.48\textwidth}
          \centering
          \includegraphics[width=\textwidth]{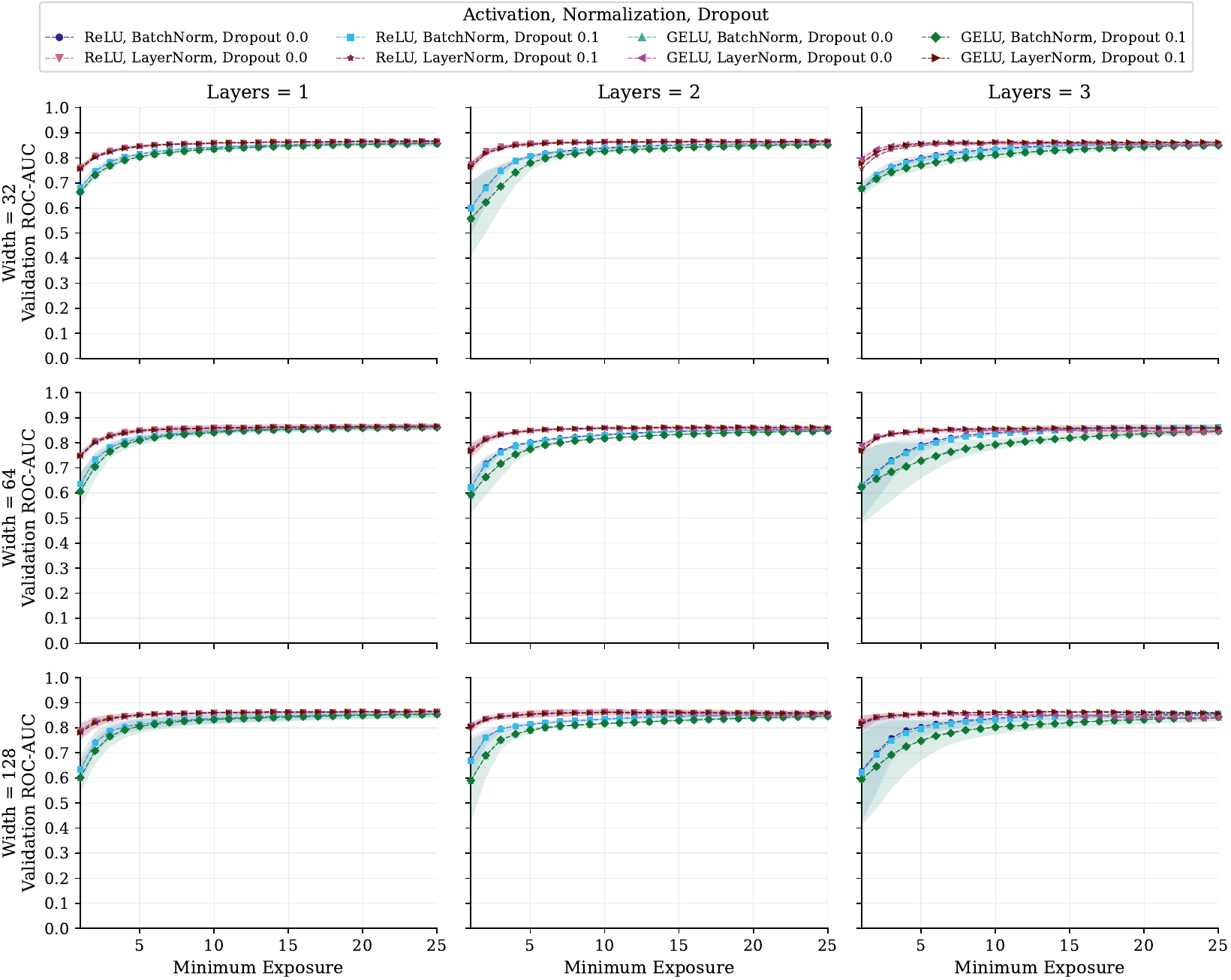}
          \caption{Utility curve ROC-AUC}
      \end{subfigure}
      \hfill
      \begin{subfigure}[t]{0.48\textwidth}
          \centering
          \includegraphics[width=\textwidth]{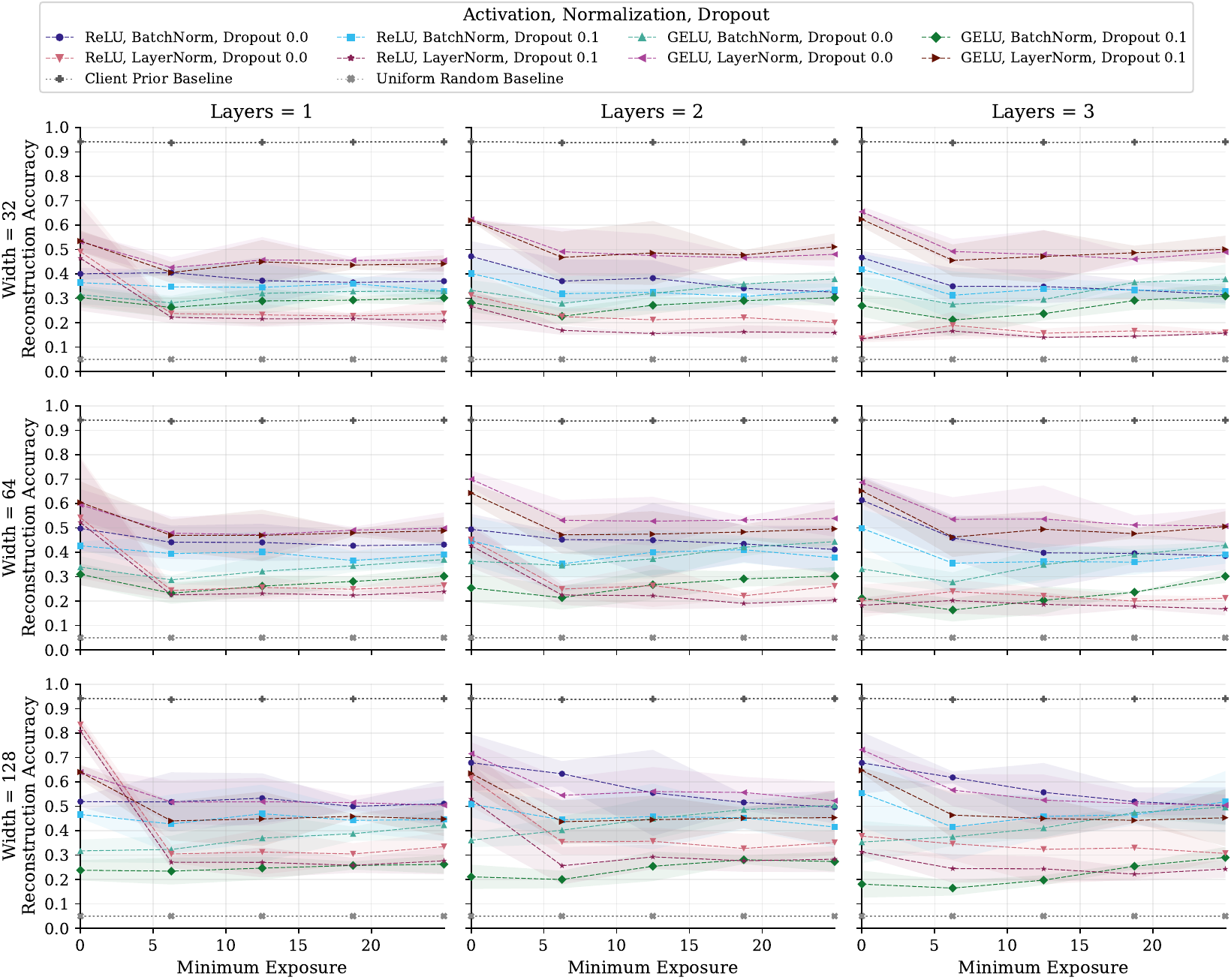}
          \caption{Reconstruction accuracy}
      \end{subfigure}

      \vspace{0.5em}

      \begin{subfigure}[t]{0.48\textwidth}
          \centering
          \includegraphics[width=\textwidth]{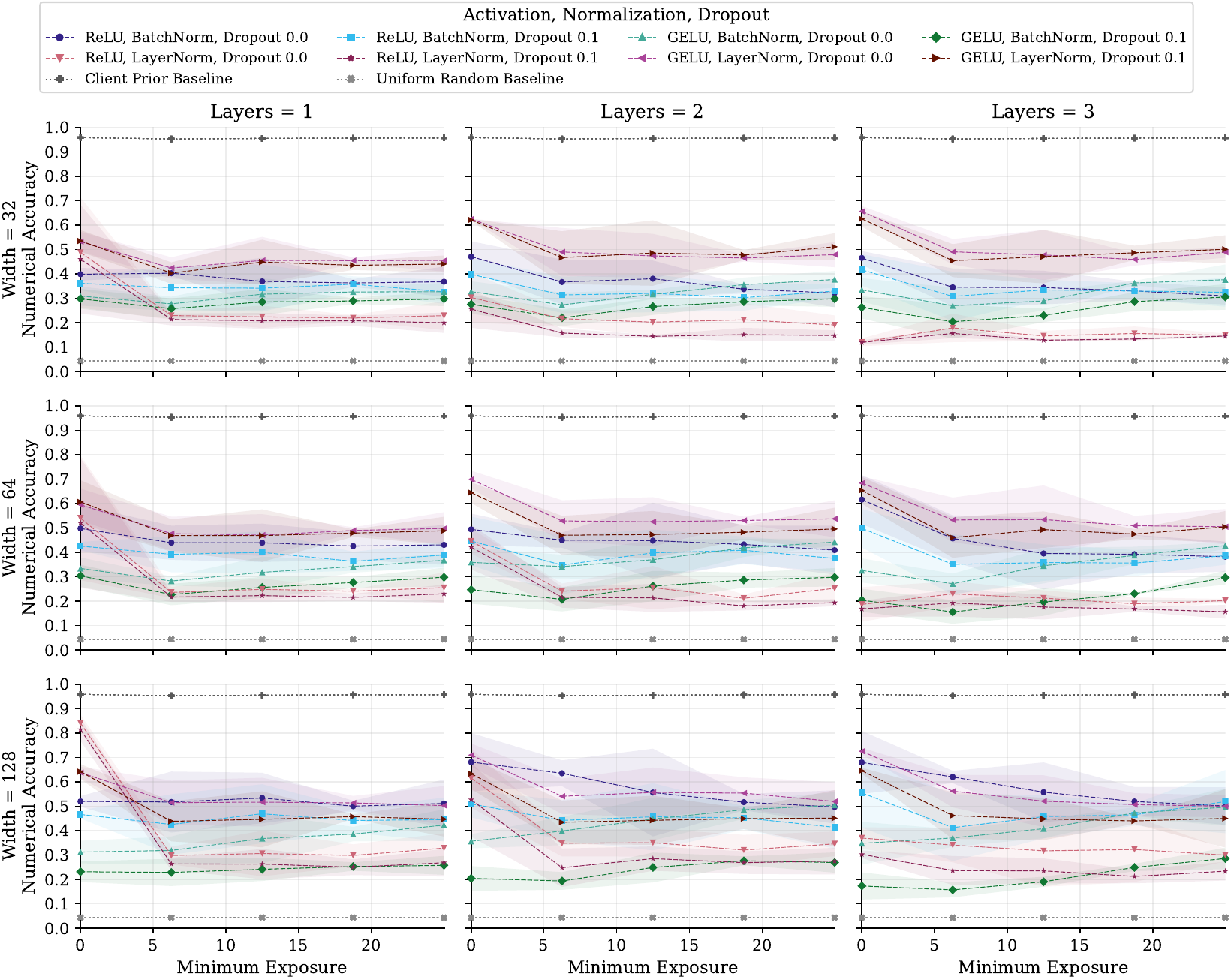}
          \caption{Numerical accuracy}
      \end{subfigure}
      \hfill
      \begin{subfigure}[t]{0.48\textwidth}
          \centering
          \includegraphics[width=\textwidth]{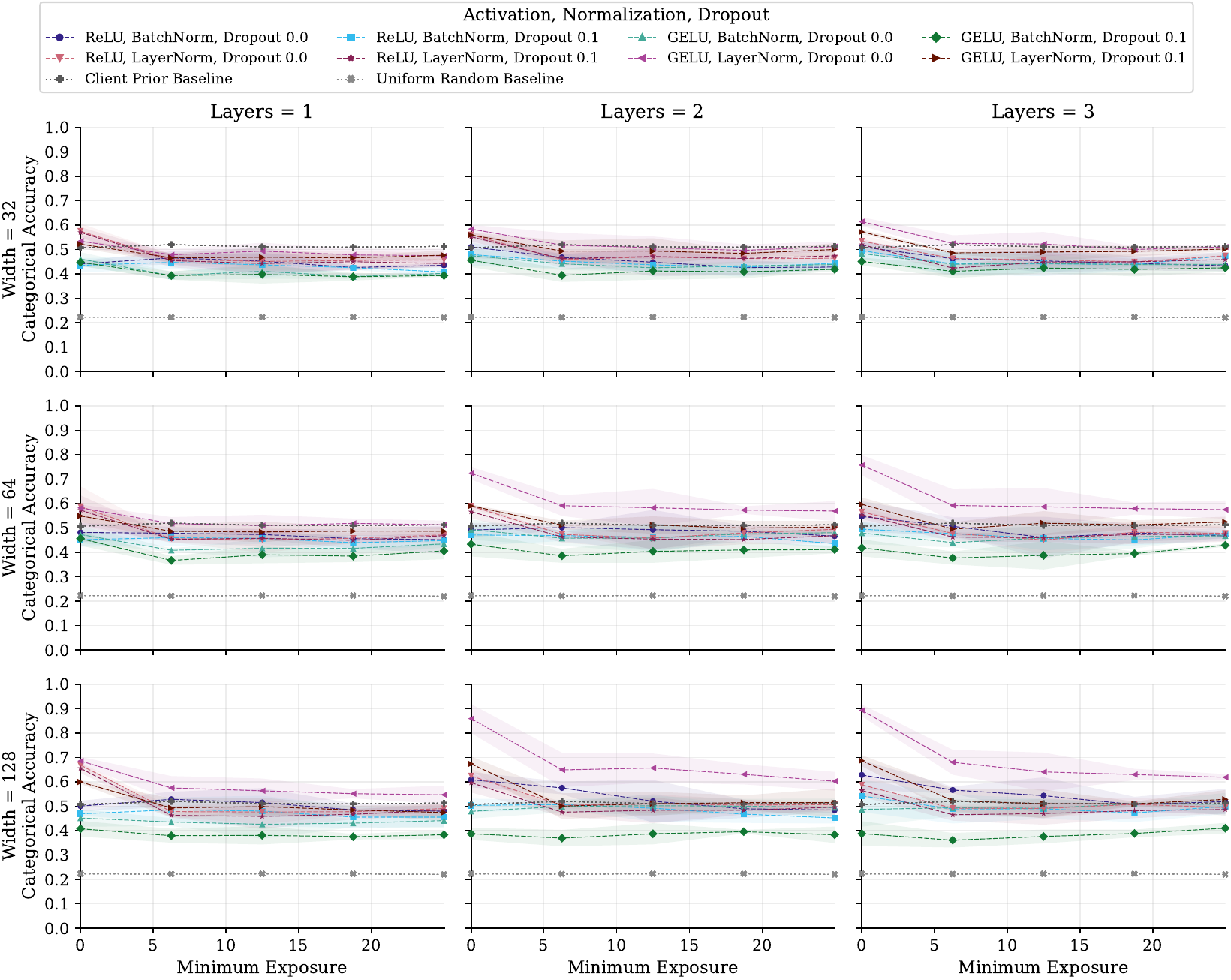}
          \caption{Categorical accuracy}
      \end{subfigure}

      \caption{Utility and inversion accuracy metrics over FL training for the MLP architecture grid using client batch size 32 on MIMIC-IV.}
      \label{fig:mimic-torch-modules-privacy-utility-b32}
\end{figure}

The MIMIC-IV MLP architecture supplement contains the grid material not included in the main text. The appendix tables report the complete module combination ranking for batch sizes \(8\) and \(32\), together with utility ranked configurations. Figures~\ref{fig:mimic-torch-modules-privacy-utility-b8} and~\ref{fig:mimic-torch-modules-privacy-utility-b32} report utility, aggregate reconstruction, numerical reconstruction, and categorical reconstruction for batch sizes \(8\) and \(32\).

\begin{table}[H]
  \centering
  \small
  \caption{Complete ranking of module combinations by final attack point reconstruction accuracy on MIMIC-IV for batch sizes 8 and 32. Rows are ordered from lowest to highest reconstruction accuracy. Reconstruction accuracy is averaged over width and depth and reported at the final attack point.}
  \label{tab:torch-modules-mimic-module-ranking}
  \begin{tabular}{ccccc}
  \hline
  \textbf{Rank} & \textbf{Normalization} & \textbf{Activation} & \textbf{Dropout} & \textbf{Recon. Acc.} \\
  \hline
  \multicolumn{5}{c}{Batch size 8} \\
  \hline
  1 & LayerNorm & ReLU & 0.1 & 0.272 $\pm$ 0.056 \\
  2 & LayerNorm & ReLU & 0.0 & 0.300 $\pm$ 0.091 \\
  3 & BatchNorm & GELU & 0.1 & 0.496 $\pm$ 0.030 \\
  4 & BatchNorm & ReLU & 0.1 & 0.539 $\pm$ 0.084 \\
  5 & BatchNorm & ReLU & 0.0 & 0.542 $\pm$ 0.074 \\
  6 & LayerNorm & GELU & 0.0 & 0.577 $\pm$ 0.098 \\
  7 & BatchNorm & GELU & 0.0 & 0.637 $\pm$ 0.041 \\
  8 & LayerNorm & GELU & 0.1 & 0.640 $\pm$ 0.034 \\
  \hline
  \multicolumn{5}{c}{Batch size 32} \\
  \hline
  1 & LayerNorm & ReLU & 0.1 & 0.215 $\pm$ 0.048 \\
  2 & LayerNorm & ReLU & 0.0 & 0.259 $\pm$ 0.064 \\
  3 & BatchNorm & GELU & 0.1 & 0.294 $\pm$ 0.016 \\
  4 & BatchNorm & ReLU & 0.1 & 0.393 $\pm$ 0.062 \\
  5 & BatchNorm & ReLU & 0.0 & 0.416 $\pm$ 0.075 \\
  6 & BatchNorm & GELU & 0.0 & 0.417 $\pm$ 0.059 \\
  7 & LayerNorm & GELU & 0.1 & 0.478 $\pm$ 0.028 \\
  8 & LayerNorm & GELU & 0.0 & 0.500 $\pm$ 0.024 \\
  \hline
  \end{tabular}
\end{table}

\begin{table}[H]
  \centering
  \small
  \caption{Top raw architectural configurations on MIMIC-IV ranked by best validation ROC-AUC for batch size 8. Scores are reported as mean $\pm$ standard deviation across seeds at the best validation checkpoint for each configuration.}
  \label{tab:torch-modules-mimic-top-utility-configs-b8}
  \begin{tabular}{ccccccc}
  \hline
  \textbf{Rank} & \textbf{Width} & \textbf{Layers} & \textbf{Normalization} & \textbf{Activation} & \textbf{Dropout} & \textbf{Val ROC-AUC} \\
  \hline
  1 & 32 & 1 & LayerNorm & ReLU & 0.1 & 0.871 $\pm$ 0.002 \\
  2 & 64 & 1 & LayerNorm & ReLU & 0.1 & 0.869 $\pm$ 0.010 \\
  3 & 64 & 1 & LayerNorm & GELU & 0.1 & 0.868 $\pm$ 0.006 \\
  4 & 32 & 1 & LayerNorm & ReLU & 0.0 & 0.867 $\pm$ 0.003 \\
  5 & 32 & 1 & LayerNorm & GELU & 0.1 & 0.867 $\pm$ 0.000 \\
  \hline
  \end{tabular}
\end{table}

\begin{table}[H]
  \centering
  \small
  \caption{Top raw architectural configurations on MIMIC-IV ranked by best validation ROC-AUC for batch size 32. Scores are reported as mean $\pm$ standard deviation across seeds at the best validation checkpoint for each configuration.}
  \label{tab:torch-modules-mimic-top-utility-configs-b32}
  \begin{tabular}{ccccccc}
  \hline
  \textbf{Rank} & \textbf{Width} & \textbf{Layers} & \textbf{Normalization} & \textbf{Activation} & \textbf{Dropout} & \textbf{Val ROC-AUC} \\
  \hline
  1 & 32 & 1 & LayerNorm & ReLU & 0.1 & 0.868 $\pm$ 0.005 \\
  2 & 32 & 2 & LayerNorm & ReLU & 0.1 & 0.868 $\pm$ 0.005 \\
  3 & 32 & 1 & LayerNorm & GELU & 0.1 & 0.867 $\pm$ 0.005 \\
  4 & 32 & 2 & LayerNorm & GELU & 0.1 & 0.867 $\pm$ 0.005 \\
  5 & 64 & 1 & LayerNorm & GELU & 0.1 & 0.866 $\pm$ 0.009 \\
  \hline
  \end{tabular}
\end{table}

\subsection{FedAvg with 64 local examples}
\label{app:fedavg-local-computation}
The 64-example FedAvg control repeats the MIMIC-IV local computation analysis with a larger fixed client dataset. It is included because increasing the fixed client dataset also changes the local mini batch size associated with each mini batch count. Table~\ref{tab:fedavg-mimic-tableak-acc-maxcap64} reports this control and should therefore be read as a FedAvg robustness check rather than as an isolated test of local epoch count.

\begin{table}[H]
      \centering
      \small
      \caption{Reconstruction accuracy at the initialized attack point and final attack point for MIMIC-IV under FedAvg with 64 local examples per client. Initialized attack point denotes the first client update computed from the initial global model, and final attack point denotes the last attacked point under the exposure budget. The row variable \(n.\) batches denotes how many local mini batches are needed to process the local client data once. Each cell reports mean $\pm$ standard deviation across 5 seeds.}
      \label{tab:fedavg-mimic-tableak-acc-maxcap64}
      \resizebox{\textwidth}{!}{%
      \begin{tabular}{lcccccc}
      \hline
      \multicolumn{1}{l}{} & \multicolumn{3}{c}{\textbf{Initialized attack point}} & \multicolumn{3}{c}{\textbf{Final attack point}} \\
      \textbf{n. batches} & \textbf{FT-Transformer} & \textbf{ResNet} & \textbf{Small MLP} & \textbf{FT-Transformer} & \textbf{ResNet} & \textbf{Small MLP} \\
      \hline
      \multicolumn{7}{c}{\textbf{1 local epoch}} \\
      \hline
      1 & 0.326 $\pm$ 0.104 & 0.420 $\pm$ 0.085 & 0.417 $\pm$ 0.035 & 0.215 $\pm$ 0.081 & 0.494 $\pm$ 0.079 & 0.373 $\pm$ 0.024 \\
      2 & 0.269 $\pm$ 0.101 & 0.456 $\pm$ 0.100 & 0.426 $\pm$ 0.041 & 0.184 $\pm$ 0.071 & 0.480 $\pm$ 0.077 & 0.364 $\pm$ 0.026 \\
      4 & 0.246 $\pm$ 0.074 & 0.441 $\pm$ 0.086 & 0.417 $\pm$ 0.055 & 0.190 $\pm$ 0.075 & 0.426 $\pm$ 0.037 & 0.362 $\pm$ 0.049 \\
      \hline
      \multicolumn{7}{c}{\textbf{2 local epochs}} \\
      \hline
      1 & 0.268 $\pm$ 0.116 & 0.480 $\pm$ 0.096 & 0.423 $\pm$ 0.047 & 0.200 $\pm$ 0.079 & 0.540 $\pm$ 0.078 & 0.367 $\pm$ 0.022 \\
      2 & 0.243 $\pm$ 0.087 & 0.495 $\pm$ 0.100 & 0.408 $\pm$ 0.049 & 0.181 $\pm$ 0.104 & 0.522 $\pm$ 0.081 & 0.362 $\pm$ 0.042 \\
      4 & 0.221 $\pm$ 0.064 & 0.508 $\pm$ 0.076 & 0.396 $\pm$ 0.041 & 0.191 $\pm$ 0.077 & 0.465 $\pm$ 0.051 & 0.358 $\pm$ 0.050 \\
      \hline
      \multicolumn{7}{c}{\textbf{4 local epochs}} \\
      \hline
      1 & 0.245 $\pm$ 0.111 & 0.523 $\pm$ 0.086 & 0.403 $\pm$ 0.039 & 0.203 $\pm$ 0.082 & 0.553 $\pm$ 0.080 & 0.365 $\pm$ 0.032 \\
      2 & 0.235 $\pm$ 0.091 & 0.524 $\pm$ 0.111 & 0.393 $\pm$ 0.045 & 0.201 $\pm$ 0.083 & 0.546 $\pm$ 0.099 & 0.361 $\pm$ 0.042 \\
      4 & 0.213 $\pm$ 0.073 & 0.527 $\pm$ 0.089 & 0.386 $\pm$ 0.039 & 0.200 $\pm$ 0.088 & 0.466 $\pm$ 0.045 & 0.358 $\pm$ 0.035 \\
      \hline
      \end{tabular}
      }
\end{table}

\subsection{MIMIC-IV non-IID client partitions}
The non-IID control tests whether client partitioning changes the MIMIC-IV leakage patterns observed under the default IID setting. Table~\ref{tab:mimic-noniid-summary} reports client batch size \(32\), and Table~\ref{tab:appendix-mimic-noniid-b8} reports the corresponding batch size \(8\) results. Together, these tables compare Dirichlet concentration values \(\alpha \in \{1.0,0.6,0.4\}\) across the three model families.

\begin{table}[H]
    \centering
    \small
    \caption{MIMIC-IV non-IID sensitivity under FedSGD at client batch size \(8\). Initialized attack point denotes the first attacked point, and final attack point denotes the last attacked point under the exposure budget. Numerical and categorical accuracies are reported at the final attack point. Each cell reports mean $\pm$ standard deviation across 3 seeds.}
    \label{tab:appendix-mimic-noniid-b8}
    \begin{tabular}{llcccc}
    \hline
    \multicolumn{2}{c}{} & \textbf{Initialized attack point} & \multicolumn{3}{c}{\textbf{Final attack point}} \\
    \textbf{Model} & \textbf{Split} & \textbf{Recon. Acc.} & \textbf{Recon. Acc.} & \textbf{Num. acc.} & \textbf{Cat. acc.} \\
    \hline
    FT-Transformer & \(\alpha=1.0\) & 0.365 $\pm$ 0.053 & 0.138 $\pm$ 0.044 & 0.132 $\pm$ 0.047 & 0.298 $\pm$ 0.025 \\
    FT-Transformer & \(\alpha=0.6\) & 0.348 $\pm$ 0.056 & 0.083 $\pm$ 0.030 & 0.077 $\pm$ 0.031 & 0.239 $\pm$ 0.050 \\
    FT-Transformer & \(\alpha=0.4\) & 0.370 $\pm$ 0.046 & 0.098 $\pm$ 0.027 & 0.091 $\pm$ 0.027 & 0.265 $\pm$ 0.028 \\
    \hline
    ResNet & \(\alpha=1.0\) & 0.764 $\pm$ 0.018 & 0.370 $\pm$ 0.030 & 0.369 $\pm$ 0.029 & 0.401 $\pm$ 0.039 \\
    ResNet & \(\alpha=0.6\) & 0.742 $\pm$ 0.128 & 0.264 $\pm$ 0.007 & 0.260 $\pm$ 0.006 & 0.361 $\pm$ 0.034 \\
    ResNet & \(\alpha=0.4\) & 0.769 $\pm$ 0.051 & 0.285 $\pm$ 0.055 & 0.282 $\pm$ 0.055 & 0.347 $\pm$ 0.057 \\
    \hline
    Small MLP & \(\alpha=1.0\) & 0.885 $\pm$ 0.036 & 0.436 $\pm$ 0.046 & 0.431 $\pm$ 0.048 & 0.555 $\pm$ 0.012 \\
    Small MLP & \(\alpha=0.6\) & 0.851 $\pm$ 0.058 & 0.340 $\pm$ 0.052 & 0.336 $\pm$ 0.052 & 0.447 $\pm$ 0.063 \\
    Small MLP & \(\alpha=0.4\) & 0.892 $\pm$ 0.033 & 0.380 $\pm$ 0.123 & 0.378 $\pm$ 0.126 & 0.447 $\pm$ 0.057 \\
    \hline
    \end{tabular}
\end{table}

\begin{table}[H]
    \centering
    \small
    \caption{MIMIC-IV non-IID sensitivity under FedSGD at client batch size \(32\). Initialized attack point denotes the first attacked point, and final attack point denotes the last attacked point under the exposure budget. Numerical and categorical accuracies are reported at the final attack point. Each cell reports mean $\pm$ standard deviation across 3 seeds.}
    \label{tab:mimic-noniid-summary}
    \begin{tabular}{llcccc}
    \hline
    \multicolumn{2}{c}{} & \textbf{Initialized attack point} & \multicolumn{3}{c}{\textbf{Final attack point}} \\
    \textbf{Model} & \textbf{Split} & \textbf{Recon. Acc.} & \textbf{Recon. Acc.} & \textbf{Num. acc.} & \textbf{Cat. acc.} \\
    \hline
    FT-Transformer & \(\alpha=1.0\) & 0.361 $\pm$ 0.114 & 0.227 $\pm$ 0.010 & 0.225 $\pm$ 0.011 & 0.267 $\pm$ 0.011 \\
    FT-Transformer & \(\alpha=0.6\) & 0.342 $\pm$ 0.023 & 0.173 $\pm$ 0.058 & 0.169 $\pm$ 0.061 & 0.290 $\pm$ 0.028 \\
    FT-Transformer & \(\alpha=0.4\) & 0.408 $\pm$ 0.049 & 0.150 $\pm$ 0.008 & 0.145 $\pm$ 0.008 & 0.282 $\pm$ 0.009 \\
    \hline
    ResNet & \(\alpha=1.0\) & 0.606 $\pm$ 0.051 & 0.374 $\pm$ 0.072 & 0.372 $\pm$ 0.074 & 0.415 $\pm$ 0.016 \\
    ResNet & \(\alpha=0.6\) & 0.541 $\pm$ 0.056 & 0.284 $\pm$ 0.019 & 0.280 $\pm$ 0.019 & 0.384 $\pm$ 0.009 \\
    ResNet & \(\alpha=0.4\) & 0.602 $\pm$ 0.030 & 0.306 $\pm$ 0.068 & 0.304 $\pm$ 0.069 & 0.368 $\pm$ 0.041 \\
    \hline
    Small MLP & \(\alpha=1.0\) & 0.612 $\pm$ 0.030 & 0.484 $\pm$ 0.035 & 0.483 $\pm$ 0.036 & 0.508 $\pm$ 0.002 \\
    Small MLP & \(\alpha=0.6\) & 0.584 $\pm$ 0.016 & 0.343 $\pm$ 0.125 & 0.338 $\pm$ 0.128 & 0.466 $\pm$ 0.036 \\
    Small MLP & \(\alpha=0.4\) & 0.610 $\pm$ 0.021 & 0.339 $\pm$ 0.087 & 0.334 $\pm$ 0.090 & 0.462 $\pm$ 0.013 \\
    \hline
    \end{tabular}
\end{table}

Client heterogeneity changes reconstruction levels, but it does not overturn the broader hierarchy of privacy factors. Averaged over models and batch sizes, final attack point reconstruction is highest for the split closest to IID, with \(0.338\) at \(\alpha=1.0\), compared with \(0.248\) at \(\alpha=0.6\) and \(0.260\) at \(\alpha=0.4\). Thus, lower \(\alpha\) does not systematically increase leakage in this setting. The model ordering is more stable than the \(\alpha\)-effect: FT-Transformer remains least vulnerable, with final attack point reconstruction of \(0.145\), followed by ResNet at \(0.314\) and the small MLP at \(0.387\). Batch size remains relevant, especially early in training, where increasing the batch size from \(8\) to \(32\) reduces initialized reconstruction from \(0.665\) to \(0.519\). By the final attack point, this difference is smaller and slightly reversed, with \(0.266\) for batch size \(8\) and \(0.298\) for batch size \(32\), because the smaller batch runs lose more reconstructive information over training. Overall, the non-IID results refine rather than contradict the main findings: client heterogeneity changes leakage in the clinical cohort, but architecture and local aggregation remain more reliable privacy factors.

\subsection{Attack optimization budget on MIMIC-IV}
Tables~\ref{tab:mimic-attack-budget-b32-tableak-acc} and~\ref{tab:appendix-mimic-attack-budget-b8-tableak-acc} report the MIMIC-IV attack budget sensitivity analysis for client batch sizes \(32\) and \(8\). The experiment varies the number of attack optimization iterations while keeping the FL training configuration fixed. Batch size \(32\) represents the less exposed intermediate setting, while batch size \(8\) provides a more exposed counterpart.

\begin{table}[H]
    \centering
    \small
    \caption{MIMIC-IV attack budget sensitivity under FedSGD at client batch size \(8\). The row variable gives the number of attack optimization iterations. Results are shown for the initialized attack point and the final attack point. Each cell reports reconstruction accuracy as mean \(\pm\) standard deviation across 3 seeds.}
    \label{tab:appendix-mimic-attack-budget-b8-tableak-acc}
    \resizebox{\textwidth}{!}{%
    \begin{tabular}{lcccccc}
    \hline
    \multicolumn{1}{l}{} & \multicolumn{3}{c}{\textbf{Initialized attack point}} & \multicolumn{3}{c}{\textbf{Final attack point}} \\
    \textbf{Attack iterations} & \textbf{FT-Transformer} & \textbf{ResNet} & \textbf{Small MLP} & \textbf{FT-Transformer} & \textbf{ResNet} & \textbf{Small MLP} \\
    \hline
    5{,}000  & 0.381 $\pm$ 0.030 & 0.801 $\pm$ 0.037 & 0.864 $\pm$ 0.039 & 0.313 $\pm$ 0.129 & 0.743 $\pm$ 0.019 & 0.631 $\pm$ 0.054 \\
    10{,}000 & 0.391 $\pm$ 0.029 & 0.785 $\pm$ 0.039 & 0.891 $\pm$ 0.035 & 0.322 $\pm$ 0.127 & 0.731 $\pm$ 0.030 & 0.620 $\pm$ 0.080 \\
    20{,}000 & 0.394 $\pm$ 0.033 & 0.787 $\pm$ 0.032 & 0.897 $\pm$ 0.035 & 0.324 $\pm$ 0.126 & 0.692 $\pm$ 0.033 & 0.613 $\pm$ 0.070 \\
    \hline
    \end{tabular}
    }
\end{table}

\begin{table}[H]
    \centering
    \small
    \caption{MIMIC-IV attack budget sensitivity under FedSGD at client batch size \(32\). The row variable gives the number of attack optimization iterations. Results are shown for the initialized attack point and the final attack point. Each cell reports reconstruction accuracy as mean \(\pm\) standard deviation across 3 seeds.}
    \label{tab:mimic-attack-budget-b32-tableak-acc}
    \resizebox{\textwidth}{!}{%
    \begin{tabular}{lcccccc}
    \hline
    \multicolumn{1}{l}{} & \multicolumn{3}{c}{\textbf{Initialized attack point}} & \multicolumn{3}{c}{\textbf{Final attack point}} \\
    \textbf{Attack iterations} & \textbf{FT-Transformer} & \textbf{ResNet} & \textbf{Small MLP} & \textbf{FT-Transformer} & \textbf{ResNet} & \textbf{Small MLP} \\
    \hline
    5{,}000  & 0.413 $\pm$ 0.089 & 0.603 $\pm$ 0.039 & 0.554 $\pm$ 0.012 & 0.266 $\pm$ 0.028 & 0.472 $\pm$ 0.067 & 0.410 $\pm$ 0.041 \\
    10{,}000 & 0.413 $\pm$ 0.095 & 0.567 $\pm$ 0.061 & 0.639 $\pm$ 0.014 & 0.262 $\pm$ 0.032 & 0.479 $\pm$ 0.068 & 0.445 $\pm$ 0.014 \\
    20{,}000 & 0.421 $\pm$ 0.101 & 0.534 $\pm$ 0.072 & 0.679 $\pm$ 0.014 & 0.262 $\pm$ 0.030 & 0.480 $\pm$ 0.072 & 0.448 $\pm$ 0.024 \\
    \hline
    \end{tabular}
    }
\end{table}

Increasing the attack budget does not materially change the main privacy conclusions. At batch size \(32\), final attack point reconstruction averaged over the three model families changes only modestly from \(0.383\) at 5{,}000 iterations to \(0.395\) at 10{,}000 and \(0.397\) at 20{,}000. At batch size \(8\), the corresponding averages do not increase, moving from \(0.562\) to \(0.558\) and \(0.543\). FT-Transformer remains the least vulnerable model family, and the one-hot models remain more exposed. The attack budget therefore functions as a robustness check rather than a primary explanation for the observed leakage patterns.

\subsection{MIMIC-IV label unknown attacks}
Table~\ref{tab:mimic-label-unknown-tableak-acc} reports MIMIC-IV FedSGD attacks without true batch labels for the completed batch size settings up to \(128\). This setting weakens the main label known attacker by replacing the true labels with dummy labels while keeping the same observed client updates, model families, and reconstruction pipeline.

\begin{table}[H]
  \centering
  \small
  \caption{Label unknown reconstruction accuracy at the initialized attack point and final attack point for MIMIC-IV under FedSGD, reported for the completed batch size settings up to \(128\). The attacker does not receive the true labels. Initialized attack point denotes the attack before any global model aggregation, and final attack point denotes the last attacked point under the exposure budget. Each cell reports mean $\pm$ standard deviation across 3 seeds.}
  \label{tab:mimic-label-unknown-tableak-acc}
  \resizebox{\textwidth}{!}{%
  \begin{tabular}{lcccccc}
  \hline
  \multicolumn{1}{l}{} & \multicolumn{3}{c}{\textbf{Initialized attack point}} & \multicolumn{3}{c}{\textbf{Final attack point}} \\
  \textbf{Batch size} & \textbf{FT-Transformer} & \textbf{ResNet} & \textbf{Small MLP} & \textbf{FT-Transformer} & \textbf{ResNet} & \textbf{Small MLP} \\
  \hline
  1   & 0.254 $\pm$ 0.028 & 0.954 $\pm$ 0.076 & 0.963 $\pm$ 0.061 & 0.226 $\pm$ 0.013 & 0.852 $\pm$ 0.119 & 0.819 $\pm$ 0.162 \\
  2   & 0.300 $\pm$ 0.031 & 0.827 $\pm$ 0.075 & 0.925 $\pm$ 0.065 & 0.202 $\pm$ 0.102 & 0.568 $\pm$ 0.027 & 0.662 $\pm$ 0.020 \\
  4   & 0.343 $\pm$ 0.020 & 0.810 $\pm$ 0.085 & 0.917 $\pm$ 0.054 & 0.326 $\pm$ 0.106 & 0.658 $\pm$ 0.004 & 0.572 $\pm$ 0.050 \\
  8   & 0.354 $\pm$ 0.042 & 0.740 $\pm$ 0.034 & 0.871 $\pm$ 0.033 & 0.331 $\pm$ 0.182 & 0.712 $\pm$ 0.037 & 0.583 $\pm$ 0.081 \\
  16  & 0.426 $\pm$ 0.042 & 0.702 $\pm$ 0.059 & 0.802 $\pm$ 0.011 & 0.468 $\pm$ 0.219 & 0.650 $\pm$ 0.052 & 0.593 $\pm$ 0.054 \\
  32  & 0.246 $\pm$ 0.003 & 0.517 $\pm$ 0.047 & 0.628 $\pm$ 0.023 & 0.246 $\pm$ 0.002 & 0.399 $\pm$ 0.060 & 0.353 $\pm$ 0.017 \\
  64  & 0.365 $\pm$ 0.089 & 0.372 $\pm$ 0.064 & 0.488 $\pm$ 0.016 & 0.240 $\pm$ 0.084 & 0.308 $\pm$ 0.011 & 0.325 $\pm$ 0.032 \\
  128 & 0.389 $\pm$ 0.112 & 0.325 $\pm$ 0.082 & 0.413 $\pm$ 0.048 & 0.215 $\pm$ 0.059 & 0.228 $\pm$ 0.024 & 0.287 $\pm$ 0.018 \\
  \hline
  \end{tabular}
  }
\end{table}

Removing label knowledge reduces initialized attack point reconstruction for the one-hot models, but the final attack point effect is not uniformly monotone. Small one-hot clinical batches can still leak strongly without labels, with final attack point reconstruction of \(0.852\) for ResNet and \(0.819\) for the small MLP at batch size \(1\). Increasing client batch size still reduces label unknown reconstruction for the one-hot models. ResNet falls from \(0.852\) at batch size \(1\) to \(0.399\) at batch size \(32\) and \(0.228\) at batch size \(128\), while the small MLP falls from \(0.819\) to \(0.353\) and \(0.287\). FT-Transformer remains lower overall, with \(0.226\) at batch size \(1\), although its batch size pattern is less monotonic and peaks at batch size \(16\). Thus, true label access strengthens the attacker, but the main aggregation and architecture patterns remain visible under the weaker label unknown setting.

\section{Additional FT-Transformer Analyses and Ablations}
\label{app:fttransformer-analyses}
The FT-Transformer analyses provide model-specific validation and supporting results for the main architecture analysis. The Adult validation tables test whether the differentiable surrogate produces the intended training update. The MIMIC-IV tables and trajectories then report reconstruction behavior under the two categorical parameterizations and dropout settings.

\subsection{Attack parameterization validation}
We run these implementation ablations on Adult because it contains both numerical and categorical features while remaining small enough for controlled repeated checks. The purpose of the ablation is not to estimate dataset specific privacy risk. Instead, it verifies that the differentiable FT-Transformer attack parameterization produces the same training update as the native ordinal forward path. Larger datasets such as MIMIC-IV and the private multiclass benchmark are therefore unnecessary for this validation. Those datasets are used in the main attack experiments where dataset specific reconstruction difficulty matters.

To verify that the FT-Transformer attack does not optimize against an unrelated surrogate, we compare the updates produced by the native ordinal FT-Transformer forward pass with the corresponding updates produced by the differentiable GIA forward pass on the same batches. For FedSGD we compare per client gradients, and for FedAvg we compare one-step model deltas. The probability simplex parameterization reproduces the native update up to numerical precision, since one-hot relaxed probabilities induce the same embedding lookup as the native categorical path. The categorical logits parameterization used in the study, with $\tau=1$ and scale $5$, is a smooth approximation to the same path. Its induced update remains nearly collinear with the native update across all tested batch sizes. The resulting agreement is summarized in Table~\ref{tab:fttransformer-gia-signal-equivalence}.

\begin{table}[H]
  \centering
  \small
  \caption{Native FT-Transformer training update versus the differentiable GIA surrogate update on Adult. Entries report the worst case over batch sizes $B\in \{1,8,32,128\}$ and three clients.}
  \label{tab:fttransformer-gia-signal-equivalence}
  \begin{tabular}{llrrrr}
  \hline
  \textbf{FL protocol} & \textbf{Attack parameterization} & \textbf{Max abs. err.} & \textbf{Mean abs. err.} & \textbf{Rel. $\ell_2$ err.} & \textbf{Cosine} \\
  \hline
  FedSGD & Probability simplex & $1.49{\times}10^{-8}$ & $4.19{\times}10^{-12}$ & $8.04{\times}10^{-9}$ & $1.000000$ \\
  FedSGD & Categorical logits & $8.86{\times}10^{-3}$ & $7.04{\times}10^{-6}$ & $2.89{\times}10^{-3}$ & $0.999996$ \\
  FedAvg & Probability simplex & $7.45{\times}10^{-9}$ & $3.47{\times}10^{-14}$ & $1.73{\times}10^{-7}$ & $1.000000$ \\
  FedAvg & Categorical logits & $2.99{\times}10^{-5}$ & $6.42{\times}10^{-8}$ & $2.17{\times}10^{-3}$ & $0.999998$ \\
  \hline
  \end{tabular}
\end{table}

Table~\ref{tab:fttransformer-gia-forward-validity} verifies that both FT-Transformer categorical parameterizations define an optimizable differentiable surrogate. Starting from the GIA space reconstruction, gradient based updates consistently reduce the gradient matching loss computed against gradients generated by the native FT-Transformer forward pass.

\begin{table}[H]
    \centering
    \small
    \caption{FT-Transformer GIA forward validity check on Adult. The table reports the mean change in gradient matching loss after 25 attack steps across three clients. Negative values indicate that the GIA space reconstruction reduced the loss against gradients produced by the native FT-Transformer forward pass. All clients improved in every setting.}
    \label{tab:fttransformer-gia-forward-validity}
    \begin{tabular}{lcc}
    \hline
    \multicolumn{1}{l}{} & \textbf{Categorical logits} & \textbf{Probability simplex} \\
    \textbf{Batch size} & \textbf{Mean loss change} & \textbf{Mean loss change} \\
    \hline
    1   & -0.017 & -0.201 \\
    8   & -0.054 & -0.460 \\
    32  & -0.027 & -0.134 \\
    128 & -0.043 & -0.191 \\
    \hline
    \end{tabular}
\end{table}

\subsection{FT-Transformer reconstruction decomposition}
The reconstruction decomposition separates FT-Transformer performance into aggregate, numerical, and categorical accuracy on MIMIC-IV. The tables summarize the initialized attack point and final attack point across client batch sizes, categorical parameterizations, and dropout settings.

Tables~\ref{tab:fttransformer-reconstruction-decomposition-initialized} and~\ref{tab:fttransformer-reconstruction-decomposition-trained} show that numerical accuracy closely tracks aggregate reconstruction, while categorical accuracy remains lower at both attack points. This supports the architecture interpretation that categorical variables are harder to recover when they are represented through the FT-Transformer embedding interface. The same pattern appears under both categorical logits and probability simplex parameterizations, which indicates that the decomposition is not an artifact of a single attack parameterization.

\begin{table}[H]
    \centering
    \small
    \caption{FT-Transformer reconstruction decomposition at the initialized attack point on MIMIC-IV. Values are aggregated over client batch sizes $1$, $2$, $4$, $8$, $16$, $32$, $64$, and $128$. Each cell reports mean $\pm$ standard deviation across seeds and client batch sizes.}
    \label{tab:fttransformer-reconstruction-decomposition-initialized}
    \resizebox{\textwidth}{!}{%
    \begin{tabular}{lcccccc}
    \hline
    \multicolumn{1}{l}{} & \multicolumn{3}{c}{\textbf{Dropout on}} & \multicolumn{3}{c}{\textbf{Dropout off}} \\
    \textbf{Attack parameterization} & \textbf{Recon. Acc.} & \textbf{Num. acc.} & \textbf{Cat. acc.} & \textbf{Recon. Acc.} & \textbf{Num. acc.} & \textbf{Cat. acc.} \\
    \hline
    Categorical logits & 0.334 $\pm$ 0.074 & 0.338 $\pm$ 0.076 & 0.231 $\pm$ 0.059 & 0.337 $\pm$ 0.068 & 0.341 $\pm$ 0.071 & 0.233 $\pm$ 0.038 \\
    Probability simplex & 0.342 $\pm$ 0.070 & 0.346 $\pm$ 0.071 & 0.227 $\pm$ 0.047 & 0.341 $\pm$ 0.072 & 0.345 $\pm$ 0.075 & 0.243 $\pm$ 0.043 \\
    \hline
    \end{tabular}
    }
\end{table}

\begin{table}[H]
    \centering
    \small
    \caption{FT-Transformer reconstruction decomposition at the final attack point on MIMIC-IV. Values are aggregated over client batch sizes $1$, $2$, $4$, $8$, $16$, $32$, $64$, and $128$. Each cell reports mean $\pm$ standard deviation across seeds and client batch sizes.}
    \label{tab:fttransformer-reconstruction-decomposition-trained}
    \resizebox{\textwidth}{!}{%
    \begin{tabular}{lcccccc}
    \hline
    \multicolumn{1}{l}{} & \multicolumn{3}{c}{\textbf{Dropout on}} & \multicolumn{3}{c}{\textbf{Dropout off}} \\
    \textbf{Attack parameterization} & \textbf{Recon. Acc.} & \textbf{Num. acc.} & \textbf{Cat. acc.} & \textbf{Recon. Acc.} & \textbf{Num. acc.} & \textbf{Cat. acc.} \\
    \hline
    Categorical logits & 0.297 $\pm$ 0.132 & 0.300 $\pm$ 0.136 & 0.228 $\pm$ 0.053 & 0.256 $\pm$ 0.114 & 0.255 $\pm$ 0.115 & 0.266 $\pm$ 0.098
  \\
    Probability simplex & 0.284 $\pm$ 0.116 & 0.287 $\pm$ 0.120 & 0.210 $\pm$ 0.043 & 0.258 $\pm$ 0.107 & 0.259 $\pm$ 0.110 & 0.237 $\pm$ 0.062 \\
    \hline
    \end{tabular}
    }
\end{table}

\subsection{FT-Transformer leakage over training exposure}
\begin{figure}[H]
      \centering

      \begin{subfigure}[t]{0.9\textwidth}
          \centering
          \includegraphics[width=\textwidth]{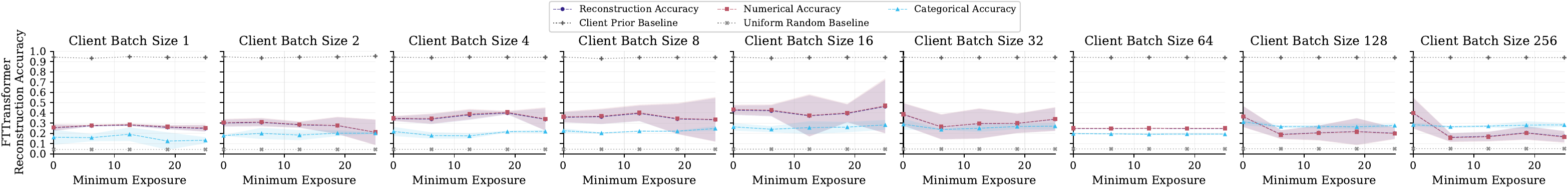}
          \caption{Categorical logits, dropout on}
      \end{subfigure}

      \vspace{0.5em}

      \begin{subfigure}[t]{0.9\textwidth}
          \centering
          \includegraphics[width=\textwidth]{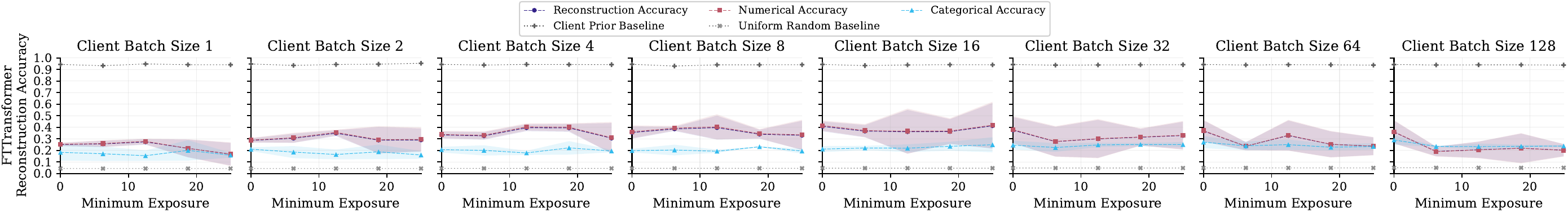}
          \caption{Probability simplex, dropout on}
      \end{subfigure}

      \vspace{0.5em}

      \begin{subfigure}[t]{0.9\textwidth}
          \centering
          \includegraphics[width=\textwidth]{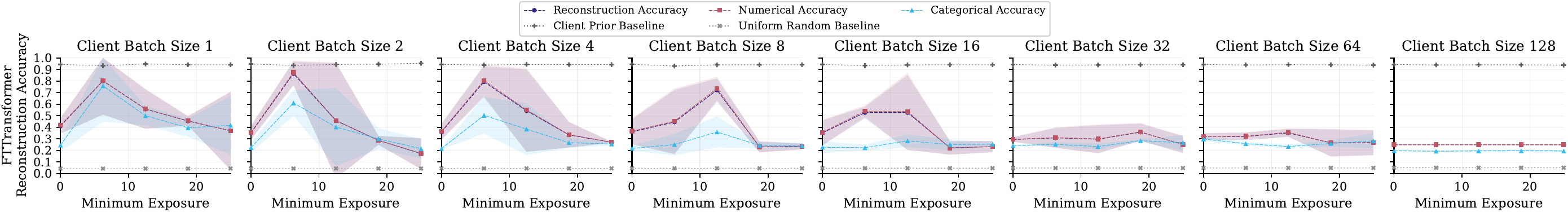}
          \caption{Categorical logits, dropout off}
      \end{subfigure}

      \vspace{0.5em}

      \begin{subfigure}[t]{0.9\textwidth}
          \centering
          \includegraphics[width=\textwidth]{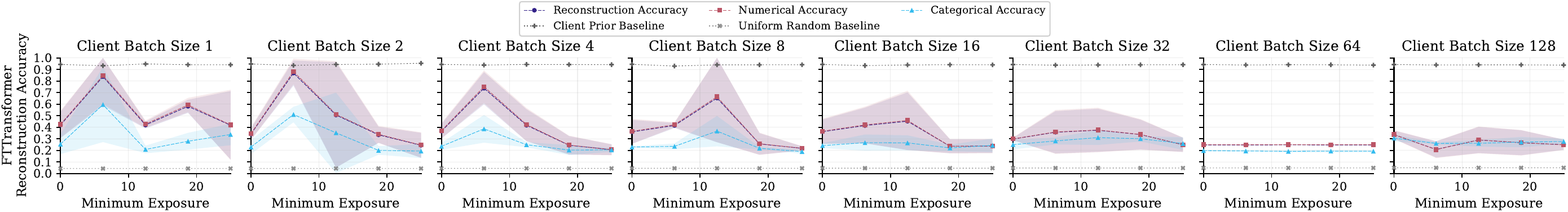}
          \caption{Probability simplex, dropout off}
      \end{subfigure}

      \caption{FT-Transformer reconstruction trajectories on MIMIC-IV over the five attack points used in the trajectory analyses. Each panel reports aggregate, numerical, and categorical reconstruction accuracy across client batch sizes for one categorical parameterization and dropout setting. The dropout disabled panels show a pronounced early leakage spike for small client batches. Panel (a) includes batch size \(256\). The remaining panels omit this batch size for computational reasons.}
      \label{fig:fttransformer-parameterizations-dropout-mimic-num-cat-trajectories}
\end{figure}

Figure~\ref{fig:fttransformer-parameterizations-dropout-mimic-num-cat-trajectories} reports FT-Transformer reconstruction over client exposure using the same five attack points as the other trajectory analyses. Each panel plots aggregate, numerical, and categorical reconstruction accuracy for one categorical parameterization and dropout setting. The comparison shows that the FT-Transformer reconstruction behavior is visible across the training trajectory and not only at the initialized attack point and final attack point.

The dropout disabled trajectories also show why initialized attack point and final attack point summaries are insufficient for this ablation. On MIMIC-IV, removing dropout produces a sharp early leakage spike for small client batches under both categorical parameterizations, followed by a later decline. This supports the interpretation that dropout affects the timing and stability of FT-Transformer leakage, not only its final attack point magnitude.

Table~\ref{tab:fixed-batch-mimic-fttransformer-attack-paths} reports the fixed client batch control for the two FT-Transformer categorical parameterizations. The fixed batch comparison preserves the main result from the resampled batch experiment. Categorical logits and probability simplex produce similar reconstruction levels, with only modest differences at batch sizes \(8\) and \(32\). This supports the main text conclusion that the lower FT-Transformer leakage is not driven by a single categorical relaxation choice.

\begin{table}[H]
  \centering
  \small
  \caption{Reconstruction accuracy at the initialized attack point and final attack point for MIMIC-IV under two FT-Transformer attack parameterizations with fixed client batches. Initialized attack point denotes the attack before any global model aggregation, and final attack point denotes the last attacked point under the exposure budget. Each cell reports mean $\pm$ standard deviation across 3 seeds.}
  \label{tab:fixed-batch-mimic-fttransformer-attack-paths}
  \begin{tabular}{lcccc}
  \hline
  \multicolumn{1}{l}{} & \multicolumn{2}{c}{\textbf{Initialized attack point}} & \multicolumn{2}{c}{\textbf{Final attack point}} \\
  \textbf{Batch size} & \textbf{Categorical logits} & \textbf{Probability simplex} & \textbf{Categorical logits} & \textbf{Probability simplex} \\
  \hline
  8 & 0.341 $\pm$ 0.020 & 0.340 $\pm$ 0.025 & 0.367 $\pm$ 0.111 & 0.330 $\pm$ 0.129 \\
  32 & 0.330 $\pm$ 0.009 & 0.334 $\pm$ 0.012 & 0.365 $\pm$ 0.059 & 0.357 $\pm$ 0.048 \\
  \hline
  \end{tabular}
\end{table}

Table~\ref{tab:batch-size-mimic-fttransformer-attack-paths-dropout0} reports the same initialized and final attack point comparison with attention and feedforward dropout disabled in the target model. Removing dropout increases leakage most clearly at batch size \(1\), where initialized attack point reconstruction rises from \(0.254\) to \(0.411\) under categorical logits and from \(0.251\) to \(0.420\) under probability simplex. Beyond the smallest batch setting, however, dropout removal does not produce a consistent increase. At batch sizes \(8\), \(16\), and \(32\), final attack point reconstruction is lower than in the corresponding dropout enabled runs for both categorical parameterizations.

\begin{table}[H]
  \centering
  \small
  \caption{Reconstruction accuracy at the initialized attack point and final attack point for MIMIC-IV under two FT-Transformer attack parameterizations with attention and feedforward dropout set to zero in the target model. Each cell reports mean $\pm$ standard deviation across 3 seeds.}
  \label{tab:batch-size-mimic-fttransformer-attack-paths-dropout0}
  \begin{tabular}{lcccc}
  \hline
  \multicolumn{1}{l}{} & \multicolumn{2}{c}{\textbf{Initialized attack point}} & \multicolumn{2}{c}{\textbf{Final attack point}} \\
  \textbf{Batch size} & \textbf{Categorical logits} & \textbf{Probability simplex} & \textbf{Categorical logits} & \textbf{Probability simplex} \\
  \hline
  1   & 0.411 $\pm$ 0.059 & 0.420 $\pm$ 0.100 & 0.370 $\pm$ 0.297 & 0.418 $\pm$ 0.261 \\
  2   & 0.351 $\pm$ 0.051 & 0.341 $\pm$ 0.044 & 0.173 $\pm$ 0.117 & 0.244 $\pm$ 0.095 \\
  4   & 0.359 $\pm$ 0.059 & 0.366 $\pm$ 0.056 & 0.270 $\pm$ 0.004 & 0.206 $\pm$ 0.041 \\
  8   & 0.361 $\pm$ 0.095 & 0.360 $\pm$ 0.091 & 0.234 $\pm$ 0.022 & 0.217 $\pm$ 0.016 \\
  16  & 0.351 $\pm$ 0.096 & 0.361 $\pm$ 0.093 & 0.233 $\pm$ 0.043 & 0.237 $\pm$ 0.052 \\
  32  & 0.294 $\pm$ 0.017 & 0.298 $\pm$ 0.009 & 0.251 $\pm$ 0.065 & 0.249 $\pm$ 0.054 \\
  64  & 0.321 $\pm$ 0.023 & 0.248 $\pm$ 0.001 & 0.267 $\pm$ 0.095 & 0.247 $\pm$ 0.001 \\
  128 & 0.249 $\pm$ 0.001 & 0.336 $\pm$ 0.031 & 0.248 $\pm$ 0.001 & 0.249 $\pm$ 0.039 \\
  \hline
  \end{tabular}
\end{table}

\section{Additional Implementation Ablations}
The implementation checks support the experimental pipeline by verifying that vectorized training and attack execution preserve the same FL updates and attack objectives as the standard client wise implementation, up to finite precision. These checks are not intended to estimate dataset specific privacy risk.

\subsection{Vectorized training and attack equivalence}
We validate the vectorized implementation by comparing it against the standard client wise implementation on Adult with ten clients and three seeds. The standard and vectorized paths implement the same mathematical computation for each client. The vectorized path does not change the FL objective, the attack objective, or the gradients used by the attack. It only changes how independent client computations are scheduled. Instead of looping over clients one at a time, the vectorized implementation evaluates the same per client computation in a batched form. This trades higher peak memory use and GPU utilization for increased throughput in the full experimental evaluation.

The observed FL updates match up to finite precision for both FedSGD and FedAvg, as summarized in Table~\ref{tab:vectorized-training-gia-equivalence}. We also compare the GIA objective after one and ten attack steps using the attack learning rate used in the main experiments. The one-step comparison checks the immediate vectorized attack update, while the ten step comparison verifies that numerical differences remain small across repeated optimization. The cosine matching loss computes the cosine similarity in double precision after flattening the compared gradient tensors. Thus, the loss comparison itself is not a low precision reduction.

\begin{table}[H]
  \centering
  \small
  \caption{Equivalence of standard and vectorized FL and GIA implementations on Adult. Values are worst cases over seeds $7$, $13$, and $42$, ten clients, and both vectorized attack comparison directions.}
  \label{tab:vectorized-training-gia-equivalence}
  \begin{tabular}{lrrrr}
  \hline
  \textbf{FL protocol} & \textbf{GIA steps} & \textbf{FL max err.} & \textbf{GIA total loss diff.} & \textbf{GIA client loss diff.} \\
  \hline
  FedSGD & $1$ & $2.38{\times}10^{-7}$ & $4.77{\times}10^{-7}$ & $2.98{\times}10^{-8}$ \\
  FedSGD & $10$ & $2.38{\times}10^{-7}$ & $3.02{\times}10^{-3}$ & $4.68{\times}10^{-4}$ \\
  FedAvg & $1$ & $1.19{\times}10^{-7}$ & $4.77{\times}10^{-7}$ & $1.49{\times}10^{-7}$ \\
  FedAvg & $10$ & $1.19{\times}10^{-7}$ & $3.23{\times}10^{-3}$ & $4.79{\times}10^{-4}$ \\
  \hline
  \end{tabular}
\end{table}

The one-step discrepancies are at the scale of single precision roundoff. This is expected because the standard and vectorized implementations evaluate the same algebraic computation with different execution orders. The ten step differences reflect the accumulation of these finite precision effects through repeated reconstruction optimizer updates. These results support using the vectorized implementation in the full experimental evaluation.

\end{document}